\pdfoutput=1
\AtBeginDocument{%
  \renewcommand{\answerYes}{\textcolor[rgb]{0,0,1}{[Yes]}}%
  \hypersetup{hidelinks}%
}

\PassOptionsToPackage{sort&compress,numbers}{natbib}

\documentclass{article}

\usepackage[square,numbers]{natbib}

\usepackage[sort&compress,numbers]{natbib}

\usepackage[english]{babel}

\usepackage[main,final]{neurips_2026}

\usepackage{amsfonts}
\usepackage{amsmath}
\usepackage{amsthm}
\usepackage{amssymb}

\newtheorem{theorem}{Theorem}[section]

\newtheorem{proposition}{Proposition}[section]
\newtheorem{lemma}[theorem]{Lemma}

\usepackage{nameref}

\usepackage{hyperref}
\usepackage{xr}
\usepackage{zref}

\usepackage{blindtext}

\usepackage{titlesec}

\usepackage{graphicx}
\graphicspath{ {./images/} }

\usepackage[utf8]{inputenc} %
\usepackage[T1]{fontenc}    %
\usepackage{hyperref}       %
\usepackage{url}            %
\usepackage{booktabs}       %
\usepackage{amsfonts}       %
\usepackage{nicefrac}       %
\usepackage{microtype}      %

\PassOptionsToPackage{table}{xcolor}
\usepackage{xcolor}         %
\usepackage{makecell}       %
\usepackage{multirow}       %
\usepackage{colortbl}       %
\usepackage{siunitx}        %
\usepackage{pdflscape}      %
\usepackage{placeins}       %
\usepackage{afterpage}      %
\usepackage{subcaption}     %
\usepackage{needspace}

\title{One Temperature to Rule Them All?}

\author{%
  Aviv Orly\thanks{Equal contribution.} \qquad
  Ori Shem-Ur\footnotemark[1] \qquad
  Yaron Oz \\
  School of Physics and Astronomy 
  \\
  Tel Aviv University, Israel \\
  \texttt{\{\href{mailto:avivorly@mail.tau.ac.il}{avivorly},
  \href{mailto:orishemor@mail.tau.ac.il}{orishemor}\}@mail.tau.ac.il}
  \quad
  \texttt{\href{mailto:yaronoz@tauex.tau.ac.il}{yaronoz@tauex.tau.ac.il}}
}

\begin{document}

\maketitle

\begin{abstract}
Inference-time sampling is a central mechanism for improving large language models, with temperature controlling the tradeoff between reliability and diversity. While recent work suggests that varying temperature across samples can improve multi-sample inference, simple fixed-temperature schemes remain the standard in practice. We investigate this discrepancy in the canonical pass@$K$ setting, asking when temperature diversity is useful, and when a single tuned temperature suffices. To this end, we analyze per-question success probabilities across multiple models and datasets, tracking how they vary with sampling temperature. We identify several consistent properties of these profiles, including that harder questions are less likely to share efficient sampling temperatures, although this trend is only logarithmic in hardness. We then derive an efficient method for evaluating the theoretically optimal multi-temperature allocation scheme, and show that although such a scheme improves performance, its gains over single-temperature baselines are typically modest when temperature is tuned as a function of the sampling budget $K$. We explain this phenomenon through mathematical mechanisms that we call dulling effects, and show how they ensure the robustness of temperature choices across different question distributions. Our results clarify when multi-temperature sampling is useful, and how temperatures should ideally be chosen.
\end{abstract}

\section{Introduction}

\label{sec:Intro}

Modern machine learning has made remarkable progress over the past decade. In particular, large language models (LLMs) have emerged as general-purpose foundations for a wide range of tasks, including question answering, summarization, program synthesis, and mathematical reasoning \cite{vaswani2017attention,bommasani2021opportunities,brown2020language,chowdhery2023palm,touvron2023llama,achiam2023gpt4,chen2021evaluating,austin2021program,li2022competition,cobbe2021training,hendrycks2021measuring,lewkowycz2022solving}. Most contemporary LLMs are trained as next-token predictors, optimizing the likelihood of human-generated text, often followed by reinforcement learning-based alignment objectives \cite{radford2019language,kaplan2020scaling,hoffmann2022training,christiano2017deep,ouyang2022training,bai2022training}. Despite the apparent simplicity of this 
paradigm, these models exhibit increasingly capable forms of inference, including multi-step reasoning and 
problem solving at human-level and beyond \cite{wei2022emergent,wei2022chainofthought,kojima2022large,zhou2023least,yao2023tree,schick2023toolformer}.

One major driver of these recent improvements in LLM performance is the shift from relying solely on training-time scaling toward investing more computation at inference time \cite{snell2024scaling,brown2024large,wu2025role}. Modern systems increasingly allocate additional resources to inference through techniques such as chain-of-thought reasoning and related structured prompting methods \cite{wei2022chainofthought,kojima2022large,zhou2023least,yao2023tree,besta2024graph}. These inference-time strategies can substantially improve accuracy without retraining the underlying model \cite{snell2024scaling,wang2023selfconsistency,uesato2022solving,lightman2023lets}. As a result, there is growing interest in \emph{test-time} (or \emph{inference-time}) compute, which trades extra sampling, reasoning, and verification for better performance \cite{snell2024scaling,brown2024large,wu2025role}.

\newpage

A central family of inference-time methods is \emph{multiple sampling}, where the model is queried several times for the same problem, and the resulting candidates are compared, filtered, or aggregated \cite{chen2021evaluating,li2022competition,cobbe2021training,wang2023selfconsistency,brown2024large}. Multiple sampling can be implemented in various ways, depending on the task structure and the available supervision signal. 

For a multiple-sampling strategy to provide meaningful benefits, rather than repeatedly producing the same result, we require a mechanism that introduces variation. A standard approach for achieving such diversity is the use of \emph{temperature}. Instead of deterministically selecting the most likely next token, sampling draws from a temperature-scaled softmax distribution, i.e., a Boltzmann-like distribution \cite{ackley1985learning,hinton2015distilling,holtzman2020curious}. In this approach, higher temperatures increase randomness and encourage the exploration of less likely continuations, which can increase the probability of discovering novel or unconventional solutions \cite{holtzman2020curious,hewitt2022truncation,meister2023locally,renze2024effect,fan2018hierarchical}. This however, comes at the cost of reliability, as the temperature increases, samples are more likely to contain errors, inconsistencies, or nonsensical outputs \cite{holtzman2020curious,hewitt2022truncation,meister2023locally,renze2024effect,du2025optimizingtemperature,fan2018hierarchical}. The effectiveness of multiple sampling thus depends on balancing exploration and exploitation.

Recent work has highlighted that this trade-off is not uniform across inputs. Different questions can exhibit substantially different \emph{optimal} sampling behavior. Some are best solved with low-temperature, highly deterministic inference, while others benefit from increased exploration \cite{renze2024effect,du2025optimizingtemperature,wu2025role,cecere2025monte,levi2025simple}. This observation motivates a natural extension of classical multiple sampling: using multiple temperatures within the same sampling process. 

Several works investigating this approach, have relied on dynamically adjusting the temperature during inference based on the properties of the input \cite{wang2020contextual,cai2024t2,du2025optimizingtemperature}, some even at a token-by-token level \cite{zhu2024hot,zhang2024edt,xie2024calibrating}. While these approaches are effective, they can also introduce a substantial computational cost. In contrast \cite{wu2025role} demonstrated that this additional cost can be avoided, as even pre-committed allocation schemes, such as sampling temperatures uniformly across the relevant temperature range, can offer substantial benefits over single-temperature strategies.

While these results suggest that multi-temperature sampling should yield substantial gains, single-temperature sampling remains the dominant approach in practical applications. The goal of this paper is to explain this apparent disparity, and to expand our understanding of temperature selection in LLM inference. 

\textbf{Contributions:}

We show that the probability of solving LLM questions as a function of temperature exhibits a simple structure, typically characterized by a single local maximum in temperature, and that questions can be classified according to whether this optimum occurs at zero or at a positive temperature (Section \ref{sec:UniMod}).

We then study question behavior as a function of hardness, defined as the number of samples required for solution (Equation \ref{eq:def:Hardness}), showing that the average optimal temperature varies significantly with hardness (Section \ref{sec:AveTemp}). We further show that both the range of efficient temperatures, and their overlap across questions shrink with hardness, albeit only logarithmically (Section \ref{sec:range}).

We then derive and prove a method for efficiently approximating the optimal multi-sampling strategy for the pass@$K$ setting (Section \ref{sec:Multiple:Ideal}), and compare it with existing single, and multi-temperature approaches. While single-temperature strategies typically underperform, much of the gap can be recovered by adapting the temperature to the number of samples (Section \ref{sec:Multiple:Comp}).

{Finally, we explain these effects, and identify practical advantages of these mechanisms. First, the optimal sampling parameters depend more strongly on the computational budget than on individual question hardness, reducing the need to estimate the latter. This robustness also allows the sampling parameters to be selected without closely matching the difficulty distributions of the tuning and test datasets (Section \ref{sec:Dulling}).}

\section{Preliminaries}

\subsection{Notation}

\label{sec:Notations}

We begin by introducing the main notation used throughout the paper. We denote by $\tau\geq0$ the temperature used during the inference process, and by $\mathcal{Q}$ the distribution over the questions, for which we study how this choice affects performance. The individual questions drawn from this distribution are then denoted by $q\sim\mathcal{Q}$.

We assign $P_q : [0,\infty) \rightarrow [0,1]$ for the probability that a question $q$ is solved in a single attempt, and write $P$ when the dependence on $q$ is clear from context. Accordingly, $P(\tau)$ denotes the probability of solving the question in a single attempt for $\tau$. We refer to the function $P(\tau)$ as the \textit{temperature profile} of the question.
Since in many of our computations the failure probability is more convenient than the success probability, we designate for it the notation $\bar{P}=1-P$.

\subsection{Pass@K}

\label{sec:passK}

We conduct our analysis of multi-sampling approaches within the canonical \emph{pass@$K$} metric. In this setting, we generate $K \in \mathbb{N}$ independent solutions, and consider the task successful if at least one of them is correct \cite{chen2021evaluating,kulal2019spoc,brown2024large,kazdan2025efficient}. Accordingly, for $K$ separate attempts, we can express the probability of failure over the corresponding $K$ temperatures $\tau_1, \dots, \tau_K$, as:
\begin{equation}
\bar{\mathcal{P}}\left(\tau_1,\dots,\tau_K\right) = \bar{P}\left(\tau_{1}\right)\cdots\bar{P}\left(\tau_{K}\right) \ .
\label{eq:MultiTempProb:Basic}
\end{equation}
In the common special case where all attempts use the same temperature $\tau$, this reduces to:
\begin{equation}
\bar{\mathcal{P}} = \bar{P}\left(\tau\right)^{K} \ .
\label{eq:PowerK}
\end{equation}
While the pass@$K$ approach is especially effective when the correctness can be verified automatically, for example by running unit tests for code generation, or checking a formal solution \cite{chen2021evaluating,austin2021program,li2022competition,lewkowycz2022solving,jain2024livecodebench}, it can also be extended using learned verifiers, including auxiliary LLMs that evaluate candidate solutions \cite{cobbe2021training,uesato2022solving,lightman2023lets,snell2024scaling}.

Furthermore, many high-performing model variants rely heavily on this general pattern of sampling multiple candidates followed by selection \cite{li2022competition,cobbe2021training,wang2023selfconsistency,lightman2023lets,snell2024scaling,brown2024large}. In fact, many of the properties of the pass@$K$ framework extend to related multi-sampling approaches, such as chain-of-thought reasoning \cite{wei2022chainofthought,kojima2022large,wang2023selfconsistency}. As such, this framework is both useful in its own right, and serves as a representative setting for understanding the general properties of more advanced multi-sampling methods.

\subsection{Empirical Setup}

We evaluate four models, Qwen3-0.6B, Qwen3-8B, Mistral 7B Instruct, and GPT-OSS 20B, on three benchmarks: LiveCodeBench (V1-V6) \cite{jain2024livecodebench} for code generation, MATH-500 \cite{hendrycks2021measuring,lightman2023lets} for mathematics, and Hi-ToM \cite{he2023hitom} for multiple-choice higher-order theory of mind. For each model and question, we draw typically 1,500 samples at each of 16 to 38 temperatures between 0.075 and 4.2, and grade them automatically by unit tests, final answers, and chosen options, respectively (Appendix \ref{zap:CompResources}).

\subsection{\texorpdfstring{Other Sampling Hyperparameters}{Other Sampling Hyperparameters}}

{
While we focus our analysis on sampling temperature, as it is the principal hyperparameter governing inference-time sampling \cite{du2025optimizingtemperature,wu2025role}, our results can be generalized to other sampling hyperparameters, including $\text{min}_p$ and $\text{top}_h$ \cite{nguyen2025minp,potraghloo2025toph}, as discussed 
for $\text{min}_p$ in Appendix \ref{zap:MinP}.}

\section{Properties of Question Profiles}

\label{sec:Profiles}

We begin by presenting our findings on the characteristics of question temperature profiles, defined in Section \ref{sec:Notations}, along with an observation on the typical overlap of the temperature ranges in which different questions can be solved efficiently. To formalize this, we introduce several characteristics that capture the fundamental properties of a question. The consistency of these patterns across all datasets and models we consider implies that they may reflect a more universal property of large language model inference.

\subsection{Unimodality and Typical Profile Shapes}

\label{sec:TypicalProfile}

\label{sec:UniMod}

Our first observation is that question profile shapes are qualitatively simple, and can be well approximated by one of two canonical forms. Typically, profiles either begin at $1$ for $\tau = 0$ and then monotonically decay toward zero, or they begin with $P(0) = 0$, increase monotonically to some optimal temperature, and then decay.

\begin{figure}[t]
   \centering
\includegraphics[width=0.75\linewidth]{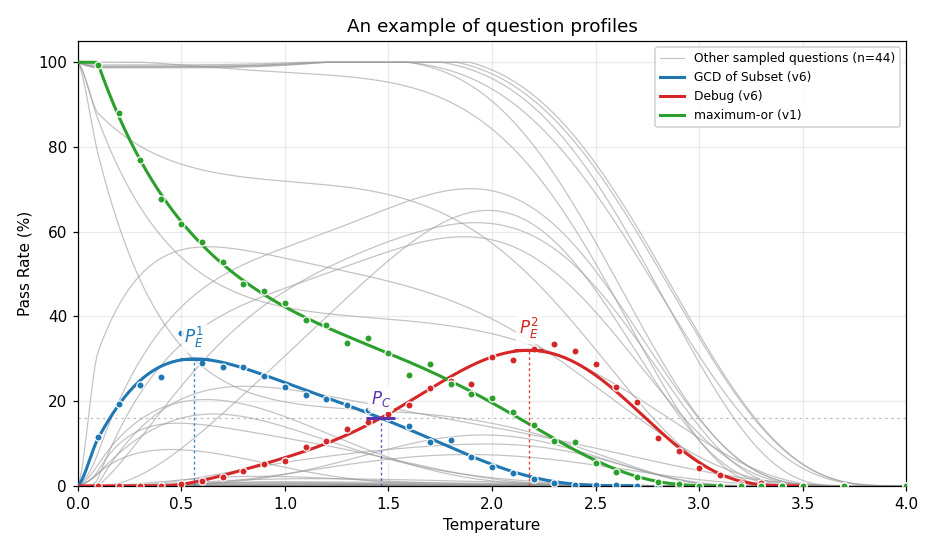}
\caption{Temperature profiles for 51 Qwen3-8B questions. The green curve, which attains its maximum at $\tau = 0$, illustrates the first canonical form, whereas the red and blue curves, which peak at positive temperatures with corresponding probabilities $P_E^{1}$ and $P_E^{2}$, illustrate the second.}
\label{fig:ManyQ}
\end{figure}

We illustrate these profiles in Figure \ref{fig:ManyQ}, which shows the probability of solving different live-code questions as a function of their temperature for the Qwen3-8B model. 

In our experiments, presented in full in {Appendix \ref{zap:QuestionsProfiles}}, only a negligible fraction of the questions considered across all models deviated from this pattern. These deviations were generally minor, with most not clearly distinguishable from statistical uncertainty.

While the two profile types appear very different, they arise from a shared underlying property, namely, unimodality, or the existence of a single local maximum. At $\tau = 0$, once external sources of randomness are removed, the model behaves deterministically, implying that $P(0) \in \{0,1\}$. At the other extreme, as $\tau \to \infty$, performance degrades toward random behavior, and $P(\tau)$ approaches near zero \cite{holtzman2020curious,hewitt2022truncation,meister2023locally,renze2024effect,du2025optimizingtemperature,fan2018hierarchical}. These boundary conditions, together with 
unimodality, imply that the two profile types we have found are the only possible qualitative forms, which explains our observation.

An important common sub-profile, consists of questions with $P(0) = 0$, but with a maximum very close to $\tau = 0$. These questions are typically easy, in the sense that they have relatively high single-sample success probability, and appear to require only a small amount of stochasticity to nudge away from the incorrect deterministic solution.

Estimating the distribution for each question required careful statistical analysis, especially for very difficult questions with only a few positive solutions, where poor estimation can dramatically distort the resulting temperature profiles. We describe this analysis in detail in Appendix \ref{zap:Stats:Methodology}.

\subsection{Question Behavior by Hardness}

A natural property to consider for a question in the context of multiple-sampling, is how many samples are required for it to be solved reliably. In the pass@$K$ setting, this can be naturally defined as the number of attempts required to achieve a success probability of at least $1 - \frac{1}{e}$. Using Equation \ref{eq:PowerK}, and relaxing the integrality constraint on $K$, this yields according to the notation in Section \ref{sec:Notations}:
\begin{equation}
H_{q}\left(\tau\right) = -\frac{1}{\ln\left(1 - P_{q}\left(\tau\right)\right)} = -\frac{1}{\ln\left(\bar{P}_{q}\left(\tau\right)\right)} \ ,
\label{eq:def:Hardness}
\end{equation}
where $H_q(\tau)$ denotes the hardness of question $q$ at temperature $\tau$. We further define the overall hardness of a question (with a slight abuse of notation) as $H_q = \inf_{\tau > 0} H_q(\tau)$. To classify questions by hardness, we denote the restriction of the distribution $\mathcal{Q}$ to questions whose hardness satisfies $H_- \leq H_q < H_+$ as: 
$\mathcal{Q}_{H_-,H_+}$.

Throughout this work, we will see the usefulness of this notion in the context of the pass@$K$ framework. We begin with the following two subsections, which present important trends in question behavior as a function of their hardness. 

\subsubsection{Average Ideal Temperature}

\label{sec:AveTemp}

Our first observation in this section is that the average optimal temperature for solving different questions varies as a function of hardness. We illustrate this effect in the left panel of Figure \ref{fig:Seperation} and in {Figures \ref{fig:AveTemp:grid}, \ref{fig:AveTemp:math500:grid}, and \ref{fig:AveTemp:hitom:grid}}, where we plot the optimal temperature for each question as a function of its hardness, along with a sliding average over all questions. To isolate the effect of the easy questions whose optimal temperature does not lie very close to $\tau = 0$, we also plot a separate sliding average for this subset.

As shown in the figure, the optimal temperature required to solve questions varies not only across individual questions, but also in its average behavior across different hardness levels. This distinction is important, as these two sources of variation have different implications for the design of multi-temperature sampling strategies, as we discuss in Section \ref{sec:Dulling:gold}.

\subsubsection{Efficient Temperature Range}

\label{sec:range}

\label{sec:QSep}

Our second observation in this section is that as hardness increases, both the variability of efficient temperatures across questions, and the fraction of questions that share a common efficient temperature decrease, albeit only logarithmically.

To formalize these observations, we define the efficient temperature range of a question, as the set of temperatures whose success probability is at least half of its optimal success probability:
\begin{equation}
E_q=\left\{\tau\geq 0 \mid P_q(\tau)\geq \frac{1}{2}\sup_{\tau'\geq 0}P_q(\tau')\right\} \ .
\label{eq:efficient range}
\end{equation}

As we discuss in Section \ref{sec:Dulling:short}, when considering two questions in the pass@$K$ setting, it is typically more efficient to use two different temperatures, rather than a single shared temperature, only when their efficient ranges do not overlap. As such, this definition provides a useful characterization of the separation between question profiles within a distribution.

\begin{figure}[t]
   \centering
\includegraphics[width=1\linewidth]{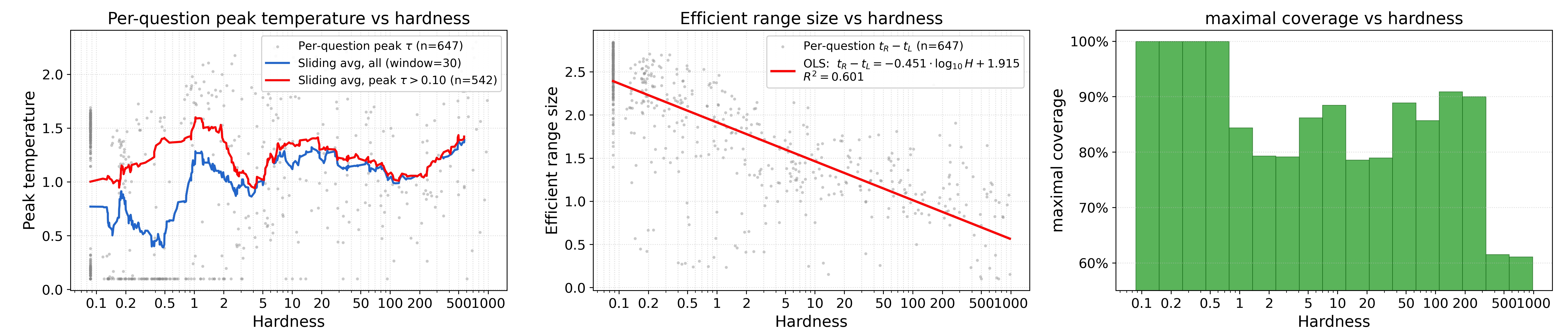}
\caption{Three hardness-based views on Qwen3-8B on LiveCodeBench: peak temperature (left), efficient range size (middle), and maximal overlap ratio of efficient ranges (right) vs. hardness (log$_{10}$ scale). Left: blue/red curves show sliding averages over all instances, with/without small ideal-temperature instances. Middle: linear fit over instances. Right: maximal coverage in log-uniform hardness bins. Temperatures vary with hardness, while range size and overlap shrink logarithmically.
\label{fig:Seperation}}

\end{figure}

In the middle panel of Figure \ref{fig:Seperation}, and in Figures \ref{fig:RangeSize:grid}, \ref{fig:RangeSize:math500:grid}, and \ref{fig:RangeSize:hitom:grid}, we examine how the sizes of these ranges vary with hardness, where we plot for each question its hardness against its efficient range size, and fit a linear trend as a function of log-hardness. The resulting relationship indicates a logarithmic decay in efficient range size as the hardness increases.

This observation suggests that a similar trend may hold for the overlap between questions' efficient ranges. To quantify this, we consider, for each hardness level $H$, the maximal fraction of questions with hardness $H$, that can be efficiently solved using the same temperature:
\begin{equation}
R_{\mathcal{Q}}(H)
=
\lim_{\Delta H \to 0}
\sup_{0 \leq \tau}
\Pr_{q \sim \mathcal{Q}_{H,H+\Delta H}}
\left(
\tau \in E_q
\right) \ .
\label{eq:SharedEfficientTemperature}
\end{equation}
In the right panel of Figure \ref{fig:Seperation}, and in Figure \ref{fig:relC:grid}, we plot an empirical approximation of this quantity using a finite hardness window $\Delta H$ on a logarithmic hardness scale over our dataset.

A logarithmic trend is still observed, though it is less pronounced than the decay in efficient range sizes. This effect is driven by the variation in the average optimal temperature across hardness levels. In particular, the cluster of easy questions whose optimal temperature lies very close to $\tau = 0$ obscures the overall trend at small $K$ for some of the models. These results support and refine the observation of \cite{wu2025role} that as questions become harder, they share fewer efficient temperatures.

\section{Multiple Temperature Sampling}

\label{sec:Multiple}

Having established the typical structure of question temperature profiles, we now analyze the relative effectiveness of pre-committed multi-temperature strategies in the pass@$K$ setting. This includes the ideal sampling strategy, which we show can be efficiently approximated.

We show that although these approaches provide a measurable advantage over fixed single-temperature strategies, this advantage is relatively limited, and a large portion of it can be recovered by using a single temperature adapted to the available computational budget. We then explain the underlying reasons for this behavior, and how it can be leveraged for practical applications.

\subsection{The Ideal Multiple-Temperature Allocation Scheme}
\label{sec:TheIdeal}
\label{sec:Multiple:Ideal}

{
We begin this section by showing that, although the space of possible multi-temperature strategies is potentially infinite, in practice, it can be approximated effectively.} 
In the pass@$K$ setting, this strategy is fully defined by the selection of temperatures that maximize the success probability, or equivalently, minimize the failure probability, which according to Equation \ref{eq:MultiTempProb:Basic}, can be written as:
\begin{equation}
\bar{\mathcal{P}}\left(\tau_{1},\ldots,\tau_{K}\right)=\mathbb{E}_{q\sim\mathcal{Q}}\bar{\mathcal{P}}_{q}\left(\tau_{1},\ldots,\tau_{K}\right) \ .
\label{eq:MultiTempProb:data}
\end{equation}
While directly optimizing over a continuous space of temperatures is challenging, we can discretize the set of candidate temperatures to a small set of $l \in \mathbb{N}$ fixed values $0 \leq \gamma_1, \dots, \gamma_l$, while still preserving near-optimal performance, due to the linear dependence on the $\bar{P}(\tau)$s in Equation \ref{eq:MultiTempProb:Basic}, as shown in Theorem \ref{zap:the:TempDis}.

This allows us to shift from optimizing over the choice of temperatures, to optimizing the allocation of samples across the fixed temperature set, denoted by $K_1,\dots,K_l \in \mathbb{N}_0$, subject to:
\begin{equation}
K_1+\cdots+K_l=K \ .
\label{eq:KSum}
\end{equation}
Under this formulation, the average failure probability becomes:
\begin{equation}
\bar{\mathcal{P}}\left(K_{1},\ldots,K_{l}\right)=\mathbb{E}_{q\sim\mathcal{Q}}\prod_{i=1}^{l}\bar{P}_{q}\left(\gamma_{i}\right)^{K_{i}}=\mathbb{E}_{q\sim\mathcal{Q}}\exp\left(\sum_{i=1}^{l}K_{i}\ln\left(\bar{P}_{q}\left(\gamma_{i}\right)\right)\right)\ .
\label{eq:ProbFailTot}
\end{equation}
As $\mathcal{Q}$ may be infinite, this optimization problem remains challenging. However, by standard concentration arguments, and the symmetry in the ordering of temperature choices, we show in Theorem \ref{zap:lem:FiniteQ} that it is sufficient to approximate the expectation using a relatively small, randomly sampled subset of questions.

We can therefore {simplify} the problem to a finite optimization task by denoting the probability of failure in a single attempt for each question $i = 1, \ldots, n$ and temperature index $j = 1, \ldots, l$ by:
\begin{equation}
\bar{P}_{ij} = \bar{P}_{q_i}\left(\gamma_j\right) \ ,
\end{equation}
and find that the optimization problem reduces to finding integers $K_1, \ldots, K_l$ satisfying Equation \ref{eq:KSum}, that minimize the average failure probability:
\begin{equation}
\bar{\mathcal{P}}\left(K_{1},\ldots,K_{l}\right) = \frac{1}{n}\sum_{i=1}^{n} \exp\left( \sum_{j=1}^{l} K_{j} \ln\left(\bar{P}_{ij}\right) \right) \ .
\label{eq:ProbFin}
\end{equation}
If we relax the integrality constraint on $K_1,\dots,K_l$, the problem becomes convex, and can be solved efficiently, even for large numbers of questions and candidate temperatures. Then, finding the actual integer solution is straightforward in this resource-allocation setting, since proximity and rounding results imply that good integer solutions can be recovered near the relaxed optimum \cite{hochbaum1990convex,moriguchi2011mconvex,cont2015optimalrounding}.

{ Because this algorithm and its associated bounds apply to any parameters satisfying the mild conditions of Theorem \ref{zap:the:TempDis}, their applicability extends beyond sampling temperature to other sampling parameters such as $\text{min}_p$, or $\text{top}_h$, and to models that do not exhibit the properties identified in Section \ref{sec:Profiles}, as we discuss further {for $\text{min}_p$ in Appendix \ref{zap:MinP}}.}

\subsection{Temperature Sampling Scheme Comparison}

\label{sec:TempSamp}

\label{sec:Multiple:Comp}

\begin{figure}[t]
   \centering
\includegraphics[width=0.75\linewidth]{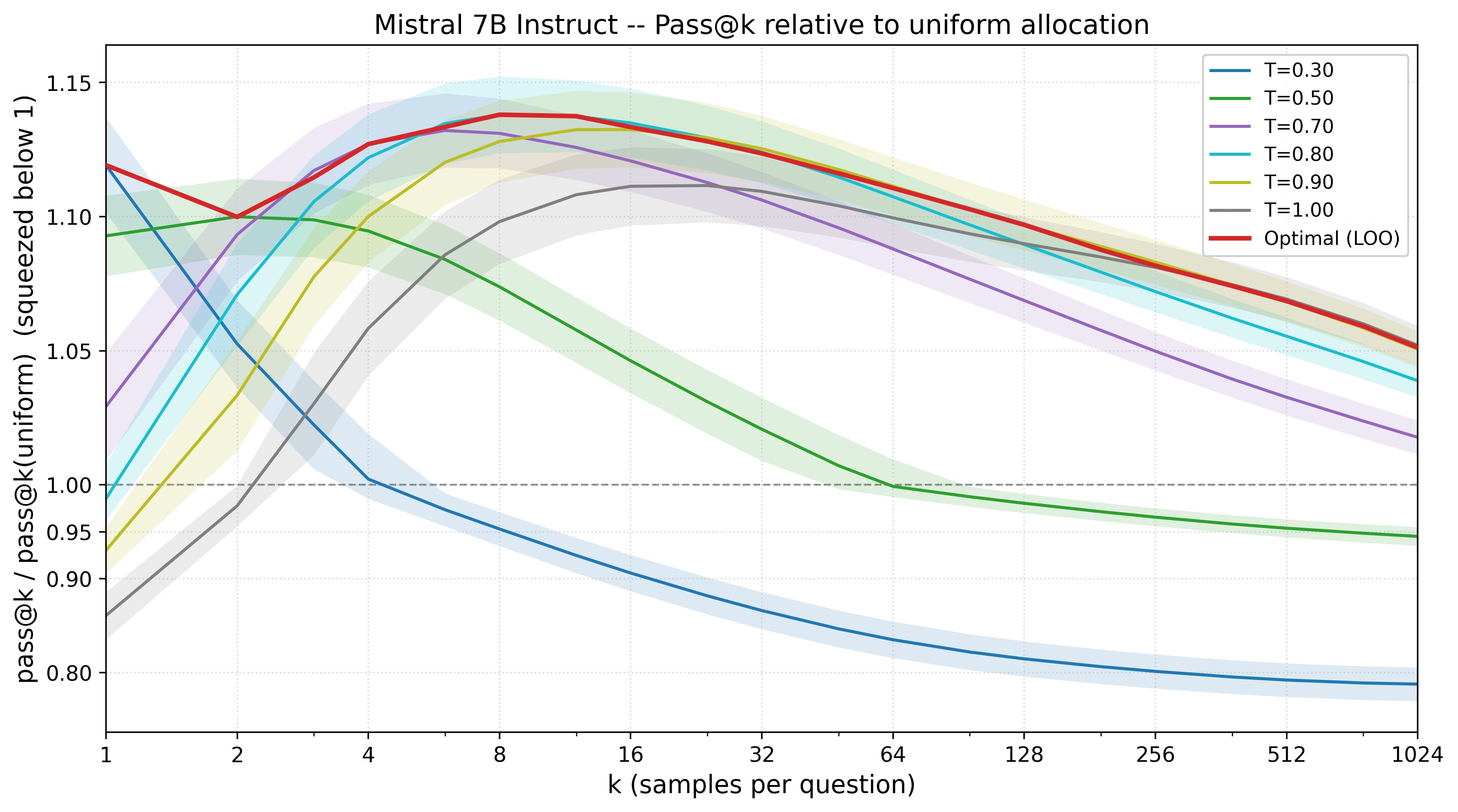}
\caption{Pass@$K$ on log$_{10}$ scale for Mistral 7B on LiveCodeBench under fixed-temperature strategies (blueish) vs.\ the optimal allocation (red), normalized by the uniform-temperature baseline (dotted). While the optimal allocation consistently outperforms fixed strategies for some $K$, for each $K$ at least one achieves comparable performance.
\label{fig:TempsCom}}
\end{figure}

Having derived the optimal temperature allocation scheme, we now compare its average pass rate against \cite{wu2025role}'s uniform sampling scheme, and against single-temperature strategies, in {Figures \ref{fig:TempsCom}, \ref{fig:rtu:grid}, \ref{fig:rtu:math500:grid}, and \ref{fig:rtu:hitom:grid}}. Even with our extensive datasets, the errors {remain} substantial, due to the differences in question difficulty and temperature sensitivity. However, after normalizing the performance {relative} to a baseline scheme, in our case, the uniform sampling scheme, these differences become significantly clearer, and the underlying trends more pronounced. We provide the details of this analysis in Appendix \ref{zap:Multiple:Comp:The}.

As shown in Figure \ref{fig:TempsCom}, {the optimal} allocation scheme has about a $10\%$ advantage over each fixed single-temperature strategy for at least some values of $K$. However, if we allow the temperature to change as a function of $K$, this advantage becomes much more limited. It is important to note that although these advantages are modest, in real-world deployment settings, even a small improvement in the pass rate can translate into substantial practical and economic value.

Another consistent finding is that while the uniform scheme underperforms relative to finely tuned single-temperature strategies, it does not deviate substantially, as it always allocates some samples to the appropriate temperature region. This observation supports the usefulness of the scheme proposed by \cite{wu2025role}, in settings where fine-tuning temperature choices is impractical.

\subsection{The Dulling Effects}

\label{sec:Dulling}

{
While the separation observed in Section \ref{sec:QSep} is relatively small, on the order of $10\%-20\%$, it is still substantially larger than the modest advantage that multi-temperature sampling methods exhibit over finely tuned single-temperature strategies. In this section, we explain this disparity by identifying two effects, which we term \emph{dulling effects}. These two effects are mechanisms that diminish the potential gains from multi-temperature sampling relative to what the apparent separation in the data would suggest.

We establish these effects under the question-profile properties identified in Section \ref{sec:Profiles}, which we have shown to capture a relatively universal phenomenon. However, the conditions underlying these effects can be generalized beyond these specific properties, particularly in the case of the second effect discussed in Section \ref{sec:Dulling:gold}. Furthermore, we show that these effects not only help explain the observed phenomenon, but also have important practical implications.}

\subsubsection{The Short-Sampling Effect}

\label{sec:Dulling:short}

\label{sec:short}

The first effect we identify, which we term the \textit{Short-Sampling Effect}, is best illustrated in a simplified setting of two questions sharing similar Gaussian-like profile shapes, as shown by the red and blue curves in Figure \ref{fig:ManyQ}. We assume these profiles have a clear maximum and rapidly decaying tails, as these assumptions are consistent with the typical behavior observed in Section \ref{sec:TypicalProfile}.

In this setting, it is straightforward to show that the number of distinct optimal temperatures in Equation \ref{eq:ProbFin} cannot exceed the number of questions, as formalized in Lemma \ref{zap:lem:TempSat}, which in this case is two. Consequently, the optimal strategy reduces to either allocating samples across two ``expert'' temperatures, each close to the optimum of one question (represented by $P_E^1, P_E^2$ in the figure), or allocating all samples to a single compromise temperature (represented by $P_C\leq P_E^1,P_E^2$, which we assume is the same for both questions). 

If we also assume for simplicity that the two profiles are identical, meaning $P_E^1 = P_E^2 = P_E$ we can capture the simplest manifestation of the short sampling effect via the following proposition:

\begin{proposition}[The Short-Sampling Effect]
\label{claim:short:ShortingEffect}
\textup
{Given the setup above, where we assume for simplicity that $K$ is even, and define:
\begin{equation}
    \Delta P= \frac{P_E - 2P_C}{2} \ ,
    \label{eq:claim:short:D}
\end{equation}
then, the difference in success probability between using the two expert temperatures, with $\frac{K}{2}$ samples assigned to each, and using a single compromise temperature with all $K$ samples assigned to it, which we denote by $\Delta\mathcal{P}$, satisfies to first order:
\begin{equation}
    \Delta \mathcal{P}
    =
    e^{-\frac{P_EK}{2}}
    \left(
    e^{K\Delta P}
    -
    1
    +
    O\left(KP_E^2\right)
    \right) \ .
    \label{eq:claim:short:DeltaP}
\end{equation}
Consequently, to the first order, the two-expert strategy outperforms the compromise strategy if and only if:
\begin{equation}
    P_E > 2P_C \ .
    \label{eq:claim:short:Res}
\end{equation}
\label{claim:short}
}
\end{proposition}
Although this proposition, which we prove in Appendix \ref{zap:sec:short} considers only a highly simplified setting, we also show that its core components extend to more realistic scenarios. In particular, we analyze settings with multiple questions (Proposition \ref{claim:short:W}), asymmetries (Proposition \ref{claim:short:Non}), and non-vanishing tails (Proposition \ref{claim:short:Gaussian}), and show that the same underlying intuition persists, highlighting the universality of the effect.

From these results, two consistent phenomena emerge. First, the threshold at which the compromise temperature begins to outperform the expert temperatures occurs when $P_C$ is on the order of $\frac{P_E}{2}$, which motivates Equation \ref{eq:efficient range} as a natural characterization of question separation.

Second, we observe that over-utilizing multi-temperature strategies incurs a much larger penalty than under-utilizing them. As seen in Equation \ref{eq:claim:short:DeltaP}, splitting resources across suboptimal temperatures leads to exponential losses, whereas the loss from using a single compromise temperature grows only linearly, since $e^{K\Delta P}-1 \simeq K\Delta P$. This asymmetry, which is analogous to short selling in financial markets both in its consequences and underlying mathematical structure, partially explains the robustness of single-temperature strategies.

\subsubsection{The Goldilocks Sampling Effect}

\label{sec:gold}

\label{sec:Dulling:gold}

Our second dulling effect is the \emph{Goldilocks sampling effect}, which states that for any number of samples $K \in \mathbb{N}$, the questions most sensitive to the choice of sampling strategy are those whose hardness is on the same scale as $K$. This observation motivates our decision to characterize questions by their hardness throughout the paper, as questions of substantially different hardness can be analyzed largely independently.

We formalize this effect by considering a question with single-sample success probability $0 < P < 1$, and analyzing how its overall success probability changes as a function of the number of samples. Relaxing $K \in \mathbb{R}$ for analytical convenience, and using Equation \ref{eq:PowerK}, we obtain:
\begin{equation}
\frac{d\mathcal{P}}{dK}
=
-\bar{P}^{K}\ln\left(\bar{P}\right) \ ,
\label{eq:marginal}
\end{equation}
which captures the \textit{marginal gain} in success probability from adding additional samples. The Goldilocks effect is then described as follows:

\begin{proposition}[The Goldilocks Sampling Effect]
\label{prop:GoldilocksSampling}
\textup
{Given the setup described above, where the number of samples $K$ is fixed, then the marginal gain in success probability is maximized as a function of the question hardness when:
\begin{equation}
    K
    =
    -\frac{1}{\ln\left(\bar{P}\right)}
    =
    H \ ,
    \label{eq:Gold:MaxMarge}
\end{equation}
where $H$ is the question hardness, as defined in Equation \ref{eq:def:Hardness}.
}

\textup
{Furthermore, if the sampling scale deviates from the hardness scale by a factor $\alpha>0$, i.e. $K=\alpha H$, then the normalized marginal gain is:
\begin{equation}
\left(\frac{d\mathcal{P}_{\mathrm{opt}}}{dK}\right)^{-1}
\frac{d\mathcal{P}_{\alpha}}{dK}
=
\alpha e^{1-\alpha} \ .
\label{eq:Gold:Dev}
\end{equation}
}
\end{proposition}

While the proof, provided in Appendix \ref{zap:gold} is a straightforward calculus exercise, it has an elegant interpretation. Questions that are significantly harder than the sampling budget $K$ are almost never solved, regardless of the temperature allocation scheme. Conversely, questions that are much easier are also almost always solved, again regardless of the temperature allocation scheme. As a result, only questions whose hardness is on the same scale as $K$ meaningfully contribute to the differences between sampling strategies. 

This transition is also quite sharp. By analyzing the range of $\alpha$ for which the expression in Equation \ref{eq:Gold:Dev} remains at least $\frac{1}{e}$, we find that $0.16 \lesssim \alpha \lesssim 3.15$ is the range in which the marginal gain remains stable. Taking the ratio between these endpoints yields an efficient range spanning roughly a factor of $20$ in hardness. Given the large variability in question difficulty observed in Section \ref{sec:Profiles}, this defines a relatively well-localized regime of relevant questions. 
{As this effect is independent of the structure of the question profiles, it applies very broadly, extending to virtually any inference-time sampling hyperparameter.}

With this effect we complete our explanation of why well-tuned single-temperature strategies are often sufficient. From Section \ref{sec:QSep}, we know that questions of similar hardness have substantially overlapping temperature profiles, meaning that by the short-sampling effect, multi-temperature sampling provides only limited additional benefit within a fixed hardness regime. Although the average optimal temperature varies with hardness, the Goldilocks sampling effect now shows that we can largely ignore hardness regimes far from the relevant scale of $K$. This explains why varying the temperature as a function of $K$ can capture most of the benefits of the ideal sampling strategy.

\begin{figure}[!t]
{\par
  \centering
  \includegraphics[width=\textwidth]{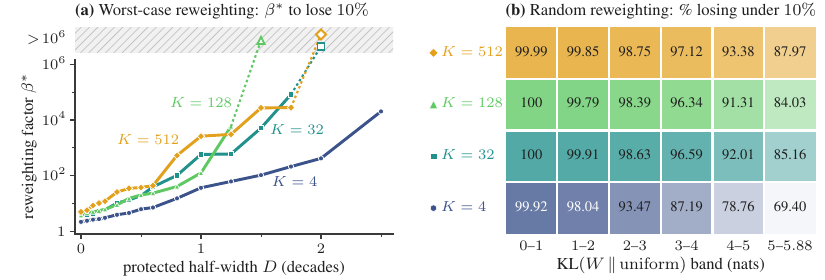}
  \caption{Sensitivity of the test pass@$K$ score to the question hardness distribution used for temperature selection (Mistral 7B Instruct, LiveCodeBench). Left: Goldilocks protection. Questions whose hardness is within a factor $10^D$ of $K$ keep their weight, a $10\%$ loss requires reweighting the others by $\beta^*$, which grows exponentially with $D$. Right: KL divergence. Even far from the test distribution in KL divergence, most random training distributions $W$ cost under $10\%$.}
  \label{fig:Prot:D:Main}
\par}
\end{figure}

{We demonstrate this effect in Figures \ref{fig:Prot:D:Main}, \ref{fig:Prot:D:LCB}, \ref{fig:Prot:D:MATH}, and \ref{fig:Prot:D:HiToM} by comparing the optimal temperature with one whose success probability is $10\%$ lower, and determining the factor $\beta$ by which questions outside the hardness range for some $0<D$:
\begin{equation}
\log K-D\leq\log H\leq\log K+D
\end{equation}
must be reweighted for the suboptimal temperature to outperform the optimal one, as explained in {Appendix \ref{zap:DataDis:Prot:The}}. As shown in Figures \ref{fig:Prot:D:Main}, \ref{fig:Prot:D:LCB}, \ref{fig:Prot:D:MATH}, and \ref{fig:Prot:D:HiToM}, the required reweighting grows approximately exponentially with $D$, demonstrating that the relative performance of different temperatures is determined primarily by questions whose hardness is close to $K$. This effect persists even for very large $K$, as shown in Appendix \ref{zap:DataDis:Prot:res}, extending its relevance to the regimes considered in \cite{li2022competition,brown2024large}.}

This effect has an important practical implication. Since performance at budget $K$ is primarily driven by questions whose hardness is on the scale of $K$, temperature-selection strategies should be relatively insensitive to mismatches between the hardness distributions of the training and deployment data. This suggests that even substantially skewed training distributions can yield temperature choices that transfer effectively, reducing the need for exact alignment between training and deployment data.

We test this prediction in Appendix \ref{zap:DataDis:RandomH}, where we randomly reweight questions according to hardness, optimize temperatures under the reweighted distribution, and then evaluate the resulting strategy on the original distribution. There we see that strategy is indeed robust to substantial distortion. Even when the KL divergence exceeds $3$, on average roughly $90\%$ of reweightings lead to less than a $10\%$ performance loss. For KL divergences between $1$ and $2$, this fraction rises to $99\%$.

Furthermore, this effect suggests that selecting  temperatures for a given question does not require knowing the question's hardness, but only the available computational budget. Although this observation is less relevant for pre-committed strategies, we believe that our results could be combined with adaptive strategies, as discussed in the \nameref{sec:Conclusion}, where this property may prove valuable.

\section{Conclusion}

\label{sec:Conclusion}

In this work, we studied pre-committed multi-temperature sampling strategies for LLM inference in the canonical pass@$K$ setting. We began by analyzing how per-question success probability varies with sampling temperature, which we refer to as a question's temperature profile. Across models and datasets, these profiles exhibit a remarkably simple structure, typically with a single local maximum occurring either at $\tau = 0$ or at a positive temperature, followed by decay.

We then analyzed these profiles through the lens of hardness, defined as the effective number of samples required for a reliable solution. This perspective revealed several consistent trends. In particular, we saw that the average optimal temperature varies significantly with hardness, while efficient temperature ranges shrink as questions become harder, although only logarithmically. As a result, we found that substantial overlap between efficient temperature ranges remains even for difficult questions.

Building on these observations, we derived an efficient approximation to the theoretically optimal multi-temperature allocation scheme. We showed that although this strategy outperforms single-temperature approaches, its gains are typically small once the temperature is tuned as a function of the sampling budget $K$. Thus, much of the benefit of temperature diversity can be recovered by selecting a single well-tuned temperature.

We explained this behavior through two dulling effects. The first shows that overusing multiple temperatures typically incurs a larger penalty than underusing them. The second shows that performance at budget $K$ is primarily governed by questions whose hardness is on the same scale as $K$. Together, these effects explain why single-temperature strategies remain strong in practice. They also suggest an important practical implication: multi-temperature strategies can be selected effectively without accurately knowing the hardness of the question, or even the full hardness distribution of the deployment data. 
This reduces the resources required to construct the training dataset, or to analyze specific questions during inference.

Overall, our results show that multi-temperature sampling is most useful when questions have well-separated efficient temperature ranges, but that such separation is limited in typical LLM settings. We discuss the limitations of our work, along with potential generalizations and broader impacts, in Appendix \ref{zapp:limitations}.

Looking forward, extending this analysis beyond the pass@$K$ setting, and combining it with adaptive or higher-dimensional inference strategies are important directions for future work. In such settings, greater separation may emerge as dimensionality increases, potentially enhancing the practical value of our approximation to the optimal sampling scheme.

\begin{ack}
This work is supported in part by the Israel Science Foundation Excellence Center, the US-Israel Binational Science Foundation, and the Israel Ministry of Science. The authors declare no competing interests.
\end{ack}

\bibliography{citations.bib}

\begin{thebibliography}{10}

\bibitem{vaswani2017attention}
Ashish Vaswani, Noam Shazeer, Niki Parmar, Jakob Uszkoreit, Llion Jones,
  Aidan~N. Gomez, Lukasz Kaiser, and Illia Polosukhin.
\newblock Attention is all you need.
\newblock In {\em Advances in Neural Information Processing Systems},
  volume~30, 2017.

\bibitem{bommasani2021opportunities}
Rishi Bommasani, Drew~A. Hudson, Ehsan Adeli, Russ Altman, Simran Arora, Sydney
  von Arx, Michael~S. Bernstein, Jeannette Bohg, Antoine Bosselut, Emma
  Brunskill, et~al.
\newblock On the opportunities and risks of foundation models.
\newblock {\em arXiv preprint arXiv:2108.07258}, 2021.

\bibitem{brown2020language}
Tom~B. Brown, Benjamin Mann, Nick Ryder, Melanie Subbiah, Jared Kaplan,
  Prafulla Dhariwal, Arvind Neelakantan, Pranav Shyam, Girish Sastry, Amanda
  Askell, et~al.
\newblock Language models are few-shot learners.
\newblock In {\em Advances in Neural Information Processing Systems},
  volume~33, pages 1877--1901, 2020.

\bibitem{chowdhery2023palm}
Aakanksha Chowdhery, Sharan Narang, Jacob Devlin, Maarten Bosma, Gaurav Mishra,
  Adam Roberts, Paul Barham, Hyung~Won Chung, Charles Sutton, Sebastian
  Gehrmann, et~al.
\newblock {{PaLM}}: Scaling language modeling with pathways.
\newblock {\em Journal of Machine Learning Research}, 24(240):1--113, 2023.

\bibitem{touvron2023llama}
Hugo Touvron, Thibaut Lavril, Gautier Izacard, Xavier Martinet, Marie-Anne
  Lachaux, Timoth{\'e}e Lacroix, Baptiste Rozi{\`e}re, Naman Goyal, Eric
  Hambro, Faisal Azhar, et~al.
\newblock {{LLaMA}}: Open and efficient foundation language models.
\newblock {\em arXiv preprint arXiv:2302.13971}, 2023.

\bibitem{achiam2023gpt4}
Josh Achiam, Steven Adler, Sandhini Agarwal, Lama Ahmad, Ilge Akkaya,
  Florencia~Leoni Aleman, Diogo Almeida, Janko Altenschmidt, Sam Altman,
  Shyamal Anadkat, et~al.
\newblock {{GPT-4}} technical report.
\newblock {\em arXiv preprint arXiv:2303.08774}, 2023.

\bibitem{chen2021evaluating}
Mark Chen, Jerry Tworek, Heewoo Jun, Qiming Yuan, Henrique Ponde de~Oliveira
  Pinto, Jared Kaplan, Harri Edwards, Yuri Burda, Nicholas Joseph, Greg
  Brockman, et~al.
\newblock Evaluating large language models trained on code.
\newblock {\em arXiv preprint arXiv:2107.03374}, 2021.

\bibitem{austin2021program}
Jacob Austin, Augustus Odena, Maxwell Nye, Maarten Bosma, Henryk Michalewski,
  David Dohan, Ellen Jiang, Carrie Cai, Michael Terry, Quoc Le, and Charles
  Sutton.
\newblock Program synthesis with large language models.
\newblock {\em arXiv preprint arXiv:2108.07732}, 2021.

\bibitem{li2022competition}
Yujia Li, David Choi, Junyoung Chung, Nate Kushman, Julian Schrittwieser,
  R{\'e}mi Leblond, Tom Eccles, James Keeling, Felix Gimeno, Agustin Dal~Lago,
  et~al.
\newblock Competition-level code generation with {{AlphaCode}}.
\newblock {\em Science}, 378(6624):1092--1097, 2022.

\bibitem{cobbe2021training}
Karl Cobbe, Vineet Kosaraju, Mohammad Bavarian, Mark Chen, Heewoo Jun, Lukasz
  Kaiser, Matthias Plappert, Jerry Tworek, Jacob Hilton, Reiichiro Nakano,
  Christopher Hesse, and John Schulman.
\newblock Training verifiers to solve math word problems.
\newblock {\em arXiv preprint arXiv:2110.14168}, 2021.

\bibitem{hendrycks2021measuring}
Dan Hendrycks, Collin Burns, Saurav Kadavath, Akul Arora, Steven Basart, Eric
  Tang, Dawn Song, and Jacob Steinhardt.
\newblock Measuring mathematical problem solving with the {{MATH}}
  dataset.
\newblock In {\em Thirty-fifth Conference on Neural Information Processing
  Systems Datasets and Benchmarks Track}, 2021.

\bibitem{lewkowycz2022solving}
Aitor Lewkowycz, Anders Andreassen, David Dohan, Ethan Dyer, Henryk
  Michalewski, Vinay Ramasesh, Ambrose Slone, Cem Anil, Imanol Schlag, Theo
  Gutman-Solo, et~al.
\newblock Solving quantitative reasoning problems with language models.
\newblock In {\em Advances in Neural Information Processing
  Systems}, volume 35, pages 3843--3857, 2022.

\bibitem{radford2019language}
Alec Radford, Jeffrey Wu, Rewon Child, David Luan, Dario Amodei, and Ilya
  Sutskever.
\newblock Language models are unsupervised multitask learners.
\newblock {\em OpenAI blog}, 1(8), 2019.

\bibitem{kaplan2020scaling}
Jared Kaplan, Sam McCandlish, Tom Henighan, Tom~B. Brown, Benjamin Chess, Rewon
  Child, Scott Gray, Alec Radford, Jeffrey Wu, and Dario Amodei.
\newblock Scaling laws for neural language models.
\newblock {\em arXiv preprint arXiv:2001.08361}, 2020.

\bibitem{hoffmann2022training}
Jordan Hoffmann, Sebastian Borgeaud, Arthur Mensch, Elena Buchatskaya, Trevor
  Cai, Eliza Rutherford, Diego de~Las~Casas, Lisa~Anne Hendricks, Johannes
  Welbl, Aidan Clark, et~al.
\newblock {{An empirical analysis of compute-optimal large language
  model training}}.
\newblock In {\em Advances in Neural Information Processing
  Systems}, volume 35, pages 30016--30030, 2022.

\bibitem{christiano2017deep}
Paul~F. Christiano, Jan Leike, Tom~B. Brown, Miljan Martic, Shane Legg, and
  Dario Amodei.
\newblock Deep reinforcement learning from human preferences.
\newblock In {\em Advances in Neural Information Processing Systems},
  volume~30, 2017.

\bibitem{ouyang2022training}
Long Ouyang, Jeffrey Wu, Xu~Jiang, Diogo Almeida, Carroll~L. Wainwright, Pamela
  Mishkin, Chong Zhang, Sandhini Agarwal, Katarina Slama, Alex Ray, John
  Schulman, Jacob Hilton, Fraser Kelton, Luke Miller, Maddie Simens, Amanda
  Askell, Peter Welinder, Paul Christiano, Jan Leike, and Ryan Lowe.
\newblock Training language models to follow instructions with human feedback.
\newblock In {\em Advances in Neural Information Processing Systems},
  volume~35, pages 27730--27744, 2022.

\bibitem{bai2022training}
Yuntao Bai, Andy Jones, Kamal Ndousse, Amanda Askell, Anna Chen, Nova DasSarma,
  Dawn Drain, Stanislav Fort, Deep Ganguli, Tom Henighan, et~al.
\newblock Training a helpful and harmless assistant with reinforcement learning
  from human feedback.
\newblock {\em arXiv preprint arXiv:2204.05862}, 2022.

\bibitem{wei2022emergent}
Jason Wei, Yi~Tay, Rishi Bommasani, Colin Raffel, Barret Zoph, Sebastian
  Borgeaud, Dani Yogatama, Maarten Bosma, Denny Zhou, Donald Metzler, et~al.
\newblock Emergent abilities of large language models.
\newblock {\em Transactions on Machine Learning Research}, 2022.

\bibitem{wei2022chainofthought}
Jason Wei, Xuezhi Wang, Dale Schuurmans, Maarten Bosma, Brian Ichter, Fei Xia,
  Ed~H. Chi, Quoc~V. Le, and Denny Zhou.
\newblock Chain-of-thought prompting elicits reasoning in large language
  models.
\newblock In {\em Advances in Neural Information Processing Systems},
  volume~35, pages 24824--24837, 2022.

\bibitem{kojima2022large}
Takeshi Kojima, Shixiang~Shane Gu, Machel Reid, Yutaka Matsuo, and Yusuke
  Iwasawa.
\newblock Large language models are zero-shot reasoners.
\newblock In {\em Advances in Neural Information Processing Systems},
  volume~35, pages 22199--22213, 2022.

\bibitem{zhou2023least}
Denny Zhou, Nathanael Sch{\"a}rli, Le~Hou, Jason Wei, Nathan Scales, Xuezhi
  Wang, Dale Schuurmans, Claire Cui, Olivier Bousquet, Quoc Le, and Ed~H. Chi.
\newblock Least-to-most prompting enables complex reasoning in large language
  models.
\newblock In {\em International Conference on Learning Representations}, 2023.

\bibitem{yao2023tree}
Shunyu Yao, Dian Yu, Jeffrey Zhao, Izhak Shafran, Thomas~L. Griffiths, Yuan
  Cao, and Karthik Narasimhan.
\newblock Tree of thoughts: Deliberate problem solving with large language
  models.
\newblock In {\em Advances in Neural Information Processing Systems},
  volume~36, 2023.

\bibitem{schick2023toolformer}
Timo Schick, Jane Dwivedi-Yu, Roberto Dess{\`i}, Roberta Raileanu, Maria
  Lomeli, {Eric Hambro}, Luke Zettlemoyer, Nicola Cancedda, and
  Thomas Scialom.
\newblock Toolformer: Language models can teach themselves to use tools.
\newblock In {\em Advances in Neural Information Processing
  Systems}, volume 36, pages 68539--68551, 2023.

\bibitem{snell2024scaling}
Charlie Snell, Jaehoon Lee, Kelvin Xu, and Aviral Kumar.
\newblock Scaling {{LLM}} test-time compute optimally can be more
  effective than scaling parameters for reasoning.
\newblock In {\em International Conference on Learning Representations}, 2025.

\bibitem{brown2024large}
Bradley Brown, Jordan Juravsky, Ryan Ehrlich, Ronald Clark, Quoc~V. Le,
  Christopher R{\'e}, and Azalia Mirhoseini.
\newblock Large language monkeys: Scaling inference compute with repeated
  sampling.
\newblock {\em arXiv preprint arXiv:2407.21787}, 2024.

\bibitem{wu2025role}
Yuheng Wu, Azalia Mirhoseini, and Thierry Tambe.
\newblock On the role of temperature sampling in test-time scaling.
\newblock {\em arXiv preprint arXiv:2510.02611}, 2025.

\bibitem{besta2024graph}
Maciej Besta, Nils Blach, Ales Kubicek, Robert Gerstenberger, Michal
  Podstawski, Lukas Gianinazzi, Joanna Gajda, Tomasz Lehmann, Hubert
  Niewiadomski, Piotr Nyczyk, and Torsten Hoefler.
\newblock Graph of thoughts: Solving elaborate problems with large language
  models.
\newblock {\em Proceedings of the AAAI Conference on Artificial Intelligence},
  38(16):17682--17690, 2024.

\bibitem{wang2023selfconsistency}
Xuezhi Wang, Jason Wei, Dale Schuurmans, Quoc Le, Ed~H. Chi, Sharan Narang,
  Aakanksha Chowdhery, and Denny Zhou.
\newblock Self-consistency improves chain of thought reasoning in language
  models.
\newblock In {\em International Conference on Learning Representations}, 2023.

\bibitem{uesato2022solving}
Jonathan Uesato, Nate Kushman, Ramana Kumar, Francis Song, Noah Siegel, Lisa
  Wang, Antonia Creswell, Geoffrey Irving, and Irina Higgins.
\newblock Solving math word problems with process- and outcome-based feedback.
\newblock {\em arXiv preprint arXiv:2211.14275}, 2022.

\bibitem{lightman2023lets}
Hunter Lightman, Vineet Kosaraju, {Yuri}~Burda,
  {Harrison}~Edwards, Bowen Baker, Teddy Lee, Jan Leike, John
  Schulman, Ilya Sutskever, and Karl Cobbe.
\newblock Let's verify step by step.
\newblock In {\em International Conference on Learning
  Representations}, pages 39578--39601, 2024.

\bibitem{ackley1985learning}
David~H. Ackley, Geoffrey~E. Hinton, and Terrence~J. Sejnowski.
\newblock A learning algorithm for {{Boltzmann}} machines.
\newblock {\em Cognitive Science}, 9(1):147--169, 1985.

\bibitem{hinton2015distilling}
Geoffrey Hinton, Oriol Vinyals, and Jeff Dean.
\newblock Distilling the knowledge in a neural network.
\newblock {\em arXiv preprint arXiv:1503.02531}, 2015.

\bibitem{holtzman2020curious}
Ari Holtzman, Jan Buys, Li~Du, Maxwell Forbes, and Yejin Choi.
\newblock The curious case of neural text degeneration.
\newblock In {\em International Conference on Learning Representations}, 2020.

\bibitem{hewitt2022truncation}
John Hewitt, Christopher~D. Manning, and Percy Liang.
\newblock Truncation sampling as language model desmoothing.
\newblock In {\em Findings of the Association for Computational Linguistics:
  EMNLP 2022}, pages 3414--3427, 2022.

\bibitem{meister2023locally}
Clara Meister, Tiago Pimentel, Gian Wiher, and Ryan Cotterell.
\newblock Locally typical sampling.
\newblock {\em Transactions of the Association for Computational Linguistics},
  11:102--121, 2023.

\bibitem{renze2024effect}
Matthew Renze and Erhan Guven.
\newblock The effect of sampling temperature on problem solving in large
  language models.
\newblock In {\em Findings of the Association for Computational Linguistics:
  EMNLP 2024}, pages 7346--7356, 2024.

\bibitem{fan2018hierarchical}
Angela Fan, Mike Lewis, and Yann Dauphin.
\newblock Hierarchical neural story generation.
\newblock In {\em Proceedings of the 56th Annual Meeting of the Association for
  Computational Linguistics}, pages 889--898, 2018.

\bibitem{du2025optimizingtemperature}
Weihua Du, Yiming Yang, and Sean Welleck.
\newblock Optimizing temperature for language models with multi-sample
  inference.
\newblock In {Aarti Singh}, {Maryam Fazel},
  {Daniel Hsu}, {Simon Lacoste-Julien}, {Felix
  Berkenkamp}, {Tegan Maharaj}, {Kiri Wagstaff}, and
  {Jerry Zhu}, editors, {\em Proceedings of the 42nd
  International Conference on Machine Learning}, volume 267 of {\em
  Proceedings of Machine Learning Research}, pages
  14648--14668. PMLR, 13--19 Jul 2025.

\bibitem{cecere2025monte}
Nicola Cecere, Andrea Bacciu, Ignacio Fern{\'a}ndez-Tob{\'i}as, and Amin
  Mantrach.
\newblock {M}onte {C}arlo temperature: A robust sampling strategy for {LLM}'s
  uncertainty quantification methods.
\newblock In {\em Proceedings of the 5th Workshop on Trustworthy NLP}, pages
  305--320, Albuquerque, New Mexico, 2025. Association for Computational
  Linguistics.

\bibitem{levi2025simple}
Noam~Itzhak Levi.
\newblock A simple model of inference scaling laws.
\newblock In Aarti Singh, Maryam Fazel, Daniel Hsu, Simon Lacoste-Julien, Felix
  Berkenkamp, Tegan Maharaj, Kiri Wagstaff, and Jerry Zhu, editors, {\em
  Proceedings of the 42nd International Conference on Machine Learning}, volume
  267 of {\em Proceedings of Machine Learning Research}, pages 33984--33998.
  PMLR, 13--19 Jul 2025.

\bibitem{wang2020contextual}
Pei-Hsin Wang, Sheng-Iou Hsieh, Shih-Chieh Chang, Yu-Ting Chen, Jia-Yu Pan, Wei
  Wei, and Da-Chang Juan.
\newblock Contextual temperature for language modeling.
\newblock {\em arXiv preprint arXiv:2012.13575}, 2020.

\bibitem{cai2024t2}
Chengkun Cai, Xu~Zhao, Yucheng Du, Haoliang Liu, and Lei Li.
\newblock {{{$T^2$}}} of thoughts: Temperature tree elicits
  reasoning in large language models.
\newblock {\em arXiv preprint arXiv:2405.14075}, 2024.

\bibitem{zhu2024hot}
Yuqi Zhu, Jia Li, Ge~Li, YunFei Zhao, Jia Li, Zhi Jin, and Hong Mei.
\newblock Hot or cold? {{Adaptive}} temperature sampling for code
  generation with large language models.
\newblock In {\em Proceedings of the AAAI Conference on Artificial
  Intelligence}, volume~38, pages 437--445, 2024.

\bibitem{zhang2024edt}
Shimao Zhang, Yu~Bao, and Shujian Huang.
\newblock {EDT}: Improving large language models' generation by entropy-based
  dynamic temperature sampling.
\newblock {\em arXiv preprint arXiv:2403.14541}, 2024.

\bibitem{xie2024calibrating}
Johnathan Xie, Annie~S. Chen, Yoonho Lee, Eric Mitchell, and Chelsea Finn.
\newblock Calibrating language models with adaptive temperature scaling.
\newblock In {\em Proceedings of the 2024 Conference on Empirical Methods in
  Natural Language Processing}, pages 18128--18138, Miami, Florida, USA, 2024.
  Association for Computational Linguistics.

\bibitem{kulal2019spoc}
Sumith Kulal, Panupong Pasupat, Kartik Chandra, Mina Lee, Oded Padon, Alex
  Aiken, and Percy Liang.
\newblock {{SPoC}}: Search-based pseudocode to code.
\newblock In {\em Advances in Neural Information Processing Systems},
  volume~32, 2019.

\bibitem{kazdan2025efficient}
Joshua Kazdan, Rylan Schaeffer, Youssef Allouah, Colin Sullivan, Kyssen Yu,
  Noam Levi, and Sanmi Koyejo.
\newblock Efficient prediction of pass@k scaling in large language models.
\newblock {\em arXiv preprint arXiv:2510.05197}, 2025.

\bibitem{jain2024livecodebench}
Naman Jain, King Han, Alex Gu, Wen-Ding Li, Fanjia Yan, Tianjun Zhang, Sida
  Wang, Armando Solar-Lezama, Koushik Sen, and Ion Stoica.
\newblock {{LiveCodeBench}}: Holistic and contamination free
  evaluation of large language models for code.
\newblock In {\em International Conference on Learning
  Representations}, pages 58791--58831, 2025.

\bibitem{he2023hitom}
Yinghui He, Yufan Wu, Yilin Jia, Rada Mihalcea, Yulong Chen, and Naihao Deng.
\newblock Hi-{T}o{M}: A benchmark for evaluating higher-order theory of mind
  reasoning in large language models.
\newblock In {\em Findings of the {A}ssociation for {C}omputational
  {L}inguistics: {EMNLP} 2023}, pages 10691--10706, Singapore, 2023.
  Association for Computational Linguistics.

\bibitem{nguyen2025minp}
Minh Nguyen, Andrew Baker, Clement Neo, Allen Roush, Andreas Kirsch, and Ravid
  Shwartz-Ziv.
\newblock Turning up the heat: Min-p sampling for creative and coherent {LLM}
  outputs.
\newblock In {\em International Conference on Learning Representations}, pages
  70333--70366, 2025.

\bibitem{potraghloo2025toph}
Erfan Baghaei~Potraghloo, Seyedarmin Azizi, Souvik Kundu, and Massoud Pedram.
\newblock Top-{H} decoding: Adapting the creativity and coherence with bounded
  entropy in text generation.
\newblock In {\em Advances in Neural Information Processing Systems},
  volume~38, pages 28482--28513, 2025.

\bibitem{hochbaum1990convex}
Dorit~S. Hochbaum and J.~George Shanthikumar.
\newblock Convex separable optimization is not much harder than linear
  optimization.
\newblock {\em Journal of the ACM}, 37(4):843--862, 1990.

\bibitem{moriguchi2011mconvex}
Satoko Moriguchi, Akiyoshi Shioura, and Nobuyuki Tsuchimura.
\newblock M-convex function minimization by continuous relaxation approach:
  Proximity theorem and algorithm.
\newblock {\em SIAM Journal on Optimization}, 21(3):633--668, 2011.

\bibitem{cont2015optimalrounding}
Rama Cont and Massoud Heidari.
\newblock Optimal rounding under integer constraints.
\newblock {\em arXiv preprint arXiv:1501.00014}, 2015.

\bibitem{rockafellar1970convex}
R.~Tyrrell Rockafellar.
\newblock {\em Convex Analysis}.
\newblock Princeton University Press, Princeton, NJ, 1970.

\bibitem{caratheodory1907}
Constantin Carath{\'e}odory.
\newblock {\"U}ber den {Variabilit{\"a}tsbereich} der {Koeffizienten} von
  {Potenzreihen}, die gegebene {Werte} nicht annehmen.
\newblock {\em Mathematische Annalen}, 64:95--115, 1907.

\bibitem{barvinok2002course}
Alexander Barvinok.
\newblock {\em A Course in Convexity}, volume~54 of {\em Graduate Studies in
  Mathematics}.
\newblock American Mathematical Society, Providence, RI, 2002.

\bibitem{jiang2023mistral}
{Albert Q. Jiang}, {Alexandre Sablayrolles},
  {Arthur Mensch}, {Chris Bamford}, {Devendra
  Singh Chaplot}, {Diego de las Casas}, et~al.
\newblock {{Mistral 7B}}.
\newblock {\em arXiv preprint arXiv:2310.06825}, 2023.

\bibitem{yang2025qwen3}
{An Yang}, {Anfeng Li}, {Baosong Yang},
  {Beichen Zhang}, {Binyuan Hui}, {Bo Zheng},
  et~al.
\newblock {{Qwen3 technical report}}.
\newblock {\em arXiv preprint arXiv:2505.09388}, 2025.

\bibitem{openai2025gptoss}
{OpenAI}.
\newblock {{gpt-oss-120b \& gpt-oss-20b model card}}.
\newblock {\em arXiv preprint arXiv:2508.10925}, 2025.

\bibitem{kwon2023vllm}
{Woosuk Kwon}, {Zhuohan Li}, {Siyuan Zhuang},
  {Ying Sheng}, {Lianmin Zheng}, {Cody Hao
  Yu}, {Joseph E. Gonzalez}, {Hao Zhang}, and
  {Ion Stoica}.
\newblock {{Efficient memory management for large language model
  serving with PagedAttention}}.
\newblock In {\em Proceedings of the 29th Symposium on Operating
  Systems Principles}, pages 611--626, 2023.

\end{thebibliography}

\newpage

\appendix
\renewcommand{\floatpagefraction}{0.8}
\renewcommand{\topfraction}{0.9}
\renewcommand{\bottomfraction}{0.8}
\renewcommand{\textfraction}{0.07}
\setcounter{topnumber}{3}
\setcounter{totalnumber}{5}

\section{Assumptions and Limitations}

\label{zapp:limitations}

We begin the appendix by discussing the main assumptions and limitations of our work, along with their potential implications and directions for future investigation.

The first limitation concerns empirical coverage. While our numerical analysis is extensive and spans multiple models and datasets, the space of LLM architectures, training procedures, and real-world tasks is extremely large. Although the consistency of our results suggests that the observed phenomena are likely to be general, we cannot fully establish their universality without significantly broader evaluation. In particular, extending this analysis to additional datasets, more recent large-scale models, and qualitatively different generative paradigms (such as diffusion-based or hybrid models) is an important direction for future work. Further empirical studies, especially on state-of-the-art systems, would help determine how broadly these effects apply.

A second limitation lies in the relationship between our theoretical framework and real-world behavior. While all of our theoretical results are formally derived and rely on explicitly stated assumptions, these assumptions are not derived rigorously from first principles. For example, properties such as unimodality of temperature profiles and the structure of success probabilities are consistently observed across models and datasets, and we provide theoretical justification for why such behavior is expected. However, they are not proven to hold universally. As a result, although our theory provides strong intuition and closely matches observed behavior, formally characterizing the extent to which it applies to all real-world LLMs remains an open problem. Bridging this remaining gap between empirically grounded theory and full theoretical guarantees is an important direction for future work.

In the case of the short-sampling dulling effect in Section \ref{sec:short} and Appendix \ref{zap:sec:short}, there is an additional limitation. While we were able to prove this effect under a variety of assumptions, and demonstrate it across diverse representative cases, we did not obtain a single unified theorem that fully characterizes the phenomenon in all settings. Instead, our analysis consists of several complementary results, each capturing a different aspect of the effect. Although these results provide strong evidence and a consistent explanation of the behavior, a more complete and general characterization remains open. 
Nevertheless, the assumptions are not arbitrary, but are supported by consistent empirical evidence and underlying structural arguments about LLM behavior. As such, we believe they capture the dominant factors governing the observed phenomena, and that the resulting conclusions remain informative even beyond the exact regimes in which they are formally justified.

More broadly, while our empirical results strongly support the theoretical analysis, additional computational validation would further strengthen the conclusions. In particular, exploring more extreme regimes, larger sample budgets, and more diverse distributions of questions could help refine the observed trends and identify potential edge cases. 
Nevertheless, the consistency of these results across multiple formulations and settings suggests that the effect captures a fundamental mechanism rather than an artifact of specific assumptions. In particular, the qualitative behavior it predicts aligns closely with both empirical observations and the structure of the optimization problem, giving us confidence that the conclusions remain valid in practice.

Overall, we believe that our work provides strong evidence for the existence and practical relevance of the phenomena we describe. However, fully understanding their generality, as well as their precise theoretical foundations, will require further empirical and theoretical investigation.

\section{Theory}

\subsection{Ideal Temperatures Sampling Approximation Derivation}

\label{zap:sec:Ideal}

In this section, we complete the analysis of Section \ref{sec:TheIdeal}, where we introduced an approach for approximating the ideal temperature sampling distribution that maximizes the total success probability. Here, we show how the approximation error scales with properties of the dataset, and under temperature discretization. We demonstrate that, in our setting, this approach provides a reliable approximation.

The optimal sampling scheme is defined for any $K \in \mathbb{N}$, by the optimal (not necessarily distinct) temperatures $0 \leq \tau_1, \ldots, \tau_K$ that maximize the success probability, or equivalently, minimize its complement, as defined in Equation \ref{eq:MultiTempProb:Basic}:
\begin{equation}
\bar{\mathcal{P}}\left(\tau_{1},\ldots,\tau_{K}\right)=\mathbb{E}_{q\sim\mathcal{Q}}\prod_{k=1}^{K}\bar{P}_{q}\left(\tau_{k}\right)
=\mathbb{E}_{q\sim\mathcal{Q}}\exp\left(\sum_{k=1}^{K}\ln\left(\bar{P}_{q}\left(\tau_{k}\right)\right)\right)\ .
\end{equation}
As discussed in the main text, the primary difficulty with this formulation, is that it involves optimization over a continuous and unbounded space of temperatures. However, under relatively mild assumptions, this problem can be discretized significantly, as formalized in the following theorem:

\begin{theorem}[Temperatures Discretization]
\ 
\label{zap:the:TempDis}
\textup
{Assume the setup presented in Section \ref{sec:Multiple:Ideal}, a probability $0<\rho<1$, a truncation temperature $T>0$, and a discretization scale $\Delta>0$, this defines a grid:
\begin{equation}
\mathcal G
=
\{0,\Delta,2\Delta,\ldots,M\Delta\} \ ,
\label{zap:eq:the:TempDis:Dis}
\end{equation}
where $M=\left\lceil \frac{T}{\Delta}\right\rceil$.}

\textup
{Suppose that within this probability $1-\rho$, over the choice of questions $q\sim\mathcal Q$, we know that the failure probability satisfies, for every $0\leq\tau,\tau'\leq T$ such that $|\tau-\tau'|\leq \Delta$, that it is proportionally Lipschitz bounded:
\begin{equation}
\bar P_q(\tau')
\leq
\exp\left(L|\tau-\tau'|\right)\bar P_q(\tau) \ ,
\label{zap:eq:the:TempDis:lip}
\end{equation}
and that beyond the truncation temperature i.e. $T<\tau$, for some $\alpha>0$ the endpoint of the truncated grid satisfies:
\begin{equation}
\bar P_q(M\Delta)
\leq
\exp(\alpha)\bar P_q(\tau) \ .
\label{zap:eq:the:TempDis:truncation}
\end{equation}
Denote the optimal failure probabilities over the continuous temperature space and over the discrete grid by:
\begin{equation}
\bar{\mathcal{P}}^{\star}=\inf_{\tau_{1},\ldots,\tau_{K}\geq0}\bar{\mathcal{P}}\left(\tau_{1},\ldots,\tau_{K}\right),\quad
\bar{\mathcal{P}}_{\mathcal{G}}^{\star}=\min_{\gamma_{1},\ldots,\gamma_{K}\in\mathcal{G}}\bar{\mathcal{P}}\left(\gamma_{1},\ldots,\gamma_{K}\right) \ .
\label{eq:ideal}
\end{equation}
Then we find that for every $K\in\mathbb N$, the error of the best assignment over the discrete temperatures is bounded by:
\begin{equation}
0\leq\bar{\mathcal{P}}_{\mathcal{G}}^{\star}-\bar{\mathcal{P}}^{\star}\leq\left(e^{K(L\Delta+\alpha)}-1\right)\bar{\mathcal{P}}^{\star}+\rho \ .
\label{zap:eq:MrMahina}
\end{equation}
}
\end{theorem}

The proof idea, presented in full in Appendix \ref{zap:sec:Ideal:exp}, is to show that any continuous temperature selection can be replaced by nearby grid temperatures while incurring only a controlled relative increase in failure probability. This relative error compounds across the $K$ temperatures, yielding a factor of $e^{K(L\Delta+\alpha)}$. Questions for which this control fails contribute at most $\rho$, since failure probabilities are bounded by $1$. Consequently, the optimal grid-based solution is close to the optimal continuous solution, up to the multiplicative discretization error and the mass of such “bad” questions.

The strength of this theorem lies in the fact that our final bound in Equation \ref{zap:eq:MrMahina} is proportional to the failure probability. Thus, although the multiplicative term grows with $K$, the overall deviation decreases as long as $L$, $\Delta$, and $\alpha$ are sufficiently small. 

In our setting, as shown in Section \ref{sec:UniMod}, the profiles tend to decay rapidly for relatively small $\tau$, so Equation \ref{zap:eq:the:TempDis:truncation} is trivially satisfied for small $\alpha$. Equation \ref{zap:eq:the:TempDis:lip} is well satisfied for small $L$, particularly for harder questions, since we consider the proportional Lipschitz continuity of $1 - P$. This means that the approximation in Equation \ref{zap:eq:MrMahina} is robust even for large $K$ and very small probabilities.

Having established that the problem can be effectively reduced to a finite set of temperatures, we can now investigate the effect of operating with a limited number of questions.

\begin{theorem}
[Finite Question Distribution Reduction]
\label{zap:lem:FiniteQ}
\normalfont
Assume the setup of Theorem \ref{zap:the:TempDis}, and let $n \in \mathbb{N}$ questions $\mathcal{Q}'=\left\{ q_{1},\ldots,q_{n}\right\}$ be drawn independently from $\mathcal Q$. For any choice of grid temperatures $\gamma_1,\ldots,\gamma_K \in \mathcal G$, define the empirical failure probability, as in Equation \ref{eq:ProbFailTot}, by:
\begin{equation}
\bar{\mathcal{P}}_{\mathcal{Q}'}\left(\gamma_{1},\ldots,\gamma_{K}\right)
=
\frac{1}{n}\sum_{i=1}^{n}\prod_{k=1}^{K}\bar{P}_{q_{i}}\left(\gamma_{k}\right)\ .
\label{eq:lemFinite:emp}
\end{equation}
Let an empirical minimizer of this objective be denoted by:
\begin{equation}
\left(\hat{\gamma}_{1},\ldots,\hat{\gamma}_{K}\right)
\in
\arg\min_{\gamma_{1},\ldots,\gamma_{K}\in\mathcal{G}}
\bar{\mathcal{P}}_{\mathcal{Q}'}\left(\gamma_{1},\ldots,\gamma_{K}\right)\ .
\label{eq:lemFinite:empMin}
\end{equation}
We denote the result of plugging these temperatures back into the population objective as:
\begin{equation}
\bar{\mathcal{P}}'_\mathcal{Q}
=
\bar{\mathcal{P}}_{\mathcal{Q}}\left(\hat{\gamma}_{1},\ldots,\hat{\gamma}_{K}\right)\ .
\label{eq:lemFinite:popPlug}
\end{equation}
Also, for every probability $0<\rho'<1$, define:
\begin{equation}
\Lambda
=
\log\binom{K+M}{M}
+
\log\left(\frac{1}{\rho'}\right)\ .
\label{eq:lem:Finite:Wiz}
\end{equation}
Then, with probability at least $1-\rho'$ over the draw of $\mathcal Q'$, the difference between this failure probability and the optimal failure probability over the grid, defined in Equation \ref{eq:ideal}, is bounded by
\begin{equation}
0
\leq
\bar{\mathcal{P}}'_\mathcal{Q}
-
\bar{\mathcal{P}}_{\mathcal G}^{\star}
\leq
2
\sqrt{
\frac{
\Lambda\bar{\mathcal{P}}_{\mathcal G}^{\star}
}{n}
}
+
\frac{10\Lambda}{3n}\ .
\label{eq:zap:lem:FiniteQ:bound}
\end{equation}
\end{theorem}
The proof idea, also presented in Appendix \ref{zap:sec:Ideal:exp}, is similar to standard bounds on the error incurred when approximating a population distribution using a finite sample over a discrete set of temperatures. However, instead of bounding the error for a fixed choice of temperatures, we need to control the error of the temperatures selected by minimizing the empirical objective. To this end, we compare every possible grid assignment to the best grid assignment under the full population objective. If an assignment has much larger population failure probability than the optimum, then it is unlikely to look better than the optimum on the sampled questions. 

A Bernstein-type concentration bound makes this comparison sharper by using the fact that the relevant random variables are bounded failure probabilities, whose variance is controlled by their mean. Finally, we take a union bound over the unordered grid assignments, whose number is $\binom{K+M}{M}$, since the order of the $K$ temperatures does not affect the objective. This shows that, with high probability, the empirical minimizer cannot be much worse under the true distribution than the best grid assignment.

The theorem shows that the gap between the true and empirical distributions grows only slowly as a function of $K$, but decreases rapidly with $n$. It also highlights which questions are most important to estimate accurately in order to minimize the error, as discussed in Section \ref{sec:Dulling:gold}. 

Combined with Theorem \ref{zap:the:TempDis}, this result implies that our evaluation of the ideal distribution, based on a relatively small dataset and a discretized set of temperatures, is very reliable.

\subsection{Detailed Analysis of the Dulling Effects}

In this part of the appendix, we complete the analysis from Section \ref{sec:Dulling}, by providing the full proofs of the main results, and elaborating on the associated discussions.

\subsubsection{The Temperature Saturation Property}

Before analyzing the dulling effects themselves, we first establish an important structural property of the optimization problem defined in Section \ref{sec:TheIdeal}. Specifically, we show that the number of distinct temperatures in the optimal solution is bounded by the number of questions in the dataset. This is formalized in the following lemma:

\begin{lemma}[The Temperature Saturation Property]
\textup
{Given the setup described in Section \ref{sec:TheIdeal}, for a finite dataset of questions with $n\in\mathbb{N}$, the number of distinct temperatures in an optimal solution is at most $n$.
 \label{zap:lem:TempSat}}
\end{lemma}

The intuition behind this result is that after sampling $n$ questions, each temperature is seen only through its $n$  log-failure probabilities on those questions. Thus, the relaxed optimization problem is a convex optimization problem of only $n$ effective dimensions, meaning we can only have up to $n$ independent temperatures. A complete and rigorous proof is provided in Appendix \ref{zap:lem:TempSat:proof}.

\subsubsection{Detailed Analysis of the Short-Sampling Effect}

\label{zap:sec:short}

We can now take a closer look at the Short-Sampling Dulling Effect, introduced in Section \ref{sec:Dulling:short}. While the high-level intuition was discussed there, a more careful analysis is required to make the argument precise.

\label{zap:Dulling:Reasources}

Having justified the reduction to the three-temperature case in Lemma \ref{zap:lem:TempSat}, we now proceed to prove Proposition \ref{claim:short}.

\begin{proof}[\textbf{Proof - Proposition \ref{claim:short}}]
The proof of this proposition follows from a direct algebraic computation. By Equation \ref{eq:ProbFin}, the probability of solving a question using the compromise temperature is
\begin{equation}
    \mathcal{P}_{\mathrm{comp}}
    =
    1-\left(1-P_C\right)^K \ .
    \label{zap:eq:claim:proof:prob:comp}
\end{equation}
For the two-expert strategy, symmetry implies that the optimal allocation is uniform, so each expert temperature receives $\frac{K}{2}$ samples. Applying Equation \ref{eq:ProbFin} again yields
\begin{equation}
    \mathcal{P}_{\mathrm{expert}}
    =
    1-\left(1-P_E\right)^{\frac{K}{2}} \ .
    \label{zap:eq:claim:proof:prob:exp}
\end{equation}
Taking the difference between the two strategies, we obtain:
\begin{equation}
    \Delta \mathcal{P}
    =
    \mathcal{P}_{\mathrm{expert}}-\mathcal{P}_{\mathrm{comp}}
    =
    \left(1-P_C\right)^K
    -
    \left(1-P_E\right)^{\frac{K}{2}} \ .
\end{equation}
Using the definition of $\Delta P$ from Equation \ref{eq:claim:short:D}, we can rewrite this as
\begin{equation}
    \Delta \mathcal{P}
    =
    \left(1-\frac{P_E}{2}+\Delta P\right)^K
    -
    \left(1-P_E\right)^{\frac{K}{2}} \ .
\end{equation}
Finally, applying the first-order approximation
\begin{equation}
    \log(1-x)=-x+O\left(x^2\right),
\end{equation}
and noting that $\Delta P=O(P_E)$ since $P_C \leq P_E$, we recover Equation \ref{eq:claim:short:DeltaP}, completing the proof.
\end{proof}

As seen from the proof, the threshold around $P_C=\frac{P_E}{2}$ arises naturally from the low-probability approximation. In this regime, the success probability appears in the exponent, scaled by the number of samples $K$. Consequently, increasing the number of samples allocated to a temperature by some factor is, to first order, equivalent to increasing the single-sample success probability by the same factor.

\begin{proposition}[The Weighted Short-Sampling Effect]
\label{claim:short:W}
\textup
{Given the setup described above, when the expert temperatures is optimized in the continuous relaxation, then up to first-order, the result is equivalent to Proposition \ref{claim:short}, with the substitution:
\begin{equation}
P_E\mapsto P_E-\frac{1}{K}\log\left(4r(1-r)\right) \ .
\label{zap:eq:snape}
\end{equation}
Or equivalently:
\begin{equation}
    \bar{P}_{E}\mapsto\sqrt[K]{4r(1-r)}\bar{P}_{E} \ .
    \label{zap:eq:Severus}
\end{equation}
}
\end{proposition}

The idea of the proof (given in full in Appendix \ref{zap:sec:short:proof}) is as follows. The compromise temperature remains unaffected, since all questions share the same effective behavior under it. In contrast, the expert allocation becomes asymmetric due to the weighting. However, since the problem still reduces to a two-variable optimization, it can be solved explicitly, yielding the $\sqrt{4r(1-r)}$ factor. Since the contribution of the number of questions appears multiplicatively in the exponent, this factor enters accordingly.

While this proposition shows that asymmetry between question groups can slightly favor expert allocations, the effect is quite limited. To see this, consider the regime $K \simeq H\simeq \frac{1}{P}$, which, as discussed in Section \ref{sec:gold}, is the relevant scaling.

In this regime, the contribution of the $\log\left(4r(1-r)\right)$ term is small. For example, to double the effective success probability, which in the Gaussian case, corresponds roughly to increasing the efficient range width by a factor of about $1.4$. This means we already require the relatively imbalanced setting of $r \simeq 10.2\%$. To double the effective efficient range, we would get that we need an extreme imbalance of $r \simeq 2.81 \cdot 10^{-6}\%$. Moreover, this gain only benefits one side. Increasing the effective efficient range of one group necessarily reduces that of the other. This further highlights the robustness of the short-sampling effect, as even significant asymmetries translate into only marginal overall gains.

This weighted setting therefore represents an extreme case in favor of multi-temperature strategies. In more realistic scenarios, introducing additional questions with intermediate profiles would only strengthen the performance of the compromise temperature, as it benefits from broader coverage. This further supports the robustness of the efficient range-based notion of separation and the generality of the short-sampling effect.

The second source of complexity is the fact that the questions' profiles may not be identical. To study this case, we analyze a setting with two questions having different expert probabilities $P_E^1, P_E^2$ and different compromise probabilities $P_C^1, P_C^2$. We restrict our attention to the two questions case, as the previous analysis shows that extending to multiple questions does not qualitatively change the conclusions.

\begin{proposition}[The Non-Symmetric Short-Sampling Effect]
\label{claim:short:Non}
\textup{Given the setup described above, the result is equivalent to Proposition
\ref{claim:short}, with
\begin{equation}
    \Delta P
    =
    \frac{\frac{1}{2}P_E^1+\frac{1}{2}P_E^2-P_C^1-P_C^2}{2} \ .
    \label{eq:claim:short:Non:D}
\end{equation}
}
\end{proposition}

We do not present the proof explicitly, as the analyses follows the same lines as that for the proof of Proposition \ref{claim:short:W}.

This result shows that even when question profiles differ, as long as their efficient ranges overlap, the compromise temperature tends to dominate. In this sense, our definition of separation via efficient range overlap provides only a lower bound. In practice, the preference for compromise temperatures can be even stronger.

Finally, we examine the validity of the simplifying assumption that the success probability outside a question's comprise region can be treated as negligible.

To this end, we consider a simplified model in which the temperature profiles are Gaussians, with identical amplitude $A$ and variance $\sigma^2$, but with shifted means:
\begin{equation}
    P_1(\tau)
    =
    A\exp\left(
    -\frac{(\tau-\tau_1)^2}{2\sigma^2}
    \right)
    ,
    \qquad
    P_2(\tau)
    =
    A\exp\left(
    -\frac{(\tau-\tau_2)^2}{2\sigma^2}
    \right) \ .
\end{equation}

We denote by $\tau_C$ the intersection point of the two profiles. Since the two
Gaussians have the same amplitude and variance, this point is given by the midpoint.

\begin{proposition}[Gaussian Tail Correction]
\label{claim:short:Gaussian}
\textup{In the Gaussian setting described above, the two-experts strategy outperforms the compromise temperature, in the first-order Gaussian-tail approximation, if and only if:
\begin{equation}
     P_E
    >
    \sqrt{e} P_C \simeq 1.65 P_C \ .
    \label{eq:claim:gaussian:basic}
\end{equation}
}
\end{proposition}

This result shows that $\frac{P_E}{2}$ is a reasonable approximation for the critical value of $P_C$ in realistic cases, where the two efficient ranges have comparable shape. Once the Gaussian tail is taken into account, the threshold shifts from $0.5P_E$ to approximately $\frac{1}{\sqrt{e}} P_E\simeq 0.61 P_E$, which changes the effective efficient range size by a factor of only about $1.18$. 

Although real temperature profiles are not exactly Gaussians, the empirical profiles we observed in Section \ref{sec:UniMod} exhibit similar behavior. Once they drop to roughly half of their maximum, they continue to decay rapidly. This supports the approximation used in the simplified short-sampling argument. The Gaussian case is solved analytically in Appendix \ref{zap:sec:short:proof}.

It is difficult to combine all of these arguments into a single, fully general theorem, as pathological counterexamples can always be constructed. Nevertheless, the results presented here demonstrate the robustness of the dulling effects. They also justify the use of efficient range overlap as a meaningful measure of separation, and explain why the cost of introducing too many temperatures typically outweighs the cost of using too few.

\subsubsection{Proof of The Goldilocks Sampling Effect}

\label{zap:gold}

\label{zap:dulling:temps}

We conclude this section by proving our main theoretical result of the paper, Proposition \ref{prop:GoldilocksSampling}.

The proof of this proposition is just a straightforward calculus exercise. We begin by inverting the definition of hardness to express the probability as a function of hardness, rather than the other way around as in Equation \ref{eq:def:Hardness}:
\begin{equation}
\bar{P}
=
e^{-1/H} \ .
\end{equation}
Substituting this into the marginal gain defined in Equation \ref{eq:marginal} yields:
\begin{equation}
    \frac{d\mathcal{P}}{dK}
    =
    \frac{1}{H}e^{-K/H} \ .
    \label{eq:Gold:Marginal:H}
\end{equation}
This expression already reveals the main behavior of the problem. If $H \ll K$, the question is easy relative to the sampling budget, and most of the success probability has already been captured. If $H \gg K$, the question is too difficult, and additional samples provide little benefit. The largest marginal gain occurs between these two regimes.

To locate this point, we fix $K$ and maximize the expression with respect to $H$. Taking the derivative and setting it to zero yields:
\begin{equation}
    H = K \ .
\end{equation}
This is the unique critical point, and it corresponds to a maximum, as we can verify by the second derivative, establishing Equation \ref{eq:Gold:MaxMarge}.

To quantify the decay away from this optimal scale, we parameterize
\begin{equation}
    K = \alpha H \quad \Longleftrightarrow \quad H = \frac{K}{\alpha} \ .
\end{equation}
Substituting this into Equation \ref{eq:Gold:Marginal:H}, we obtain:
\begin{equation}
    \frac{d\mathcal{P}_{\alpha}}{dK}
    =
    \frac{\alpha}{K}e^{-\alpha} \ .
\end{equation}
At the optimal scale, where $H = K$ (i.e., $\alpha = 1$), the marginal gain is:
\begin{equation}
    \frac{d\mathcal{P}_{\mathrm{opt}}}{dK}
    =
    \frac{1}{K}e^{-1} \ .
\end{equation}
Normalizing by this optimal value gives
\begin{equation}
\left(\frac{d\mathcal{P}_{\mathrm{opt}}}{dK}\right)^{-1}
\frac{d\mathcal{P}_{\alpha}}{dK}
=
\alpha e^{1-\alpha} \ ,
\end{equation}
which proves Equation \ref{eq:Gold:Dev} and completes the proof.

\section{Computational Resources}
\label{zap:CompResources}

We used the \texttt{vLLM} inference engine for batched sampling from each of the four base models, on NVIDIA L40S, A40, GeForce RTX 5090, and GeForce RTX 4090 GPUs (24 to 48~GB of memory each), for a total of about 14,000 GPU-hours.

\paragraph{Code and reproducibility.} Our code is available at \url{https://github.com/avivorly/one-temperature}. It contains a vLLM sampling script over a temperature grid, the MATH-500 grader, the unbiased pass@$K$ estimator, and the temperature-allocation comparison (best single temperature, uniform mixing, and the optimal allocation) with leave-one-out and 10-fold held-out evaluation. Its README gives the exact commands and lists the parts it does not include. All benchmarks and models are public; their sources, versions and licenses are listed in the paper checklist.

\subsection{GPT OSS 20B}
\label{zap:DataDis:gpt_oss_20b}

\begin{center}
\small
\begin{tabular}{@{}lc>{\raggedright\arraybackslash}p{0.25\linewidth}>{\raggedright\arraybackslash}p{0.38\linewidth}@{}}
  \toprule
  Dataset & \makecell{Included\\questions} & Samples per $(q, \tau)$ & Temperature grid \\
  \midrule
  LiveCodeBench & $766$ & ${\sim}1.5\mathrm{k}$; harder questions ${\sim}10\mathrm{k}$, sometimes more & $16$: $0.1$, $0.2$, $\ldots$, $1.2$, $1.4$, $1.5$, $1.7$, $2.0$ \\
  MATH-500 & $500$ & $1500$ & the same $16$ \\
  Hi-ToM & $678$ & $1500$ & $31$: $0.075$, $0.15$, $\ldots$, $1.95$, $2.0$, $2.2$, $2.5$, $3.0$, $4.0$ \\
  \bottomrule
\end{tabular}
\end{center}

\subsection{Mistral 7B Instruct}
\label{zap:DataDis:mistral_7b_instruct}

\begin{center}
\small
\begin{tabular}{@{}lc>{\raggedright\arraybackslash}p{0.25\linewidth}>{\raggedright\arraybackslash}p{0.38\linewidth}@{}}
  \toprule
  Dataset & \makecell{Included\\questions} & Samples per $(q, \tau)$ & Temperature grid \\
  \midrule
  LiveCodeBench & $359$ & ${\sim}1.5\mathrm{k}$; harder questions ${\sim}10\mathrm{k}$, sometimes more & $16$: $0.1$, $0.2$, $\ldots$, $1.2$, $1.4$, $1.5$, $1.7$, $2.0$ \\
  MATH-500 & $455$ & $1500$ & the same $16$ \\
  Hi-ToM & $615$ & $1500$ & $31$: $0.075$, $0.15$, $\ldots$, $1.95$, $2.0$, $2.2$, $2.5$, $3.0$, $4.0$ \\
  \bottomrule
\end{tabular}
\end{center}

\subsection{Qwen3-8B}
\label{zap:DataDis:qwen3_8b}

\begin{center}
\small
\begin{tabular}{@{}lc>{\raggedright\arraybackslash}p{0.25\linewidth}>{\raggedright\arraybackslash}p{0.38\linewidth}@{}}
  \toprule
  Dataset & \makecell{Included\\questions} & Samples per $(q, \tau)$ & Temperature grid \\
  \midrule
  LiveCodeBench & $646$ & ${\sim}1.5\mathrm{k}$; harder questions ${\sim}10\mathrm{k}$, sometimes more & $38$: $0.1$, $0.2$, $\ldots$, $3.5$, $3.7$, $4.0$, $4.2$ \\
  MATH-500 & $496$ & $750$ & $20$: $0.1$, $0.2$, $\ldots$, $2.0$ \\
  Hi-ToM & $466$ & $1500$ & $31$: $0.075$, $0.15$, $\ldots$, $1.95$, $2.0$, $2.2$, $2.5$, $3.0$, $4.0$ \\
  \bottomrule
\end{tabular}
\end{center}

\subsection{Qwen3-0.6B}
\label{zap:DataDis:qwen3_0_6b}

\begin{center}
\small
\begin{tabular}{@{}lc>{\raggedright\arraybackslash}p{0.25\linewidth}>{\raggedright\arraybackslash}p{0.38\linewidth}@{}}
  \toprule
  Dataset & \makecell{Included\\questions} & Samples per $(q, \tau)$ & Temperature grid \\
  \midrule
  LiveCodeBench & $85$ & approximately doubled on some particularly hard questions & $20$: $0.1$, $0.2$, $\ldots$, $2.0$ \\
  MATH-500 & $492$ & $1500$ & the same $20$ \\
  Hi-ToM & $743$ & $1500$ & $31$: $0.075$, $0.15$, $\ldots$, $1.95$, $2.0$, $2.2$, $2.5$, $3.0$, $4.0$ \\
  \bottomrule
\end{tabular}
\end{center}

\subsection{\texorpdfstring{Datasets}{Datasets}}
\label{zap:Datasets}
\label{zap:MATH500}
\label{zap:HiToM}

From LiveCodeBench \cite{jain2024livecodebench}, we use the six datasets V1 to V6, with 975 coding questions, and Qwen3-0.6B uses 110 of them. MATH-500 is the 500-question test split of the MATH benchmark \cite{hendrycks2021measuring} defined for PRM800K \cite{lightman2023lets}. Hi-ToM \cite{he2023hitom} is a multiple-choice benchmark of higher-order theory-of-mind reasoning with 1,200 questions. On Hi-ToM, we analyze the questions whose best single-temperature pass rate was below $0.30$ in a first stage of $500$ samples per question and temperature ($1000$ for Mistral 7B Instruct).

The token limit is $4096$ new tokens on MATH-500 ($2048$ for Qwen3-8B) and $1000$ on Hi-ToM ($500$ for Mistral 7B Instruct). On MATH-500, every model received the instruction ``Solve the following math problem efficiently and clearly. Reason step by step, and put your final answer within \texttt{\textbackslash boxed\{\}}.'' in its chat template, and Qwen3-8B received it followed by a sentence that states its limit. On Hi-ToM, GPT OSS 20B and the two Qwen3 models received each question's prompt exactly as the dataset gives it, and Mistral 7B Instruct received it followed by a fixed instruction that states its limit. On MATH-500 and Hi-ToM, GPT OSS 20B used the Harmony chat format with reasoning effort \texttt{medium} and the response primed to open in the \texttt{final} channel, and ran in bfloat16 with the checkpoint's native MXFP4 quantization. On these two datasets, Mistral 7B Instruct used the Mistral chat template and ran in float16, and the two Qwen3 models used the Qwen3 chat template with thinking mode disabled and ran in bfloat16 (Qwen3-8B) and float16 (Qwen3-0.6B).

On MATH-500, the answer is the content of the last \texttt{\textbackslash boxed\{\}} span of the completion, and a sample with no such span fails. A sample passes if its answer equals the reference answer after normalization, or if an equivalence layer of generic rules and a time-bounded symbolic check with SymPy accepts it. On Hi-ToM, where the prompt lists all $15$ options, the answer is the first valid option in the completion, that is, the first mention of one of the question's own options, by its name or by its letter given as an answer label. A completion that names no valid option fails.

\section{Question Profile Analyses}

\label{zap:QuestionsProfiles}

We begin this appendix with a detailed exposition of our analysis of the most fundamental objects we consider, which underpin the remainder of the text.

\label{zap:Stats:Methodology}

\textbf{Goal and choice of basis}

Our starting point is a set of discrete pass-rate measurements on a temperature grid, one curve per sampled question $q_i$, and our goal is to recover the smooth profile $P_{q_i}(\tau)$ that best denoises those measurements. Concretely, for question $q_i$ we observe $s_{ij}$ successes out of $\nu$ trials at each sampled temperature $\tau_j$, and we model their source as an underlying smooth function $P_{q_i}(\tau)$ that we wish to estimate faithfully.

We do not want to assume a specific functional form as a sigmoid, an exponential decay, a Gaussian peak, because any such ansatz would bias the recovered peak position, peak height, and tail behavior toward what the family can express, as conclusions drawn afterwards would reflect the chosen form as much as the data.

We therefore fit in a \textbf{general basis}: a polynomial space rich enough to express any sufficiently smooth $P_{q_i}$ up to a chosen degree, with \textbf{shape constraints} that encode minimal physical priors but no functional form. The data, rather than an ansatz, determines the curve.

\textbf{Confidence intervals for each measurement}

Each measurement is binomial, $s_{ij} \sim \mathrm{Binomial}(\nu, P_{q_i}(\tau_j))$. For the per-point error bars we use \textbf{exact Clopper-Pearson $1\sigma$ intervals}: at the two-sided tail probability $\alpha_{\mathrm{CP}} = 1 - \mathrm{erf}(1/\sqrt{2}) \approx 0.317$ corresponding to $1\sigma$, the interval $[\mathrm{lo}, \mathrm{hi}]$ for a binomial proportion is given by quantiles of the Beta distribution,
\begin{equation}
\mathrm{lo} \;=\; \mathrm{Beta}^{-1}_{s,\, \nu-s+1}\!\left(\tfrac{\alpha_{\mathrm{CP}}}{2}\right), \qquad
\mathrm{hi} \;=\; \mathrm{Beta}^{-1}_{s+1,\, \nu-s}\!\left(1 - \tfrac{\alpha_{\mathrm{CP}}}{2}\right) \ ,
\end{equation}
with $\mathrm{lo} = 0$ when $s = 0$ and $\mathrm{hi} = 1$ when $s = \nu$. We prefer this exact construction over Wald or Wilson approximations because Clopper-Pearson has guaranteed coverage at the boundary counts ($s$ near $0$ or $\nu$) where Gaussian-style intervals systematically under-cover.

Clopper-Pearson intervals are asymmetric around $\hat{p} = s / \nu$, especially near the boundary. We keep the asymmetry rather than collapse it: each datum carries both half-widths $\sigma_- = \hat{p} - \mathrm{lo}$ and $\sigma_+ = \mathrm{hi} - \hat{p}$. Symmetrizing to a single $\sigma_{\mathrm{eff}} = (\mathrm{hi} - \mathrm{lo}) / 2$ would discard information that the Clopper-Pearson computation already produced and would shift the implied $1\sigma$ band off-center.

\textbf{Loss: weighted least squares with side-dependent variance}

We fit by \textbf{weighted least squares (WLS)}, with each residual normalized by the actual CP half-width on its side. Without side-dependent weights, the asymmetric Clopper-Pearson information preserved in the previous subsection would simply be discarded. Concretely:
\begin{equation}
\mathcal{L}(P_{q_i}) \;=\; \sum_{j} \left( \frac{P_{q_i}(\tau_j) - \hat{p}_{ij}}{\sigma_\pm(P_{q_i}, \hat{p}_{ij})} \right)^{\!2} \ ,
\qquad
\sigma_\pm \;=\; \begin{cases}\sigma_+ & \text{if } P_{q_i}(\tau_j) > \hat{p}_{ij}, \\[2pt] \sigma_- & \text{otherwise} \ . \end{cases}
\end{equation}
This loss is convex in the polynomial coefficients, $C^1$ at $P_{q_i}(\tau_j) = \hat{p}_{ij}$ (the gradient from both sides is zero regardless of which $\sigma$ applies, so only the second derivative jumps), and faithful to the binomial geometry near $s = 0$ and $s = \nu$ where the CP asymmetry is most pronounced.

The full binomial negative-log-likelihood would be statistically more rigorous, but its curvature diverges as $p \to 0$ or $1$, breaking the closed-form step size used below. The two-piece Gaussian above is a tractable proxy.

\textbf{Shape constraints}

We use the \textbf{Bernstein basis of degree $10$} on a per-question domain bracket $[\tau_{\mathrm{lo}}, \tau_{\mathrm{hi}}]$, taken as the support of the non-zero observations padded by one temperature step on each side; $P_{q_i} \equiv 0$ outside. Bernstein is chosen over the monomial or Legendre bases because shape constraints transfer cleanly from the function to the coefficients $\{c_j\}$: non-negative coefficients yield a non-negative function, and unimodal coefficient sequences yield a unimodal function, which makes constrained optimization tractable. Degree $10$ is large enough to express the observed shapes and small enough to avoid the conditioning blow-up that plagues higher-degree polynomial bases on a fine grid. Moreover, degree $10$ is feasible because the shape regularization itself (the non-negativity and unimodality projections below) reduces the effective number of free polynomial parameters; without those constraints a lower degree would be safer, but with them $10$ is empirically optimal.

We impose two minimal physical priors on $P_{q_i}$:
\begin{itemize}
  \item \textbf{Non-negativity.} $P_{q_i}$ is a probability.
  \item \textbf{Unimodality.} Sampling temperature trades exploration against exploitation, so on physical grounds each question has a single optimal regime; multimodal $P_{q_i}$ would correspond to two distinct mechanisms favoring different temperatures, which we do not see in the data.
\end{itemize}

\begin{table}[t]
\centering
\caption{Shape constraints translated from the function level to the Bernstein-coefficient level. The $\{c_j\}$ are the Bernstein-basis weights of the fitted profile.}
\label{tab:shape-constraints}
\begin{tabular}{@{}lll@{}}
\toprule
\textbf{Constraint} & \textbf{Function-level meaning} & \textbf{Bernstein-coefficient meaning} \\
\midrule
Non-negativity & $P_{q_i}(\tau) \ge 0$ for all $\tau$ & All coefficients $c_j \ge 0$ \\
Unimodality    & $P_{q_i}(\tau)$ has a single peak & Sequence $\{c_j\}$ is unimodal \\
\bottomrule
\end{tabular}
\end{table}

Both constraints are enforced as \textbf{projections during optimization}, not as additive penalties: the current Bernstein coefficient vector is periodically projected onto the cone of non-negative unimodal sequences via isotonic regression around each candidate peak position, retaining the projection closest in Euclidean distance. We deliberately do not add an explicit smoothness penalty: the bounded polynomial degree combined with the two priors already produces visibly smooth curves, and any explicit smoothness term would compete with the data fit at the peak and bias the recovered peak height downward.

\textbf{Optimization}

The WLS loss is minimized by projected gradient descent on the Bernstein coefficients, with a Lipschitz-safe step size derived from the largest eigenvalue of the WLS Hessian (estimated by power iteration). Initialization is at zero, which biases the iterates toward small-norm solutions before the projections engage. In the early iterations the projection is soft-blended into the iterate so the data fit can settle before the shape priors take effect; later iterations apply it as a hard projection. The optimization runs for a fixed budget chosen generously enough that further iteration changes the coefficients negligibly.

The output, per question, is a smooth, non-negative, unimodal $P_{q_i}(\tau)$: a degree-$10$ polynomial inside its bracket and zero outside.

\textbf{Question filtering and sample-budget conventions}

\paragraph{Minimum peak success probability (applies to every dataset and every plot).}
A single global filter is applied to every dataset and every figure in
Appendices \ref{zap:Hard} to \ref{zap:DataDis:Prot}: questions whose fitted peak success probability
$\sup_{\tau} P_q(\tau)$ is below $10^{-3}$ are excluded. This filter applies
once, in advance, before any plot is drawn or any table is built.

We have observed questions whose hardness $H_q$
(Equation \ref{eq:def:Hardness}) extends another order of magnitude beyond
this cutoff, but including them would be misleading for two reasons:
\begin{enumerate}
  \item \emph{Fit instability.} Below $10^{-3}$ peak success probability,
        the available data per question is so sparse that the polynomial
        fit becomes chaotic and is no longer representative of the
        underlying response curve.
  \item \emph{Goldilocks irrelevance.} By the Goldilocks sampling
        effect (Proposition \ref{prop:GoldilocksSampling}), questions
        whose hardness $H$ (Equation \ref{eq:def:Hardness}) is much larger
        than $K_{\max}$ are negligible at the $K$ values reported in
        Appendices \ref{zap:Hard} to \ref{zap:DataDis:Prot}.
\end{enumerate}

\paragraph{Unsolvable questions.}
Questions that no temperature ever solves (peak success probability
exactly zero) are excluded everywhere, from every plot, every table,
and every denominator. They contribute nothing to any number reported
below.

\paragraph{Maximum sample budget.}
All pass@$K$ traces, leave-one-out (LOO) points, and ratio-to-uniform
curves in Appendices \ref{zap:Hard} to \ref{zap:DataDis:Prot} are evaluated up to $K_{\max} = 1024$.
Larger values of $K$ are uninformative for the deployed setting and were
clipped at the report-build stage to keep the optimal-allocation LOO
solve tractable.

\section{Section \ref{sec:Profiles} Expanded Results}
\label{zap:Hard}

In this section, we present the analyses and results from Section \ref{sec:Profiles} after deriving the question profiles, for the four models on the three datasets, LiveCodeBench, MATH-500, and Hi-ToM. We begin by describing the analysis procedure and then present the results. 

\subsection{\texorpdfstring{Statistical Analysis Approach}{Statistical Analysis Approach}}
\label{zap:Hard:The}

\paragraph{Sliding window.}
\label{zap:DataDis:setup}
Throughout this paper, ``sliding window of size $N$'' (where $N$ is even)
refers to the $N$ questions closest in hardness to a given anchor point:
the $N/2$ questions with the next-higher hardness and the $N/2$ questions
with the next-lower hardness. At the edges of the data the window is
clipped symmetrically (so the window may shrink, but never becomes
asymmetric). Quantities reported ``per window'' average over those
$\le N$ neighbors.

\paragraph{How to read the per-question peak temperature plot.}
\label{zap:DataDis:AveTemp}
Each grey dot is one included question, plotted at its hardness
$H_q$ (Equation \ref{eq:def:Hardness}, $x$-axis on a log scale)
versus its peak temperature $\tau_q^\star = \arg\max_\tau P_q(\tau)$
($y$-axis, linear).

\medskip\noindent
\textbf{Sliding averages.} The blue line is a symmetric sliding mean of
$\tau_q^\star$ taken over a window of $30$ consecutive questions in
the $H$-sorted list (the same definition as the global ``sliding
window of size $N$'' in Section \ref{zap:DataDis:setup}, with
$N = 30$). At the edges the window is clipped symmetrically so it
shrinks but never becomes asymmetric. The red line is the same
running mean, but restricted to the subset of questions with
$\tau_q^\star > 0.1$; this isolates the trend in $\tau_q^\star$ once
the spike of near-zero-peak questions has been removed.
Each line leaves out the $15$ questions at each end of its $H$-sorted list, where its window is not full. The figures compute $H_q$ with the peak pass rate capped at $99.999\%$, so every question at or above that rate sits at $H = 0.0869$. On LiveCodeBench and MATH-500, each line then starts at the first remaining question whose next question is at least $5\%$ harder. This removes questions at $H = 0.0869$ and at most two other questions per line. The Hi-ToM lines use every full window, because no Hi-ToM question sits at $H = 0.0869$.

\medskip\noindent
\textbf{What to look for.} A trend in either curve as $H$ increases
indicates that the hardness of a question carries information about
its preferred sampling temperature. The red curve is the more
informative of the two when the dataset contains a large near-zero
cluster, because the all-questions blue line is otherwise dominated
by it.

\paragraph{How to read the efficient-range-size plot.}
\label{zap:DataDis:RangeSize}
Each grey dot is one included question, plotted at its hardness $H_q$
(Equation \ref{eq:def:Hardness}, $x$-axis on a log scale) versus the
size $|E_q|$ of its efficient temperature range
(Equation \ref{eq:efficient range}, $y$-axis on a linear scale). For a
unimodal profile the efficient range is approximately an interval
$[\,t_L,\, t_R\,]$, so we report $|E_q| = t_R - t_L$ directly.

\medskip\noindent
\textbf{OLS fit.} The red line is the ordinary least-squares
regression of $|E_q|$ on $\log_{10} H_q$; the slope, intercept, and
$R^2$ are printed in the legend. The slope is the rate at which the
efficient range shrinks per decade of hardness.

\medskip\noindent
\textbf{What to look for.} A clearly negative slope confirms the
shrinking-range trend stated in the main text
(Section \ref{sec:range}): harder questions have narrower efficient
temperature windows. The $R^2$ summarizes how tight that
log-hardness trend is across the dataset.

\paragraph{How to read the maximum-coverage plot.}
\label{zap:DataDis:relC}
We split the hardness range $[\min_q H_q,\ \max_q H_q]$ into
log-uniform bins (by default $\lfloor n/20 \rfloor$ bins, where $n$
is the number of included questions). For each bin $b$, let $M_b$
be the number of in-bin questions, and let $\mathcal{C}(\tau)$ be
the number of those questions whose efficient temperature range $E$
(Equation \ref{eq:efficient range}) contains temperature $\tau$. We compute
$\max_{\tau} \mathcal{C}(\tau)$ via a sweep-line over the range
endpoints and plot $\max_{\tau} \mathcal{C}(\tau) / M_b$ as a bar
spanning the bin's width.

\medskip\noindent
\textbf{Bar position and height.} Each bar is centered on its bin
center and spans the bin's width on the $H$ axis. Height $= 1$
means \emph{some} single temperature lies inside the half-max
range of every in-bin question simultaneously; height $= 1/M_b$
means at most one in-bin question is covered by any given
temperature.

\medskip\noindent
\textbf{What to look for.} Bars should stay near $1$ at low $H$
(broad, mutually-overlapping ranges) and fall toward $0$ at high
$H$ (narrow ranges, no shared sweet spot), the same
shrinking-range trend the previous two plots quantify
question-by-question.

\subsection{\texorpdfstring{Extended Results}{Extended Results}}
\label{zap:Hard:res}

\begin{figure}[t]
  \centering
  \begin{subfigure}[t]{0.48\textwidth}
    \centering
    \includegraphics[width=\linewidth]{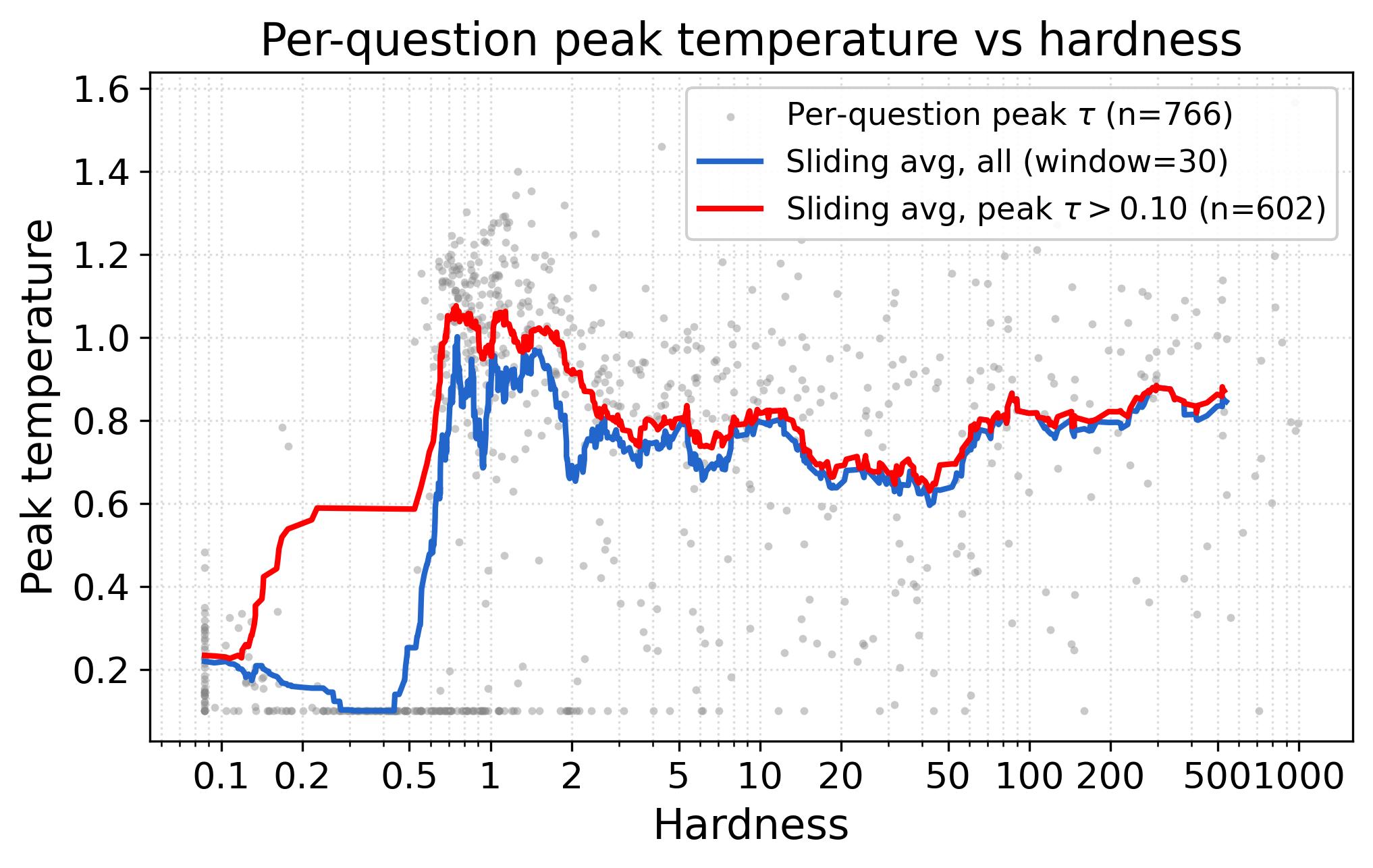}
    \caption{GPT OSS 20B, $n = 766$}
    \label{fig:AveTemp:gpt_oss_20b}
  \end{subfigure}\hfill  \begin{subfigure}[t]{0.48\textwidth}
    \centering
    \includegraphics[width=\linewidth]{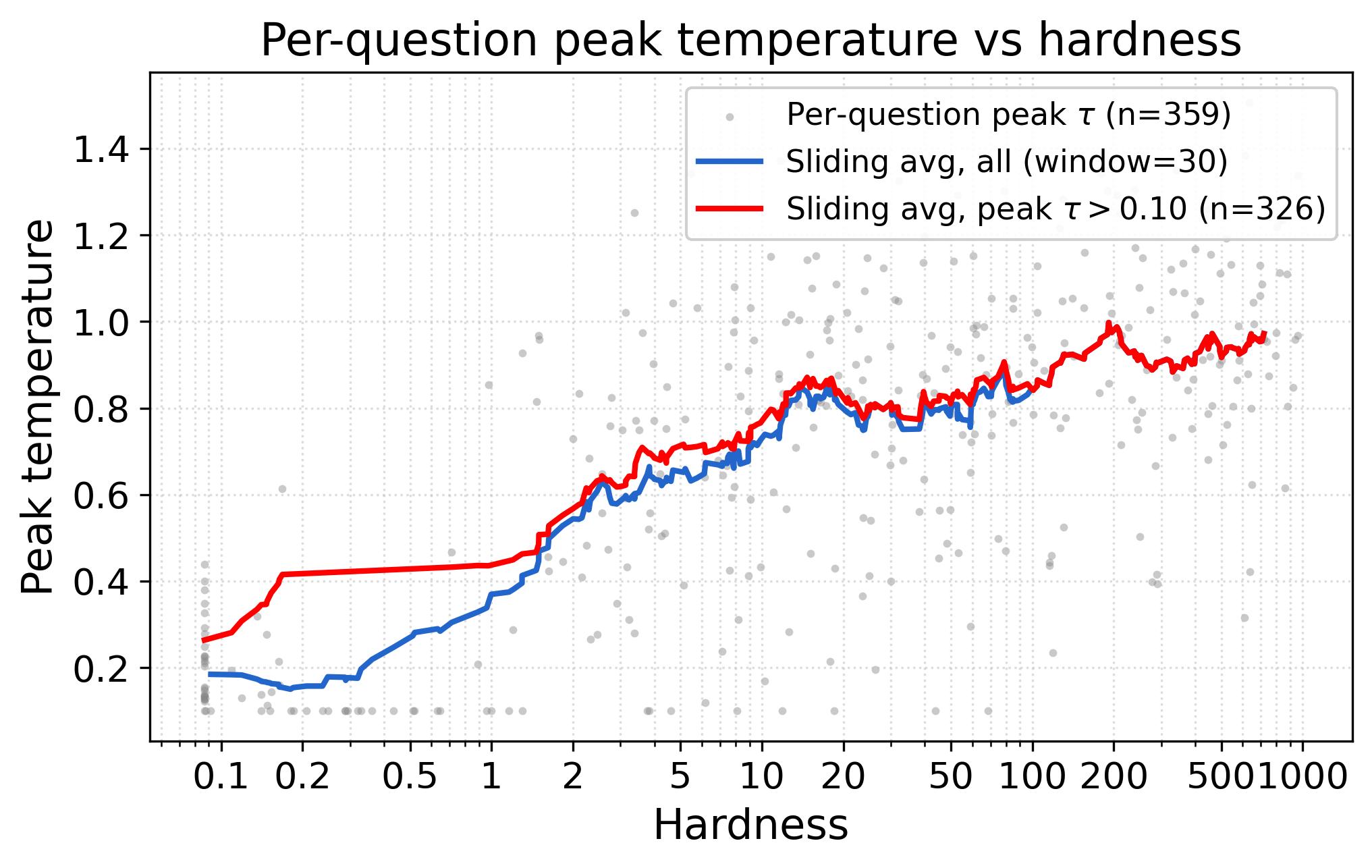}
    \caption{Mistral 7B Instruct, $n = 359$}
    \label{fig:AveTemp:mistral_7b_instruct}
  \end{subfigure}
  \vspace{8pt}

  \begin{subfigure}[t]{0.48\textwidth}
    \centering
    \includegraphics[width=\linewidth]{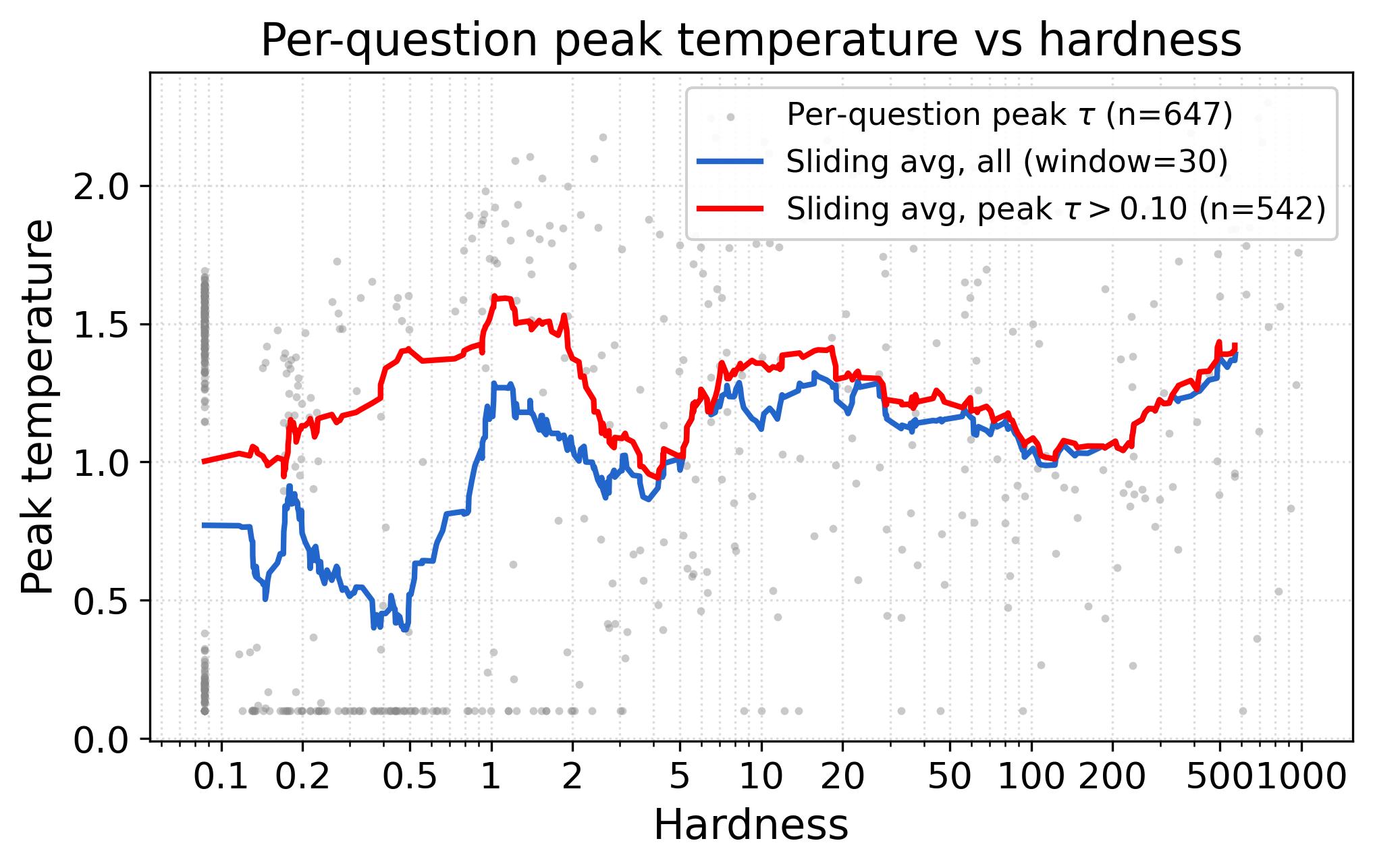}
    \caption{Qwen3-8B, $n = 647$}
    \label{fig:AveTemp:qwen3_8b}
  \end{subfigure}\hfill  \begin{subfigure}[t]{0.48\textwidth}
    \centering
    \includegraphics[width=\linewidth]{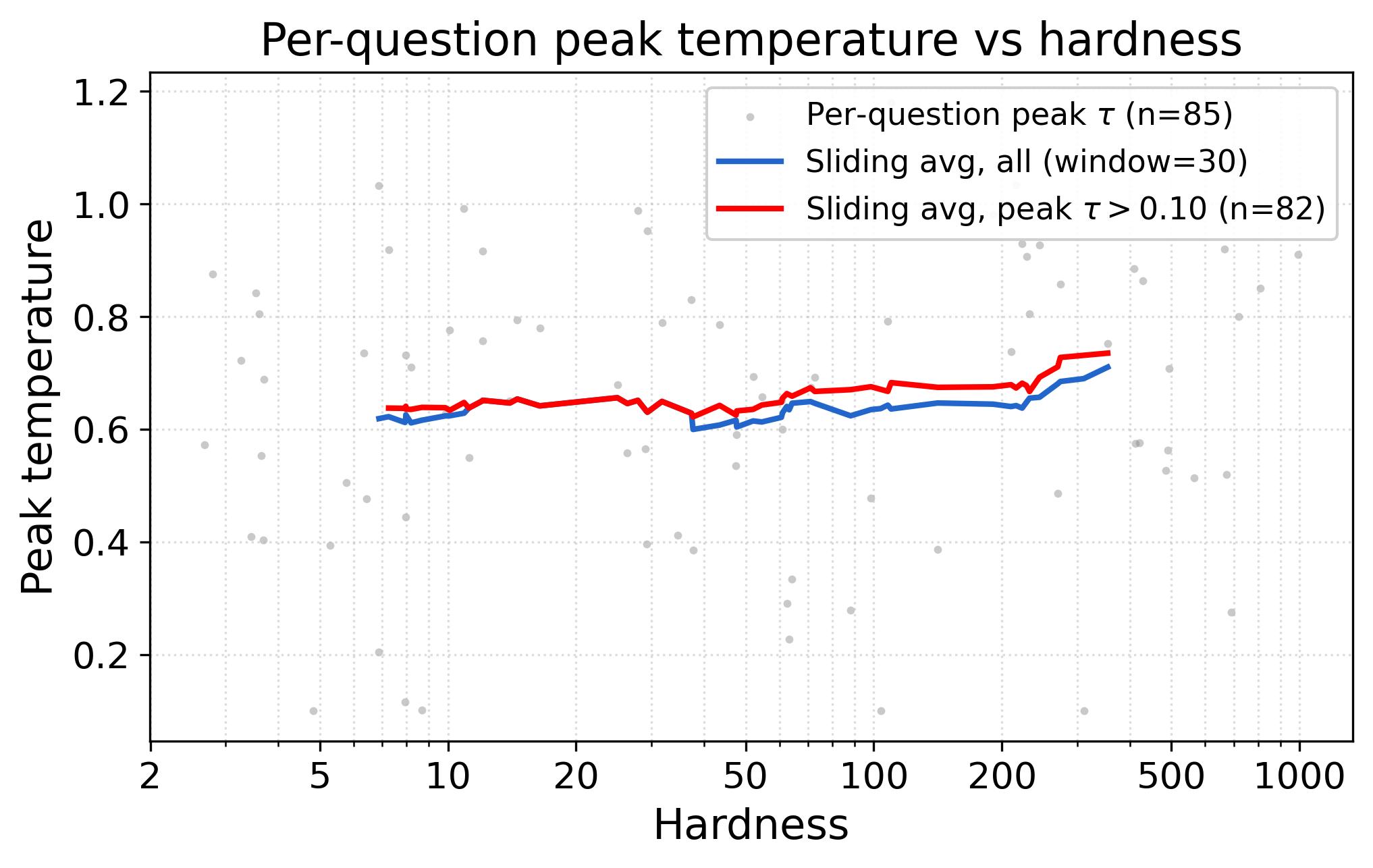}
    \caption{Qwen3-0.6B, $n = 85$}
    \label{fig:AveTemp:qwen3_0_6b}
  \end{subfigure}  \caption{Per-question peak temperature vs.\ hardness on LiveCodeBench, with sliding averages over all questions (blue) and over the large ideal temperature subset (red), across all four models. $n$ is the number of included questions.}
  \label{fig:AveTemp:grid}
\end{figure}

\begin{figure}[p]
  \centering
  \begin{subfigure}[t]{0.48\textwidth}
    \centering
    \includegraphics[width=\linewidth]{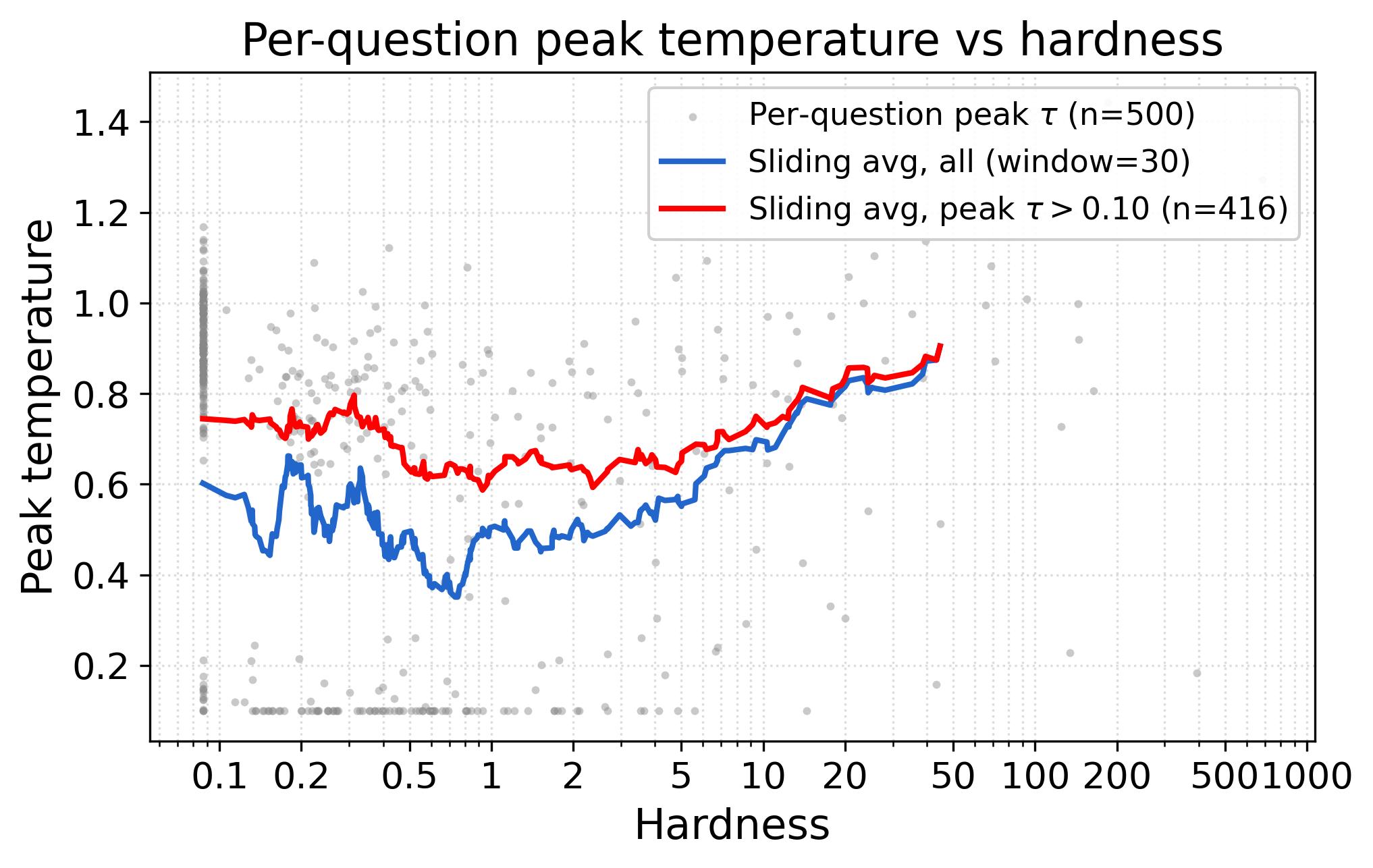}
    \caption{GPT OSS 20B, $n = 500$}
    \label{fig:AveTemp:math500:gpt_oss_20b}
  \end{subfigure}\hfill  \begin{subfigure}[t]{0.48\textwidth}
    \centering
    \includegraphics[width=\linewidth]{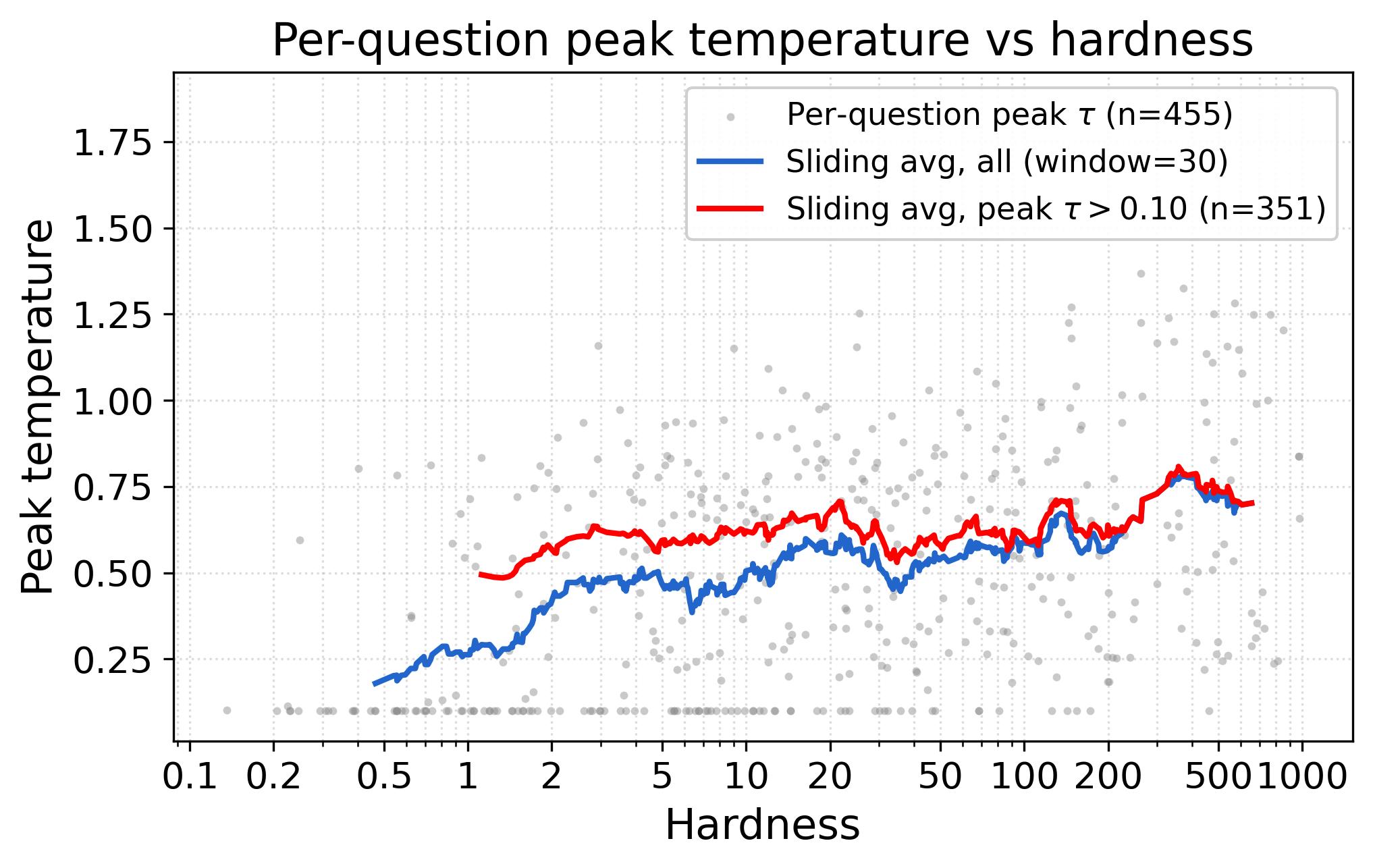}
    \caption{Mistral 7B Instruct, $n = 455$}
    \label{fig:AveTemp:math500:mistral_7b_instruct}
  \end{subfigure}
  \vspace{8pt}

  \begin{subfigure}[t]{0.48\textwidth}
    \centering
    \includegraphics[width=\linewidth]{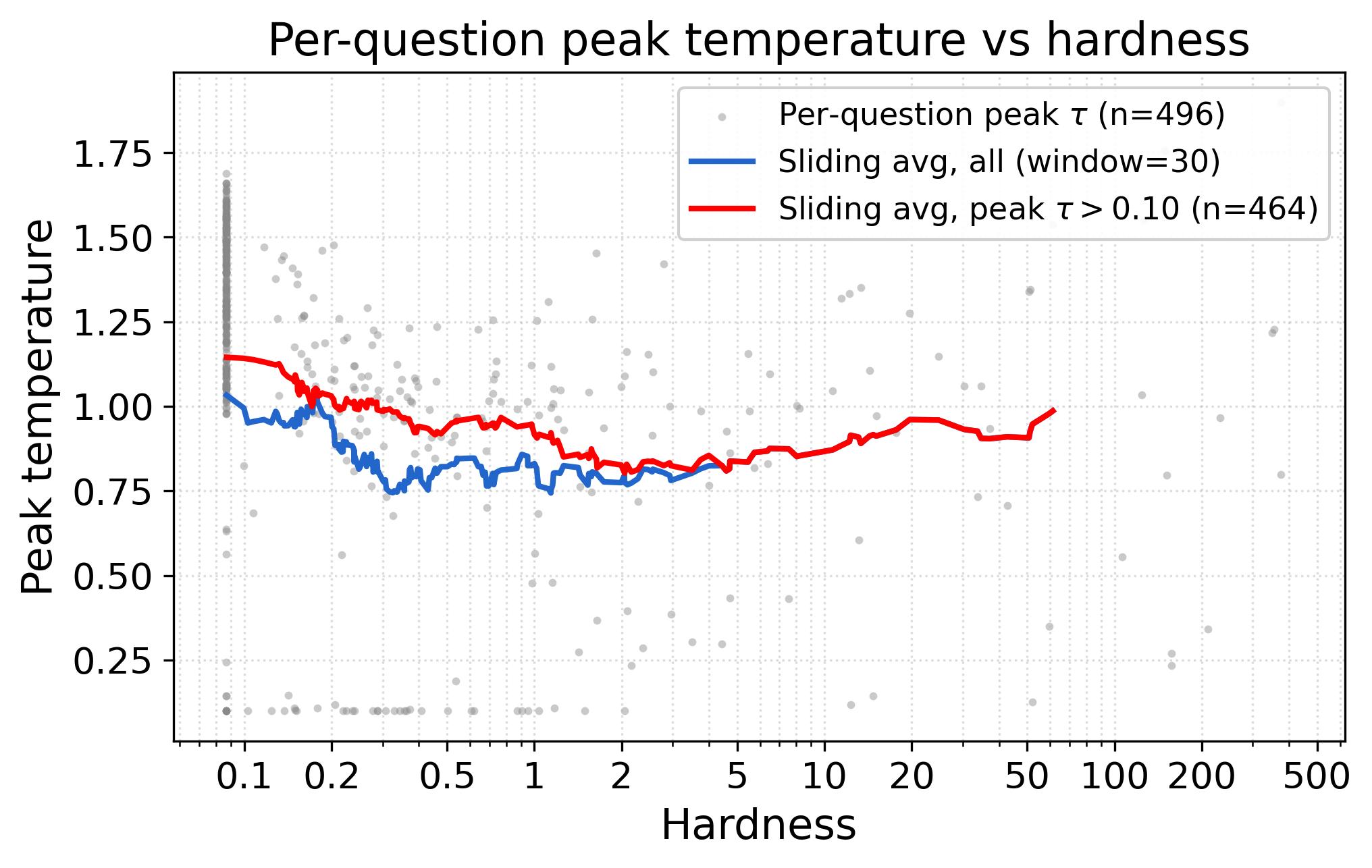}
    \caption{Qwen3-8B, $n = 496$}
    \label{fig:AveTemp:math500:qwen3_8b}
  \end{subfigure}\hfill  \begin{subfigure}[t]{0.48\textwidth}
    \centering
    \includegraphics[width=\linewidth]{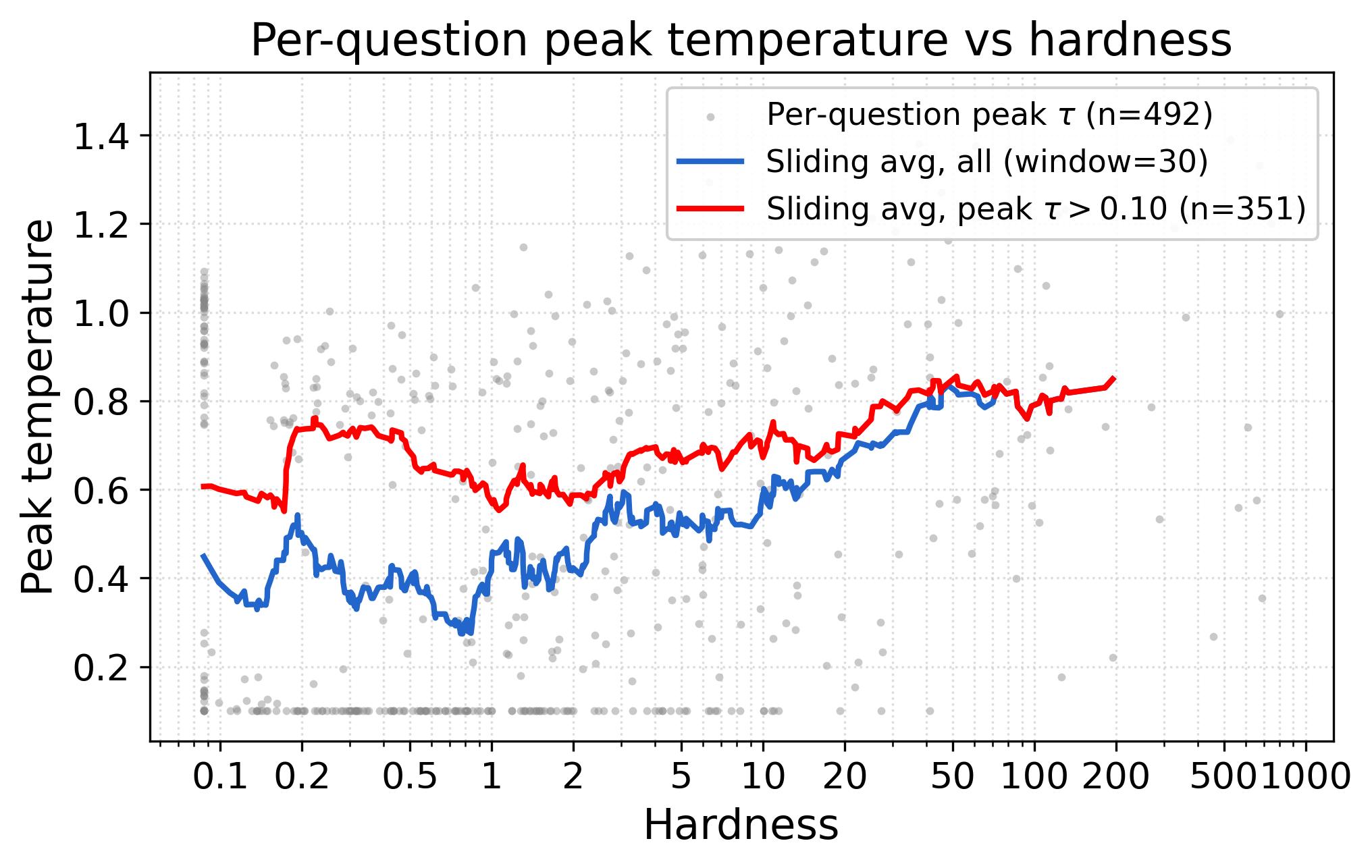}
    \caption{Qwen3-0.6B, $n = 492$}
    \label{fig:AveTemp:math500:qwen3_0_6b}
  \end{subfigure}
  \caption{Per-question peak temperature vs.~hardness on MATH-500, drawn as in Figure \ref{fig:AveTemp:grid}. $n$ is the number of included questions.}
  \label{fig:AveTemp:math500:grid}
  \par\vspace{10pt}

  \centering
  \begin{subfigure}[t]{0.48\textwidth}
    \centering
    \includegraphics[width=\linewidth]{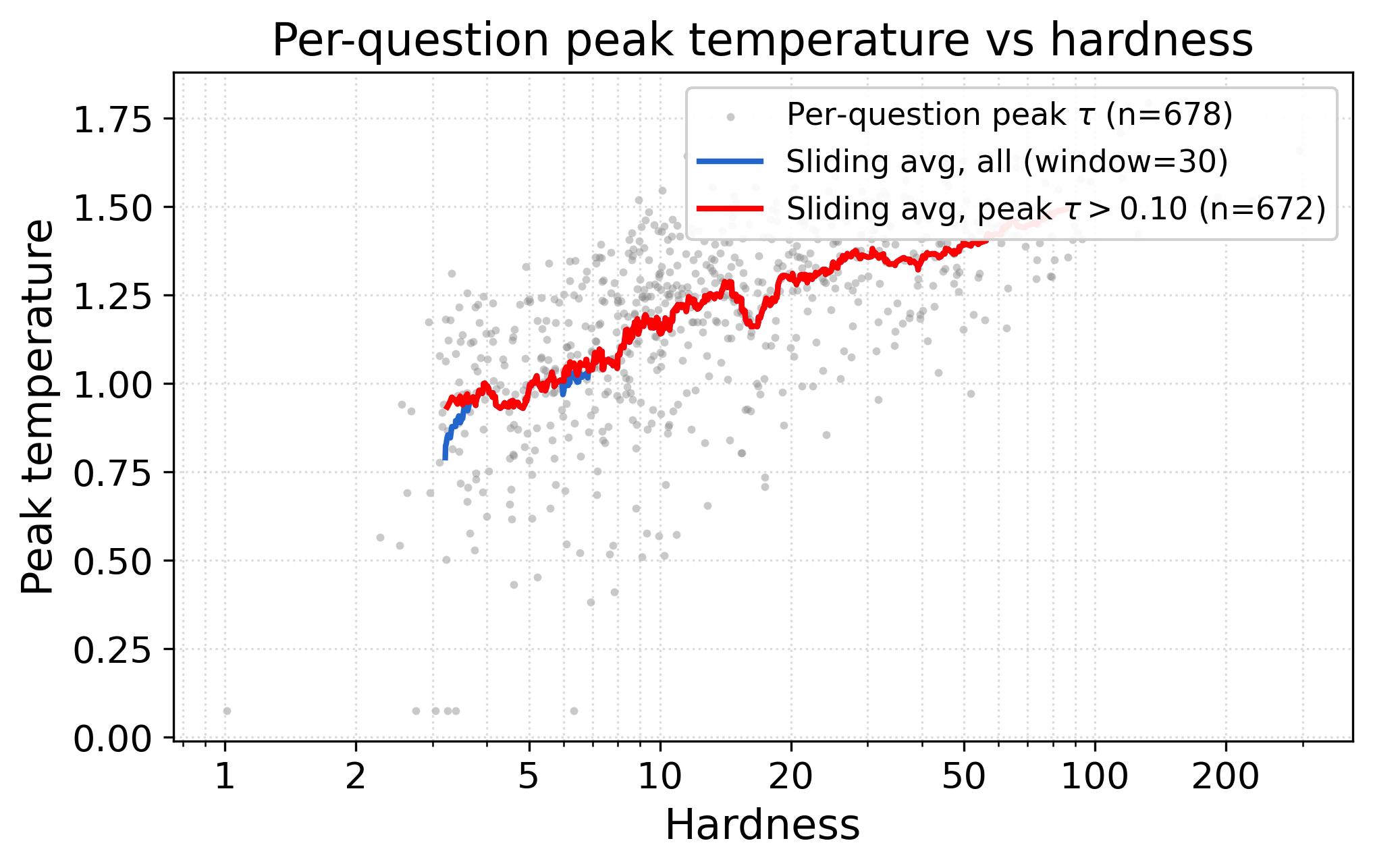}
    \caption{GPT OSS 20B, $n = 678$}
    \label{fig:AveTemp:hitom:gpt_oss_20b}
  \end{subfigure}\hfill  \begin{subfigure}[t]{0.48\textwidth}
    \centering
    \includegraphics[width=\linewidth]{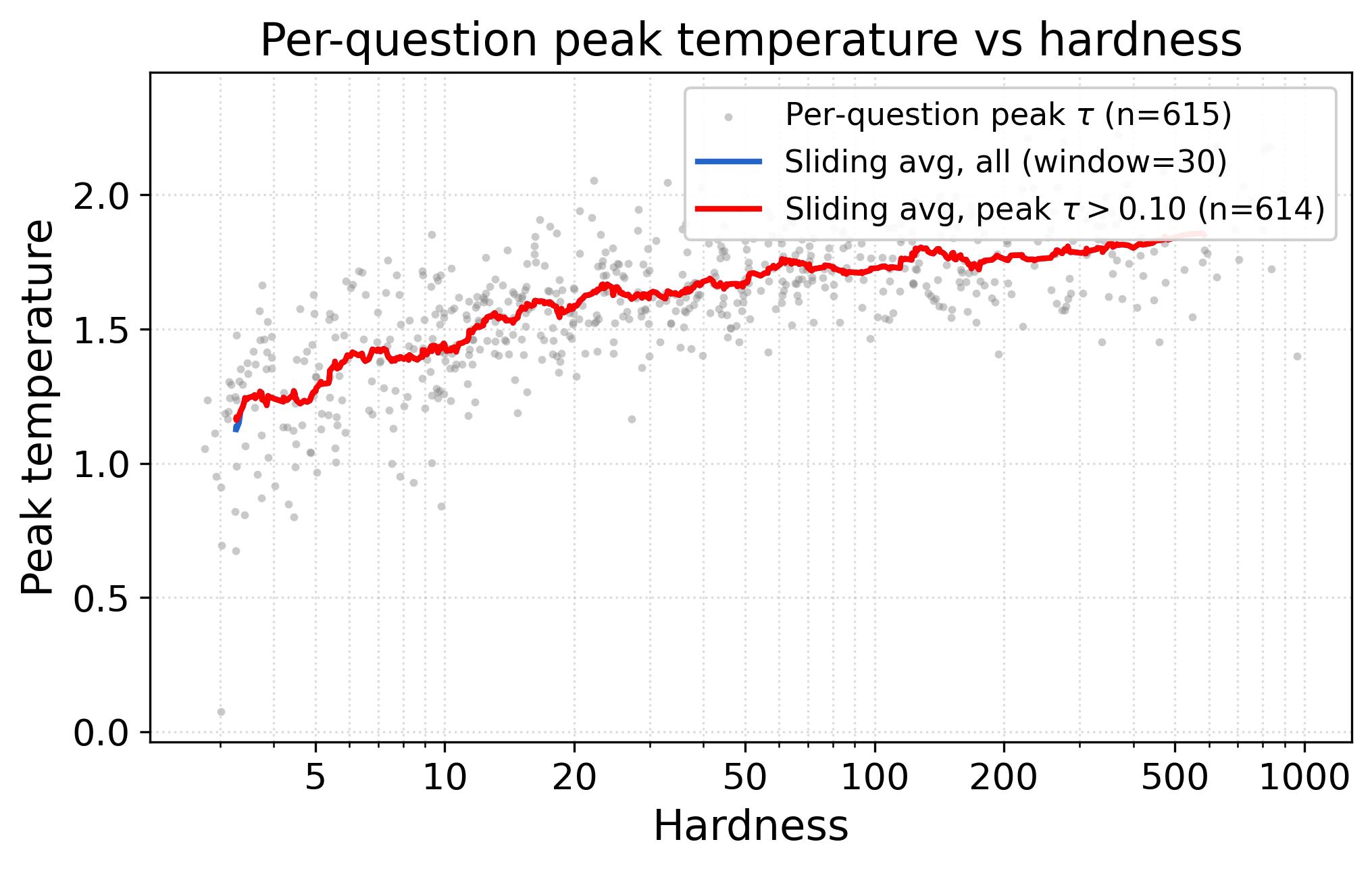}
    \caption{Mistral 7B Instruct, $n = 615$}
    \label{fig:AveTemp:hitom:mistral_7b_instruct}
  \end{subfigure}
  \vspace{8pt}

  \begin{subfigure}[t]{0.48\textwidth}
    \centering
    \includegraphics[width=\linewidth]{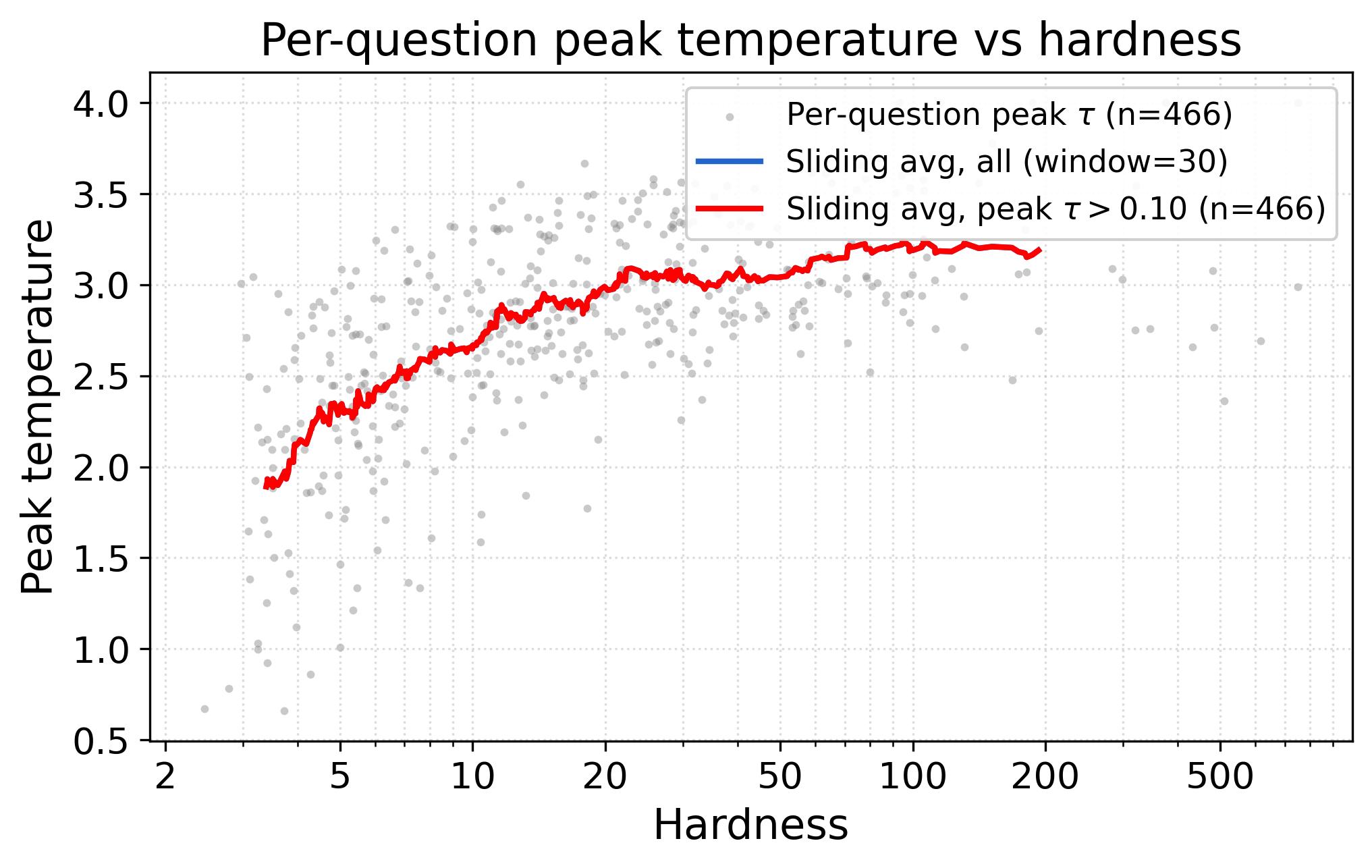}
    \caption{Qwen3-8B, $n = 466$}
    \label{fig:AveTemp:hitom:qwen3_8b}
  \end{subfigure}\hfill  \begin{subfigure}[t]{0.48\textwidth}
    \centering
    \includegraphics[width=\linewidth]{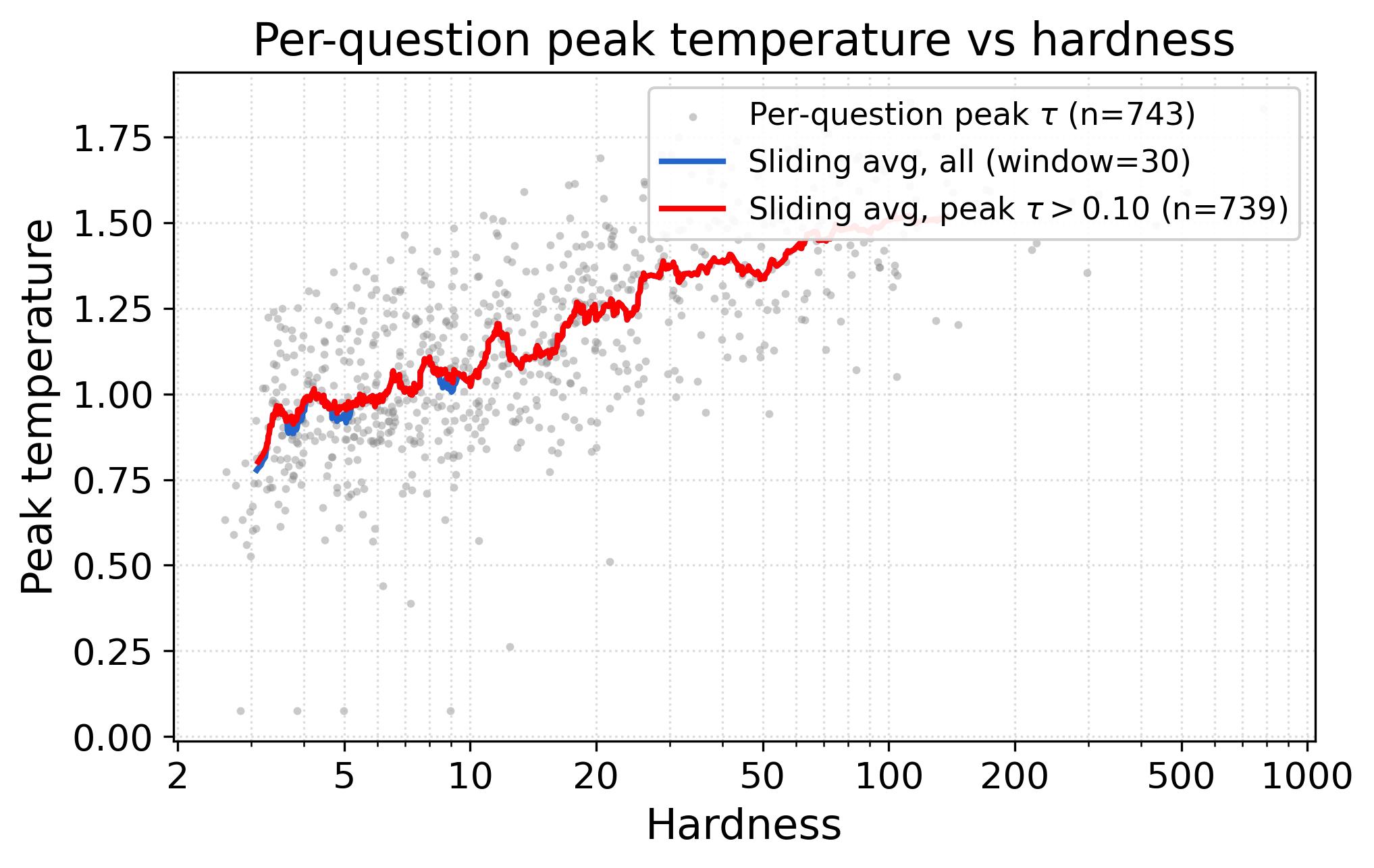}
    \caption{Qwen3-0.6B, $n = 743$}
    \label{fig:AveTemp:hitom:qwen3_0_6b}
  \end{subfigure}
  \caption{Per-question peak temperature vs.~hardness on Hi-ToM, drawn as in Figure \ref{fig:AveTemp:grid}, with the sliding averages over every full window (Appendix \ref{zap:Hard:The}). $n$ is the number of included questions.}
  \label{fig:AveTemp:hitom:grid}
\end{figure}

\begin{figure}[p]
  \centering
  \begin{subfigure}[t]{0.48\textwidth}
    \centering
    \includegraphics[width=\linewidth]{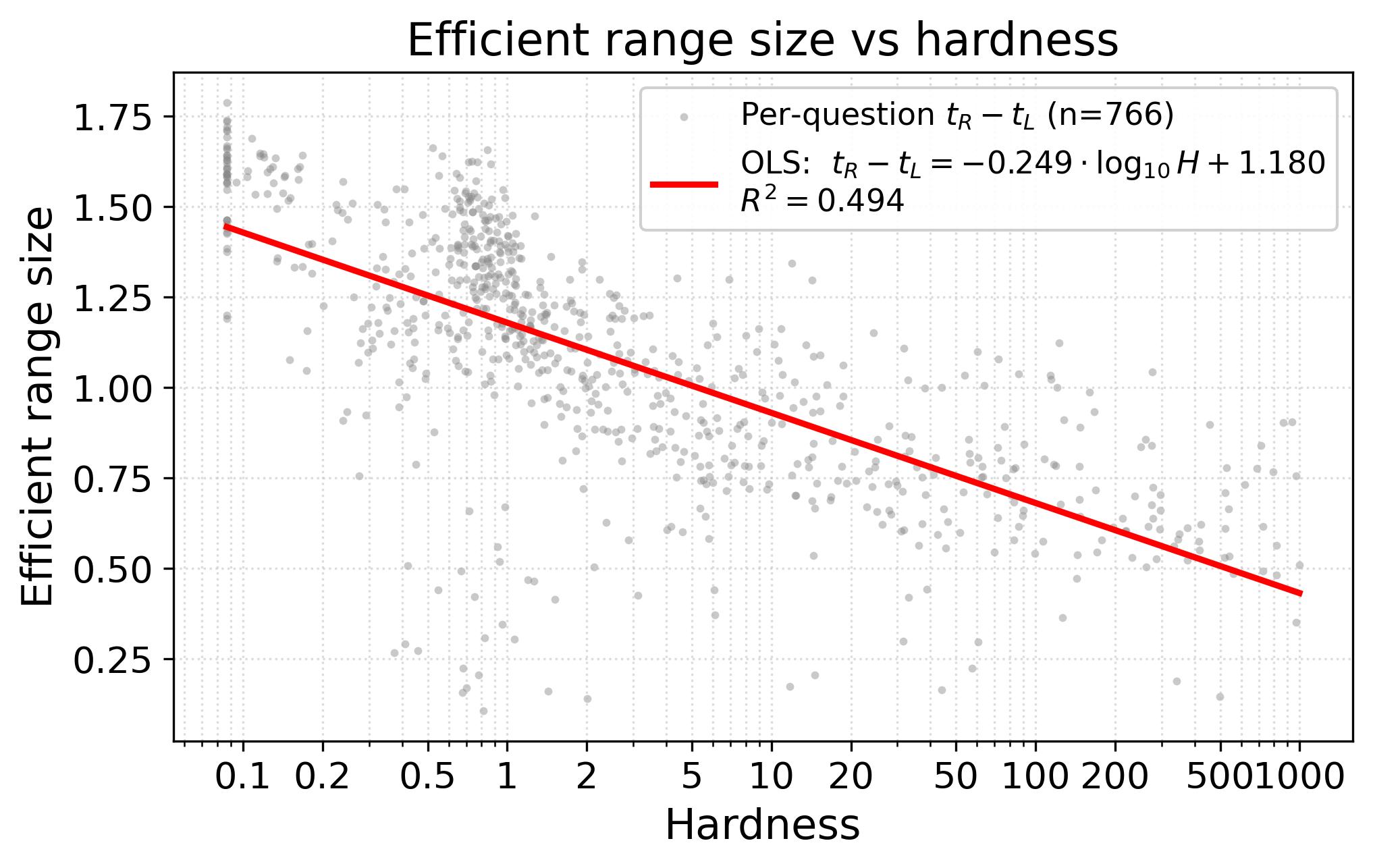}
    \caption{GPT OSS 20B, $n = 766$}
    \label{fig:RangeSize:gpt_oss_20b}
  \end{subfigure}\hfill  \begin{subfigure}[t]{0.48\textwidth}
    \centering
    \includegraphics[width=\linewidth]{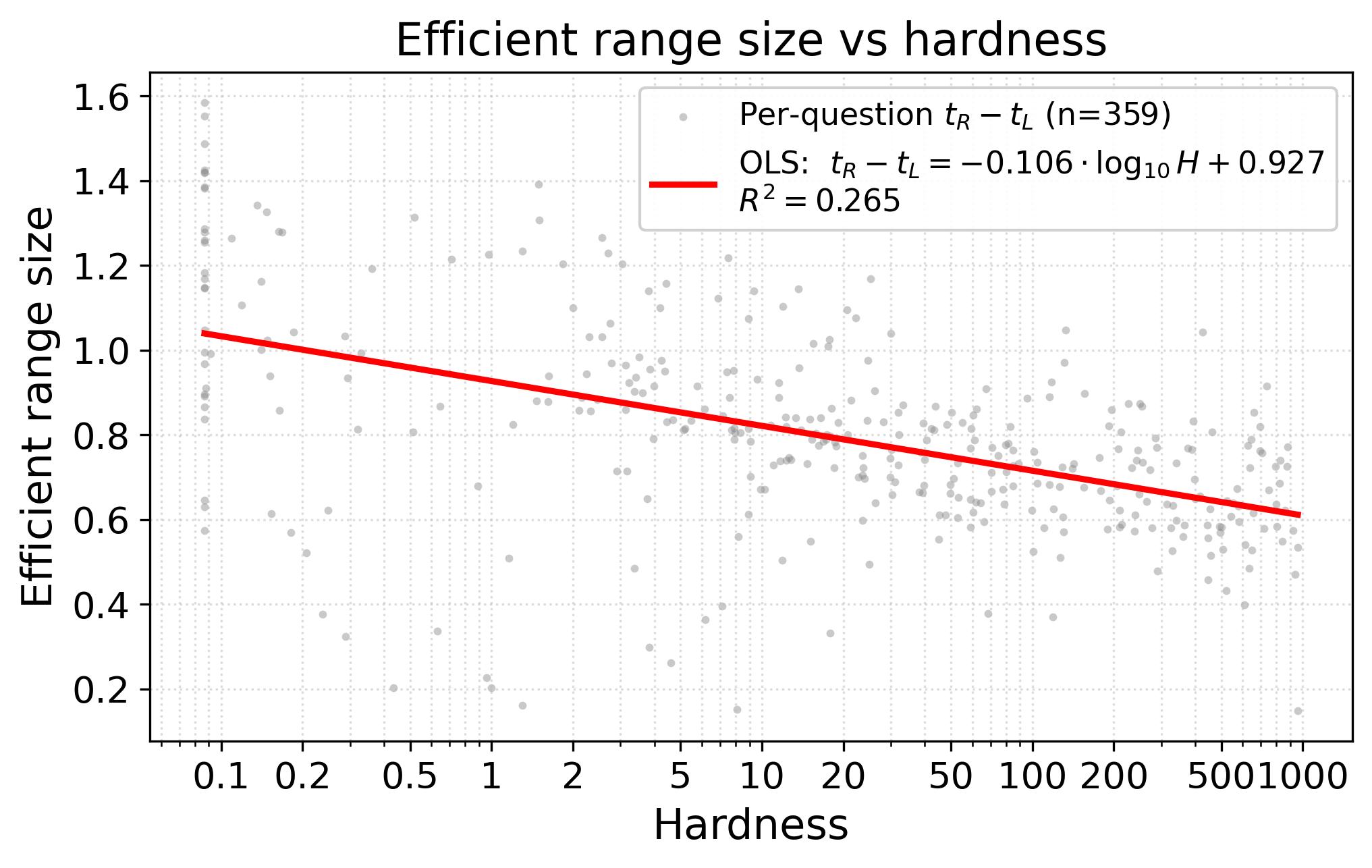}
    \caption{Mistral 7B Instruct, $n = 359$}
    \label{fig:RangeSize:mistral_7b_instruct}
  \end{subfigure}
  \vspace{8pt}

  \begin{subfigure}[t]{0.48\textwidth}
    \centering
    \includegraphics[width=\linewidth]{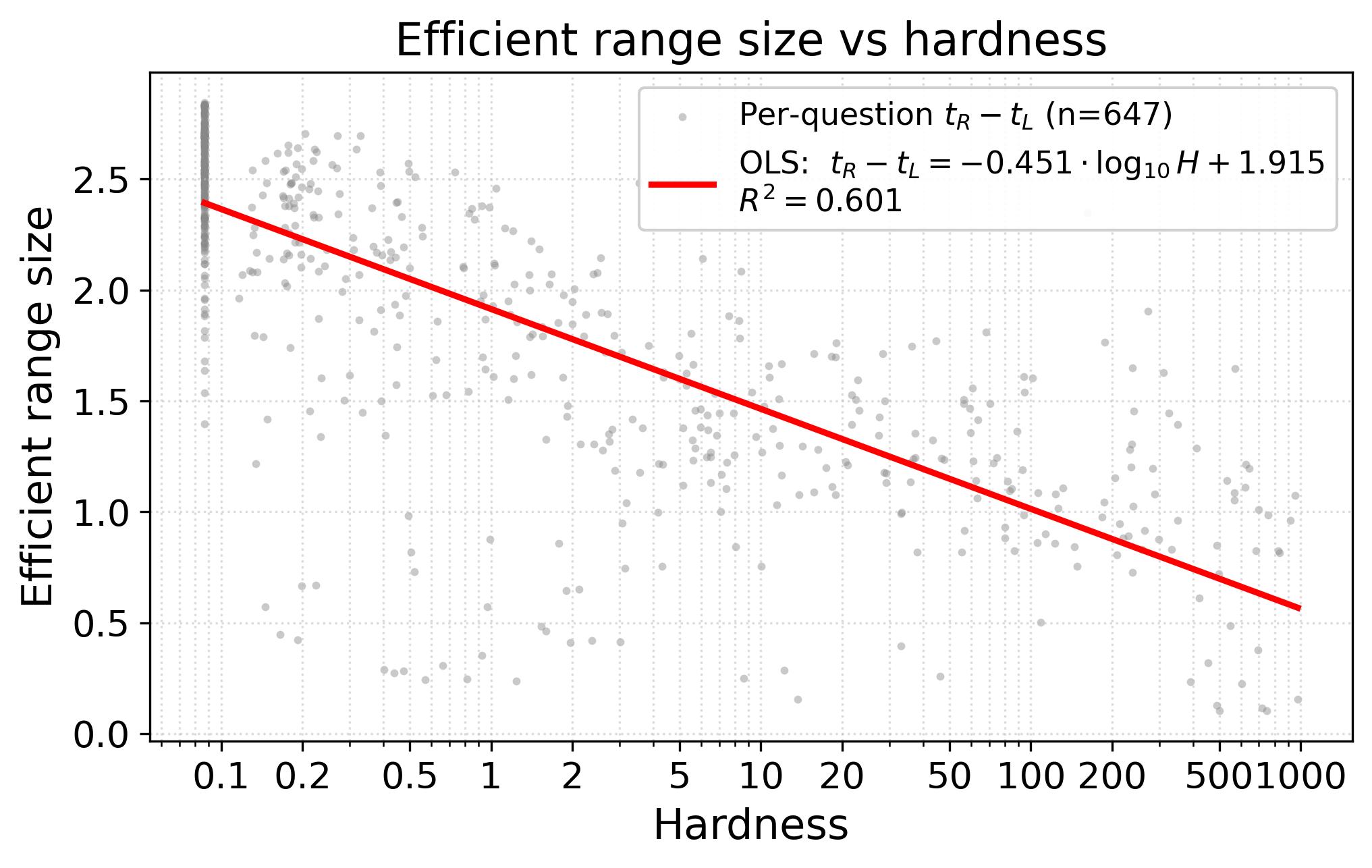}
    \caption{Qwen3-8B, $n = 647$}
    \label{fig:RangeSize:qwen3_8b}
  \end{subfigure}\hfill  \begin{subfigure}[t]{0.48\textwidth}
    \centering
    \includegraphics[width=\linewidth]{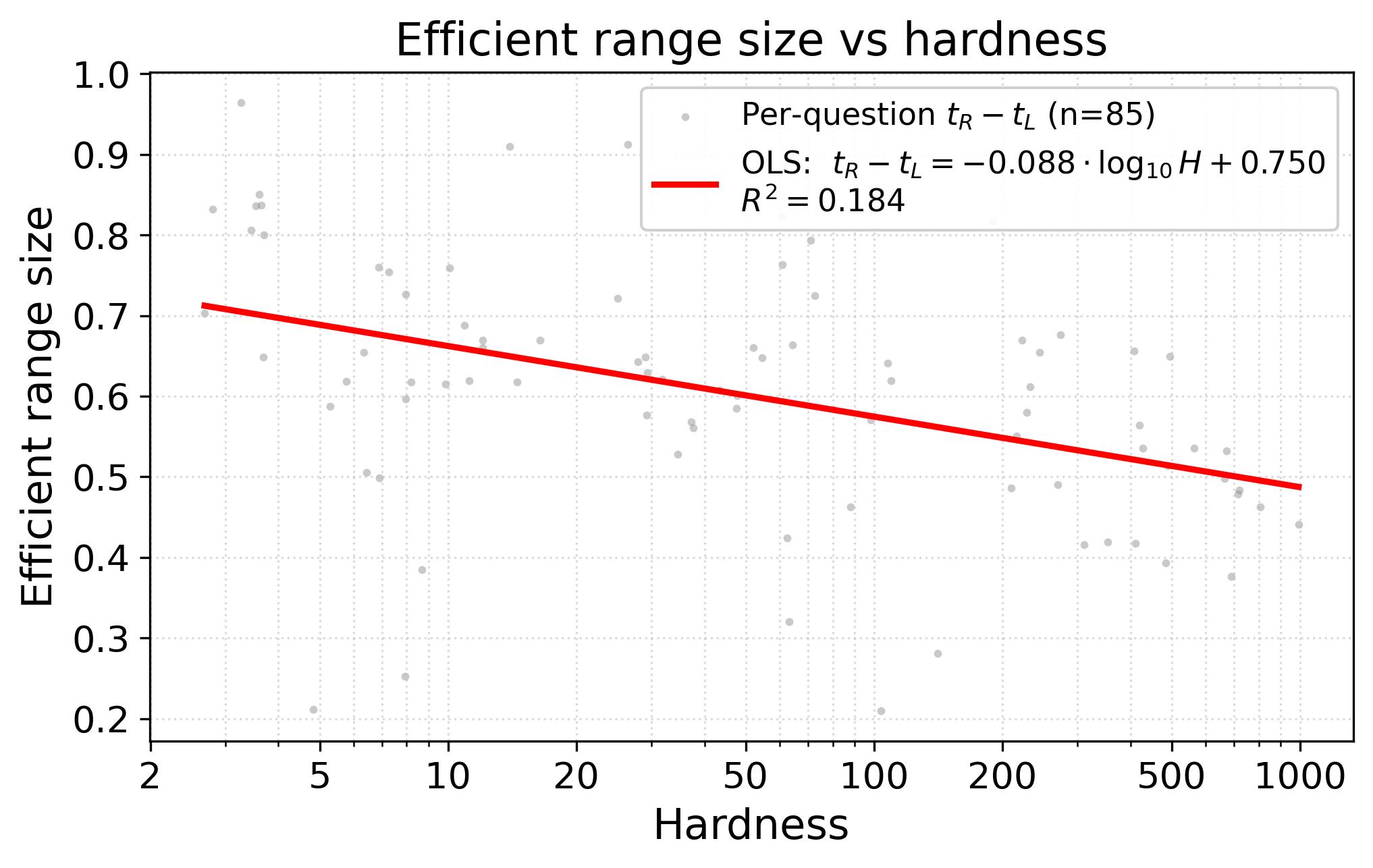}
    \caption{Qwen3-0.6B, $n = 85$}
    \label{fig:RangeSize:qwen3_0_6b}
  \end{subfigure}  \caption{Per-question efficient range size vs.\ hardness on LiveCodeBench, with the OLS fit on $\log_{10} H$ (slope and $R^2$ in each legend), across all four models. $n$ is the number of included questions.}
  \label{fig:RangeSize:grid}
  \par\vspace{10pt}

  \centering
  \begin{subfigure}[t]{0.48\textwidth}
    \centering
    \includegraphics[width=\linewidth]{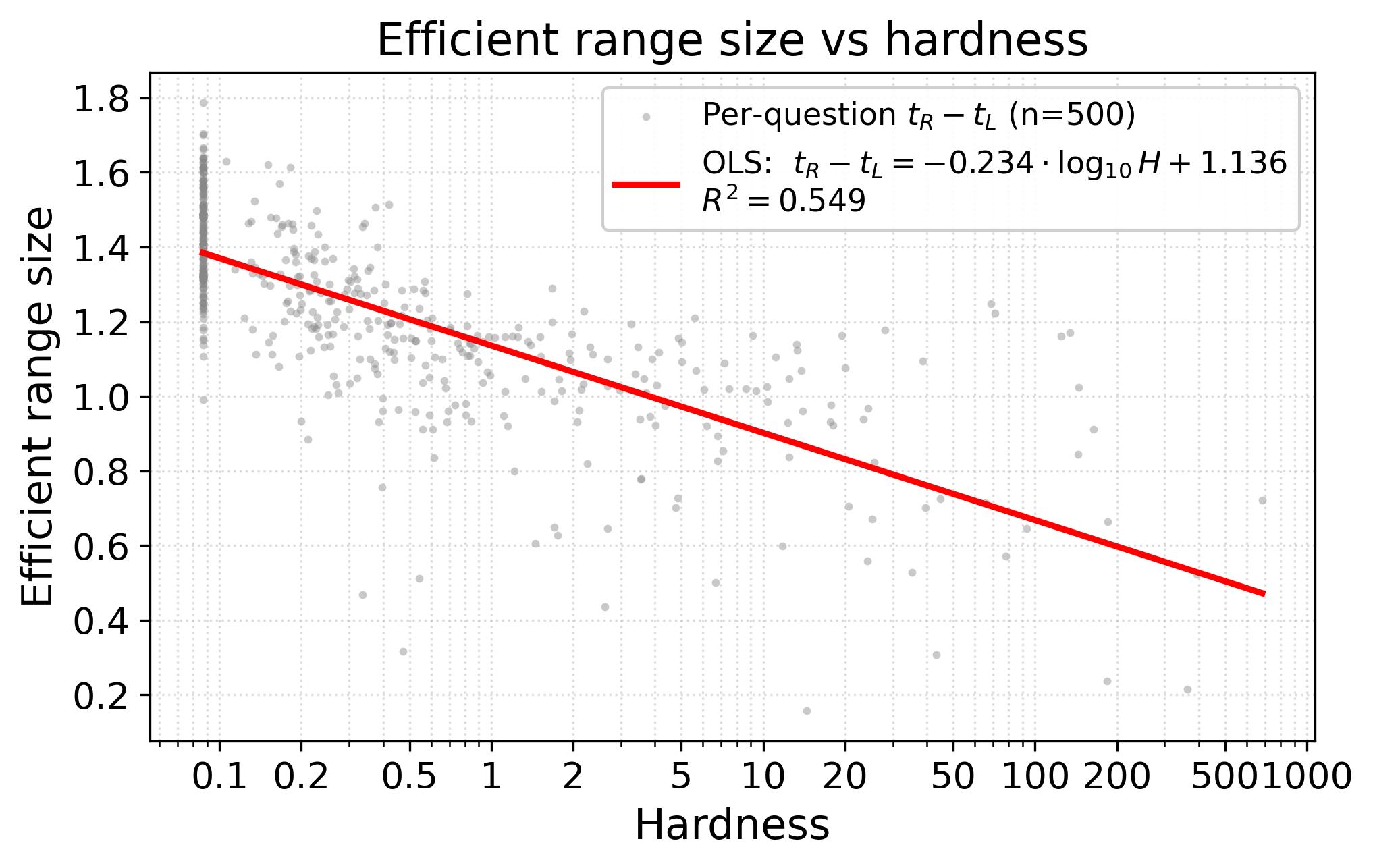}
    \caption{GPT OSS 20B, $n = 500$}
    \label{fig:RangeSize:math500:gpt_oss_20b}
  \end{subfigure}\hfill  \begin{subfigure}[t]{0.48\textwidth}
    \centering
    \includegraphics[width=\linewidth]{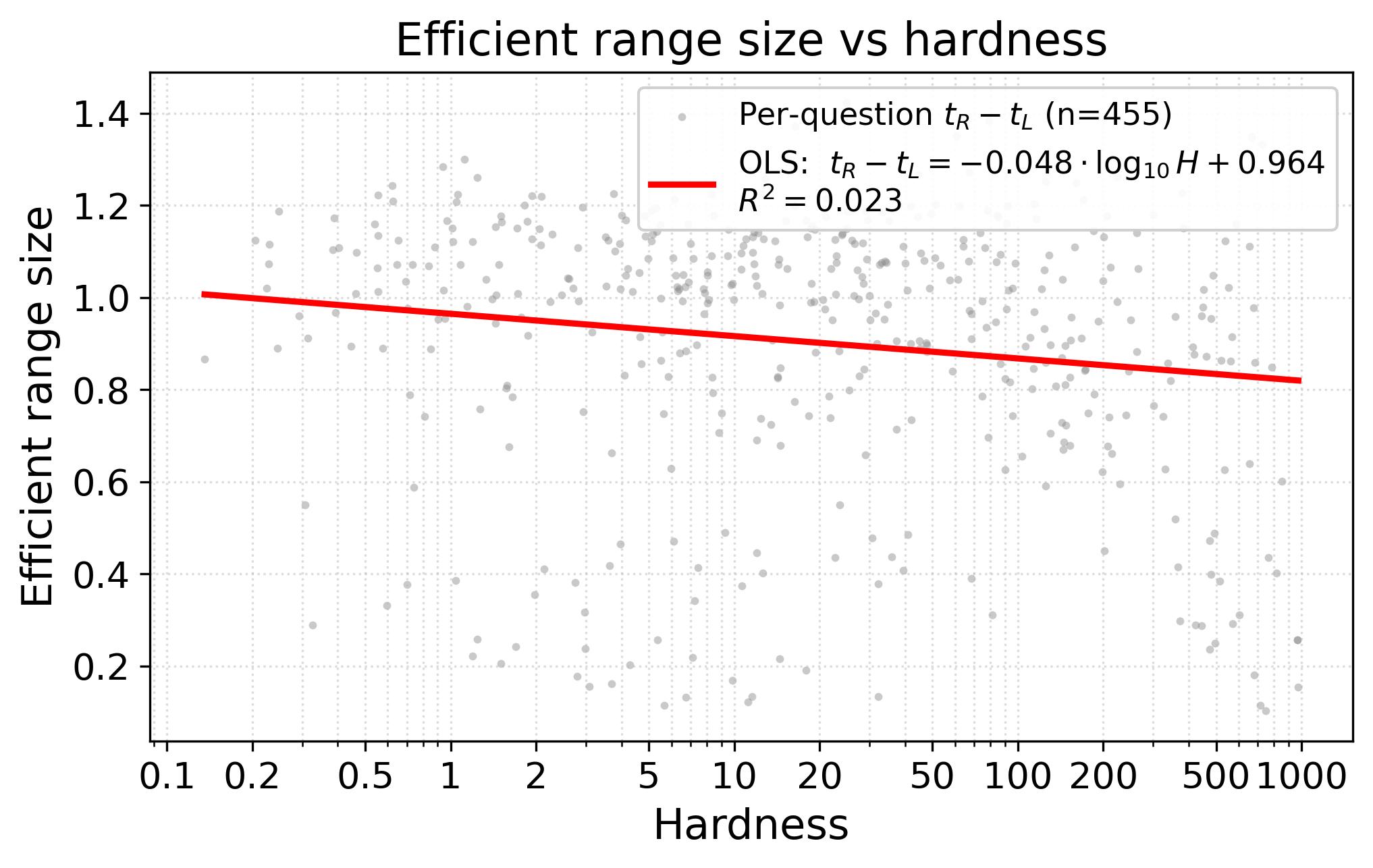}
    \caption{Mistral 7B Instruct, $n = 455$}
    \label{fig:RangeSize:math500:mistral_7b_instruct}
  \end{subfigure}
  \vspace{8pt}

  \begin{subfigure}[t]{0.48\textwidth}
    \centering
    \includegraphics[width=\linewidth]{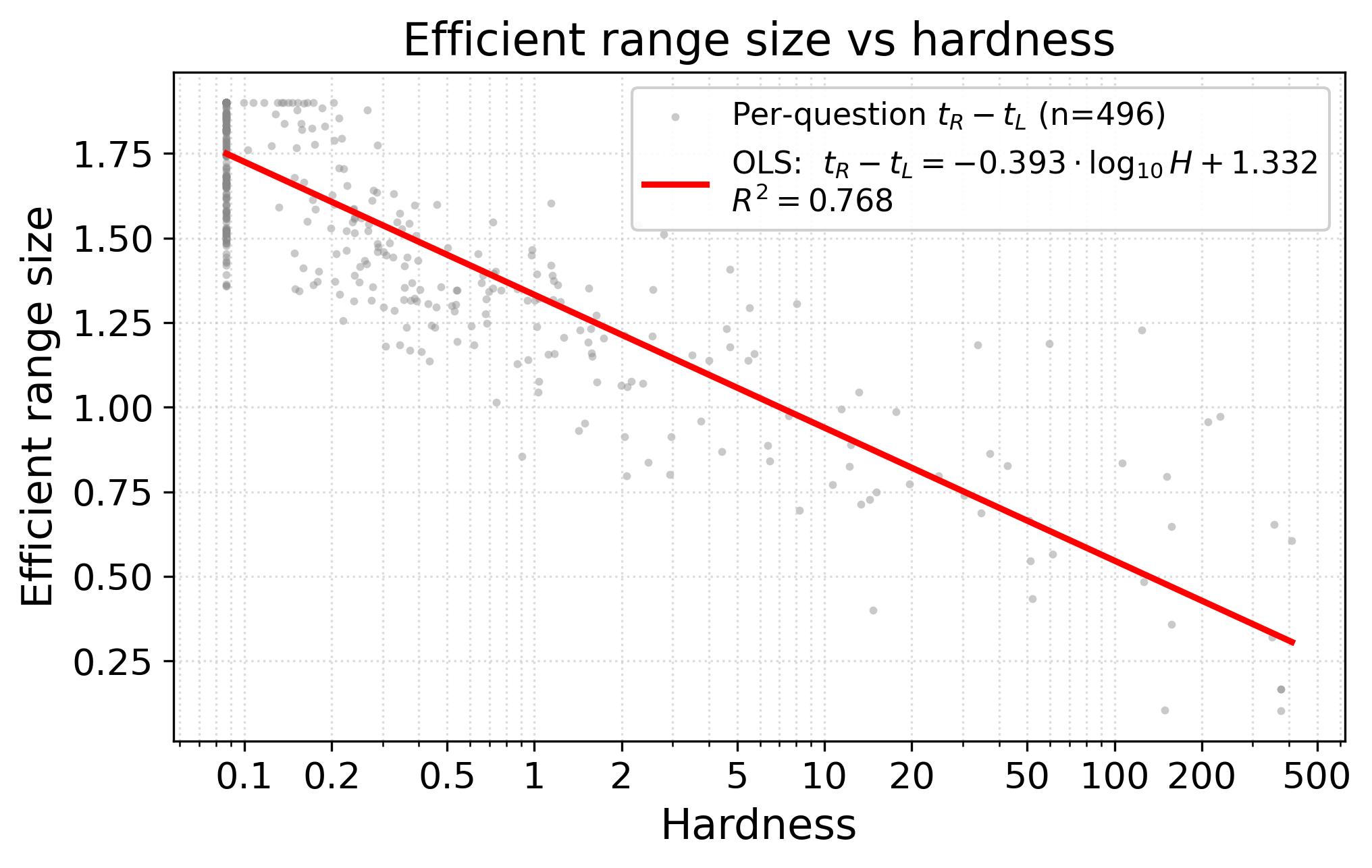}
    \caption{Qwen3-8B, $n = 496$}
    \label{fig:RangeSize:math500:qwen3_8b}
  \end{subfigure}\hfill  \begin{subfigure}[t]{0.48\textwidth}
    \centering
    \includegraphics[width=\linewidth]{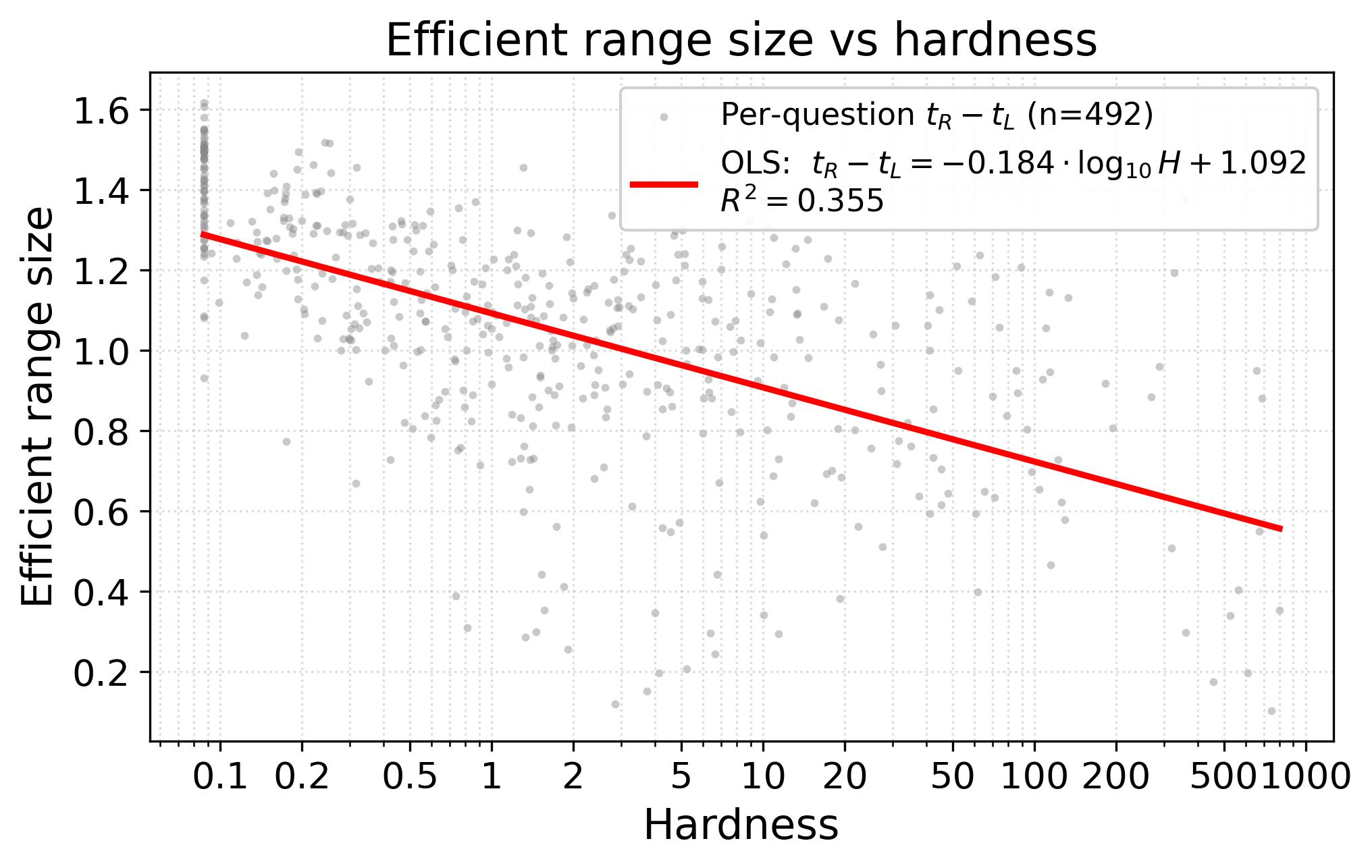}
    \caption{Qwen3-0.6B, $n = 492$}
    \label{fig:RangeSize:math500:qwen3_0_6b}
  \end{subfigure}
  \caption{Per-question efficient range size vs.~hardness on MATH-500, drawn as in Figure \ref{fig:RangeSize:grid}. $n$ is the number of included questions.}
  \label{fig:RangeSize:math500:grid}
\end{figure}

\begin{figure}[p]
  \centering
  \begin{subfigure}[t]{0.48\textwidth}
    \centering
    \includegraphics[width=\linewidth]{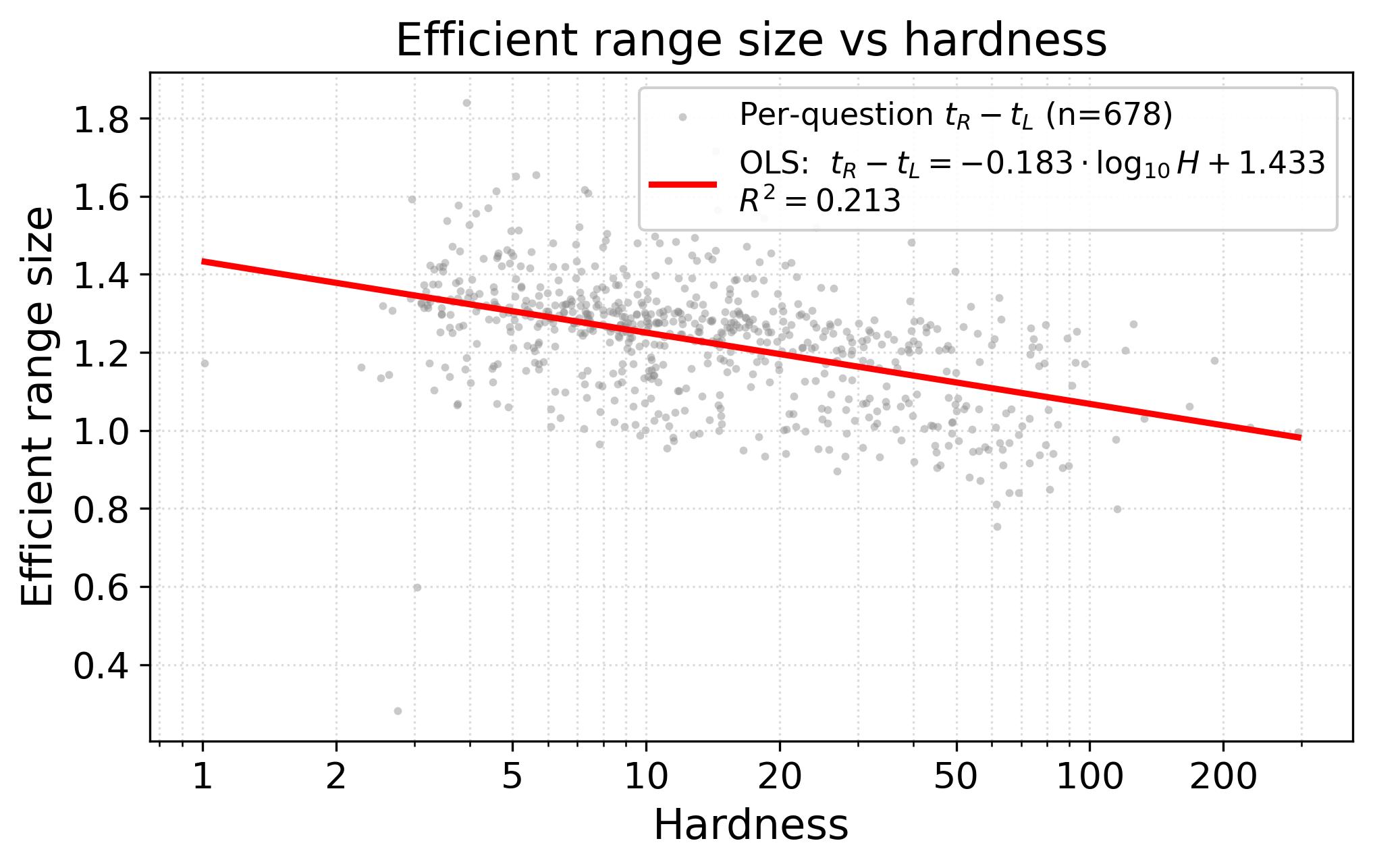}
    \caption{GPT OSS 20B, $n = 678$}
    \label{fig:RangeSize:hitom:gpt_oss_20b}
  \end{subfigure}\hfill  \begin{subfigure}[t]{0.48\textwidth}
    \centering
    \includegraphics[width=\linewidth]{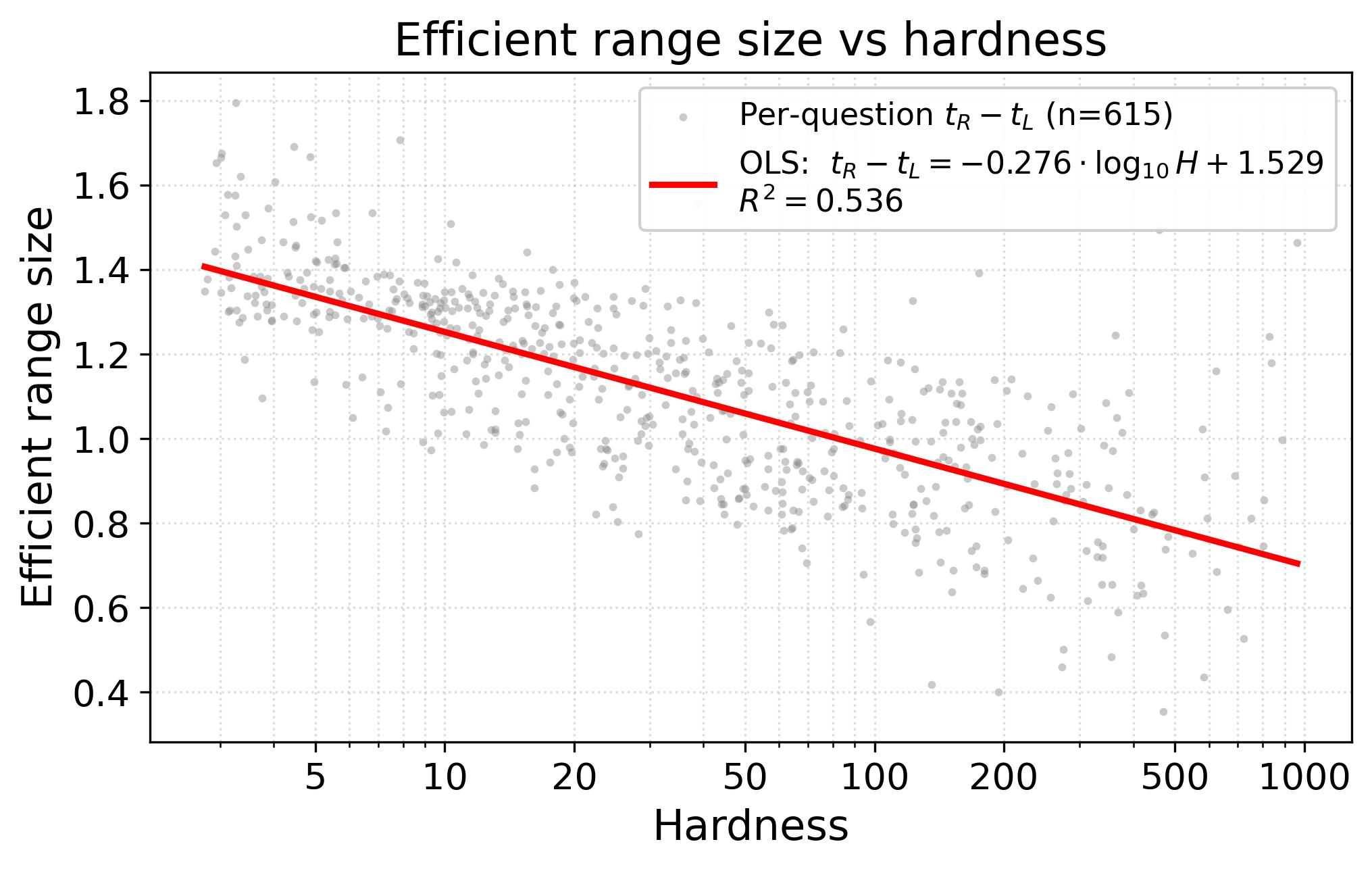}
    \caption{Mistral 7B Instruct, $n = 615$}
    \label{fig:RangeSize:hitom:mistral_7b_instruct}
  \end{subfigure}
  \vspace{8pt}

  \begin{subfigure}[t]{0.48\textwidth}
    \centering
    \includegraphics[width=\linewidth]{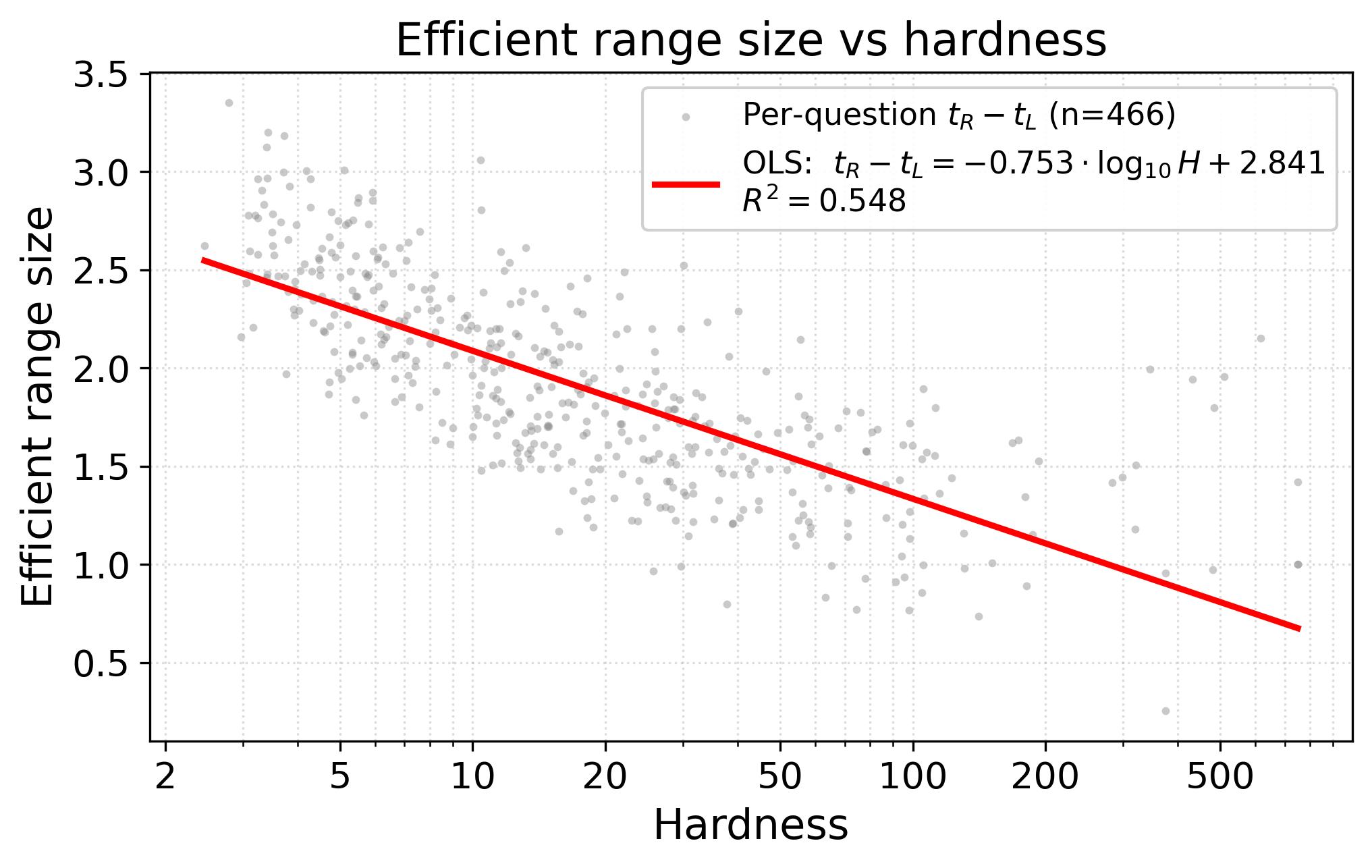}
    \caption{Qwen3-8B, $n = 466$}
    \label{fig:RangeSize:hitom:qwen3_8b}
  \end{subfigure}\hfill  \begin{subfigure}[t]{0.48\textwidth}
    \centering
    \includegraphics[width=\linewidth]{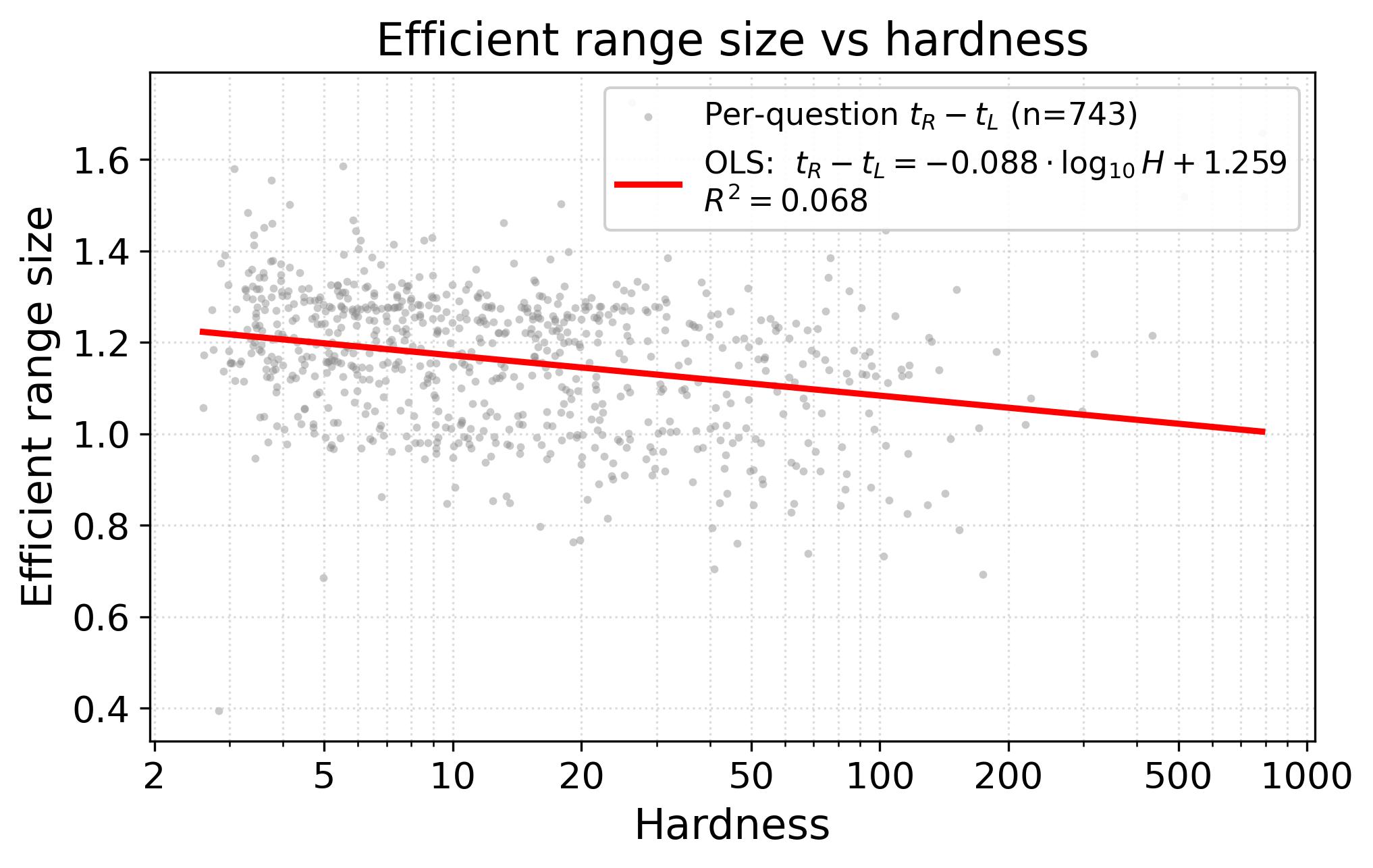}
    \caption{Qwen3-0.6B, $n = 743$}
    \label{fig:RangeSize:hitom:qwen3_0_6b}
  \end{subfigure}
  \caption{Per-question efficient range size vs.~hardness on Hi-ToM, drawn as in Figure \ref{fig:RangeSize:grid}. $n$ is the number of included questions.}
  \label{fig:RangeSize:hitom:grid}
  \par\vspace{10pt}

  \begin{subfigure}[t]{0.48\textwidth}
    \centering
    \includegraphics[width=\linewidth]{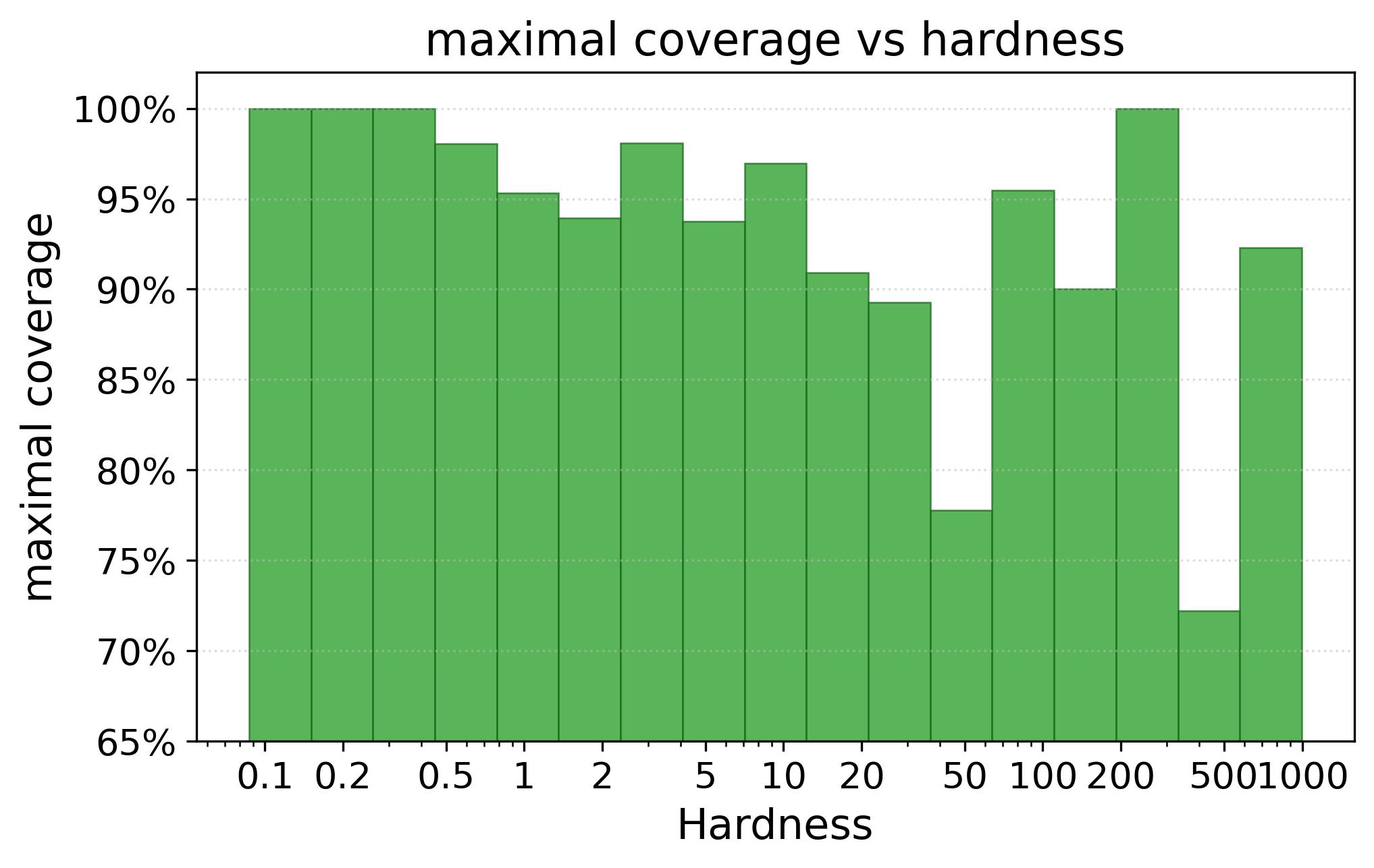}
    \caption{GPT OSS 20B, $n = 766$}
    \label{fig:relC:gpt_oss_20b}
  \end{subfigure}\hfill  \begin{subfigure}[t]{0.48\textwidth}
    \centering
    \includegraphics[width=\linewidth]{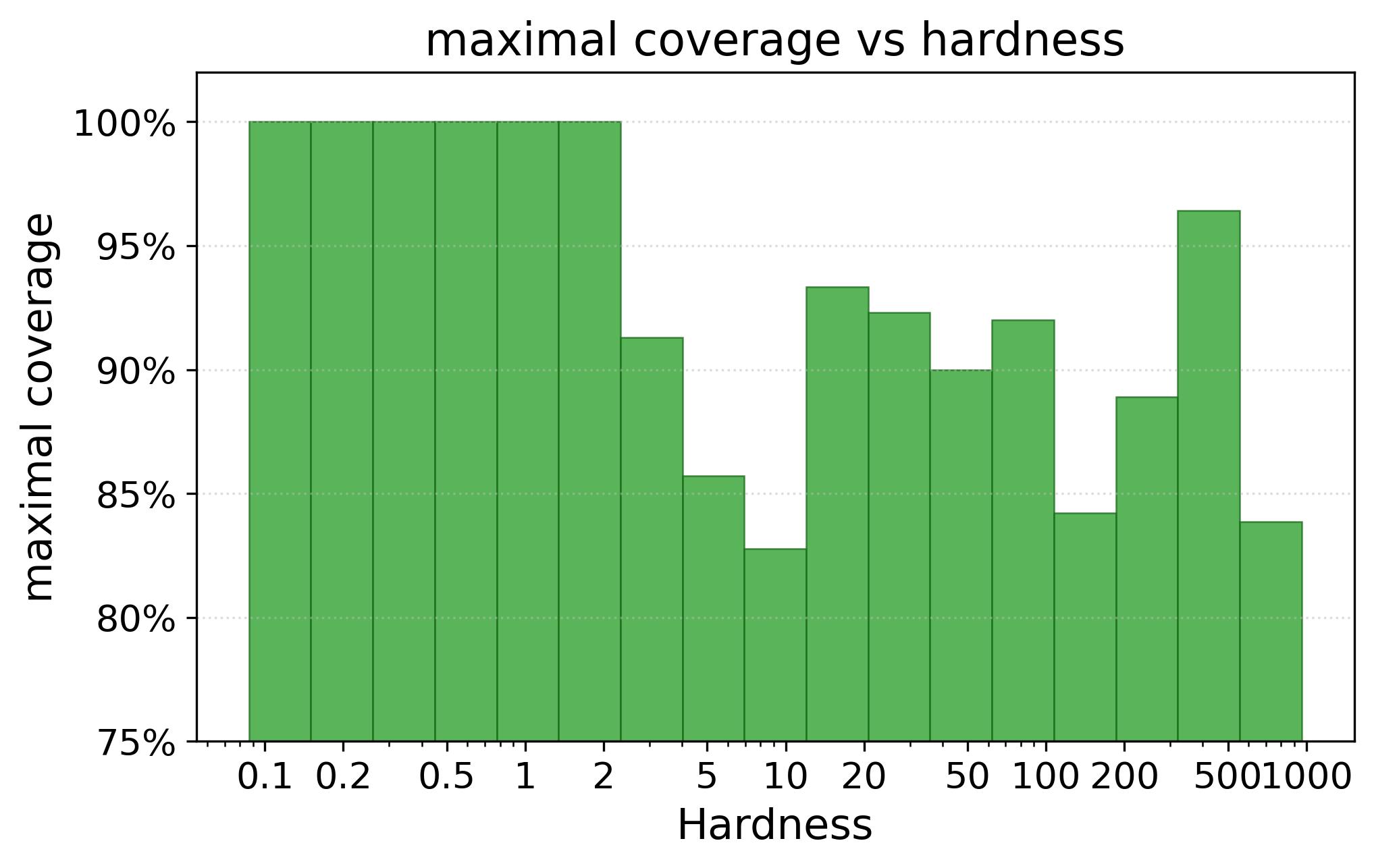}
    \caption{Mistral 7B Instruct, $n = 359$}
    \label{fig:relC:mistral_7b_instruct}
  \end{subfigure}
  \vspace{8pt}

  \begin{subfigure}[t]{0.48\textwidth}
    \centering
    \includegraphics[width=\linewidth]{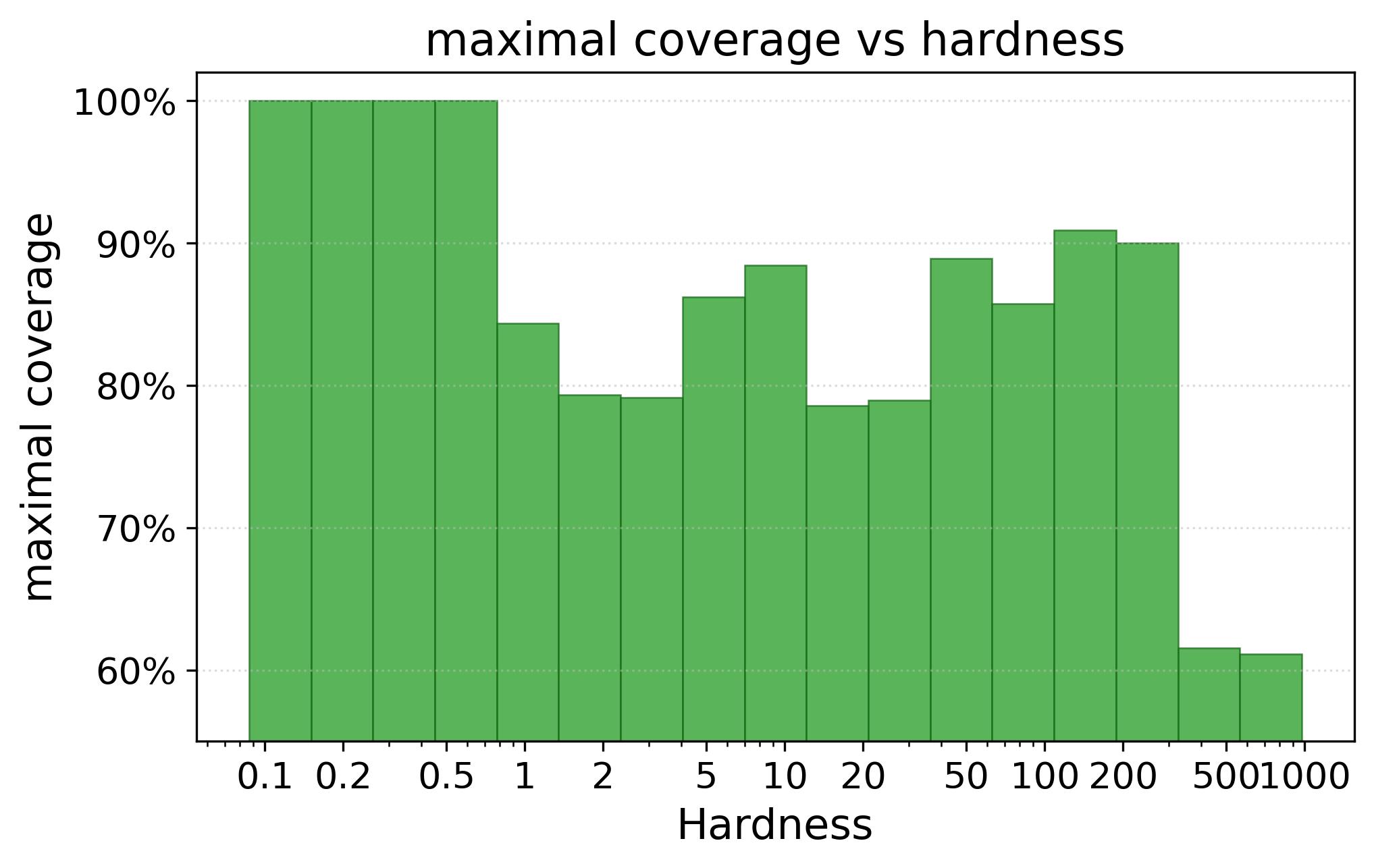}
    \caption{Qwen3-8B, $n = 647$}
    \label{fig:relC:qwen3_8b}
  \end{subfigure}\hfill  \begin{subfigure}[t]{0.48\textwidth}
    \centering
    \includegraphics[width=\linewidth]{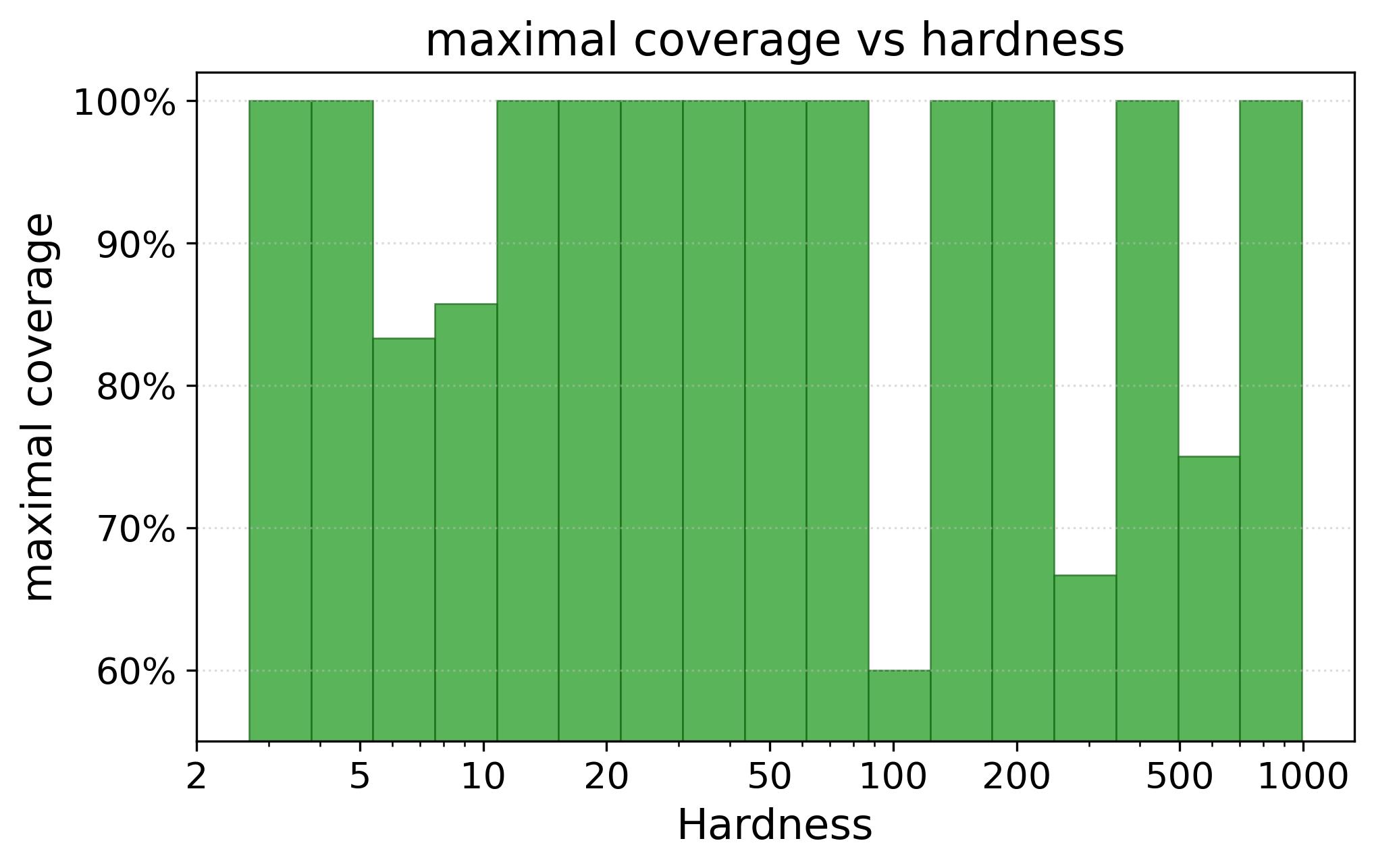}
    \caption{Qwen3-0.6B, $n = 85$}
    \label{fig:relC:qwen3_0_6b}
  \end{subfigure}  \caption{Empirical $R_{\mathcal{Q}}(H)$ from Eq. \ref{eq:SharedEfficientTemperature} on LiveCodeBench for all four models. Questions are placed in $17$ log-uniform hardness bins. $n$ is the number of included questions.}
  \label{fig:relC:grid}
\end{figure}

\begin{figure}[p]
  \centering
  \begin{subfigure}[t]{0.48\textwidth}
    \centering
    \includegraphics[width=\linewidth]{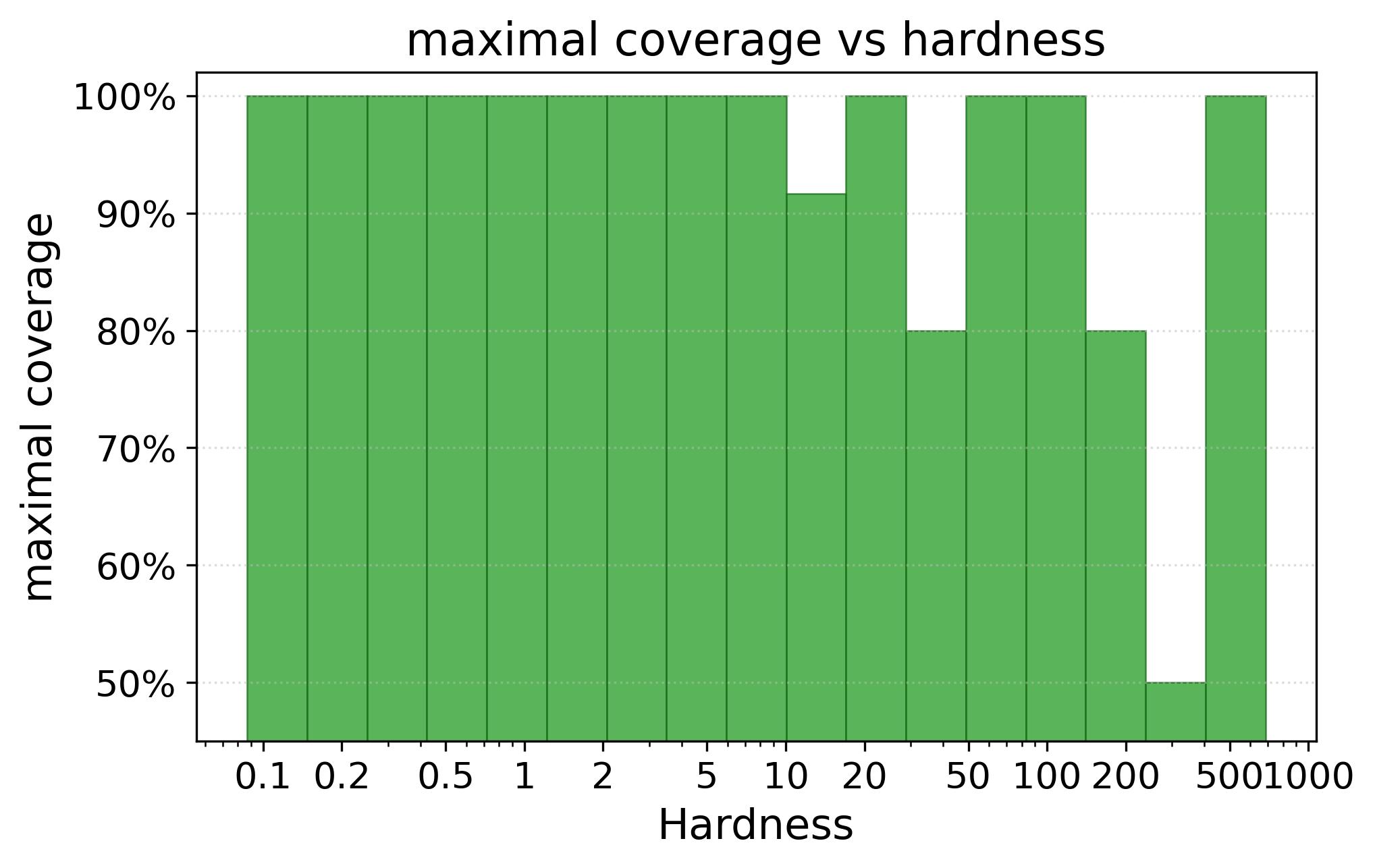}
    \caption{GPT OSS 20B, $n = 500$}
    \label{fig:relC:math500:gpt_oss_20b}
  \end{subfigure}\hfill  \begin{subfigure}[t]{0.48\textwidth}
    \centering
    \includegraphics[width=\linewidth]{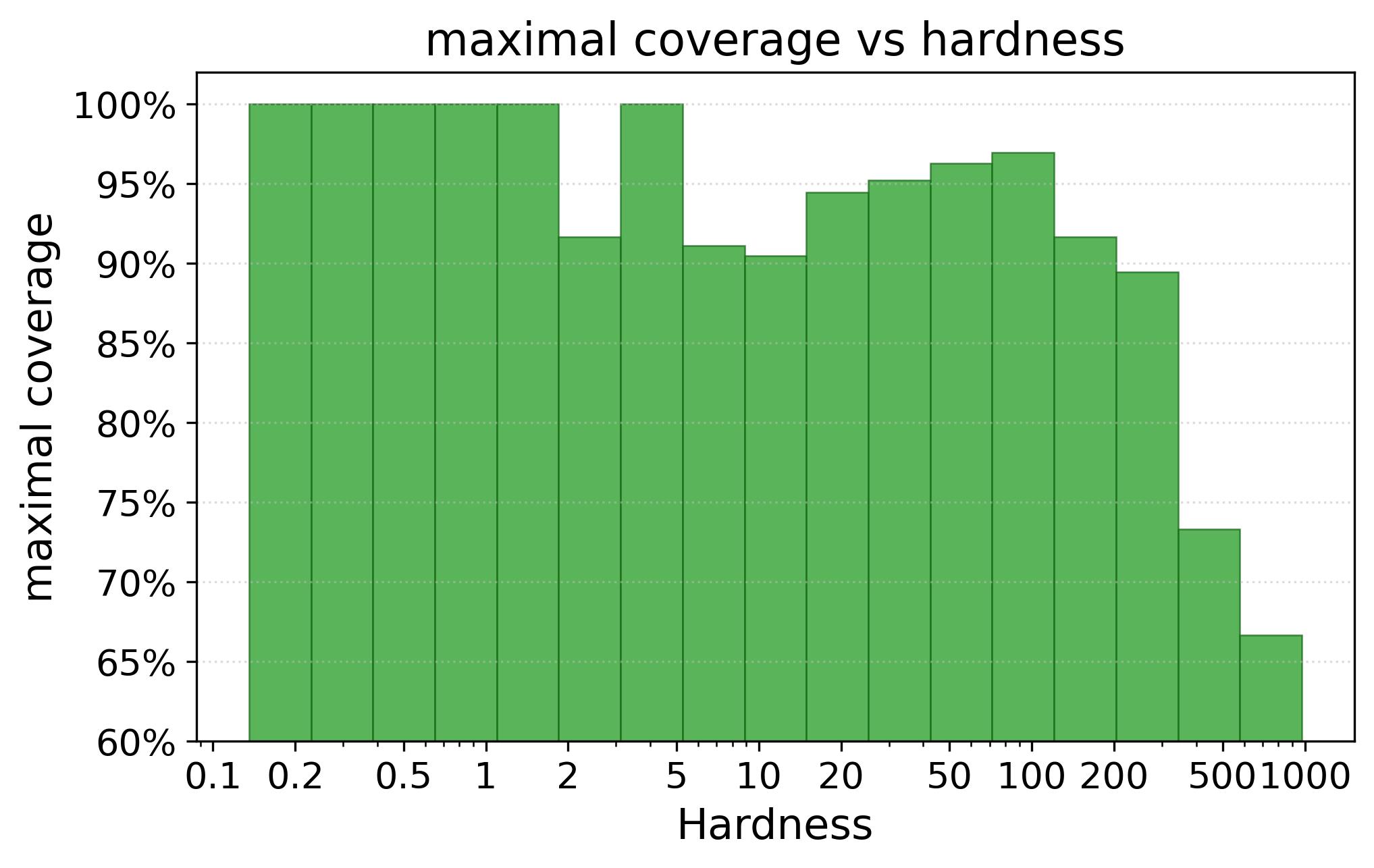}
    \caption{Mistral 7B Instruct, $n = 455$}
    \label{fig:relC:math500:mistral_7b_instruct}
  \end{subfigure}
  \vspace{8pt}

  \begin{subfigure}[t]{0.48\textwidth}
    \centering
    \includegraphics[width=\linewidth]{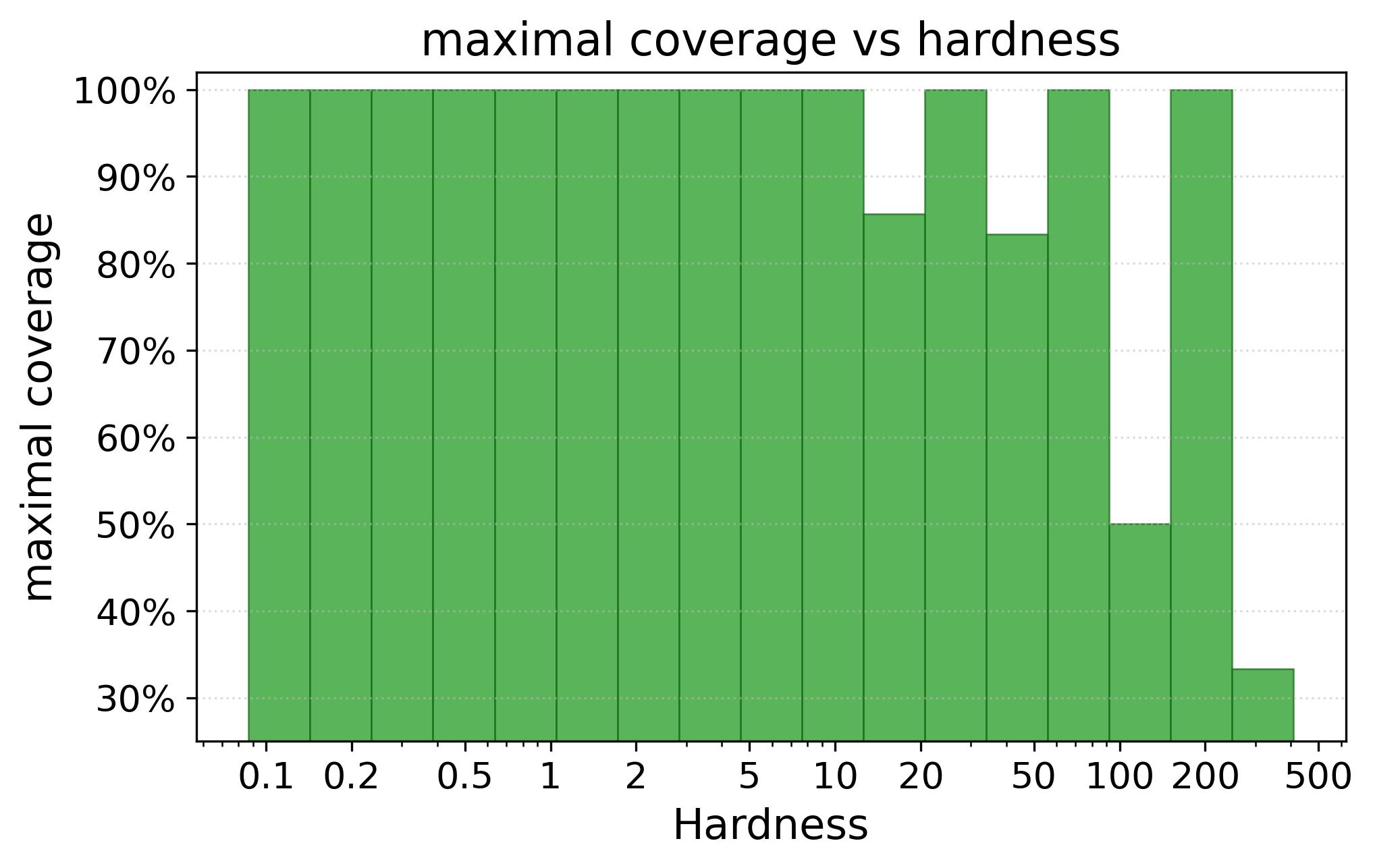}
    \caption{Qwen3-8B, $n = 496$}
    \label{fig:relC:math500:qwen3_8b}
  \end{subfigure}\hfill  \begin{subfigure}[t]{0.48\textwidth}
    \centering
    \includegraphics[width=\linewidth]{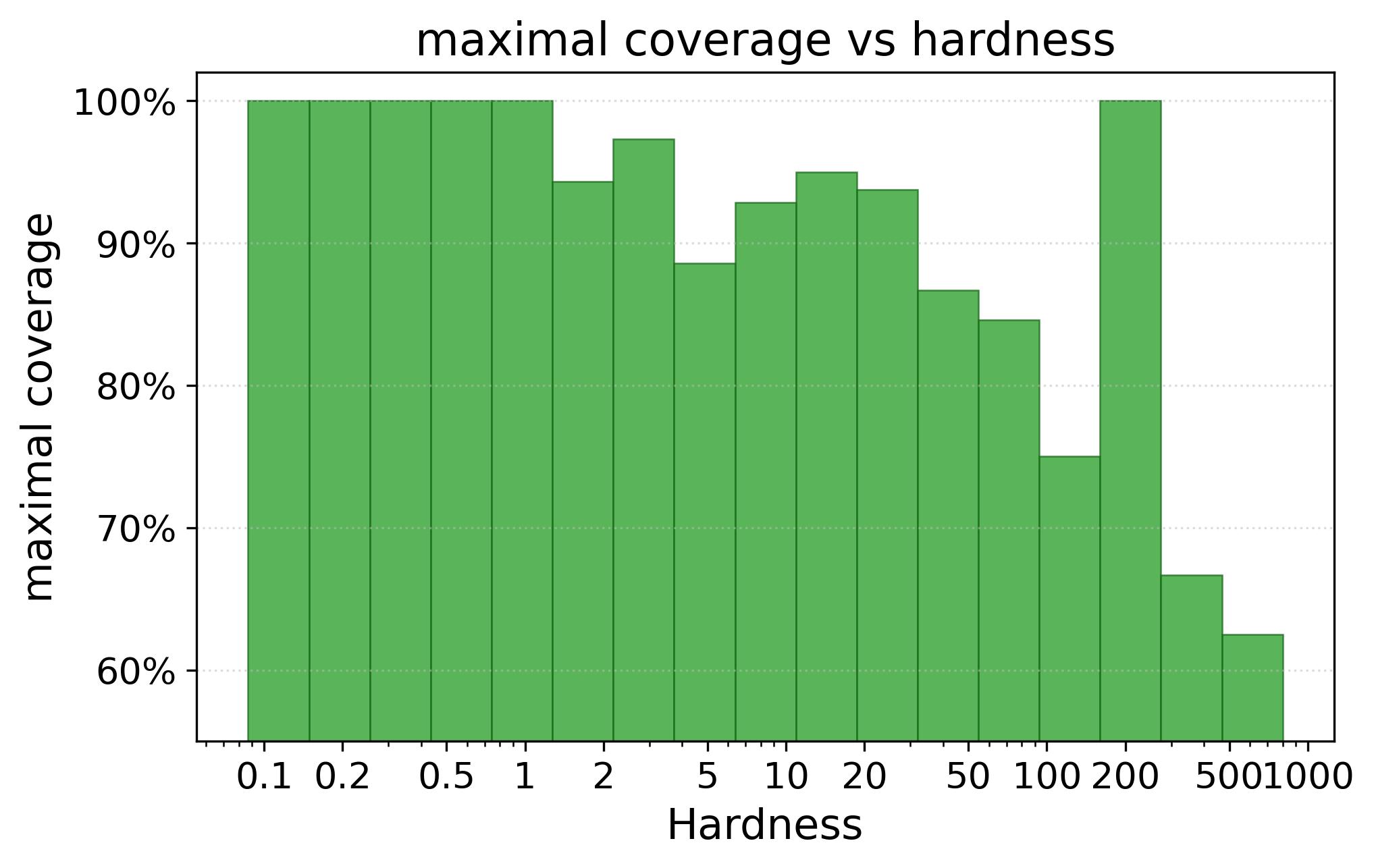}
    \caption{Qwen3-0.6B, $n = 492$}
    \label{fig:relC:math500:qwen3_0_6b}
  \end{subfigure}
  \caption{Empirical $R_{\mathcal{Q}}(H)$ on MATH-500, drawn as in Figure \ref{fig:relC:grid}. $n$ is the number of included questions.}
  \label{fig:relC:math500:grid}
  \par\vspace{10pt}

  \centering
  \begin{subfigure}[t]{0.48\textwidth}
    \centering
    \includegraphics[width=\linewidth]{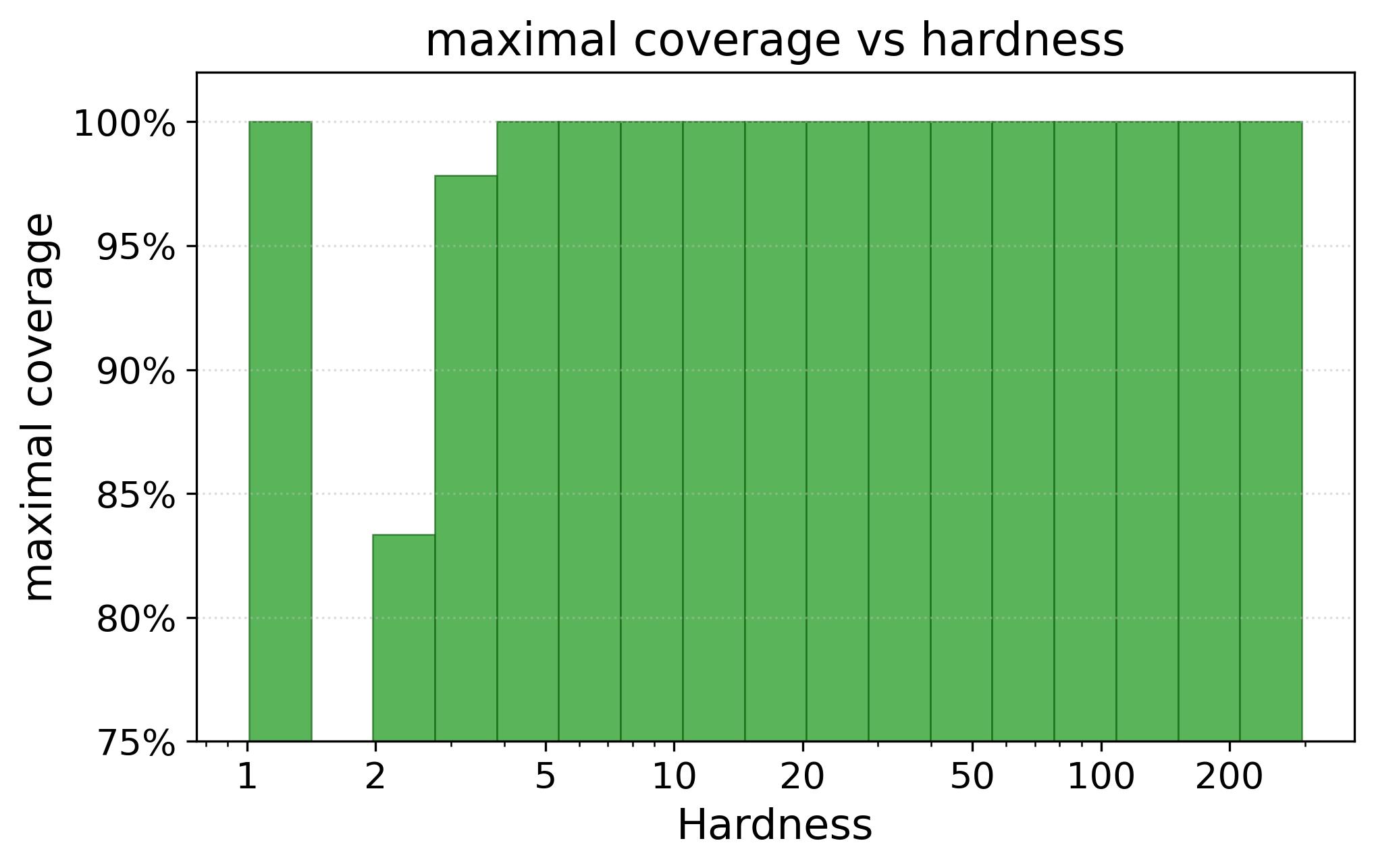}
    \caption{GPT OSS 20B, $n = 678$}
    \label{fig:relC:hitom:gpt_oss_20b}
  \end{subfigure}\hfill  \begin{subfigure}[t]{0.48\textwidth}
    \centering
    \includegraphics[width=\linewidth]{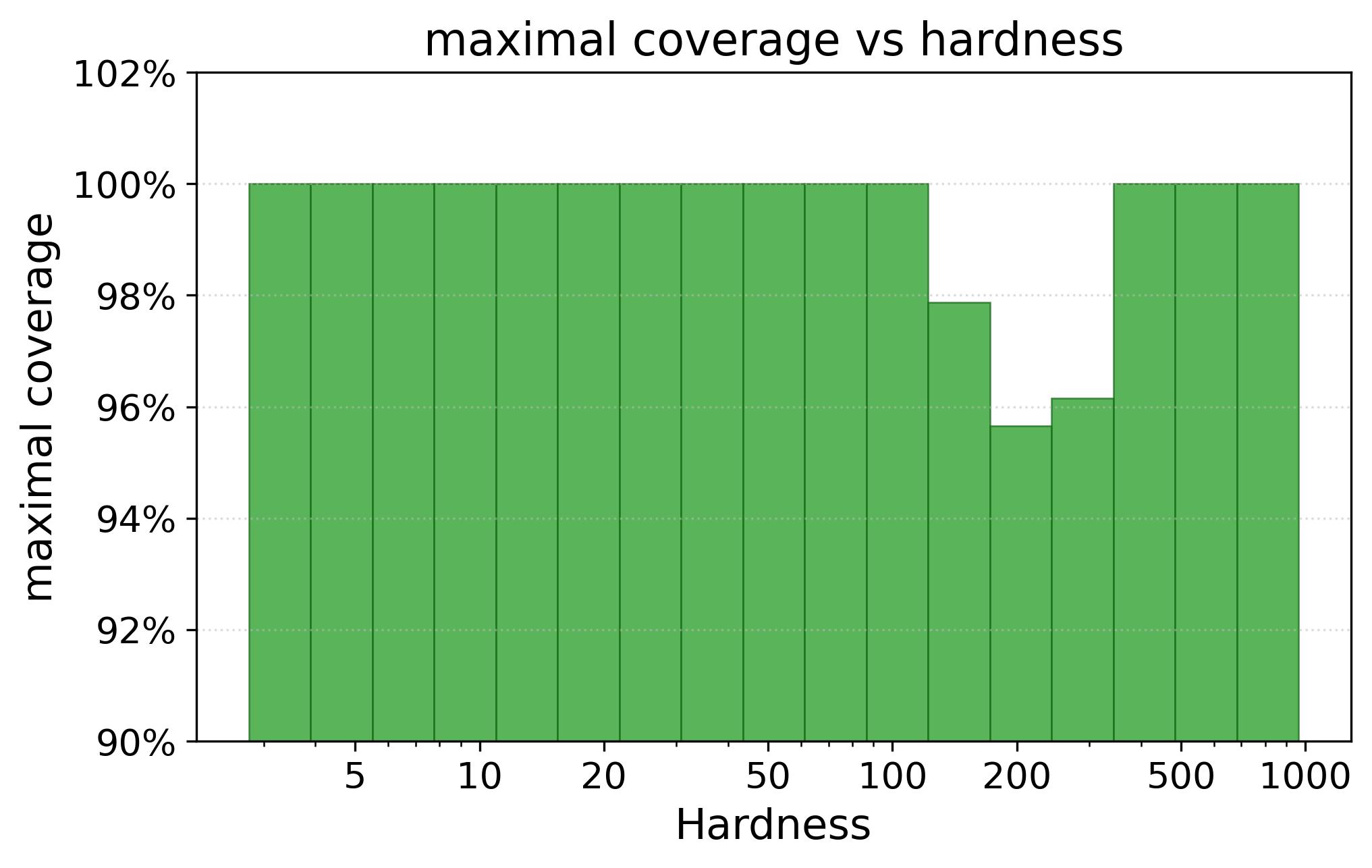}
    \caption{Mistral 7B Instruct, $n = 615$}
    \label{fig:relC:hitom:mistral_7b_instruct}
  \end{subfigure}
  \vspace{8pt}

  \begin{subfigure}[t]{0.48\textwidth}
    \centering
    \includegraphics[width=\linewidth]{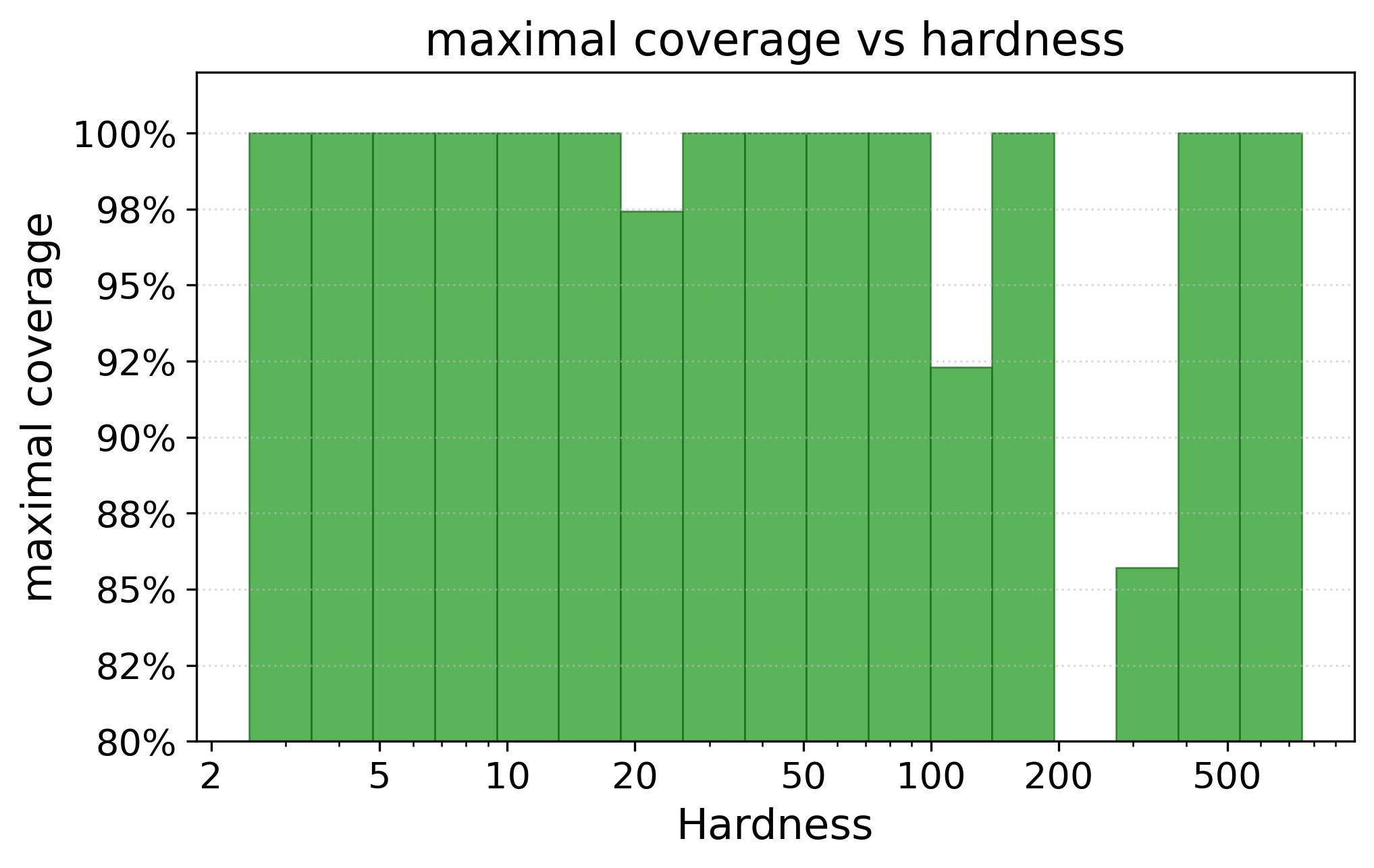}
    \caption{Qwen3-8B, $n = 466$}
    \label{fig:relC:hitom:qwen3_8b}
  \end{subfigure}\hfill  \begin{subfigure}[t]{0.48\textwidth}
    \centering
    \includegraphics[width=\linewidth]{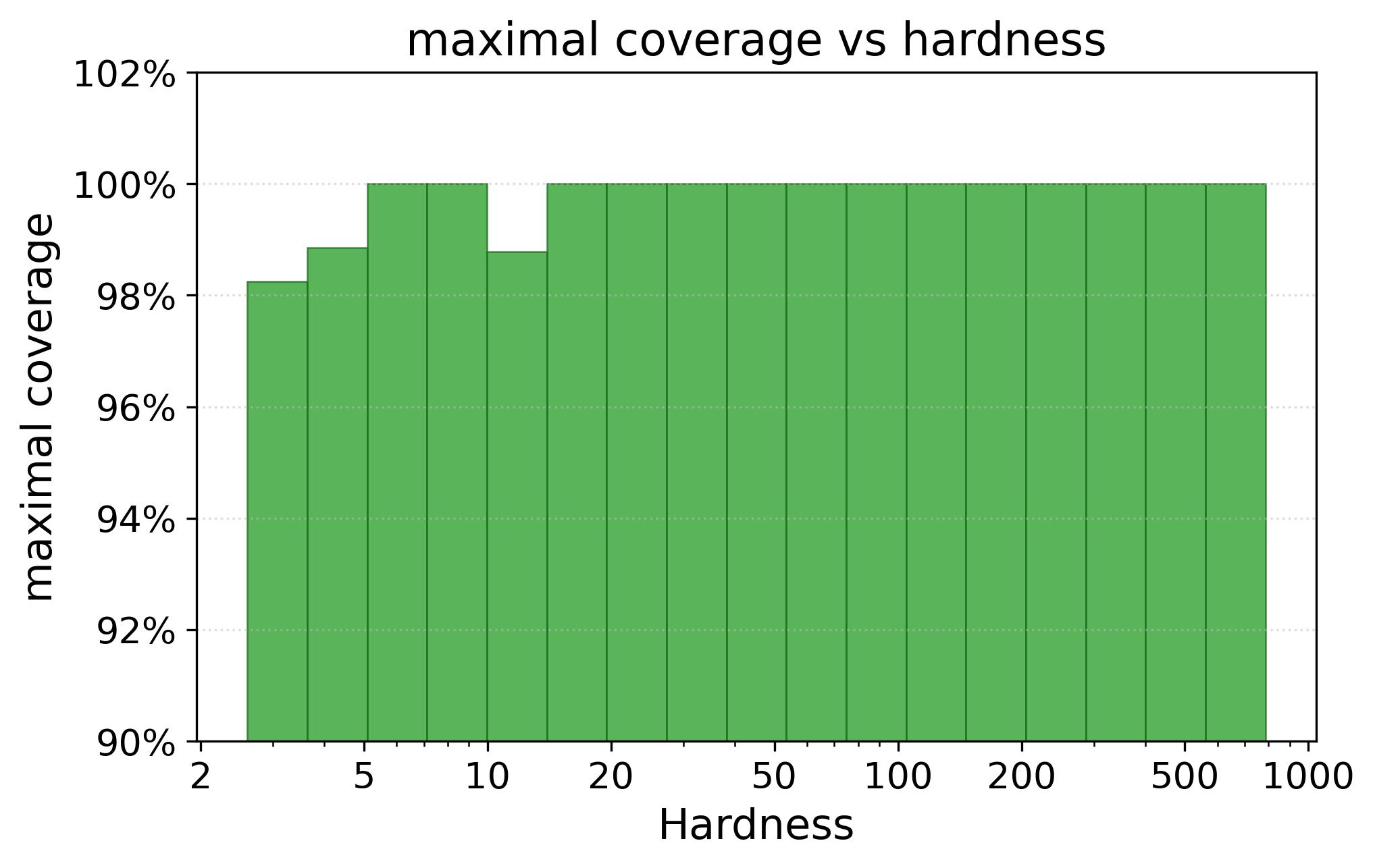}
    \caption{Qwen3-0.6B, $n = 743$}
    \label{fig:relC:hitom:qwen3_0_6b}
  \end{subfigure}
  \caption{Empirical $R_{\mathcal{Q}}(H)$ on Hi-ToM, drawn as in Figure \ref{fig:relC:grid}. $n$ is the number of included questions. A bin with no question has no bar, so the GPT OSS 20B and Qwen3-8B panels have $16$ bars.}
  \label{fig:relC:hitom:grid}
\end{figure}

Here, we present all results from the analysis of the questions in Section~\ref{sec:Profiles}, as explained in Appendix \ref{zap:Hard:The}.

On LiveCodeBench, across all four models, the peak temperature trends rise gently with hardness once near-zero peaks are excluded (Figure \ref{fig:AveTemp:grid}); the efficient temperature range shrinks roughly log-linearly with $H$ with a clearly negative OLS slope for every model (Figure \ref{fig:RangeSize:grid}); and the maximum-coverage bars correspondingly fall from near $1$ at low $H$ toward $0$ at high $H$ (Figure \ref{fig:relC:grid}), so harder questions become both narrower and harder to cover with a single shared temperature. The intermediate-$H$ deviations in Figure \ref{fig:relC:grid} are consistent with the near-zero-$\tau_q^\star$ easy cluster discussed in Section \ref{sec:QSep}.

On MATH-500, the red sliding average of the peak temperature ends higher than it starts for GPT OSS 20B ($0.74$ to $0.90$), Mistral 7B Instruct ($0.49$ to $0.70$), and Qwen3-0.6B ($0.61$ to $0.85$), and lower for Qwen3-8B ($1.14$ to $0.99$) (Figure \ref{fig:AveTemp:math500:grid}); the OLS slope of the efficient range size is negative for every model, from $-0.393$ ($R^2 = 0.768$) for Qwen3-8B to $-0.048$ ($R^2 = 0.023$) for Mistral 7B Instruct (Figure \ref{fig:RangeSize:math500:grid}); and the maximum-coverage bars start at $1$ and reach their lowest value in one of the two highest-$H$ bins, $0.500$, $0.667$, $0.333$, and $0.625$ in the order of the panels (Figure \ref{fig:relC:math500:grid}).

On Hi-ToM, the red sliding average of the peak temperature rises with hardness for every model, from $0.93$ to $1.50$ for GPT OSS 20B, $1.17$ to $1.85$ for Mistral 7B Instruct, $1.89$ to $3.19$ for Qwen3-8B, and $0.80$ to $1.52$ for Qwen3-0.6B (Figure \ref{fig:AveTemp:hitom:grid}); the OLS slope of the efficient range size is negative for every model, from $-0.753$ ($R^2 = 0.548$) for Qwen3-8B to $-0.088$ ($R^2 = 0.068$) for Qwen3-0.6B (Figure \ref{fig:RangeSize:hitom:grid}); and every maximum-coverage bar lies between $0.83$ and $1$ (Figure \ref{fig:relC:hitom:grid}).

\FloatBarrier
\section{Temperature Sampling Schemes Comparisons}
\label{zap:Multiple:Comp}

In this section, we explain how we compared the different temperature-sampling schemes. We begin by describing the analysis procedure and then present the results.

\subsection{\texorpdfstring{Statistical Analysis Approach}{Statistical Analysis Approach}}

\label{zap:Multiple:Comp:The}

\textbf{Per-temperature pass@$K$ score}

Once each profile $P_{q_i}(\tau)$ has been fitted, we evaluate sampling strategies via the pass@$K$ metric: the expected fraction of questions solved when each question receives $K$ independent attempts under a given allocation across temperatures. For a single temperature $\tau_j$ used for all $K$ attempts, the probability that question $q_i$ is solved at least once is $1 - \bar{P}_{q_i}(\tau_j)^{K}$, where $\bar{P}_{q_i}(\tau) = 1 - P_{q_i}(\tau)$ is the single-attempt failure probability. The dataset-level score at temperature $\tau_j$ is
\begin{equation}
\mathcal{P}_{K}(\tau_j) \;=\; 1 - \bar{\mathcal{P}}_{K}(\tau_j) \;=\; \frac{1}{n}\sum_{i=1}^{n} \left[1 - \bar{P}_{q_i}(\tau_j)^{K}\right] \ ,
\end{equation}

\textbf{Bootstrap error bars}

We use bootstrap because the scores reported here are averages over a sampled question set, and we want a confidence interval that reflects how those averages would change if a different set of questions had been drawn. Resampling questions with replacement gives a consistent estimate of that variability without any parametric assumption on the per-question score distribution.

We attach a confidence band to each per-temperature score using the \textbf{paired nonparametric bootstrap}. We draw $R_{\mathrm{boot}}$ resamples of the question set with replacement; in each resample $b$, we recompute $\mathcal{P}_{K}(\tau_j)$ for every $\tau_j$ from the same resampled question set, and the standard error is the empirical standard deviation of $\mathcal{P}_{K}^{(b)}(\tau_j)$ across replicates. Pairing across temperatures, that is, using one common resample for all $\tau_j$, removes question-level variation that would otherwise inflate every interval; the resulting ratios between temperatures are especially stable. For this reason, we display bootstrap bands only on the ratio-to-uniform plots, where the band is attached to the superiority proposition itself. An absolute per-temperature band answers a different question: how uncertain that one temperature's pass rate is in isolation. Without pairing, such bands would be wide and largely overlap, hiding genuine strategy differences. The comparison of interest is whether one allocation strategy outperforms another at the same $K$. Paired resampling keeps each question's difficulty common to both sides of the ratio, so the interval reflects uncertainty in the relative advantage rather than in the absolute score.

Computationally, we exploit a count-then-matmul identity that reduces the bootstrap inner loop to a single matrix multiplication: for any per-question quantity $f_{i}$, the bootstrap sum over the $n$ slots of resample $b$ equals $\sum_{i=1}^{n} f_{i} \cdot c_{b,i}$, where $c_{b,i}$ is the number of times question $q_i$ was drawn in resample $b$. This collapses each bootstrap pass into a single BLAS-optimized matrix multiplication, orders of magnitude faster than direct loop-based indexing.

\textbf{Optimal temperature allocation}

For a fixed total budget of $K$ attempts, we ask how those attempts should be distributed across the $l$ candidate temperatures $\tau_1, \ldots, \tau_l$ to maximize the dataset-level pass@$K$ score according to Equation \ref{eq:ProbFin}.

\paragraph{Convexity and uniqueness.}
Each summand $\exp\!\left(\sum_{j} K_j \ln \bar{P}_{ij}\right)$ is the composition of a convex nondecreasing outer function with an affine map of the allocation, hence convex. The objective is a finite sum of convex terms over a polyhedron, so the relaxation of Equation \ref{eq:ProbFin} is convex. The Hessian has entries
\begin{equation}
\frac{\partial^{2}\!\left(n\,\bar{\mathcal{P}}\right)}{\partial K_{j}\,\partial K_{j'}}
\;=\;
\sum_{i=1}^{n} \ln \bar{P}_{ij} \cdot \ln \bar{P}_{ij'} \cdot \exp\!\left(\sum_{k=1}^{l} K_k \ln \bar{P}_{ik}\right) \ ,
\end{equation}
which is positive definite whenever the matrix $(\ln \bar{P}_{ij})_{i,j}$ has full column rank. On our data this matrix has rank $l$ (smallest singular value $\approx 3.3 \times 10^{-2}$), so the objective is strictly convex and the minimizer is unique. Slater's condition holds at the uniform sampling scheme $K_j = K / l$, which makes the KKT conditions necessary and sufficient for global optimality.

\paragraph{Numerical solver.}
We solve this continuous relaxation with SLSQP (sequential least-squares programming), supplying the analytic gradient
\begin{equation}
\frac{\partial\!\left(n\,\bar{\mathcal{P}}\right)}{\partial K_{j}}
\;=\;
\sum_{i=1}^{n} \ln \bar{P}_{ij} \cdot \exp\!\left(\sum_{k=1}^{l} K_k \ln \bar{P}_{ik}\right) \ .
\end{equation}
Iterates are warm-started across $K$ by carrying the previous continuous optimum forward. Because each $K_j$ is in practice a non-negative integer, we follow the SLSQP step with a combinatorial integer search inside a radius-$2$ neighborhood of the rounded continuous optimum.

\paragraph{Certificate of global optimality.}
At every reported $K$, the continuous optimum is verified by three independent diagnostics:
\begin{itemize}
\item \textbf{KKT residuals.} Equality residual $\left|\sum_{j} K_j - K\right|$, bound violations, stationarity gap on the active set, and dual slack on the inactive set, evaluated against the gradient above.
\item \textbf{First-order gap bound.} The a posteriori convexity bound on the objective gap obtained by linearizing $\bar{\mathcal{P}}$ at the candidate allocation and minimizing the linearization over the simplex.
\item \textbf{Independent solver cross-check.} The same problem is re-solved with the interior-point conic solver CLARABEL (via CVXPY), and the gap between the two reported objective values is recorded.
\end{itemize}
On the dataset used for the headline results, all three quantities remain below $10^{-7}$ across the entire $K$ grid, certifying that the reported allocation achieves the unique global optimum to floating-point tolerance.

\paragraph{How to read the ratio-to-uniform plot.}
\label{zap:DataDis:rtu}
Every $y$ value is a pass@$K$ score divided by the same dataset's
uniform-scheme pass@$K$ at the same $K$. The uniform scheme spreads $K$ equally over the temperatures of the model's grid on that dataset (Appendix \ref{zap:CompResources}). The dashed grey line
at $y = 1$ is the uniform baseline: values above $1$ outperform
uniform at that $K$ and values below $1$ underperform. The
$x$-axis is the total budget $K$ on a log scale.

\medskip\noindent
\textbf{Colored curves.} One curve per single-temperature
allocation: all $K$ attempts placed at one fixed $\tau$. To keep
the plot readable, only temperatures that are the dominant choice
in the optimal allocation at some $K$ are drawn; the full
per-temperature pass-rates for every $\tau$ in the dataset's grid
are tabulated in Section \ref{zap:DataDis:passk_tables} for LiveCodeBench. The
shaded band around each curve is the paired-bootstrap $\pm 1\sigma$
interval on the ratio (see
the \emph{Bootstrap error bars} paragraph above); pairing keeps each question's
difficulty common to numerator and denominator, so the band
measures uncertainty in the relative advantage rather than in the
absolute score. The ratios of Figures \ref{fig:rtu:math500:grid} and \ref{fig:rtu:hitom:grid} come from the paper's rendering code. Figure \ref{fig:rtu:hitom:grid} is drawn with dashed reference curves, colors by temperature, exact legend values, and tick labels over the full range.

\medskip\noindent
\textbf{Red trace.} The leave-one-out
out-of-sample optimal allocation. At each $K$ the convex
relaxation of Equation \ref{eq:ProbFin} is fit on the held-in questions
and scored on the held-out one; the score is then divided by the same
in-sample uniform-scheme total used for the colored curves.
Figures \ref{fig:rtu:grid}, \ref{fig:rtu:math500:grid}, and \ref{fig:rtu:hitom:grid} draw the red trace without a band.

\medskip\noindent
\textbf{What to look for.} Compare the red trace against the
highest colored curve at each $K$. Where they coincide, no
allocation beats that single temperature; where the red trace
sits clearly above, mixing temperatures pays off out-of-sample.

\paragraph{\texorpdfstring{How to read the summary tables.}{How to read the summary tables.}}
The summary table of Appendix \ref{zap:Multiple:Comp:res} gives, for each benchmark, two quantities per model at $K \in \{1, 16, 256, 1024\}$, both from the same ratio-to-uniform computation as the figures. These are the Optimal (LOO) / uniform ratio of the red trace, and the best single temperature, followed in parentheses by its ratio to uniform. The best single temperature at $K$ is the temperature, among all temperatures of the model's grid, whose single-$\tau$ allocation has the highest ratio to uniform at that $K$ on the model's full included question set. It is therefore selected and scored in-sample. The LOO optimum is scored on held-out questions. This temperature is not necessarily drawn in the figures, whose colored curves (the solid curves in Figure \ref{fig:rtu:hitom:grid}) show only the dominant temperatures of the optimal allocation.

\paragraph{\texorpdfstring{How to read the pass@$K$ tables.}{How to read the pass@K tables.}}
Tables \ref{tab:passk_gpt_oss_20b} to \ref{tab:passk_qwen3_0_6b} report, for LiveCodeBench, per-$K$ pass-rates for each allocation strategy:
the per-temperature single-$\tau$ allocations, and three row-aggregate
strategies, namely the best single $\tau$, the uniform sampling scheme,
and the optimal allocation. \textbf{Bold} marks the row-best aggregate
strategy among the three; per-temperature columns are read column-wise
to locate each $K$'s peak temperature and are never bolded.
The Best $\tau$ and Optimal columns each show two numbers per row.
\emph{In-sample} (top): the strategy (best single $\tau$ or convex
allocation, Eq. \ref{eq:ProbFin}) chosen on all included questions
and scored on the same set, optimistic since the strategy saw its
own test data. \emph{Out-of-sample}
(bottom, smaller): the leave-one-out mean, for each question we re-fit
the strategy on the remaining $n_{\text{inc}}-1$ questions and score
the held-out one, then average. Close numbers indicate a robust
choice; a large gap indicates the in-sample value is fitting noise.
Each per-model table places the same $K$ rows in two stacked sub-tables
to fit a portrait page: low temperatures above, high temperatures
together with the three row-aggregate columns below.

\subsection{\texorpdfstring{Extended Results}{Extended Results}}
\label{zap:Multiple:Comp:res}

\begin{figure}[p]
  \centering
  \begin{subfigure}[t]{0.48\textwidth}
    \centering
    \includegraphics[width=\linewidth]{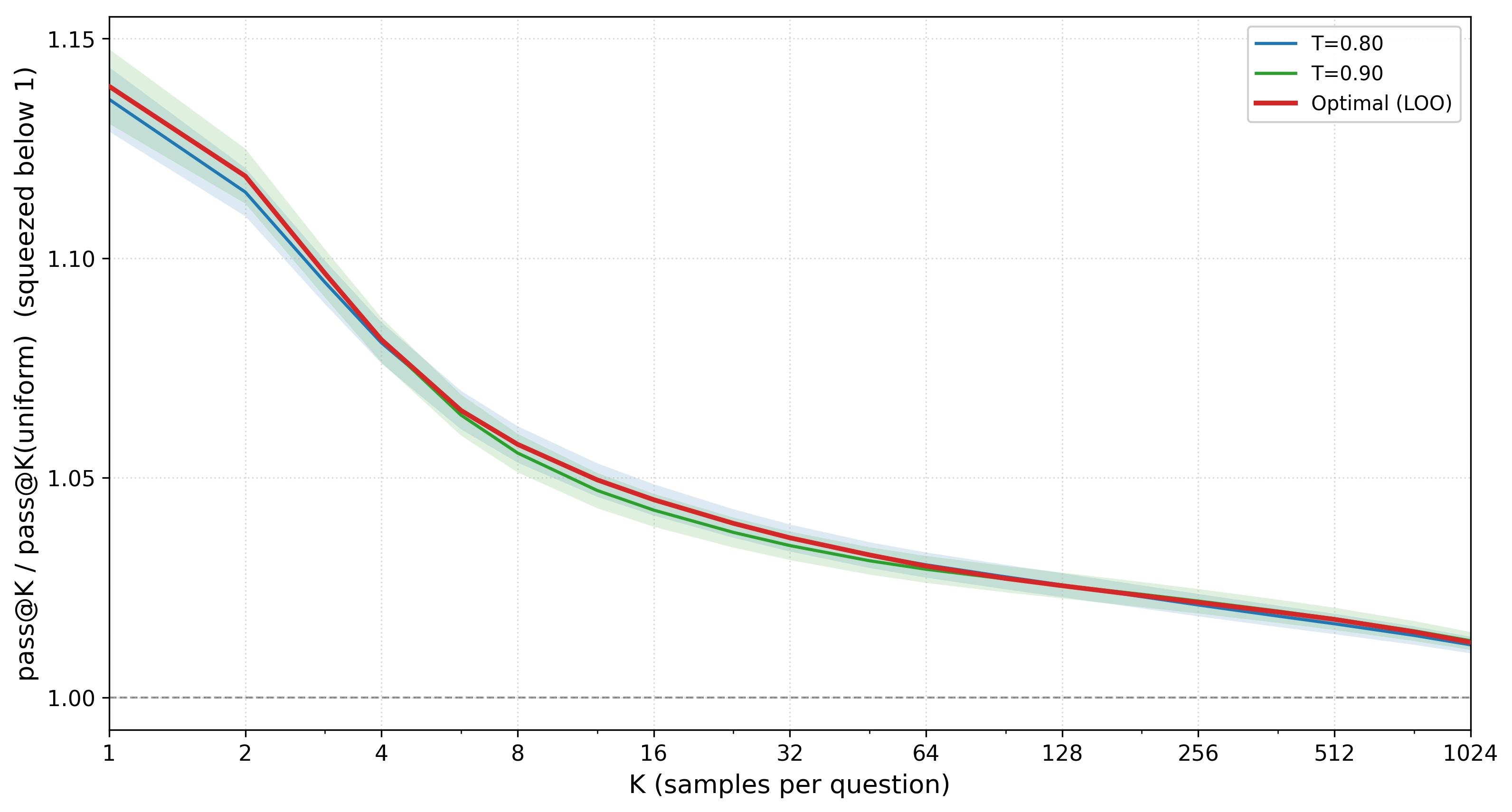}
    \caption{GPT OSS 20B, $n = 766$}
    \label{fig:rtu:gpt_oss_20b}
  \end{subfigure}\hfill  \begin{subfigure}[t]{0.48\textwidth}
    \centering
    \includegraphics[width=\linewidth]{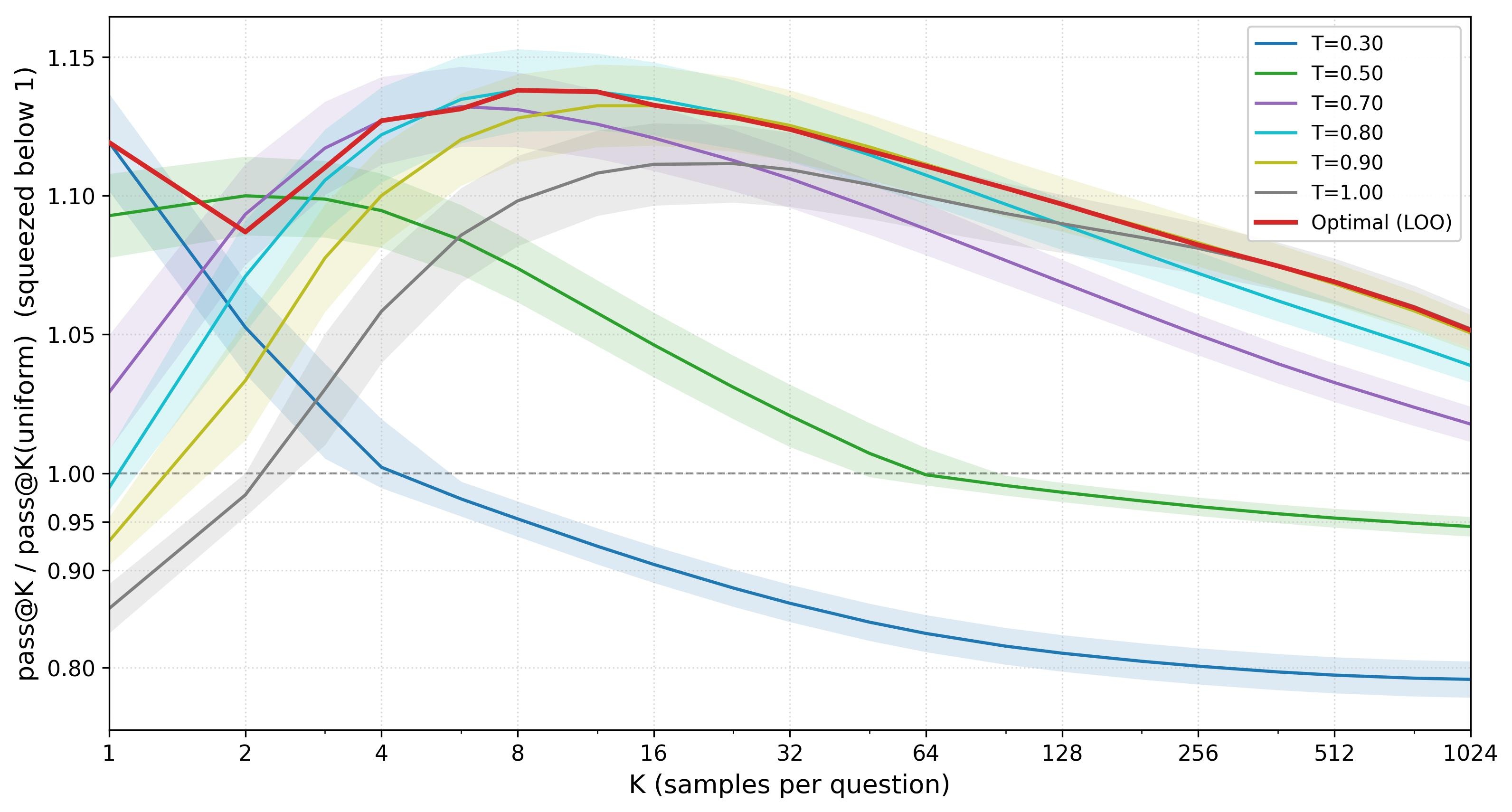}
    \caption{Mistral 7B Instruct, $n = 359$}
    \label{fig:rtu:mistral_7b_instruct}
  \end{subfigure}
  \vspace{8pt}

  \begin{subfigure}[t]{0.48\textwidth}
    \centering
    \includegraphics[width=\linewidth]{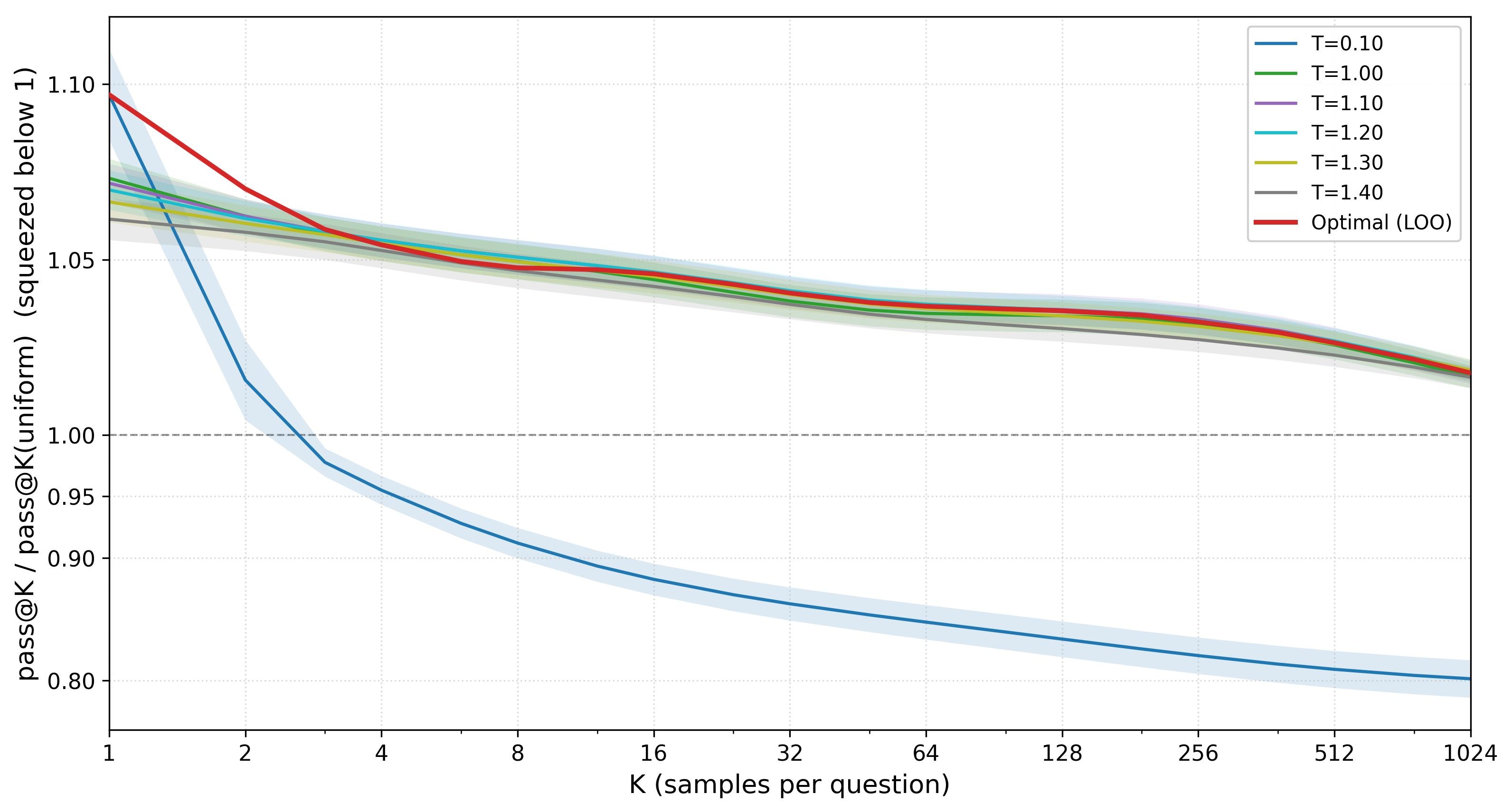}
    \caption{Qwen3-8B, $n = 647$}
    \label{fig:rtu:qwen3_8b}
  \end{subfigure}\hfill  \begin{subfigure}[t]{0.48\textwidth}
    \centering
    \includegraphics[width=\linewidth]{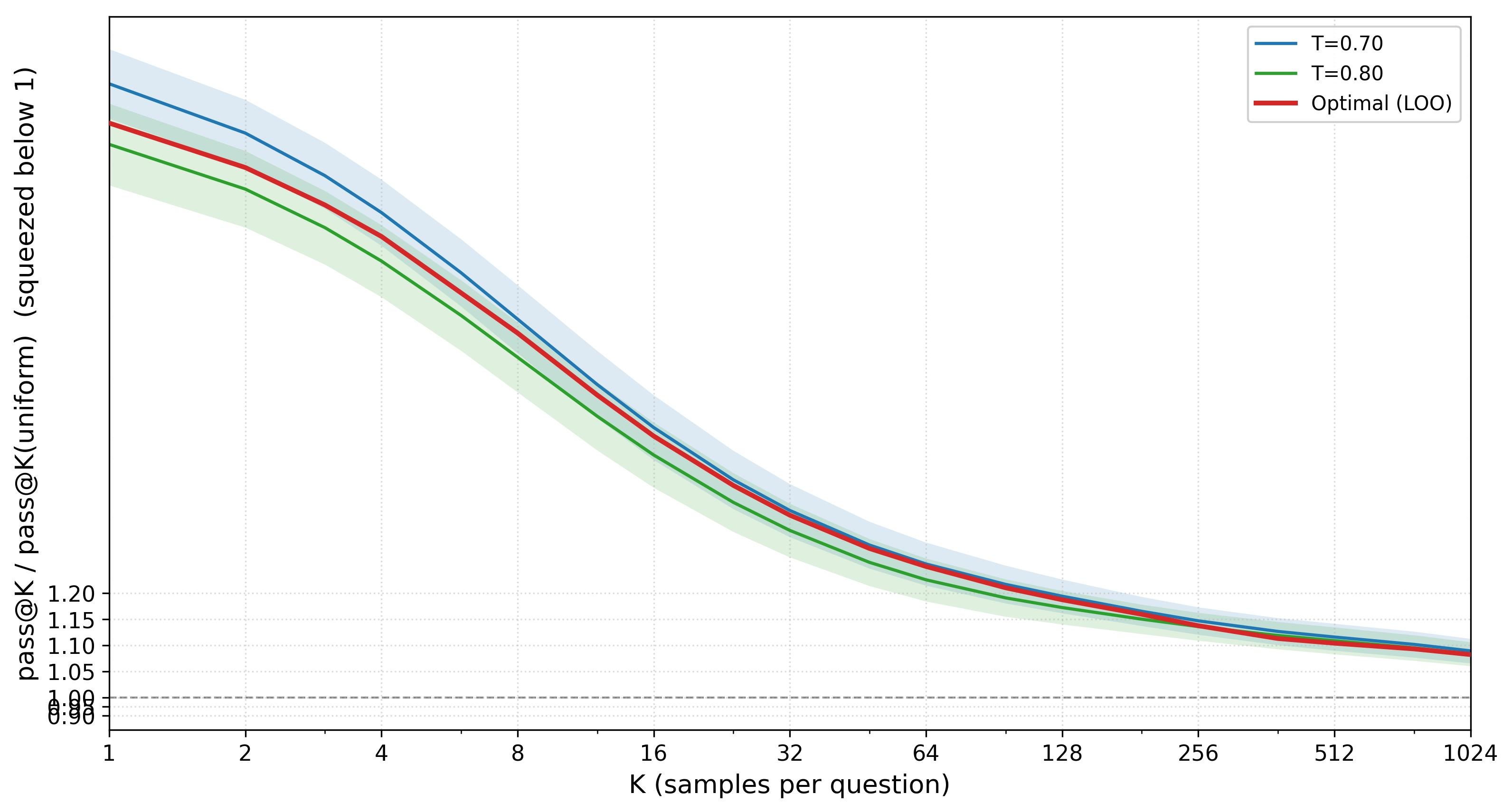}
    \caption{Qwen3-0.6B, $n = 85$}
    \label{fig:rtu:qwen3_0_6b}
  \end{subfigure}  \caption{Ratio to uniform pass@$K$, $\mathrm{rtu}(K) = \mathrm{pass}@K(\mathrm{strategy}) / \mathrm{pass}@K(\mathrm{uniform})$, on LiveCodeBench for all four models, read as described in Appendix \ref{zap:Multiple:Comp:The}. The red trace is the Section \ref{sec:Multiple:Ideal} optimum, evaluated leave-one-out (LOO) at each $K$ of the colored-curve grid. $n$ is the number of included questions.}
  \label{fig:rtu:grid}
  \par\vspace{10pt}

{\par
  \centering
  \begin{subfigure}[t]{0.48\textwidth}
    \centering
    \includegraphics[width=\linewidth]{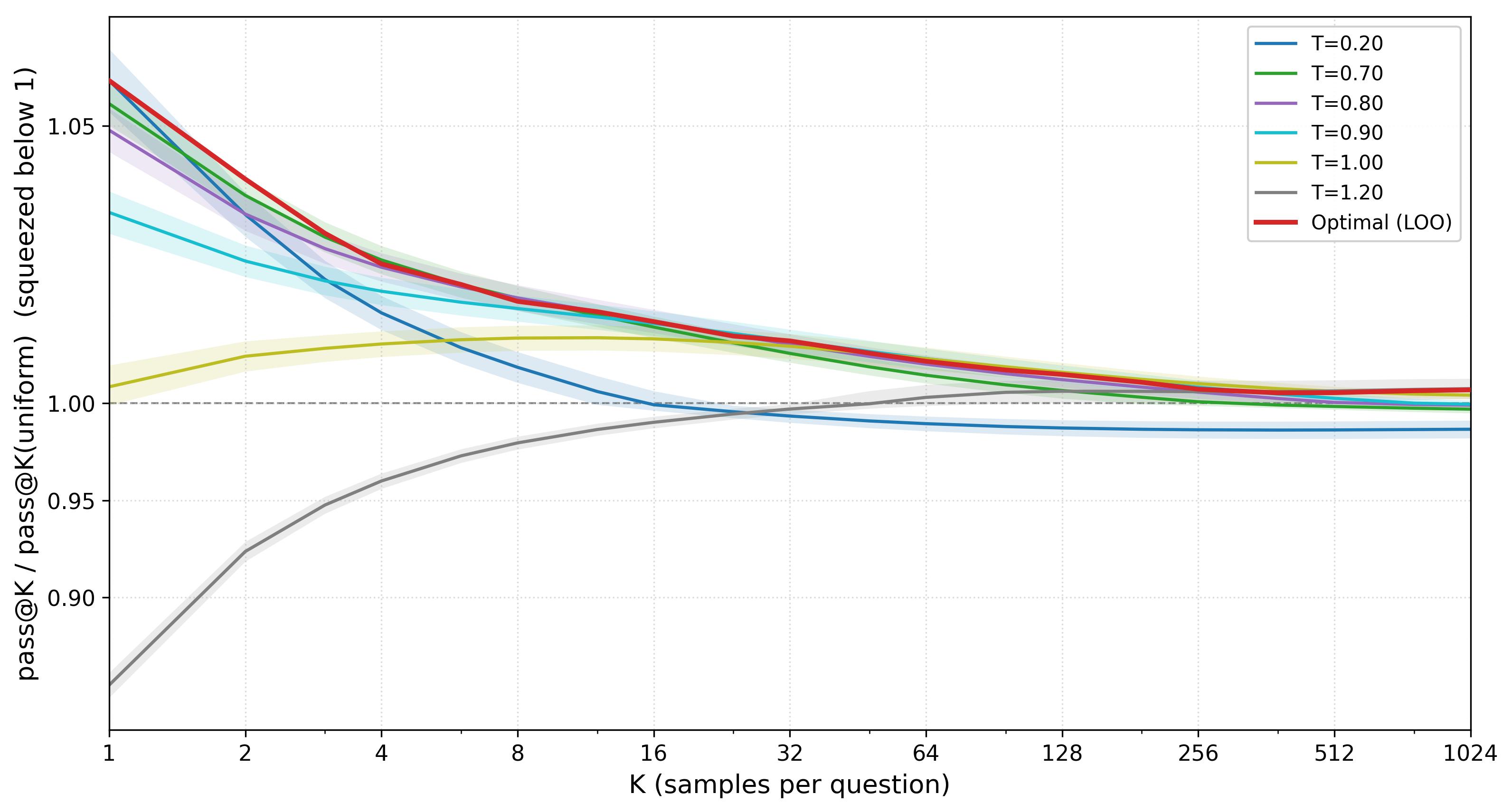}
    \caption{GPT OSS 20B, $n = 500$}
    \label{fig:rtu:math500:gpt_oss_20b}
  \end{subfigure}\hfill  \begin{subfigure}[t]{0.48\textwidth}
    \centering
    \includegraphics[width=\linewidth]{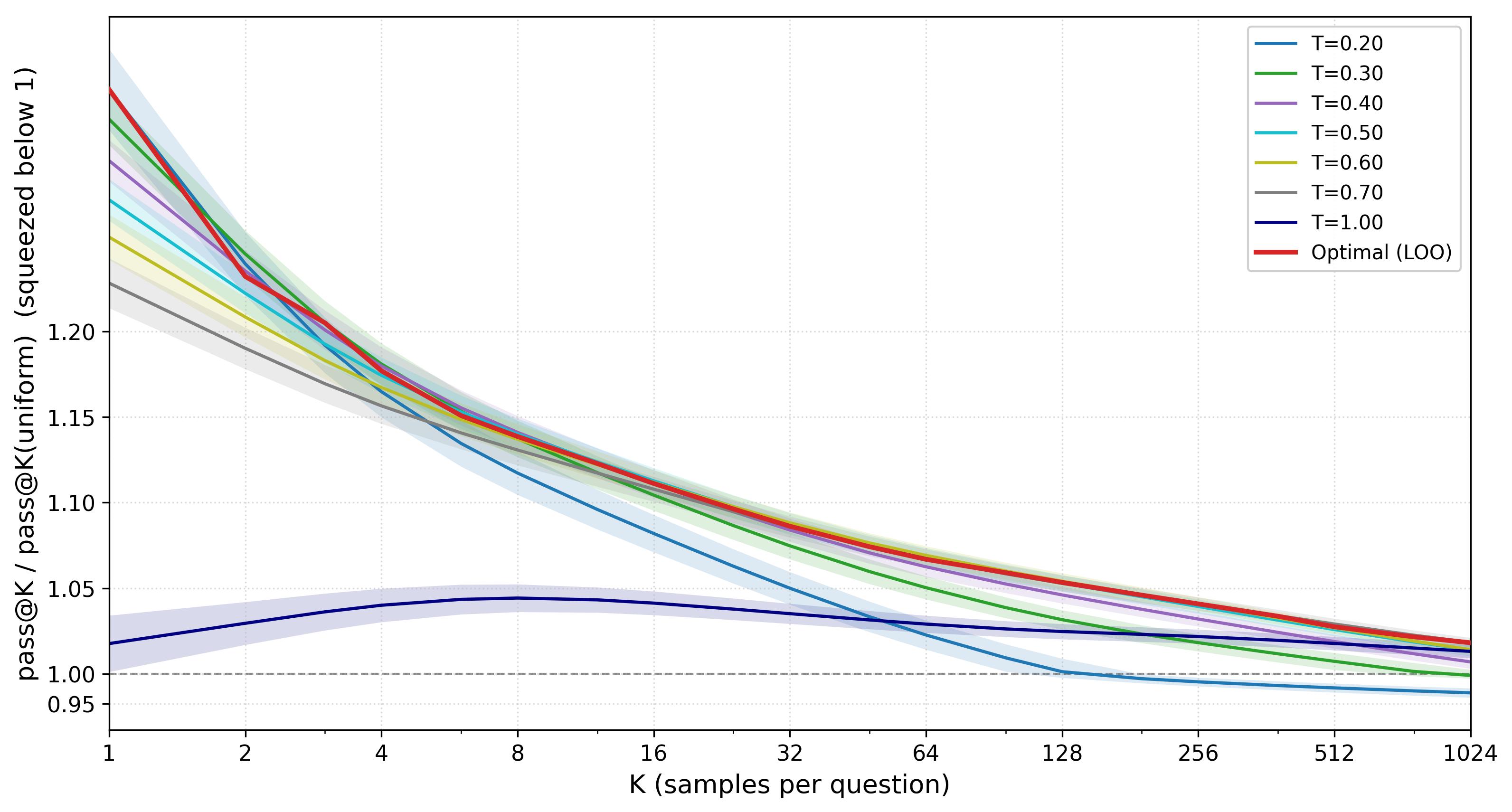}
    \caption{Mistral 7B Instruct, $n = 455$}
    \label{fig:rtu:math500:mistral_7b_instruct}
  \end{subfigure}
  \vspace{8pt}

  \begin{subfigure}[t]{0.48\textwidth}
    \centering
    \includegraphics[width=\linewidth]{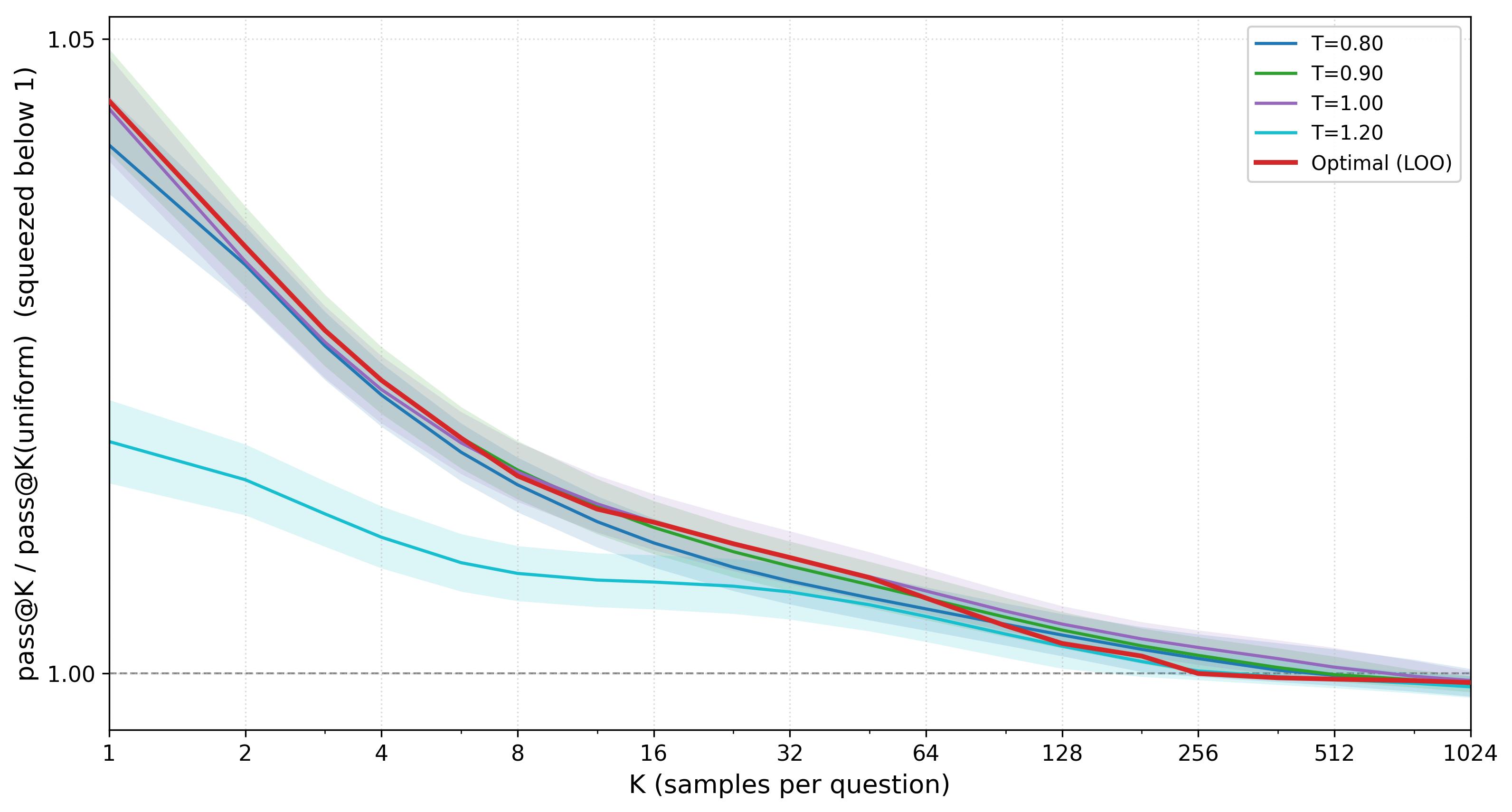}
    \caption{Qwen3-8B, $n = 496$}
    \label{fig:rtu:math500:qwen3_8b}
  \end{subfigure}\hfill  \begin{subfigure}[t]{0.48\textwidth}
    \centering
    \includegraphics[width=\linewidth]{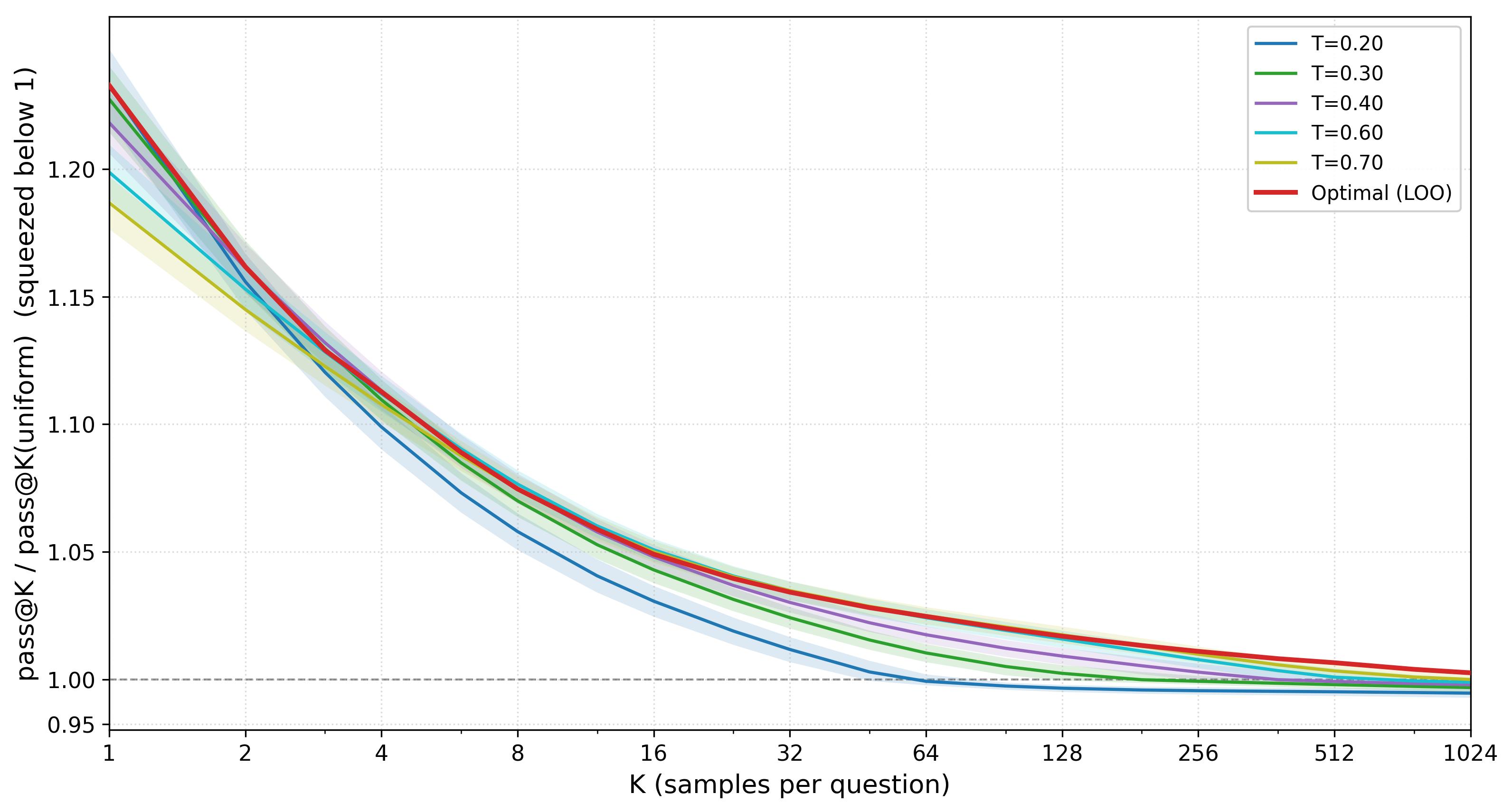}
    \caption{Qwen3-0.6B, $n = 492$}
    \label{fig:rtu:math500:qwen3_0_6b}
  \end{subfigure}
  \caption{Ratio to uniform pass@$K$ on MATH-500 for all four models, defined as in Figure \ref{fig:rtu:grid}; $n$ is the number of included questions. For Qwen3-8B, the uniform scheme spreads $K$ over $20$ temperatures here and over $38$ in Figure \ref{fig:rtu:qwen3_8b}, so the absolute ratios of the two panels are not comparable.}
  \label{fig:rtu:math500:grid}
\par}
\end{figure}

\begin{figure}[t]
{\par
  \centering
  \begin{subfigure}[t]{0.48\textwidth}
    \centering
    \includegraphics[width=\linewidth]{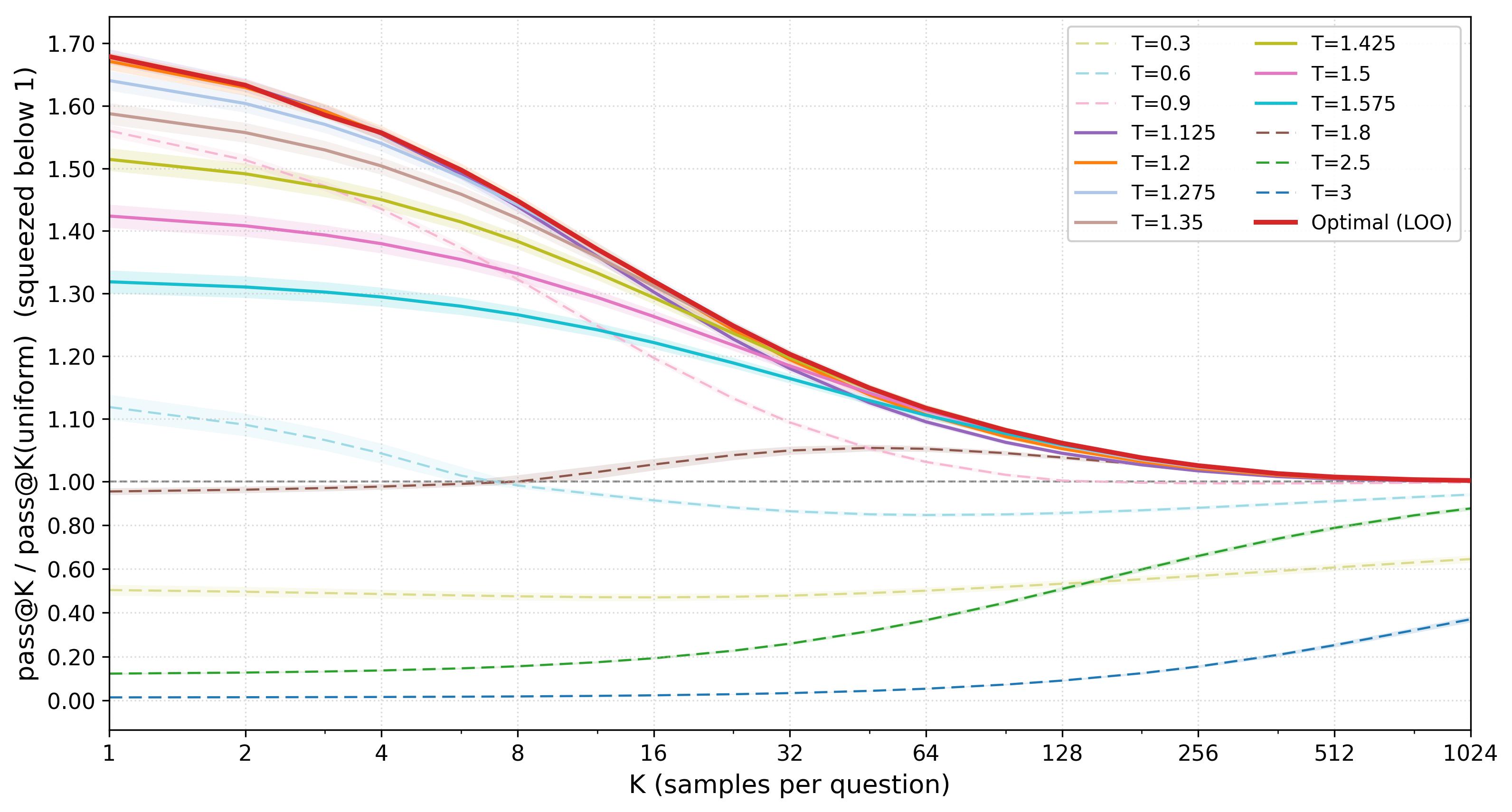}
    \caption{GPT OSS 20B, $n = 678$}
    \label{fig:rtu:hitom:gpt_oss_20b}
  \end{subfigure}\hfill  \begin{subfigure}[t]{0.48\textwidth}
    \centering
    \includegraphics[width=\linewidth]{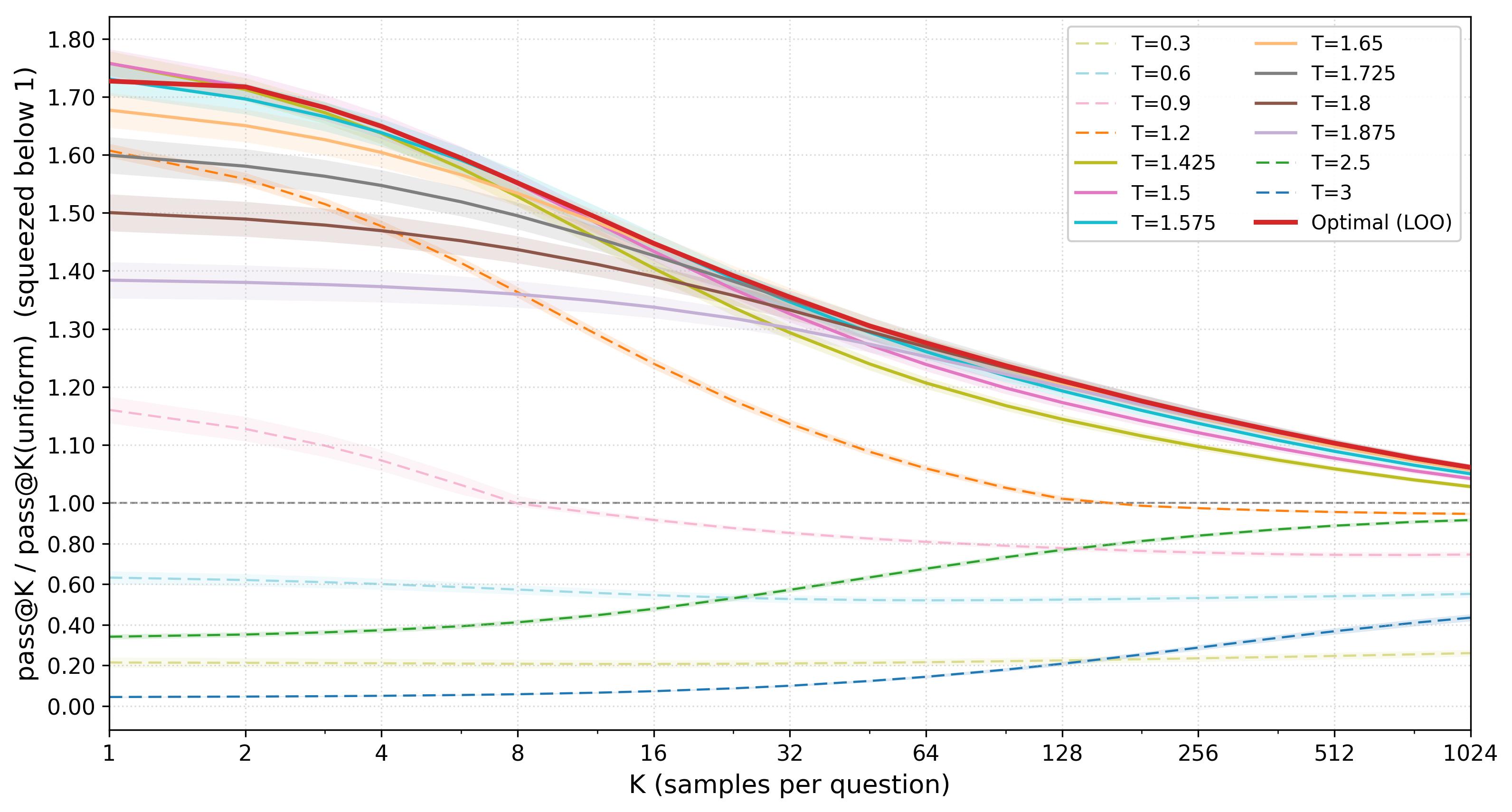}
    \caption{Mistral 7B Instruct, $n = 615$}
    \label{fig:rtu:hitom:mistral_7b_instruct}
  \end{subfigure}
  \vspace{8pt}

  \begin{subfigure}[t]{0.48\textwidth}
    \centering
    \includegraphics[width=\linewidth]{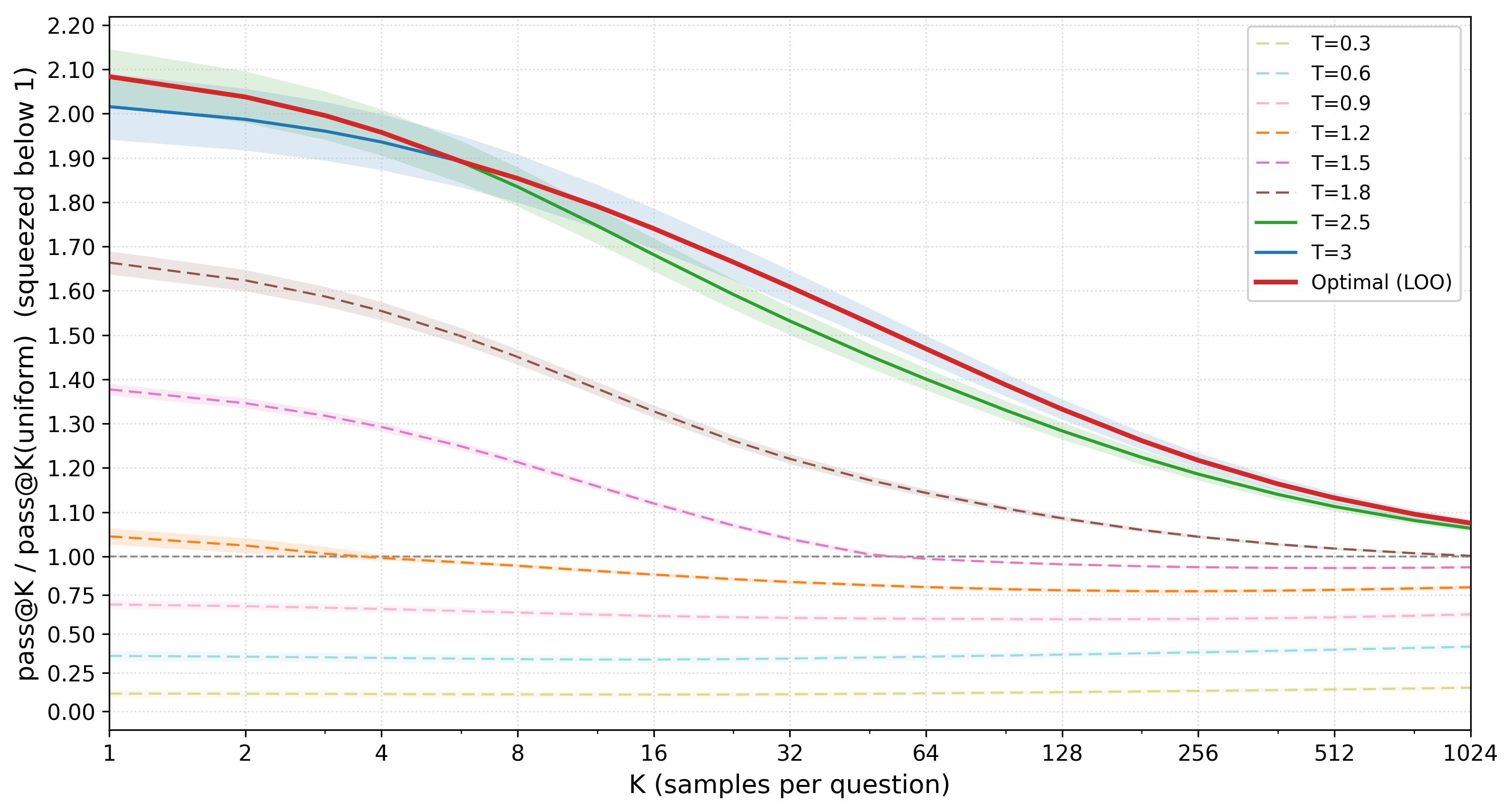}
    \caption{Qwen3-8B, $n = 466$}
    \label{fig:rtu:hitom:qwen3_8b}
  \end{subfigure}\hfill  \begin{subfigure}[t]{0.48\textwidth}
    \centering
    \includegraphics[width=\linewidth]{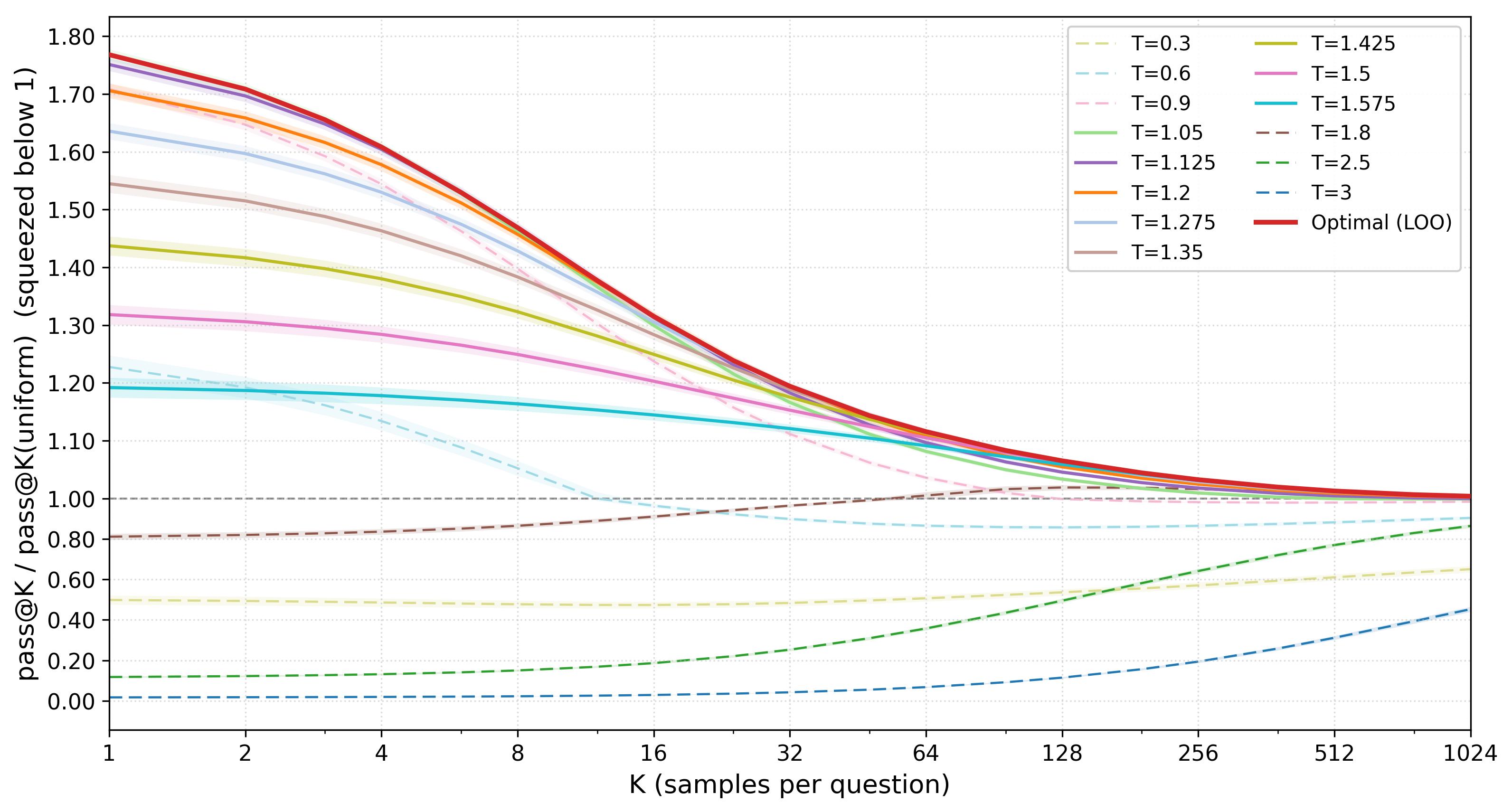}
    \caption{Qwen3-0.6B, $n = 743$}
    \label{fig:rtu:hitom:qwen3_0_6b}
  \end{subfigure}
  \caption{Ratio to uniform pass@$K$ on Hi-ToM for all four models, defined as in Figure \ref{fig:rtu:grid}; $n$ is the number of included questions. Solid curves show the temperatures that are the dominant choice in the optimal allocation at some $K$, namely every grid temperature from $1.125$ to $1.575$ for GPT OSS 20B, from $1.425$ to $1.875$ for Mistral 7B Instruct, $2.5$ and $3.0$ for Qwen3-8B, and from $1.05$ to $1.575$ for Qwen3-0.6B. Dashed curves show the other temperatures of a reference set drawn in every panel, $\tau \in \{0.3, 0.6, 0.9, 1.2, 1.5, 1.8, 2.5, 3.0\}$. A temperature has the same color in every panel.}
  \label{fig:rtu:hitom:grid}
\par}
\end{figure}

We show the ratio-to-uniform curves of the four models in Figure \ref{fig:rtu:grid} for LiveCodeBench, in Figure \ref{fig:rtu:math500:grid} for MATH-500, and in Figure \ref{fig:rtu:hitom:grid} for Hi-ToM. The table below gives, for each benchmark and model, the two quantities defined in Appendix \ref{zap:Multiple:Comp:The} at $K \in \{1, 16, 256, 1024\}$, and Tables \ref{tab:passk_gpt_oss_20b} to \ref{tab:passk_qwen3_0_6b} give the complete pass@$K$ values on LiveCodeBench.\label{zap:DataDis:passk_tables}

On all three benchmarks, the red trace tracks the colored upper envelope, which shows the limited gain of multi-temperature allocation over a $K$-tuned single temperature, the central observation of Section \ref{sec:Multiple:Comp}. We evaluate all curves from the fitted profiles up to $K_{\max} = 1024$ (Appendix \ref{zap:Stats:Methodology}), which exceeds the per-question, per-temperature sample count only for Qwen3-8B on MATH-500, where $\nu = 750$, so its points at $K = 768$ and $K = 1024$ are fit-based.

{\par
\centering
\footnotesize
\setlength{\tabcolsep}{4pt}
\begin{tabular}{@{}cllcccc@{}}
  \toprule
  & \textsc{Model} & & $K = 1$ & $K = 16$ & $K = 256$ & $K = 1024$ \\
  \midrule
   & GPT OSS 20B & Optimal (LOO) / uniform & $1.139$ & $1.045$ & $1.022$ & $1.013$ \\
  & & best single $\tau$ (in-sample) & $0.9$ ($1.139$) & $0.8$ ($1.045$) & $0.9$ ($1.022$) & $0.9$ ($1.013$) \\
   & Mistral 7B Instruct & Optimal (LOO) / uniform & $1.119$ & $1.133$ & $1.082$ & $1.052$ \\
  & & best single $\tau$ (in-sample) & $0.3$ ($1.119$) & $0.8$ ($1.135$) & $0.9$ ($1.083$) & $1.0$ ($1.052$) \\
  \smash{\raisebox{3.9pt}{\rotatebox[origin=c]{90}{\textit{LiveCodeBench}}}} & Qwen3-8B & Optimal (LOO) / uniform & $1.097$ & $1.046$ & $1.032$ & $1.018$ \\
  & & best single $\tau$ (in-sample) & $0.1$ ($1.097$) & $1.2$ ($1.046$) & $1.1$ ($1.033$) & $1.3$ ($1.019$) \\
   & Qwen3-0.6B & Optimal (LOO) / uniform & $2.099$ & $1.500$ & $1.137$ & $1.082$ \\
  & & best single $\tau$ (in-sample) & $0.7$ ($2.174$) & $0.7$ ($1.516$) & $0.7$ ($1.147$) & $0.7$ ($1.089$) \\
  \midrule
   & GPT OSS 20B & Optimal (LOO) / uniform & $1.058$ & $1.015$ & $1.003$ & $1.002$ \\
  & & best single $\tau$ (in-sample) & $0.2$ ($1.058$) & $0.8$ ($1.015$) & $1.0$ ($1.004$) & $1.2$ ($1.003$) \\
   & Mistral 7B Instruct & Optimal (LOO) / uniform & $1.341$ & $1.111$ & $1.041$ & $1.018$ \\
  & & best single $\tau$ (in-sample) & $0.2$ ($1.341$) & $0.5$ ($1.113$) & $0.7$ ($1.041$) & $0.7$ ($1.018$) \\
  \smash{\raisebox{3.9pt}{\rotatebox[origin=c]{90}{\textit{MATH-500}}}} & Qwen3-8B & Optimal (LOO) / uniform & $1.045$ & $1.012$ & $1.000$ & $0.998$ \\
  & & best single $\tau$ (in-sample) & $0.9$ ($1.045$) & $1.0$ ($1.012$) & $1.0$ ($1.002$) & $1.0$ ($0.998$) \\
   & Qwen3-0.6B & Optimal (LOO) / uniform & $1.233$ & $1.049$ & $1.011$ & $1.003$ \\
  & & best single $\tau$ (in-sample) & $0.2$ ($1.233$) & $0.6$ ($1.051$) & $0.9$ ($1.011$) & $1.0$ ($1.003$) \\
  \midrule
   & GPT OSS 20B & Optimal (LOO) / uniform & $1.679$ & $1.320$ & $1.025$ & $1.001$ \\
  & & best single $\tau$ (in-sample) & $1.125$ ($1.679$) & $1.275$ ($1.320$) & $1.5$ ($1.025$) & $1.575$ ($1.001$) \\
   & Mistral 7B Instruct & Optimal (LOO) / uniform & $1.727$ & $1.447$ & $1.153$ & $1.061$ \\
  & & best single $\tau$ (in-sample) & $1.425$ ($1.758$) & $1.575$ ($1.447$) & $1.8$ ($1.153$) & $1.875$ ($1.062$) \\
  \smash{\raisebox{3.9pt}{\rotatebox[origin=c]{90}{\textit{Hi-ToM}}}} & Qwen3-8B & Optimal (LOO) / uniform & $2.084$ & $1.740$ & $1.217$ & $1.075$ \\
  & & best single $\tau$ (in-sample) & $2.5$ ($2.084$) & $3.0$ ($1.740$) & $3.0$ ($1.217$) & $3.0$ ($1.075$) \\
   & Qwen3-0.6B & Optimal (LOO) / uniform & $1.768$ & $1.315$ & $1.033$ & $1.004$ \\
  & & best single $\tau$ (in-sample) & $1.05$ ($1.768$) & $1.2$ ($1.315$) & $1.5$ ($1.033$) & $1.575$ ($1.004$) \\
  \bottomrule
\end{tabular}
\par
\par}

\afterpage{\clearpage}
\begin{table}[p]
\centering
\footnotesize
\setlength{\tabcolsep}{3pt}
\renewcommand{\arraystretch}{1.1}
\caption{Pass@$K$ for GPT OSS 20B. Low temperatures $\tau \in [0.1,\,0.8]$ above, high temperatures $\tau \in [0.9,\,2.0]$ together with the row-aggregate columns below.}
\label{tab:passk_gpt_oss_20b}

\rowcolors{3}{}{gray!8}
\begin{tabular}{rS[table-format=1.3]S[table-format=1.3]S[table-format=1.3]S[table-format=1.3]S[table-format=1.3]S[table-format=1.3]S[table-format=1.3]S[table-format=1.3]}
\toprule
\textbf{$K$} & \multicolumn{1}{c}{\textbf{0.1}} & \multicolumn{1}{c}{\textbf{0.2}} & \multicolumn{1}{c}{\textbf{0.3}} & \multicolumn{1}{c}{\textbf{0.4}} & \multicolumn{1}{c}{\textbf{0.5}} & \multicolumn{1}{c}{\textbf{0.6}} & \multicolumn{1}{c}{\textbf{0.7}} & \multicolumn{1}{c}{\textbf{0.8}} \\
\midrule
  $\mathbf{1}$ & 0.344 & 0.361 & 0.372 & 0.381 & 0.389 & 0.396 & 0.402 & 0.407 \\
  $\mathbf{2}$ & 0.424 & 0.466 & 0.491 & 0.508 & 0.521 & 0.532 & 0.542 & 0.549 \\
  $\mathbf{3}$ & 0.467 & 0.522 & 0.555 & 0.575 & 0.590 & 0.602 & 0.612 & 0.619 \\
  $\mathbf{4}$ & 0.495 & 0.560 & 0.595 & 0.617 & 0.633 & 0.645 & 0.654 & 0.660 \\
  \addlinespace[2pt]
  $\mathbf{6}$ & 0.531 & 0.607 & 0.646 & 0.669 & 0.685 & 0.696 & 0.704 & 0.709 \\
  $\mathbf{8}$ & 0.556 & 0.638 & 0.678 & 0.701 & 0.716 & 0.727 & 0.734 & 0.738 \\
  $\mathbf{12}$ & 0.589 & 0.678 & 0.718 & 0.740 & 0.754 & 0.764 & 0.771 & 0.774 \\
  $\mathbf{16}$ & 0.611 & 0.704 & 0.743 & 0.765 & 0.778 & 0.788 & 0.794 & 0.796 \\
  \addlinespace[2pt]
  $\mathbf{24}$ & 0.641 & 0.737 & 0.775 & 0.795 & 0.807 & 0.816 & 0.822 & 0.824 \\
  $\mathbf{32}$ & 0.662 & 0.759 & 0.795 & 0.814 & 0.826 & 0.835 & 0.840 & 0.843 \\
  $\mathbf{48}$ & 0.691 & 0.787 & 0.820 & 0.838 & 0.850 & 0.858 & 0.864 & 0.866 \\
  $\mathbf{64}$ & 0.709 & 0.804 & 0.836 & 0.854 & 0.866 & 0.874 & 0.879 & 0.882 \\
  \addlinespace[2pt]
  $\mathbf{96}$ & 0.734 & 0.826 & 0.856 & 0.874 & 0.886 & 0.894 & 0.899 & 0.902 \\
  $\mathbf{128}$ & 0.750 & 0.840 & 0.869 & 0.886 & 0.899 & 0.907 & 0.913 & 0.916 \\
  $\mathbf{192}$ & 0.771 & 0.857 & 0.885 & 0.903 & 0.915 & 0.924 & 0.930 & 0.933 \\
  $\mathbf{256}$ & 0.784 & 0.868 & 0.896 & 0.914 & 0.926 & 0.935 & 0.941 & 0.945 \\
  \addlinespace[2pt]
  $\mathbf{384}$ & 0.802 & 0.882 & 0.910 & 0.928 & 0.940 & 0.949 & 0.955 & 0.959 \\
  $\mathbf{512}$ & 0.814 & 0.891 & 0.919 & 0.937 & 0.949 & 0.958 & 0.964 & 0.968 \\
\bottomrule
\end{tabular}

\par\vspace{8pt}

\rowcolors{3}{}{gray!8}
\begin{tabular}{rS[table-format=1.3]S[table-format=1.3]S[table-format=1.3]S[table-format=1.3]S[table-format=1.3]S[table-format=1.3]S[table-format=1.3]S[table-format=1.3]@{\hskip 0.8em}cS[table-format=1.3]c}
\toprule
\textbf{$K$} & \multicolumn{1}{c}{\textbf{0.9}} & \multicolumn{1}{c}{\textbf{1.0}} & \multicolumn{1}{c}{\textbf{1.1}} & \multicolumn{1}{c}{\textbf{1.2}} & \multicolumn{1}{c}{\textbf{1.4}} & \multicolumn{1}{c}{\textbf{1.5}} & \multicolumn{1}{c}{\textbf{1.7}} & \multicolumn{1}{c}{\textbf{2.0}} & \multicolumn{1}{c}{\textbf{Best $\tau$}} & \multicolumn{1}{c}{\textbf{Uniform}} & \multicolumn{1}{c}{\textbf{\makecell{Optimal\\(in / out)}}} \\
\midrule
  $\mathbf{1}$ & 0.408 & 0.402 & 0.387 & 0.360 & 0.270 & 0.214 & 0.101 & 0.007 & {\makecell{0.408 \\ \footnotesize 0.408}} & 0.358 & {\makecell{\textbf{0.408} \\ \footnotesize \textbf{0.408}}} \\
  $\mathbf{2}$ & 0.551 & 0.545 & 0.528 & 0.498 & 0.394 & 0.323 & 0.167 & 0.013 & {\makecell{0.551 \\ \footnotesize 0.551}} & 0.492 & {\makecell{\textbf{0.551} \\ \footnotesize \textbf{0.551}}} \\
  $\mathbf{3}$ & 0.620 & 0.613 & 0.597 & 0.568 & 0.464 & 0.390 & 0.215 & 0.019 & {\makecell{0.620 \\ \footnotesize 0.620}} & 0.565 & {\makecell{\textbf{0.620} \\ \footnotesize \textbf{0.620}}} \\
  $\mathbf{4}$ & 0.660 & 0.654 & 0.638 & 0.610 & 0.509 & 0.434 & 0.251 & 0.024 & {\makecell{0.660 \\ \footnotesize 0.660}} & 0.611 & {\makecell{\textbf{0.660} \\ \footnotesize \textbf{0.660}}} \\
  \addlinespace[2pt]
  $\mathbf{6}$ & 0.708 & 0.701 & 0.686 & 0.660 & 0.565 & 0.491 & 0.303 & 0.033 & {\makecell{0.709 \\ \footnotesize \textbf{0.709}}} & 0.665 & {\makecell{\textbf{0.709} \\ \footnotesize 0.709}} \\
  $\mathbf{8}$ & 0.736 & 0.729 & 0.714 & 0.690 & 0.599 & 0.527 & 0.339 & 0.040 & {\makecell{0.738 \\ \footnotesize 0.738}} & 0.698 & {\makecell{\textbf{0.738} \\ \footnotesize \textbf{0.738}}} \\
  $\mathbf{12}$ & 0.772 & 0.765 & 0.750 & 0.727 & 0.641 & 0.571 & 0.388 & 0.052 & {\makecell{0.774 \\ \footnotesize 0.774}} & 0.737 & {\makecell{\textbf{0.774} \\ \footnotesize \textbf{0.774}}} \\
  $\mathbf{16}$ & 0.794 & 0.787 & 0.774 & 0.751 & 0.667 & 0.599 & 0.420 & 0.061 & {\makecell{0.796 \\ \footnotesize 0.796}} & 0.762 & {\makecell{\textbf{0.796} \\ \footnotesize \textbf{0.796}}} \\
  \addlinespace[2pt]
  $\mathbf{24}$ & 0.823 & 0.816 & 0.803 & 0.782 & 0.700 & 0.634 & 0.460 & 0.075 & {\makecell{0.824 \\ \footnotesize 0.824}} & 0.793 & {\makecell{\textbf{0.824} \\ \footnotesize \textbf{0.824}}} \\
  $\mathbf{32}$ & 0.841 & 0.835 & 0.822 & 0.802 & 0.722 & 0.656 & 0.485 & 0.084 & {\makecell{0.843 \\ \footnotesize 0.843}} & 0.813 & {\makecell{\textbf{0.843} \\ \footnotesize \textbf{0.843}}} \\
  $\mathbf{48}$ & 0.865 & 0.859 & 0.848 & 0.828 & 0.749 & 0.685 & 0.515 & 0.097 & {\makecell{0.866 \\ \footnotesize 0.866}} & 0.839 & {\makecell{\textbf{0.866} \\ \footnotesize \textbf{0.866}}} \\
  $\mathbf{64}$ & 0.881 & 0.876 & 0.864 & 0.845 & 0.767 & 0.703 & 0.533 & 0.107 & {\makecell{0.882 \\ \footnotesize \textbf{0.882}}} & 0.856 & {\makecell{\textbf{0.882} \\ \footnotesize 0.881}} \\
  \addlinespace[2pt]
  $\mathbf{96}$ & 0.902 & 0.897 & 0.887 & 0.868 & 0.791 & 0.727 & 0.554 & 0.119 & {\makecell{0.902 \\ \footnotesize \textbf{0.902}}} & 0.878 & {\makecell{\textbf{0.902} \\ \footnotesize 0.902}} \\
  $\mathbf{128}$ & 0.916 & 0.911 & 0.901 & 0.884 & 0.807 & 0.742 & 0.566 & 0.128 & {\makecell{0.916 \\ \footnotesize 0.913}} & 0.893 & {\makecell{\textbf{0.916} \\ \footnotesize \textbf{0.916}}} \\
  $\mathbf{192}$ & 0.934 & 0.930 & 0.921 & 0.904 & 0.827 & 0.762 & 0.580 & 0.141 & {\makecell{0.934 \\ \footnotesize \textbf{0.934}}} & 0.912 & {\makecell{\textbf{0.934} \\ \footnotesize 0.933}} \\
  $\mathbf{256}$ & 0.945 & 0.942 & 0.934 & 0.918 & 0.841 & 0.776 & 0.588 & 0.149 & {\makecell{0.945 \\ \footnotesize \textbf{0.945}}} & 0.925 & {\makecell{\textbf{0.945} \\ \footnotesize 0.945}} \\
  \addlinespace[2pt]
  $\mathbf{384}$ & 0.960 & 0.957 & 0.949 & 0.935 & 0.858 & 0.793 & 0.597 & 0.161 & {\makecell{0.960 \\ \footnotesize \textbf{0.960}}} & 0.941 & {\makecell{\textbf{0.960} \\ \footnotesize 0.960}} \\
  $\mathbf{512}$ & 0.969 & 0.967 & 0.959 & 0.946 & 0.870 & 0.805 & 0.602 & 0.169 & {\makecell{0.969 \\ \footnotesize \textbf{0.969}}} & 0.952 & {\makecell{\textbf{0.969} \\ \footnotesize 0.969}} \\
\bottomrule
\end{tabular}

\end{table}

\begin{table}[p]
\centering
\footnotesize
\setlength{\tabcolsep}{3pt}
\renewcommand{\arraystretch}{1.1}
\caption{Pass@$K$ for Mistral 7B Instruct. Low temperatures $\tau \in [0.1,\,0.8]$ above, high temperatures $\tau \in [0.9,\,2.0]$ together with the row-aggregate columns below.}
\label{tab:passk_mistral_7b_instruct}

\rowcolors{3}{}{gray!8}
\begin{tabular}{rS[table-format=1.3]S[table-format=1.3]S[table-format=1.3]S[table-format=1.3]S[table-format=1.3]S[table-format=1.3]S[table-format=1.3]S[table-format=1.3]}
\toprule
\textbf{$K$} & \multicolumn{1}{c}{\textbf{0.1}} & \multicolumn{1}{c}{\textbf{0.2}} & \multicolumn{1}{c}{\textbf{0.3}} & \multicolumn{1}{c}{\textbf{0.4}} & \multicolumn{1}{c}{\textbf{0.5}} & \multicolumn{1}{c}{\textbf{0.6}} & \multicolumn{1}{c}{\textbf{0.7}} & \multicolumn{1}{c}{\textbf{0.8}} \\
\midrule
  $\mathbf{1}$ & 0.178 & 0.184 & 0.186 & 0.185 & 0.181 & 0.177 & 0.171 & 0.164 \\
  $\mathbf{2}$ & 0.198 & 0.216 & 0.227 & 0.234 & 0.238 & 0.238 & 0.236 & 0.231 \\
  $\mathbf{3}$ & 0.210 & 0.236 & 0.254 & 0.265 & 0.273 & 0.277 & 0.277 & 0.274 \\
  $\mathbf{4}$ & 0.219 & 0.252 & 0.274 & 0.289 & 0.299 & 0.305 & 0.308 & 0.306 \\
  \addlinespace[2pt]
  $\mathbf{6}$ & 0.233 & 0.275 & 0.303 & 0.323 & 0.338 & 0.347 & 0.352 & 0.353 \\
  $\mathbf{8}$ & 0.243 & 0.292 & 0.325 & 0.349 & 0.366 & 0.378 & 0.386 & 0.388 \\
  $\mathbf{12}$ & 0.256 & 0.316 & 0.357 & 0.386 & 0.408 & 0.424 & 0.434 & 0.439 \\
  $\mathbf{16}$ & 0.266 & 0.334 & 0.380 & 0.413 & 0.438 & 0.457 & 0.470 & 0.475 \\
  \addlinespace[2pt]
  $\mathbf{24}$ & 0.278 & 0.360 & 0.413 & 0.452 & 0.482 & 0.505 & 0.521 & 0.529 \\
  $\mathbf{32}$ & 0.287 & 0.378 & 0.436 & 0.480 & 0.514 & 0.540 & 0.557 & 0.566 \\
  $\mathbf{48}$ & 0.298 & 0.405 & 0.470 & 0.520 & 0.560 & 0.589 & 0.609 & 0.619 \\
  $\mathbf{64}$ & 0.307 & 0.424 & 0.495 & 0.549 & 0.591 & 0.623 & 0.644 & 0.656 \\
  \addlinespace[2pt]
  $\mathbf{96}$ & 0.318 & 0.451 & 0.529 & 0.588 & 0.635 & 0.669 & 0.692 & 0.706 \\
  $\mathbf{128}$ & 0.327 & 0.470 & 0.553 & 0.616 & 0.665 & 0.701 & 0.725 & 0.739 \\
  $\mathbf{192}$ & 0.339 & 0.497 & 0.586 & 0.653 & 0.705 & 0.743 & 0.768 & 0.784 \\
  $\mathbf{256}$ & 0.349 & 0.516 & 0.608 & 0.679 & 0.733 & 0.771 & 0.797 & 0.814 \\
  \addlinespace[2pt]
  $\mathbf{384}$ & 0.363 & 0.542 & 0.639 & 0.714 & 0.770 & 0.808 & 0.835 & 0.853 \\
  $\mathbf{512}$ & 0.373 & 0.561 & 0.660 & 0.738 & 0.795 & 0.833 & 0.860 & 0.879 \\
\bottomrule
\end{tabular}

\par\vspace{8pt}

\rowcolors{3}{}{gray!8}
\begin{tabular}{rS[table-format=1.3]S[table-format=1.3]S[table-format=1.3]S[table-format=1.3]S[table-format=1.3]S[table-format=1.3]S[table-format=1.3]S[table-format=1.3]@{\hskip 0.8em}cS[table-format=1.3]c}
\toprule
\textbf{$K$} & \multicolumn{1}{c}{\textbf{0.9}} & \multicolumn{1}{c}{\textbf{1.0}} & \multicolumn{1}{c}{\textbf{1.1}} & \multicolumn{1}{c}{\textbf{1.2}} & \multicolumn{1}{c}{\textbf{1.4}} & \multicolumn{1}{c}{\textbf{1.5}} & \multicolumn{1}{c}{\textbf{1.7}} & \multicolumn{1}{c}{\textbf{2.0}} & \multicolumn{1}{c}{\textbf{Best $\tau$}} & \multicolumn{1}{c}{\textbf{Uniform}} & \multicolumn{1}{c}{\textbf{\makecell{Optimal\\(in / out)}}} \\
\midrule
  $\mathbf{1}$ & 0.154 & 0.143 & 0.129 & 0.112 & 0.072 & 0.052 & 0.018 & 0.000 & {\makecell{0.186 \\ \footnotesize 0.186}} & 0.166 & {\makecell{\textbf{0.186} \\ \footnotesize \textbf{0.186}}} \\
  $\mathbf{2}$ & 0.223 & 0.211 & 0.195 & 0.173 & 0.119 & 0.088 & 0.034 & 0.001 & {\makecell{0.238 \\ \footnotesize \textbf{0.238}}} & 0.216 & {\makecell{\textbf{0.238} \\ \footnotesize 0.235}} \\
  $\mathbf{3}$ & 0.267 & 0.256 & 0.238 & 0.215 & 0.153 & 0.116 & 0.047 & 0.001 & {\makecell{0.277 \\ \footnotesize \textbf{0.277}}} & 0.248 & {\makecell{\textbf{0.277} \\ \footnotesize 0.275}} \\
  $\mathbf{4}$ & 0.300 & 0.289 & 0.271 & 0.247 & 0.180 & 0.139 & 0.058 & 0.002 & {\makecell{0.308 \\ \footnotesize 0.308}} & 0.273 & {\makecell{\textbf{0.308} \\ \footnotesize \textbf{0.308}}} \\
  \addlinespace[2pt]
  $\mathbf{6}$ & 0.349 & 0.338 & 0.320 & 0.295 & 0.221 & 0.174 & 0.077 & 0.003 & {\makecell{0.353 \\ \footnotesize \textbf{0.353}}} & 0.311 & {\makecell{\textbf{0.353} \\ \footnotesize 0.352}} \\
  $\mathbf{8}$ & 0.385 & 0.374 & 0.356 & 0.330 & 0.252 & 0.202 & 0.093 & 0.003 & {\makecell{0.388 \\ \footnotesize 0.388}} & 0.341 & {\makecell{\textbf{0.388} \\ \footnotesize \textbf{0.388}}} \\
  $\mathbf{12}$ & 0.437 & 0.427 & 0.409 & 0.382 & 0.297 & 0.242 & 0.118 & 0.005 & {\makecell{0.439 \\ \footnotesize 0.439}} & 0.386 & {\makecell{\textbf{0.439} \\ \footnotesize \textbf{0.439}}} \\
  $\mathbf{16}$ & 0.474 & 0.466 & 0.448 & 0.420 & 0.331 & 0.273 & 0.138 & 0.006 & {\makecell{0.475 \\ \footnotesize \textbf{0.475}}} & 0.419 & {\makecell{\textbf{0.475} \\ \footnotesize 0.475}} \\
  \addlinespace[2pt]
  $\mathbf{24}$ & 0.529 & 0.520 & 0.502 & 0.474 & 0.380 & 0.317 & 0.169 & 0.009 & {\makecell{0.529 \\ \footnotesize 0.519}} & 0.468 & {\makecell{\textbf{0.529} \\ \footnotesize \textbf{0.528}}} \\
  $\mathbf{32}$ & 0.567 & 0.559 & 0.541 & 0.513 & 0.416 & 0.350 & 0.192 & 0.011 & {\makecell{0.567 \\ \footnotesize \textbf{0.567}}} & 0.504 & {\makecell{\textbf{0.567} \\ \footnotesize 0.566}} \\
  $\mathbf{48}$ & 0.621 & 0.613 & 0.596 & 0.567 & 0.467 & 0.397 & 0.227 & 0.014 & {\makecell{0.621 \\ \footnotesize \textbf{0.621}}} & 0.556 & {\makecell{\textbf{0.621} \\ \footnotesize 0.620}} \\
  $\mathbf{64}$ & 0.658 & 0.651 & 0.634 & 0.605 & 0.503 & 0.430 & 0.253 & 0.017 & {\makecell{0.658 \\ \footnotesize \textbf{0.658}}} & 0.592 & {\makecell{\textbf{0.658} \\ \footnotesize 0.658}} \\
  \addlinespace[2pt]
  $\mathbf{96}$ & 0.709 & 0.703 & 0.687 & 0.658 & 0.552 & 0.477 & 0.290 & 0.021 & {\makecell{0.709 \\ \footnotesize 0.709}} & 0.643 & {\makecell{\textbf{0.709} \\ \footnotesize \textbf{0.709}}} \\
  $\mathbf{128}$ & 0.744 & 0.739 & 0.724 & 0.695 & 0.587 & 0.509 & 0.316 & 0.025 & {\makecell{0.744 \\ \footnotesize 0.744}} & 0.678 & {\makecell{\textbf{0.744} \\ \footnotesize \textbf{0.744}}} \\
  $\mathbf{192}$ & 0.791 & 0.788 & 0.774 & 0.746 & 0.636 & 0.554 & 0.354 & 0.030 & {\makecell{0.791 \\ \footnotesize \textbf{0.791}}} & 0.726 & {\makecell{\textbf{0.791} \\ \footnotesize 0.790}} \\
  $\mathbf{256}$ & 0.822 & 0.821 & 0.808 & 0.781 & 0.671 & 0.585 & 0.381 & 0.033 & {\makecell{0.822 \\ \footnotesize \textbf{0.822}}} & 0.759 & {\makecell{\textbf{0.822} \\ \footnotesize 0.821}} \\
  \addlinespace[2pt]
  $\mathbf{384}$ & 0.863 & 0.863 & 0.852 & 0.828 & 0.719 & 0.629 & 0.419 & 0.039 & {\makecell{0.863 \\ \footnotesize 0.854}} & 0.803 & {\makecell{\textbf{0.864} \\ \footnotesize \textbf{0.863}}} \\
  $\mathbf{512}$ & 0.890 & 0.891 & 0.881 & 0.858 & 0.752 & 0.659 & 0.445 & 0.044 & {\makecell{0.891 \\ \footnotesize \textbf{0.891}}} & 0.833 & {\makecell{\textbf{0.891} \\ \footnotesize 0.890}} \\
\bottomrule
\end{tabular}

\end{table}

\begin{table}[p]
\centering
\footnotesize
\setlength{\tabcolsep}{3pt}
\renewcommand{\arraystretch}{1.1}
\caption{Pass@$K$ for Qwen3-8B, low-temperature half ($\tau \in [0.1,\,1.9]$). $\tau \in [0.1,\,1.0]$ above, $\tau \in [1.1,\,1.9]$ below. Continued in Table \ref{tab:passk_qwen3_8b_high}; the row-aggregate columns are computed across the full 38-temperature sweep and appear there.}
\label{tab:passk_qwen3_8b_low}

\rowcolors{3}{}{gray!8}
\begin{tabular}{rS[table-format=1.3]S[table-format=1.3]S[table-format=1.3]S[table-format=1.3]S[table-format=1.3]S[table-format=1.3]S[table-format=1.3]S[table-format=1.3]S[table-format=1.3]S[table-format=1.3]}
\toprule
\textbf{$K$} & \multicolumn{1}{c}{\textbf{0.1}} & \multicolumn{1}{c}{\textbf{0.2}} & \multicolumn{1}{c}{\textbf{0.3}} & \multicolumn{1}{c}{\textbf{0.4}} & \multicolumn{1}{c}{\textbf{0.5}} & \multicolumn{1}{c}{\textbf{0.6}} & \multicolumn{1}{c}{\textbf{0.7}} & \multicolumn{1}{c}{\textbf{0.8}} & \multicolumn{1}{c}{\textbf{0.9}} & \multicolumn{1}{c}{\textbf{1.0}} \\
\midrule
  $\mathbf{1}$ & 0.585 & 0.585 & 0.583 & 0.582 & 0.580 & 0.578 & 0.576 & 0.575 & 0.574 & 0.573 \\
  $\mathbf{2}$ & 0.615 & 0.626 & 0.632 & 0.637 & 0.639 & 0.641 & 0.642 & 0.643 & 0.643 & 0.643 \\
  $\mathbf{3}$ & 0.628 & 0.645 & 0.656 & 0.663 & 0.668 & 0.672 & 0.675 & 0.676 & 0.678 & 0.679 \\
  $\mathbf{4}$ & 0.636 & 0.658 & 0.671 & 0.681 & 0.688 & 0.693 & 0.696 & 0.699 & 0.701 & 0.702 \\
  \addlinespace[2pt]
  $\mathbf{6}$ & 0.647 & 0.675 & 0.693 & 0.705 & 0.713 & 0.720 & 0.725 & 0.729 & 0.731 & 0.733 \\
  $\mathbf{8}$ & 0.655 & 0.688 & 0.707 & 0.720 & 0.730 & 0.738 & 0.744 & 0.748 & 0.751 & 0.753 \\
  $\mathbf{12}$ & 0.666 & 0.705 & 0.726 & 0.741 & 0.752 & 0.761 & 0.768 & 0.774 & 0.778 & 0.780 \\
  $\mathbf{16}$ & 0.674 & 0.716 & 0.738 & 0.754 & 0.766 & 0.776 & 0.784 & 0.790 & 0.795 & 0.798 \\
  \addlinespace[2pt]
  $\mathbf{24}$ & 0.686 & 0.732 & 0.755 & 0.772 & 0.785 & 0.796 & 0.805 & 0.812 & 0.817 & 0.821 \\
  $\mathbf{32}$ & 0.694 & 0.742 & 0.767 & 0.784 & 0.798 & 0.809 & 0.819 & 0.827 & 0.832 & 0.836 \\
  $\mathbf{48}$ & 0.706 & 0.756 & 0.783 & 0.801 & 0.816 & 0.829 & 0.839 & 0.847 & 0.853 & 0.856 \\
  $\mathbf{64}$ & 0.713 & 0.766 & 0.793 & 0.812 & 0.829 & 0.842 & 0.853 & 0.861 & 0.867 & 0.870 \\
  \addlinespace[2pt]
  $\mathbf{96}$ & 0.722 & 0.779 & 0.808 & 0.829 & 0.847 & 0.861 & 0.872 & 0.881 & 0.886 & 0.890 \\
  $\mathbf{128}$ & 0.728 & 0.788 & 0.818 & 0.840 & 0.860 & 0.875 & 0.886 & 0.894 & 0.900 & 0.903 \\
  $\mathbf{192}$ & 0.736 & 0.801 & 0.833 & 0.856 & 0.877 & 0.893 & 0.904 & 0.912 & 0.918 & 0.921 \\
  $\mathbf{256}$ & 0.742 & 0.810 & 0.843 & 0.868 & 0.889 & 0.904 & 0.916 & 0.924 & 0.929 & 0.933 \\
  \addlinespace[2pt]
  $\mathbf{384}$ & 0.749 & 0.821 & 0.856 & 0.882 & 0.905 & 0.920 & 0.931 & 0.939 & 0.944 & 0.947 \\
  $\mathbf{512}$ & 0.754 & 0.830 & 0.865 & 0.892 & 0.915 & 0.930 & 0.940 & 0.948 & 0.953 & 0.955 \\
\bottomrule
\end{tabular}

\par\vspace{8pt}

\rowcolors{3}{}{gray!8}
\begin{tabular}{rS[table-format=1.3]S[table-format=1.3]S[table-format=1.3]S[table-format=1.3]S[table-format=1.3]S[table-format=1.3]S[table-format=1.3]S[table-format=1.3]S[table-format=1.3]}
\toprule
\textbf{$K$} & \multicolumn{1}{c}{\textbf{1.1}} & \multicolumn{1}{c}{\textbf{1.2}} & \multicolumn{1}{c}{\textbf{1.3}} & \multicolumn{1}{c}{\textbf{1.4}} & \multicolumn{1}{c}{\textbf{1.5}} & \multicolumn{1}{c}{\textbf{1.6}} & \multicolumn{1}{c}{\textbf{1.7}} & \multicolumn{1}{c}{\textbf{1.8}} & \multicolumn{1}{c}{\textbf{1.9}} \\
\midrule
  $\mathbf{1}$ & 0.572 & 0.571 & 0.569 & 0.567 & 0.563 & 0.557 & 0.549 & 0.538 & 0.522 \\
  $\mathbf{2}$ & 0.643 & 0.643 & 0.642 & 0.641 & 0.638 & 0.635 & 0.630 & 0.623 & 0.613 \\
  $\mathbf{3}$ & 0.679 & 0.679 & 0.679 & 0.677 & 0.675 & 0.672 & 0.667 & 0.661 & 0.654 \\
  $\mathbf{4}$ & 0.703 & 0.703 & 0.702 & 0.701 & 0.699 & 0.695 & 0.691 & 0.685 & 0.678 \\
  \addlinespace[2pt]
  $\mathbf{6}$ & 0.734 & 0.733 & 0.733 & 0.731 & 0.729 & 0.725 & 0.721 & 0.715 & 0.708 \\
  $\mathbf{8}$ & 0.754 & 0.754 & 0.753 & 0.751 & 0.749 & 0.745 & 0.740 & 0.734 & 0.727 \\
  $\mathbf{12}$ & 0.781 & 0.781 & 0.780 & 0.778 & 0.775 & 0.771 & 0.766 & 0.760 & 0.753 \\
  $\mathbf{16}$ & 0.799 & 0.799 & 0.798 & 0.796 & 0.793 & 0.789 & 0.784 & 0.777 & 0.770 \\
  \addlinespace[2pt]
  $\mathbf{24}$ & 0.823 & 0.823 & 0.822 & 0.820 & 0.816 & 0.812 & 0.807 & 0.801 & 0.793 \\
  $\mathbf{32}$ & 0.838 & 0.838 & 0.837 & 0.835 & 0.832 & 0.828 & 0.822 & 0.816 & 0.808 \\
  $\mathbf{48}$ & 0.858 & 0.859 & 0.858 & 0.855 & 0.852 & 0.848 & 0.842 & 0.836 & 0.828 \\
  $\mathbf{64}$ & 0.872 & 0.873 & 0.872 & 0.869 & 0.865 & 0.861 & 0.855 & 0.849 & 0.842 \\
  \addlinespace[2pt]
  $\mathbf{96}$ & 0.892 & 0.892 & 0.890 & 0.887 & 0.883 & 0.879 & 0.873 & 0.867 & 0.859 \\
  $\mathbf{128}$ & 0.905 & 0.905 & 0.903 & 0.900 & 0.896 & 0.891 & 0.884 & 0.878 & 0.871 \\
  $\mathbf{192}$ & 0.922 & 0.922 & 0.921 & 0.917 & 0.912 & 0.907 & 0.901 & 0.894 & 0.887 \\
  $\mathbf{256}$ & 0.934 & 0.933 & 0.932 & 0.928 & 0.924 & 0.919 & 0.912 & 0.905 & 0.897 \\
  \addlinespace[2pt]
  $\mathbf{384}$ & 0.948 & 0.948 & 0.946 & 0.943 & 0.939 & 0.934 & 0.927 & 0.920 & 0.912 \\
  $\mathbf{512}$ & 0.956 & 0.956 & 0.955 & 0.953 & 0.948 & 0.944 & 0.937 & 0.931 & 0.922 \\
\bottomrule
\end{tabular}

\end{table}

\begin{table}[p]
\centering
\footnotesize
\setlength{\tabcolsep}{3pt}
\renewcommand{\arraystretch}{1.1}
\caption{Pass@$K$ for Qwen3-8B, high-temperature half ($\tau \in [2.0,\,4.2]$); continuation of Table \ref{tab:passk_qwen3_8b_low}. $\tau \in [2.0,\,2.9]$ above, $\tau \in [3.0,\,4.2]$ together with the row-aggregate columns below. Aggregates are computed across the full 38-temperature sweep.}
\label{tab:passk_qwen3_8b_high}
\label{tab:passk_qwen3_8b}

\rowcolors{3}{}{gray!8}
\begin{tabular}{rS[table-format=1.3]S[table-format=1.3]S[table-format=1.3]S[table-format=1.3]S[table-format=1.3]S[table-format=1.3]S[table-format=1.3]S[table-format=1.3]S[table-format=1.3]S[table-format=1.3]}
\toprule
\textbf{$K$} & \multicolumn{1}{c}{\textbf{2.0}} & \multicolumn{1}{c}{\textbf{2.1}} & \multicolumn{1}{c}{\textbf{2.2}} & \multicolumn{1}{c}{\textbf{2.3}} & \multicolumn{1}{c}{\textbf{2.4}} & \multicolumn{1}{c}{\textbf{2.5}} & \multicolumn{1}{c}{\textbf{2.6}} & \multicolumn{1}{c}{\textbf{2.7}} & \multicolumn{1}{c}{\textbf{2.8}} & \multicolumn{1}{c}{\textbf{2.9}} \\
\midrule
  $\mathbf{1}$ & 0.501 & 0.475 & 0.444 & 0.409 & 0.369 & 0.326 & 0.281 & 0.236 & 0.191 & 0.150 \\
  $\mathbf{2}$ & 0.601 & 0.584 & 0.563 & 0.536 & 0.502 & 0.462 & 0.415 & 0.363 & 0.307 & 0.250 \\
  $\mathbf{3}$ & 0.643 & 0.630 & 0.613 & 0.592 & 0.564 & 0.530 & 0.488 & 0.438 & 0.381 & 0.319 \\
  $\mathbf{4}$ & 0.669 & 0.657 & 0.642 & 0.624 & 0.600 & 0.570 & 0.532 & 0.486 & 0.431 & 0.369 \\
  \addlinespace[2pt]
  $\mathbf{6}$ & 0.699 & 0.689 & 0.676 & 0.660 & 0.640 & 0.615 & 0.583 & 0.543 & 0.493 & 0.434 \\
  $\mathbf{8}$ & 0.719 & 0.709 & 0.696 & 0.682 & 0.663 & 0.640 & 0.612 & 0.575 & 0.530 & 0.475 \\
  $\mathbf{12}$ & 0.744 & 0.734 & 0.722 & 0.708 & 0.691 & 0.671 & 0.645 & 0.613 & 0.572 & 0.523 \\
  $\mathbf{16}$ & 0.761 & 0.751 & 0.739 & 0.725 & 0.709 & 0.690 & 0.665 & 0.635 & 0.597 & 0.550 \\
  \addlinespace[2pt]
  $\mathbf{24}$ & 0.784 & 0.773 & 0.762 & 0.748 & 0.732 & 0.713 & 0.691 & 0.662 & 0.626 & 0.583 \\
  $\mathbf{32}$ & 0.799 & 0.789 & 0.777 & 0.763 & 0.747 & 0.729 & 0.707 & 0.679 & 0.645 & 0.603 \\
  $\mathbf{48}$ & 0.819 & 0.809 & 0.797 & 0.782 & 0.766 & 0.749 & 0.728 & 0.701 & 0.668 & 0.628 \\
  $\mathbf{64}$ & 0.833 & 0.822 & 0.810 & 0.796 & 0.779 & 0.762 & 0.741 & 0.716 & 0.683 & 0.644 \\
  \addlinespace[2pt]
  $\mathbf{96}$ & 0.850 & 0.840 & 0.828 & 0.814 & 0.797 & 0.779 & 0.759 & 0.734 & 0.703 & 0.665 \\
  $\mathbf{128}$ & 0.862 & 0.853 & 0.841 & 0.826 & 0.809 & 0.790 & 0.771 & 0.747 & 0.715 & 0.678 \\
  $\mathbf{192}$ & 0.878 & 0.869 & 0.857 & 0.843 & 0.826 & 0.806 & 0.786 & 0.763 & 0.731 & 0.695 \\
  $\mathbf{256}$ & 0.889 & 0.880 & 0.868 & 0.855 & 0.837 & 0.817 & 0.797 & 0.775 & 0.742 & 0.706 \\
  \addlinespace[2pt]
  $\mathbf{384}$ & 0.903 & 0.894 & 0.883 & 0.870 & 0.852 & 0.832 & 0.811 & 0.790 & 0.758 & 0.721 \\
  $\mathbf{512}$ & 0.913 & 0.904 & 0.893 & 0.880 & 0.862 & 0.842 & 0.821 & 0.800 & 0.768 & 0.731 \\
\bottomrule
\end{tabular}

\par\vspace{8pt}

\rowcolors{3}{}{gray!8}
\begin{tabular}{rS[table-format=1.3]S[table-format=1.3]S[table-format=1.3]S[table-format=1.3]S[table-format=1.3]S[table-format=1.3]S[table-format=1.3]S[table-format=1.3]S[table-format=1.3]@{\hskip 0.8em}cS[table-format=1.3]c}
\toprule
\textbf{$K$} & \multicolumn{1}{c}{\textbf{3.0}} & \multicolumn{1}{c}{\textbf{3.1}} & \multicolumn{1}{c}{\textbf{3.2}} & \multicolumn{1}{c}{\textbf{3.3}} & \multicolumn{1}{c}{\textbf{3.4}} & \multicolumn{1}{c}{\textbf{3.5}} & \multicolumn{1}{c}{\textbf{3.7}} & \multicolumn{1}{c}{\textbf{4.0}} & \multicolumn{1}{c}{\textbf{4.2}} & \multicolumn{1}{c}{\textbf{Best $\tau$}} & \multicolumn{1}{c}{\textbf{Uniform}} & \multicolumn{1}{c}{\textbf{\makecell{Optimal\\(in / out)}}} \\
\midrule
  $\mathbf{1}$ & 0.113 & 0.081 & 0.055 & 0.035 & 0.021 & 0.012 & 0.003 & 0.000 & 0.000 & {\makecell{0.585 \\ \footnotesize 0.585}} & 0.534 & {\makecell{\textbf{0.585} \\ \footnotesize \textbf{0.585}}} \\
  $\mathbf{2}$ & 0.195 & 0.144 & 0.101 & 0.066 & 0.041 & 0.023 & 0.005 & 0.000 & 0.000 & {\makecell{0.643 \\ \footnotesize 0.642}} & 0.606 & {\makecell{\textbf{0.648} \\ \footnotesize \textbf{0.648}}} \\
  $\mathbf{3}$ & 0.256 & 0.194 & 0.139 & 0.093 & 0.058 & 0.033 & 0.008 & 0.000 & 0.000 & {\makecell{0.679 \\ \footnotesize 0.677}} & 0.642 & {\makecell{\textbf{0.681} \\ \footnotesize \textbf{0.680}}} \\
  $\mathbf{4}$ & 0.302 & 0.235 & 0.172 & 0.117 & 0.074 & 0.043 & 0.010 & 0.000 & 0.000 & {\makecell{0.703 \\ \footnotesize 0.701}} & 0.666 & {\makecell{\textbf{0.704} \\ \footnotesize \textbf{0.702}}} \\
  \addlinespace[2pt]
  $\mathbf{6}$ & 0.367 & 0.295 & 0.223 & 0.157 & 0.102 & 0.060 & 0.015 & 0.001 & 0.000 & {\makecell{0.734 \\ \footnotesize 0.730}} & 0.697 & {\makecell{\textbf{0.734} \\ \footnotesize \textbf{0.731}}} \\
  $\mathbf{8}$ & 0.410 & 0.337 & 0.261 & 0.188 & 0.126 & 0.076 & 0.020 & 0.001 & 0.000 & {\makecell{0.754 \\ \footnotesize 0.752}} & 0.718 & {\makecell{\textbf{0.754} \\ \footnotesize \textbf{0.752}}} \\
  $\mathbf{12}$ & 0.463 & 0.392 & 0.314 & 0.236 & 0.163 & 0.102 & 0.029 & 0.001 & 0.000 & {\makecell{0.781 \\ \footnotesize 0.779}} & 0.745 & {\makecell{\textbf{0.781} \\ \footnotesize \textbf{0.780}}} \\
  $\mathbf{16}$ & 0.494 & 0.427 & 0.350 & 0.269 & 0.191 & 0.124 & 0.037 & 0.002 & 0.000 & {\makecell{0.799 \\ \footnotesize \textbf{0.799}}} & 0.764 & {\makecell{\textbf{0.799} \\ \footnotesize 0.799}} \\
  \addlinespace[2pt]
  $\mathbf{24}$ & 0.531 & 0.469 & 0.394 & 0.313 & 0.232 & 0.156 & 0.050 & 0.003 & 0.001 & {\makecell{0.823 \\ \footnotesize \textbf{0.823}}} & 0.789 & {\makecell{\textbf{0.823} \\ \footnotesize 0.822}} \\
  $\mathbf{32}$ & 0.553 & 0.494 & 0.422 & 0.341 & 0.259 & 0.180 & 0.062 & 0.004 & 0.001 & {\makecell{0.838 \\ \footnotesize \textbf{0.838}}} & 0.805 & {\makecell{\textbf{0.838} \\ \footnotesize 0.838}} \\
  $\mathbf{48}$ & 0.580 & 0.523 & 0.456 & 0.377 & 0.294 & 0.213 & 0.081 & 0.005 & 0.001 & {\makecell{0.859 \\ \footnotesize \textbf{0.859}}} & 0.827 & {\makecell{\textbf{0.859} \\ \footnotesize 0.858}} \\
  $\mathbf{64}$ & 0.596 & 0.542 & 0.477 & 0.399 & 0.317 & 0.235 & 0.096 & 0.007 & 0.002 & {\makecell{0.873 \\ \footnotesize \textbf{0.873}}} & 0.841 & {\makecell{\textbf{0.873} \\ \footnotesize 0.872}} \\
  \addlinespace[2pt]
  $\mathbf{96}$ & 0.618 & 0.564 & 0.502 & 0.428 & 0.345 & 0.264 & 0.118 & 0.011 & 0.002 & {\makecell{0.892 \\ \footnotesize 0.890}} & 0.860 & {\makecell{\textbf{0.892} \\ \footnotesize \textbf{0.891}}} \\
  $\mathbf{128}$ & 0.633 & 0.579 & 0.518 & 0.446 & 0.363 & 0.284 & 0.134 & 0.014 & 0.003 & {\makecell{0.905 \\ \footnotesize 0.905}} & 0.874 & {\makecell{\textbf{0.905} \\ \footnotesize \textbf{0.905}}} \\
  $\mathbf{192}$ & 0.652 & 0.597 & 0.539 & 0.468 & 0.386 & 0.309 & 0.156 & 0.020 & 0.004 & {\makecell{0.922 \\ \footnotesize \textbf{0.922}}} & 0.892 & {\makecell{\textbf{0.923} \\ \footnotesize 0.922}} \\
  $\mathbf{256}$ & 0.665 & 0.610 & 0.552 & 0.482 & 0.402 & 0.326 & 0.170 & 0.025 & 0.006 & {\makecell{0.934 \\ \footnotesize \textbf{0.934}}} & 0.904 & {\makecell{\textbf{0.934} \\ \footnotesize 0.933}} \\
  \addlinespace[2pt]
  $\mathbf{384}$ & 0.680 & 0.628 & 0.571 & 0.501 & 0.423 & 0.347 & 0.187 & 0.034 & 0.008 & {\makecell{0.948 \\ \footnotesize \textbf{0.948}}} & 0.920 & {\makecell{\textbf{0.948} \\ \footnotesize 0.947}} \\
  $\mathbf{512}$ & 0.691 & 0.640 & 0.583 & 0.514 & 0.438 & 0.360 & 0.196 & 0.041 & 0.010 & {\makecell{0.956 \\ \footnotesize 0.954}} & 0.931 & {\makecell{\textbf{0.957} \\ \footnotesize \textbf{0.956}}} \\
\bottomrule
\end{tabular}

\end{table}

\begin{table}[p]
\centering
\footnotesize
\setlength{\tabcolsep}{3pt}
\renewcommand{\arraystretch}{1.1}
\caption{Pass@$K$ for Qwen3-0.6B. Low temperatures $\tau \in [0.1,\,1.0]$ above, high temperatures $\tau \in [1.1,\,2.0]$ together with the row-aggregate columns below.}
\label{tab:passk_qwen3_0_6b}

\rowcolors{3}{}{gray!8}
\begin{tabular}{rS[table-format=1.3]S[table-format=1.3]S[table-format=1.3]S[table-format=1.3]S[table-format=1.3]S[table-format=1.3]S[table-format=1.3]S[table-format=1.3]S[table-format=1.3]S[table-format=1.3]}
\toprule
\textbf{$K$} & \multicolumn{1}{c}{\textbf{0.1}} & \multicolumn{1}{c}{\textbf{0.2}} & \multicolumn{1}{c}{\textbf{0.3}} & \multicolumn{1}{c}{\textbf{0.4}} & \multicolumn{1}{c}{\textbf{0.5}} & \multicolumn{1}{c}{\textbf{0.6}} & \multicolumn{1}{c}{\textbf{0.7}} & \multicolumn{1}{c}{\textbf{0.8}} & \multicolumn{1}{c}{\textbf{0.9}} & \multicolumn{1}{c}{\textbf{1.0}} \\
\midrule
  $\mathbf{1}$ & 0.011 & 0.022 & 0.032 & 0.041 & 0.047 & 0.051 & 0.051 & 0.048 & 0.043 & 0.036 \\
  $\mathbf{2}$ & 0.020 & 0.041 & 0.061 & 0.077 & 0.087 & 0.093 & 0.094 & 0.089 & 0.079 & 0.066 \\
  $\mathbf{3}$ & 0.029 & 0.059 & 0.086 & 0.107 & 0.122 & 0.130 & 0.130 & 0.124 & 0.111 & 0.093 \\
  $\mathbf{4}$ & 0.036 & 0.075 & 0.108 & 0.134 & 0.152 & 0.161 & 0.162 & 0.154 & 0.139 & 0.117 \\
  \addlinespace[2pt]
  $\mathbf{6}$ & 0.050 & 0.102 & 0.146 & 0.179 & 0.201 & 0.213 & 0.214 & 0.204 & 0.185 & 0.158 \\
  $\mathbf{8}$ & 0.061 & 0.125 & 0.177 & 0.215 & 0.240 & 0.254 & 0.255 & 0.244 & 0.222 & 0.191 \\
  $\mathbf{12}$ & 0.078 & 0.162 & 0.226 & 0.271 & 0.299 & 0.315 & 0.317 & 0.305 & 0.279 & 0.242 \\
  $\mathbf{16}$ & 0.091 & 0.190 & 0.263 & 0.312 & 0.343 & 0.360 & 0.362 & 0.350 & 0.322 & 0.282 \\
  \addlinespace[2pt]
  $\mathbf{24}$ & 0.110 & 0.232 & 0.316 & 0.372 & 0.406 & 0.424 & 0.426 & 0.413 & 0.384 & 0.340 \\
  $\mathbf{32}$ & 0.124 & 0.264 & 0.355 & 0.415 & 0.451 & 0.469 & 0.471 & 0.458 & 0.428 & 0.382 \\
  $\mathbf{48}$ & 0.145 & 0.310 & 0.410 & 0.476 & 0.515 & 0.532 & 0.533 & 0.520 & 0.490 & 0.442 \\
  $\mathbf{64}$ & 0.162 & 0.345 & 0.449 & 0.519 & 0.559 & 0.576 & 0.576 & 0.563 & 0.534 & 0.485 \\
  \addlinespace[2pt]
  $\mathbf{96}$ & 0.189 & 0.395 & 0.504 & 0.578 & 0.619 & 0.634 & 0.635 & 0.622 & 0.594 & 0.546 \\
  $\mathbf{128}$ & 0.210 & 0.430 & 0.541 & 0.618 & 0.658 & 0.673 & 0.675 & 0.662 & 0.636 & 0.589 \\
  $\mathbf{192}$ & 0.240 & 0.476 & 0.591 & 0.670 & 0.709 & 0.723 & 0.727 & 0.718 & 0.693 & 0.648 \\
  $\mathbf{256}$ & 0.261 & 0.507 & 0.625 & 0.704 & 0.744 & 0.757 & 0.763 & 0.756 & 0.733 & 0.689 \\
  \addlinespace[2pt]
  $\mathbf{384}$ & 0.291 & 0.548 & 0.668 & 0.748 & 0.789 & 0.804 & 0.812 & 0.807 & 0.786 & 0.743 \\
  $\mathbf{512}$ & 0.312 & 0.575 & 0.697 & 0.778 & 0.821 & 0.836 & 0.846 & 0.841 & 0.820 & 0.779 \\
\bottomrule
\end{tabular}

\par\vspace{8pt}

\rowcolors{3}{}{gray!8}
\begin{tabular}{rS[table-format=1.3]S[table-format=1.3]S[table-format=1.3]S[table-format=1.3]S[table-format=1.3]S[table-format=1.3]S[table-format=1.3]S[table-format=1.3]S[table-format=1.3]S[table-format=1.3]@{\hskip 0.6em}cS[table-format=1.3]c}
\toprule
\textbf{$K$} & \multicolumn{1}{c}{\textbf{1.1}} & \multicolumn{1}{c}{\textbf{1.2}} & \multicolumn{1}{c}{\textbf{1.3}} & \multicolumn{1}{c}{\textbf{1.4}} & \multicolumn{1}{c}{\textbf{1.5}} & \multicolumn{1}{c}{\textbf{1.6}} & \multicolumn{1}{c}{\textbf{1.7}} & \multicolumn{1}{c}{\textbf{1.8}} & \multicolumn{1}{c}{\textbf{1.9}} & \multicolumn{1}{c}{\textbf{2.0}} & \multicolumn{1}{c}{\textbf{Best $\tau$}} & \multicolumn{1}{c}{\textbf{Uniform}} & \multicolumn{1}{c}{\textbf{\makecell{Optimal\\(in / out)}}} \\
\midrule
  $\mathbf{1}$ & 0.027 & 0.019 & 0.012 & 0.007 & 0.003 & 0.001 & 0.000 & 0.000 & 0.000 & 0.000 & {\makecell{0.051 \\ \footnotesize 0.049}} & 0.023 & {\makecell{\textbf{0.051} \\ \footnotesize \textbf{0.049}}} \\
  $\mathbf{2}$ & 0.051 & 0.036 & 0.023 & 0.013 & 0.006 & 0.002 & 0.001 & 0.000 & 0.000 & 0.000 & {\makecell{0.094 \\ \footnotesize \textbf{0.091}}} & 0.045 & {\makecell{\textbf{0.094} \\ \footnotesize 0.091}} \\
  $\mathbf{3}$ & 0.073 & 0.052 & 0.033 & 0.019 & 0.009 & 0.003 & 0.001 & 0.000 & 0.000 & 0.000 & {\makecell{0.130 \\ \footnotesize 0.127}} & 0.065 & {\makecell{\textbf{0.130} \\ \footnotesize \textbf{0.127}}} \\
  $\mathbf{4}$ & 0.092 & 0.066 & 0.043 & 0.024 & 0.012 & 0.005 & 0.001 & 0.000 & 0.000 & 0.000 & {\makecell{0.162 \\ \footnotesize \textbf{0.158}}} & 0.084 & {\makecell{\textbf{0.162} \\ \footnotesize 0.158}} \\
  \addlinespace[2pt]
  $\mathbf{6}$ & 0.125 & 0.092 & 0.060 & 0.035 & 0.017 & 0.007 & 0.002 & 0.000 & 0.000 & 0.000 & {\makecell{0.214 \\ \footnotesize \textbf{0.213}}} & 0.118 & {\makecell{\textbf{0.214} \\ \footnotesize 0.209}} \\
  $\mathbf{8}$ & 0.153 & 0.113 & 0.076 & 0.044 & 0.022 & 0.009 & 0.003 & 0.001 & 0.000 & 0.000 & {\makecell{0.255 \\ \footnotesize \textbf{0.253}}} & 0.148 & {\makecell{\textbf{0.255} \\ \footnotesize 0.251}} \\
  $\mathbf{12}$ & 0.197 & 0.149 & 0.102 & 0.061 & 0.031 & 0.013 & 0.004 & 0.001 & 0.000 & 0.000 & {\makecell{0.317 \\ \footnotesize \textbf{0.317}}} & 0.198 & {\makecell{\textbf{0.317} \\ \footnotesize 0.313}} \\
  $\mathbf{16}$ & 0.232 & 0.176 & 0.123 & 0.076 & 0.039 & 0.017 & 0.005 & 0.001 & 0.000 & 0.000 & {\makecell{0.362 \\ \footnotesize \textbf{0.362}}} & 0.239 & {\makecell{\textbf{0.362} \\ \footnotesize 0.358}} \\
  \addlinespace[2pt]
  $\mathbf{24}$ & 0.283 & 0.219 & 0.155 & 0.099 & 0.054 & 0.023 & 0.008 & 0.001 & 0.000 & 0.000 & {\makecell{0.426 \\ \footnotesize \textbf{0.426}}} & 0.301 & {\makecell{\textbf{0.426} \\ \footnotesize 0.423}} \\
  $\mathbf{32}$ & 0.321 & 0.251 & 0.180 & 0.117 & 0.065 & 0.029 & 0.010 & 0.002 & 0.000 & 0.000 & {\makecell{0.471 \\ \footnotesize \textbf{0.471}}} & 0.347 & {\makecell{\textbf{0.472} \\ \footnotesize 0.468}} \\
  $\mathbf{48}$ & 0.376 & 0.298 & 0.217 & 0.144 & 0.083 & 0.039 & 0.014 & 0.003 & 0.000 & 0.000 & {\makecell{0.533 \\ \footnotesize 0.527}} & 0.413 & {\makecell{\textbf{0.534} \\ \footnotesize \textbf{0.531}}} \\
  $\mathbf{64}$ & 0.417 & 0.332 & 0.245 & 0.165 & 0.097 & 0.048 & 0.018 & 0.004 & 0.000 & 0.000 & {\makecell{0.576 \\ \footnotesize 0.564}} & 0.459 & {\makecell{\textbf{0.578} \\ \footnotesize \textbf{0.574}}} \\
  \addlinespace[2pt]
  $\mathbf{96}$ & 0.475 & 0.384 & 0.285 & 0.195 & 0.118 & 0.060 & 0.024 & 0.006 & 0.001 & 0.000 & {\makecell{0.635 \\ \footnotesize 0.626}} & 0.522 & {\makecell{\textbf{0.636} \\ \footnotesize \textbf{0.632}}} \\
  $\mathbf{128}$ & 0.517 & 0.421 & 0.315 & 0.217 & 0.134 & 0.070 & 0.029 & 0.007 & 0.001 & 0.000 & {\makecell{0.675 \\ \footnotesize \textbf{0.675}}} & 0.565 & {\makecell{\textbf{0.676} \\ \footnotesize 0.671}} \\
  $\mathbf{192}$ & 0.577 & 0.476 & 0.358 & 0.249 & 0.157 & 0.084 & 0.038 & 0.010 & 0.001 & 0.000 & {\makecell{0.727 \\ \footnotesize \textbf{0.727}}} & 0.624 & {\makecell{\textbf{0.728} \\ \footnotesize 0.724}} \\
  $\mathbf{256}$ & 0.619 & 0.516 & 0.391 & 0.272 & 0.174 & 0.094 & 0.044 & 0.013 & 0.002 & 0.000 & {\makecell{0.763 \\ \footnotesize \textbf{0.763}}} & 0.665 & {\makecell{\textbf{0.764} \\ \footnotesize 0.757}} \\
  \addlinespace[2pt]
  $\mathbf{384}$ & 0.675 & 0.571 & 0.440 & 0.305 & 0.198 & 0.109 & 0.053 & 0.018 & 0.002 & 0.000 & {\makecell{0.812 \\ \footnotesize \textbf{0.812}}} & 0.721 & {\makecell{\textbf{0.814} \\ \footnotesize 0.802}} \\
  $\mathbf{512}$ & 0.712 & 0.608 & 0.476 & 0.329 & 0.215 & 0.121 & 0.059 & 0.021 & 0.003 & 0.000 & {\makecell{0.846 \\ \footnotesize \textbf{0.846}}} & 0.758 & {\makecell{\textbf{0.849} \\ \footnotesize 0.837}} \\
\bottomrule
\end{tabular}

\end{table}

\section{Sensitivity of optimal temperature sampling detection to dataset hardness-distribution variance}
\label{zap:DataDis:RandomH}

\subsection{\texorpdfstring{Statistical Analysis Approach}{Statistical Analysis Approach}}
\label{zap:DataDis:RandomH:The}

The pass@$K$ scores reported in this paper are population averages
over a particular set of questions with a particular hardness
profile. A natural concern is whether the optimal sampling
temperature is an artifact of \emph{this} hardness profile. Would
a different temperature win if the question population were
reweighted, say by shifting weight toward harder questions or by
concentrating on a narrower difficulty band?

The Goldilocks sampling effect
(Proposition \ref{prop:GoldilocksSampling}) predicts that the
answer is \emph{no}, with exponential margin. The contribution of
a question with hardness $H$ to pass@$K$ is suppressed by the
factor $\alpha\,e^{1-\alpha}$ with $\alpha = K/H$
(Eq. \ref{eq:Gold:Dev}), which falls off exponentially as $\alpha$
moves away from $1$. Questions outside the band
$\alpha \approx 1$ are therefore exponentially silent, and any
reweighting that redistributes mass between out-of-scale questions
should have an exponentially suppressed effect on the argmax over
$\tau$. The optimal temperature is therefore expected to be robust
to all but the most surgical reweightings, and increasingly so as
$K$ grows. The purpose of this section is to quantify that
prediction empirically. We perturb the per-question weights,
measure how much the chosen temperature degrades, and repeat the
experiment across $K \in \{4,\,32,\,128,\,512\}$ to see how the
empirical sensitivity scales with the pass@$K$ regime.

\begin{figure}[tbp]
  \centering
  \includegraphics[width=\linewidth]{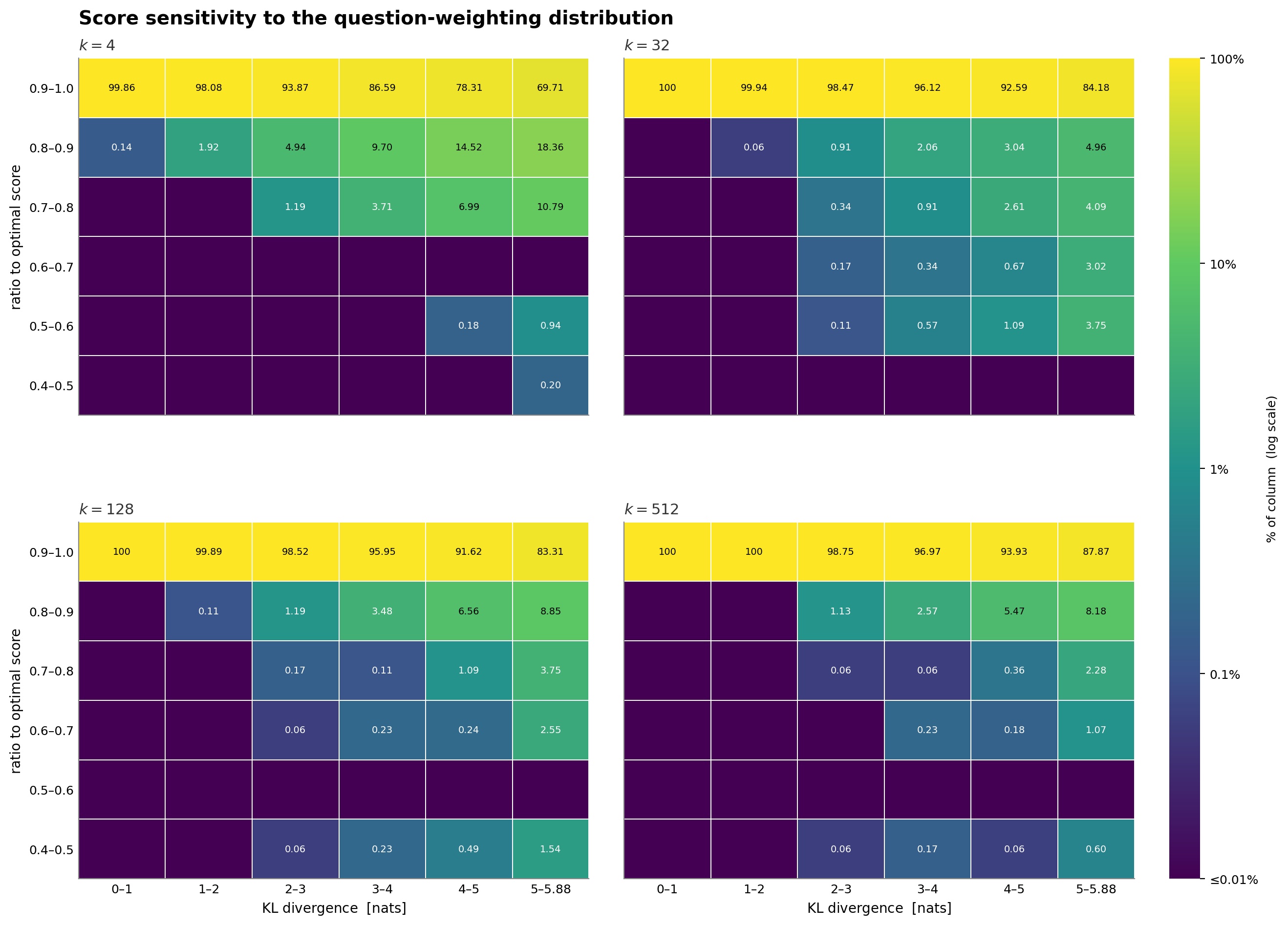}
  \caption{Distribution of $\mathrm{score}_B / \mathrm{score}_A$
           as a function of $\mathrm{KL}(W \,\|\, \mathrm{uniform})$
           on the Mistral 7B dataset ($359$ included questions,
           peak pass-rate $\ge 0.1\%$),
           with $W$ drawn from a Dirichlet smooth-base sampler
           over the questions. Each panel corresponds to one value
           of $K \in \{4,\,32,\,128,\,512\}$; the same $10{,}000$
           random $W$'s are used in every panel, so differences
           reflect only the change of $K$. Each panel is a
           $6 \times 6$ column-normalized heatmap (KL bands on
           $x$, ratio bands on $y$); the cell label reports the
           percent of $W$'s in that band, and the fill is on a
           log-scaled viridis from $\le 0.01\%$ to $100\%$. The
           cell label $100$ is reserved for columns in which
           \emph{every} $W$ lands in that ratio band; otherwise the
           label is the two-decimal percentage. At $K = 4$ the
           argmax is brittle (top-row occupancy falls to $69.71\%$
           at the highest KL band); at $K \in \{32,\,128\}$ the
           rightmost band still keeps $\sim\!84\%$ of $W$'s
           within $10\%$ of $\mathrm{score}_A$; at $K = 512$ the rightmost band keeps
           $87.87\%$, marginally the most robust regime.}
  \label{fig:RandomH:KGrid}
\end{figure}

\paragraph{Experiment.}
We work on the data of the four models on each of our three datasets, LiveCodeBench, MATH-500, and Hi-ToM, restricted to the included questions whose peak pass-rate is at least $0.1\%$ on the temperature grid.
On LiveCodeBench, these are $766$ questions for GPT OSS 20B, $359$ for Mistral 7B Instruct, $647$ for Qwen3-8B, and $85$ for Qwen3-0.6B.
In the same order, they are $500$, $455$, $496$, and $492$ on MATH-500, and $678$, $615$, $466$, and $743$ on Hi-ToM.
For Qwen3-8B, we use a later release of its LiveCodeBench data, with $119{,}228{,}250$ samples, in which $647$ questions pass the filter.
The table of Appendix \ref{zap:DataDis:qwen3_8b} counts the earlier release, with $100{,}568{,}250$ samples, in which $646$ questions pass it.
For each dataset, each model, and each $K \in \{4,\,32,\,128,\,512\}$, we compare two ways of choosing the best single temperature:
\begin{itemize}
  \item \textbf{Reference $A$.} The best single $\tau_A$ as defined
        in Section \ref{sec:Multiple:Comp}, picked under uniform
        per-question weights.
  \item \textbf{Reweighted $B$.} A random per-question weight vector $W$ is drawn from a Dirichlet smooth-base sampler over the included questions of the model; $\tau_B$ is the same argmax as in Reference $A$, but with each per-question term reweighted by $W$.
\end{itemize}
Crucially, only the \emph{choice} of $\tau_B$ uses $W$. Once
$\tau_B$ is chosen, $\mathrm{score}_B$ is computed back under
uniform per-question weights, so $\mathrm{score}_B$ and
$\mathrm{score}_A$ live on the same scale and are directly
comparable. We report the ratio
$\mathrm{score}_B / \mathrm{score}_A \in (0,\,1]$ as a function of
$\mathrm{KL}(W \,\|\, \mathrm{uniform})$.
For each dataset and model, we draw $10{,}000$ random $W$'s with the same Dirichlet smooth-base sampler and seed, in log space so that no $W$ underflows to zero, elect $\tau_B$ in log space so that no pass@$K$ rounds to $1$, and reuse the $W$'s across the four $K$ panels of its figure, so any difference between panels is attributable to $K$ alone. Figure \ref{fig:Prot:D:Main}(b) repeats the top row of Figure \ref{fig:RandomH:KGrid} with $24$ million $W$'s; this moves its cells by at most $1.0$ points.

The Dirichlet smooth-base sampler sweeps $\mathrm{KL}(W \,\|\, \mathrm{uniform})$ from $0$ nats up to the entropy ceiling $\ln n$, where $n$ is the number of included questions, so every panel covers the full reachable KL range.
On LiveCodeBench, the ceilings are $\ln(766) \approx 6.64$ nats for GPT OSS 20B, $\ln(359) \approx 5.88$ for Mistral 7B Instruct, $\ln(647) \approx 6.47$ for Qwen3-8B, and $\ln(85) \approx 4.44$ for Qwen3-0.6B.
On MATH-500 and Hi-ToM, they run from $\ln(455) \approx 6.12$ to $\ln(743) \approx 6.61$ nats, and the last KL band of each figure ends at its own.

\begin{figure}[tbp]
{\par
  \centering
  \includegraphics[width=\linewidth]{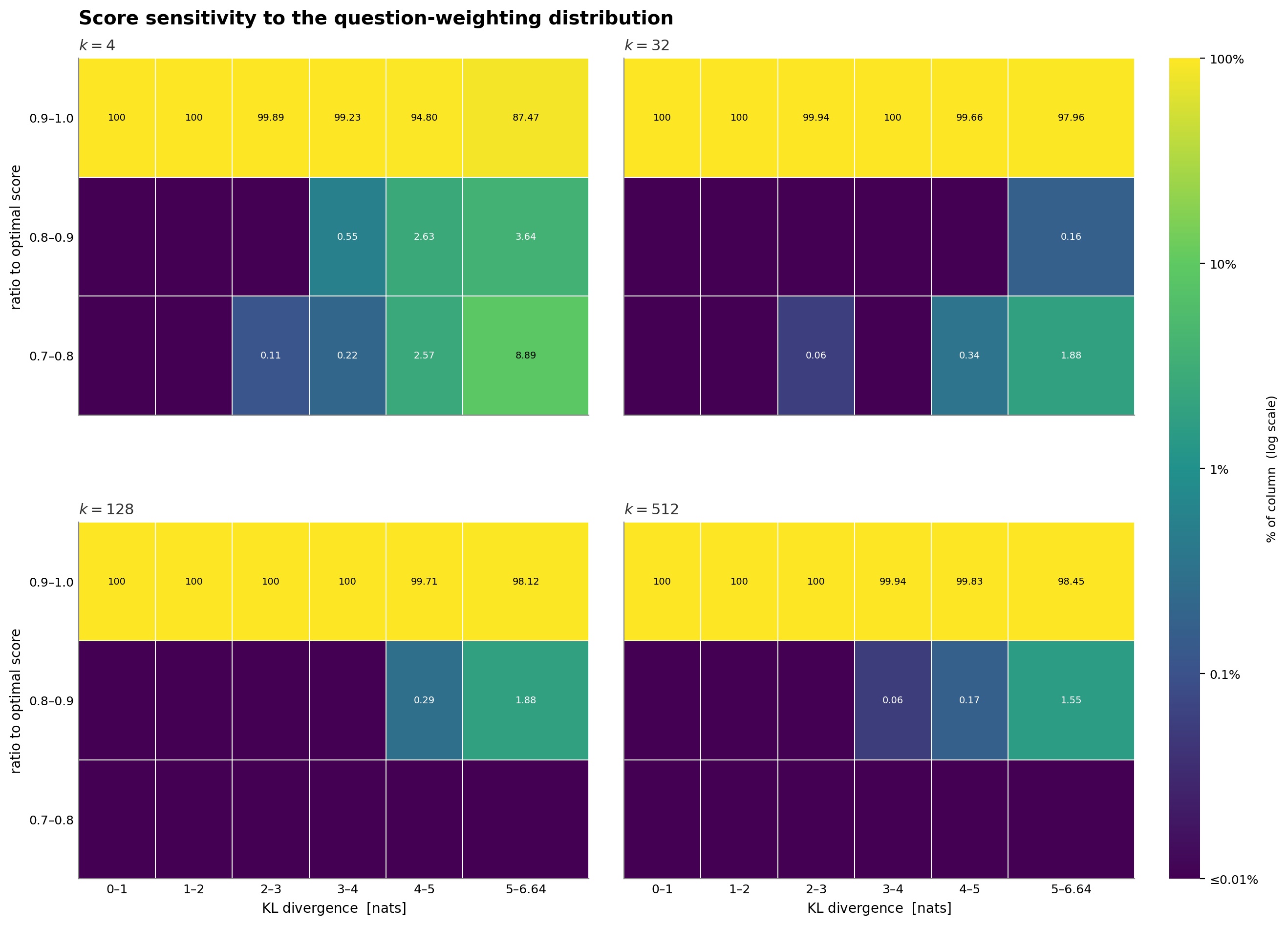}
  \caption{Distribution of $\mathrm{score}_B / \mathrm{score}_A$ as a function of $\mathrm{KL}(W \,\|\, \mathrm{uniform})$ for GPT OSS 20B ($766$ included questions, entropy ceiling $\ln(766) \approx 6.64$ nats), constructed as in Figure \ref{fig:RandomH:KGrid}.}
  \label{fig:RandomH:KGrid:GPToss}
\par}
\end{figure}

\begin{figure}[tbp]
{\par
  \centering
  \includegraphics[width=\linewidth]{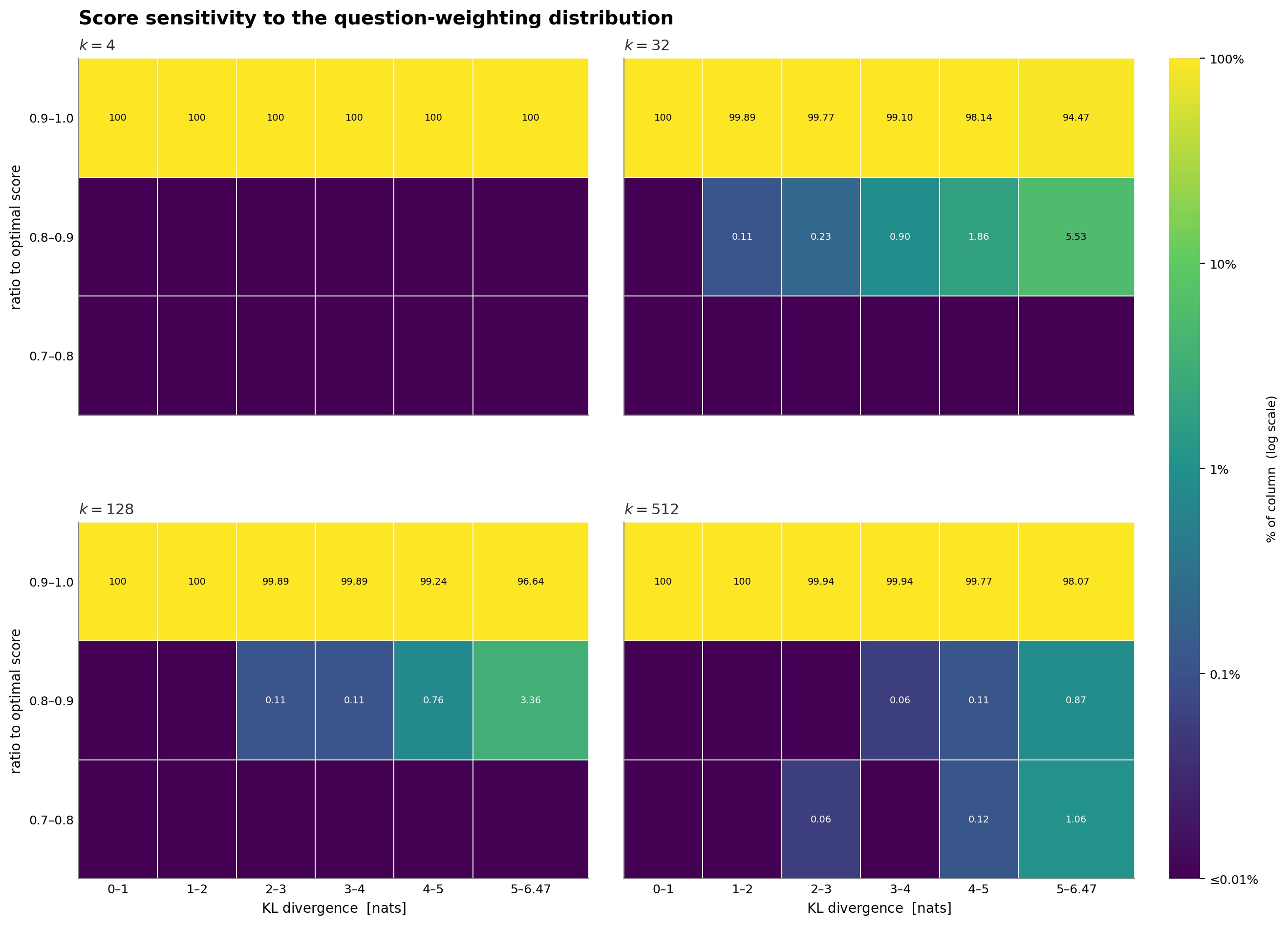}
  \caption{Distribution of $\mathrm{score}_B / \mathrm{score}_A$ as a function of $\mathrm{KL}(W \,\|\, \mathrm{uniform})$ for Qwen3-8B ($647$ included questions, entropy ceiling $\ln(647) \approx 6.47$ nats, later data release, see text), constructed as in Figure \ref{fig:RandomH:KGrid}.}
  \label{fig:RandomH:KGrid:Qwen8B}
\par}
\end{figure}

\begin{figure}[tbp]
{\par
  \centering
  \includegraphics[width=\linewidth]{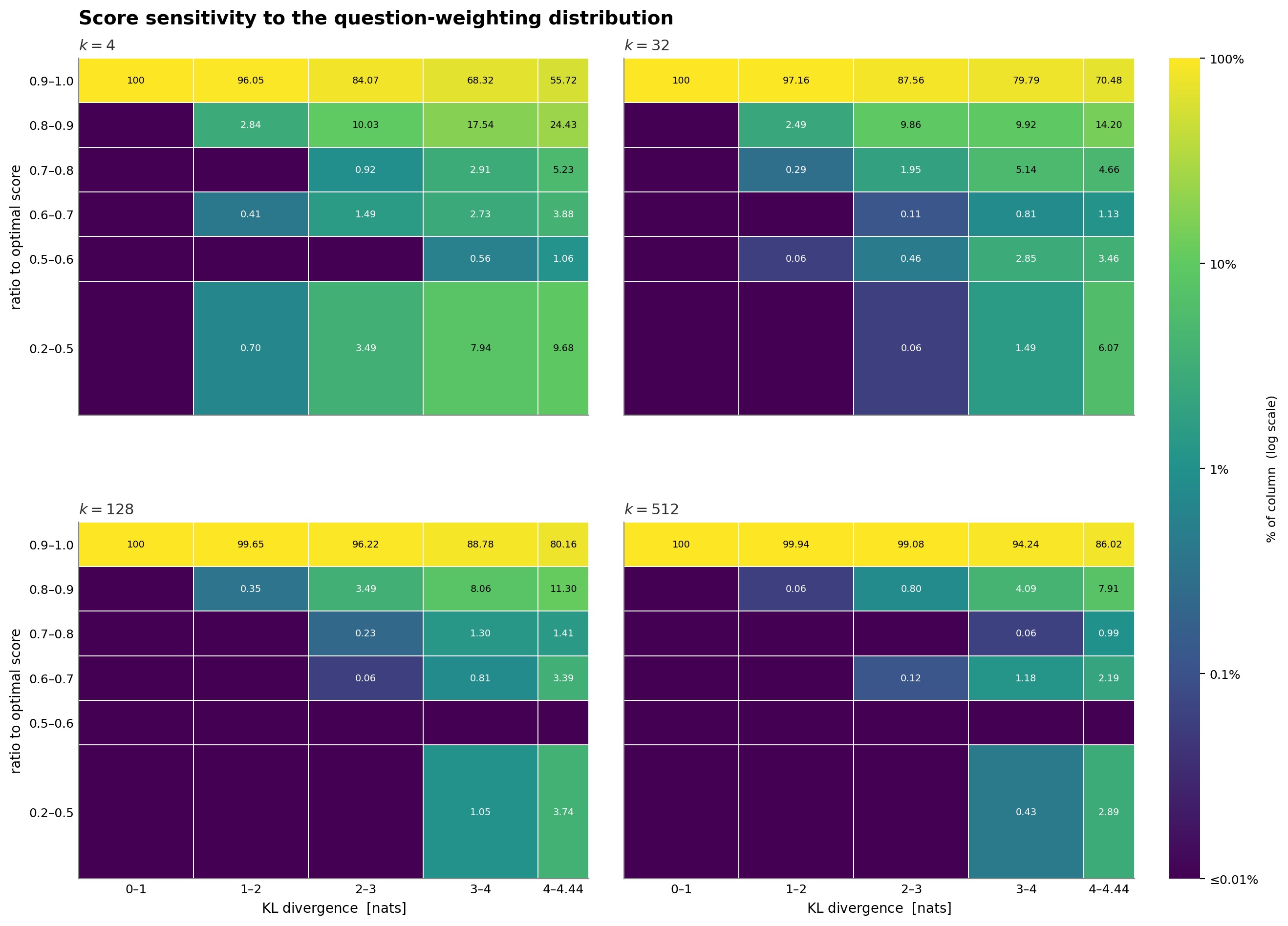}
  \caption{Distribution of $\mathrm{score}_B / \mathrm{score}_A$ as a function of $\mathrm{KL}(W \,\|\, \mathrm{uniform})$ for Qwen3-0.6B ($85$ included questions, entropy ceiling $\ln(85) \approx 4.44$ nats), constructed as in Figure \ref{fig:RandomH:KGrid}.}
  \label{fig:RandomH:KGrid:Qwen06B}
\par}
\end{figure}

\begin{figure}[p]
  \centering
  \includegraphics[width=\linewidth]{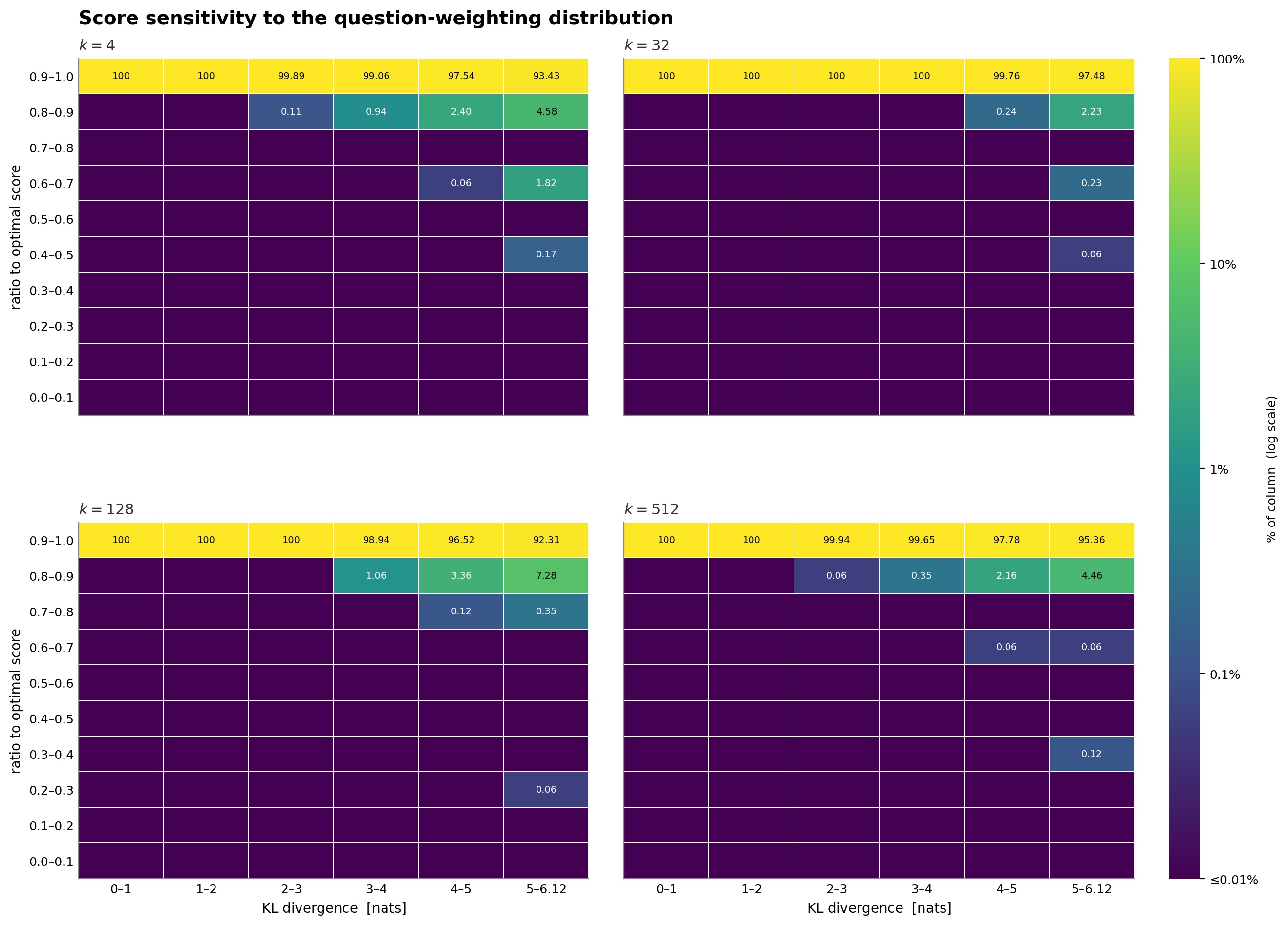}
  \caption{Mistral 7B Instruct on MATH-500 ($455$ included questions), constructed as in Figure \ref{fig:RandomH:KGrid}.}
  \label{fig:RandomH:KGrid:MATH:Mistral}
  \par\vspace{10pt}

  \centering
  \includegraphics[width=\linewidth]{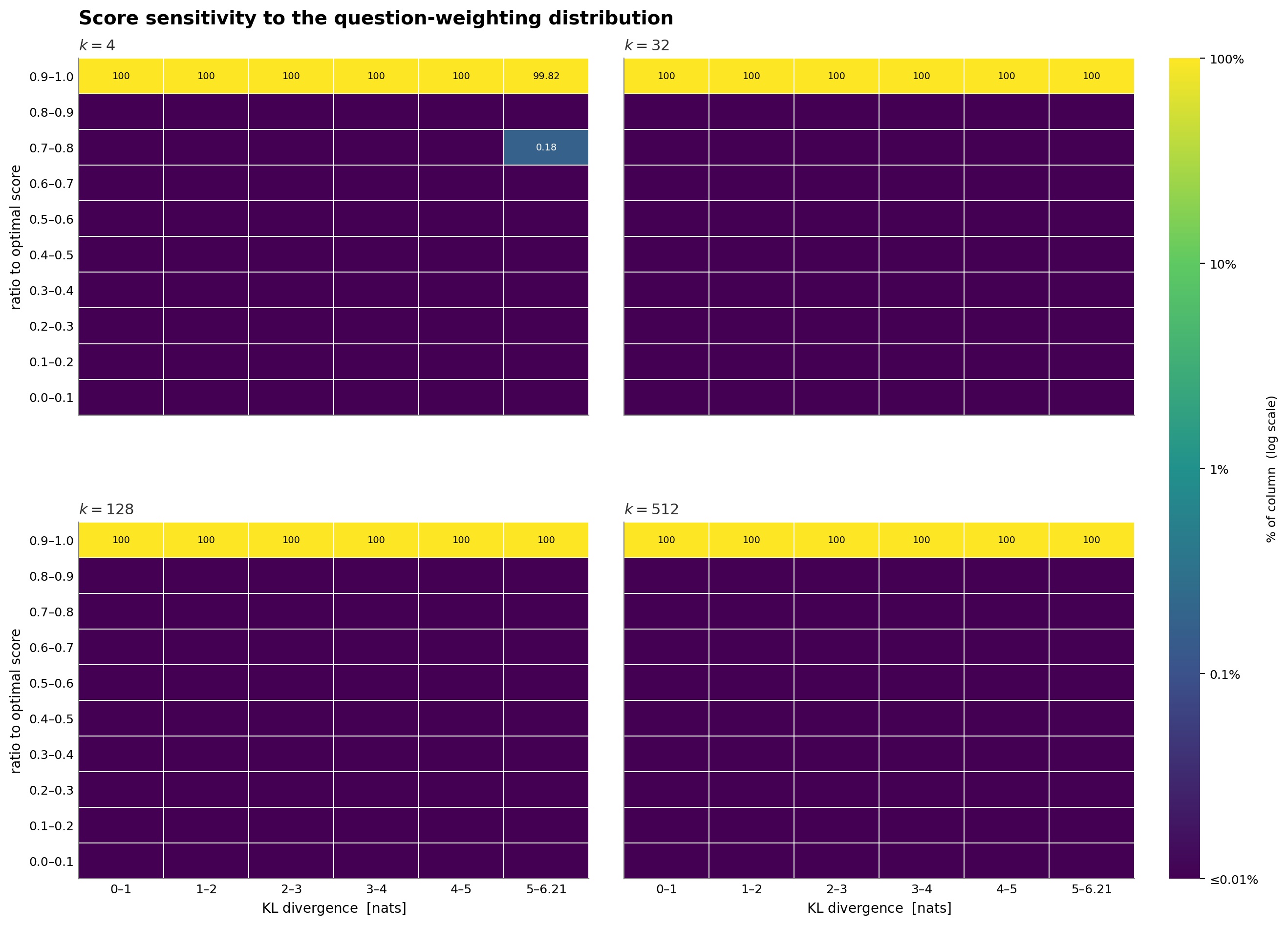}
  \caption{GPT OSS 20B on MATH-500 ($500$ included questions), constructed as in Figure \ref{fig:RandomH:KGrid}.}
  \label{fig:RandomH:KGrid:MATH:GPToss}
\end{figure}

\begin{figure}[p]
  \centering
  \includegraphics[width=\linewidth]{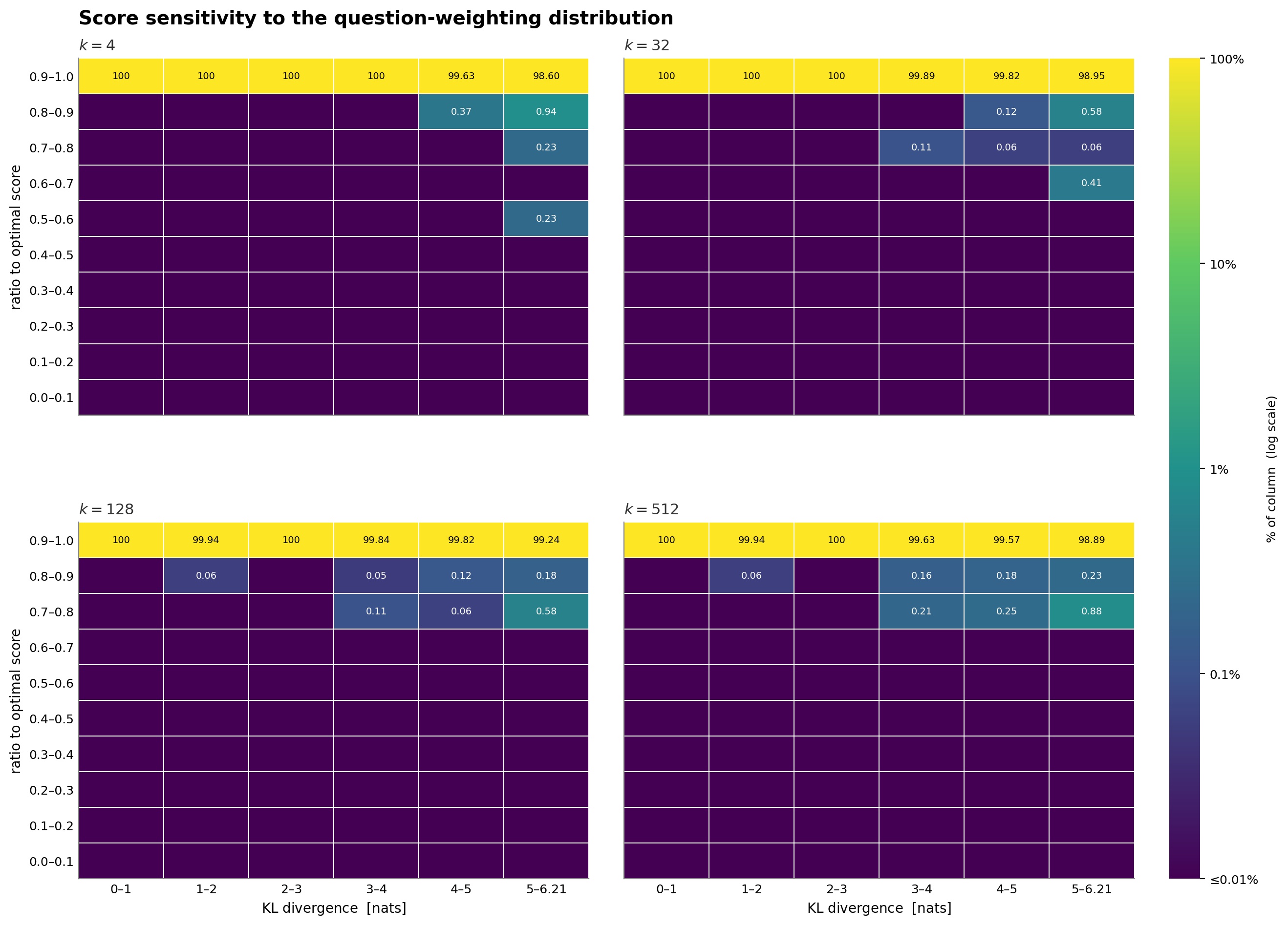}
  \caption{Qwen3-8B on MATH-500 ($496$ included questions), constructed as in Figure \ref{fig:RandomH:KGrid}.}
  \label{fig:RandomH:KGrid:MATH:Qwen8B}
  \par\vspace{10pt}

  \centering
  \includegraphics[width=\linewidth]{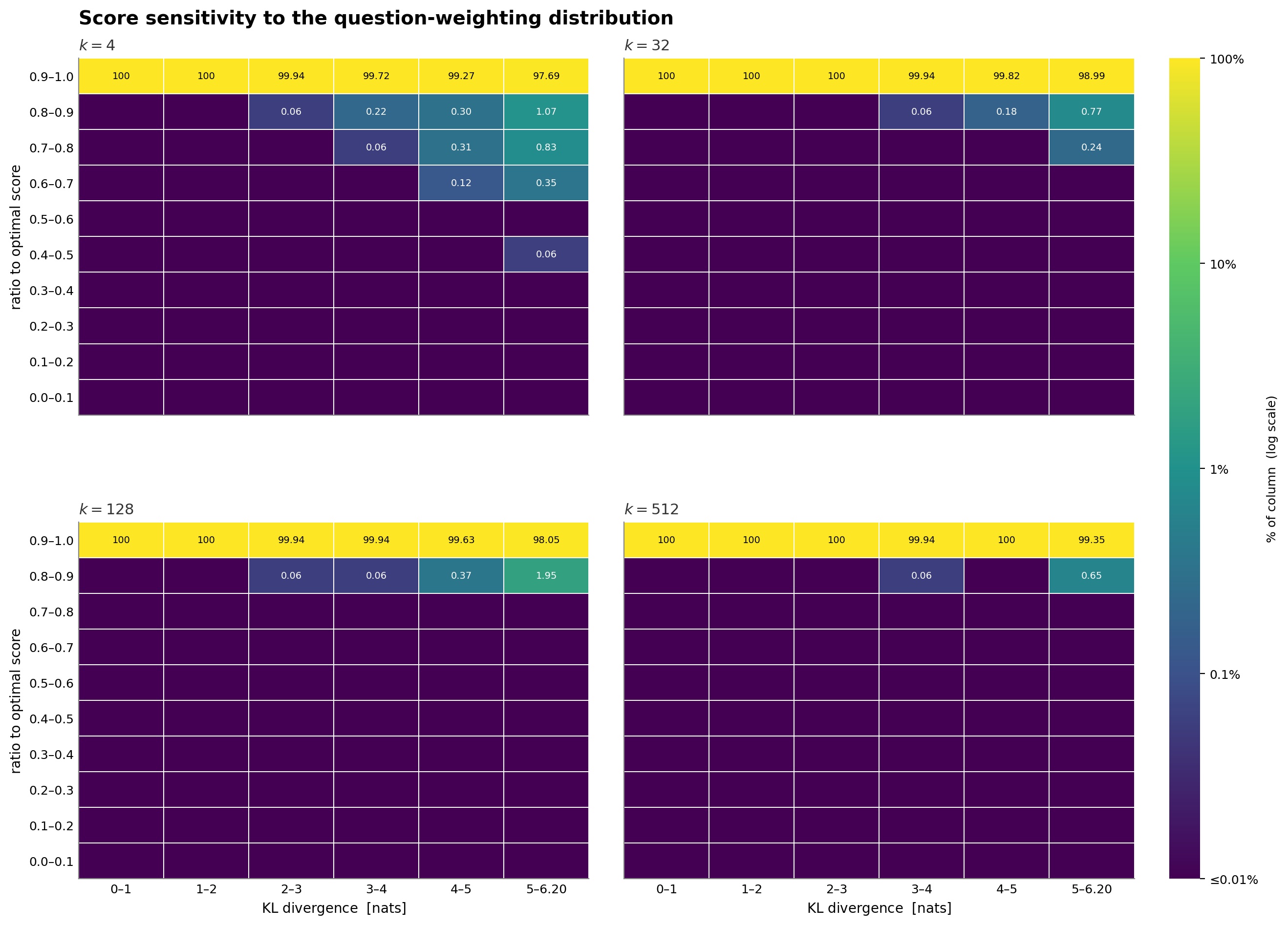}
  \caption{Qwen3-0.6B on MATH-500 ($492$ included questions), constructed as in Figure \ref{fig:RandomH:KGrid}.}
  \label{fig:RandomH:KGrid:MATH:Qwen06B}
\end{figure}

\begin{figure}[p]
  \centering
  \includegraphics[width=\linewidth]{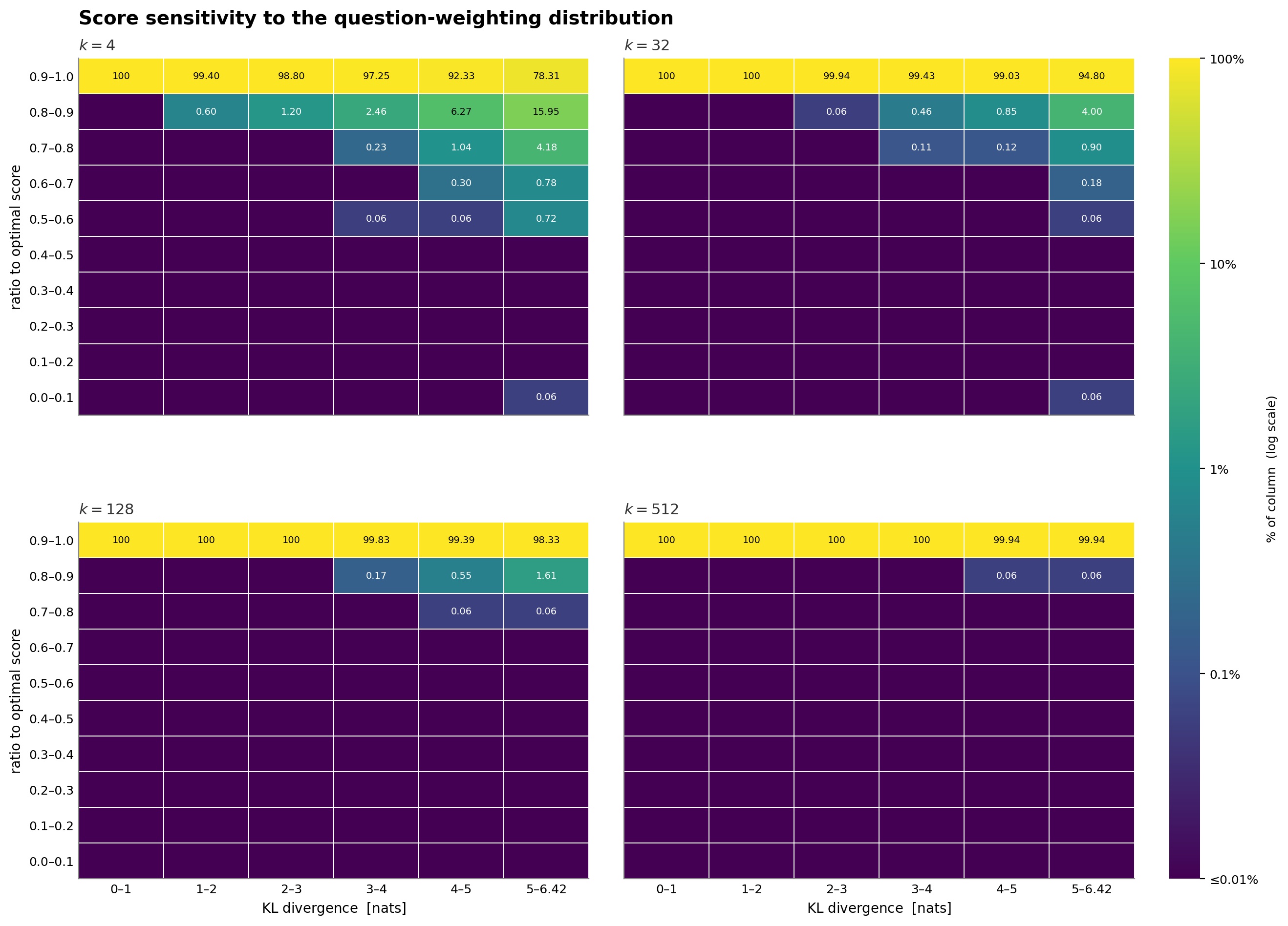}
  \caption{Mistral 7B Instruct on Hi-ToM ($615$ included questions), constructed as in Figure \ref{fig:RandomH:KGrid}.}
  \label{fig:RandomH:KGrid:HiToM:Mistral}
  \par\vspace{10pt}

  \centering
  \includegraphics[width=\linewidth]{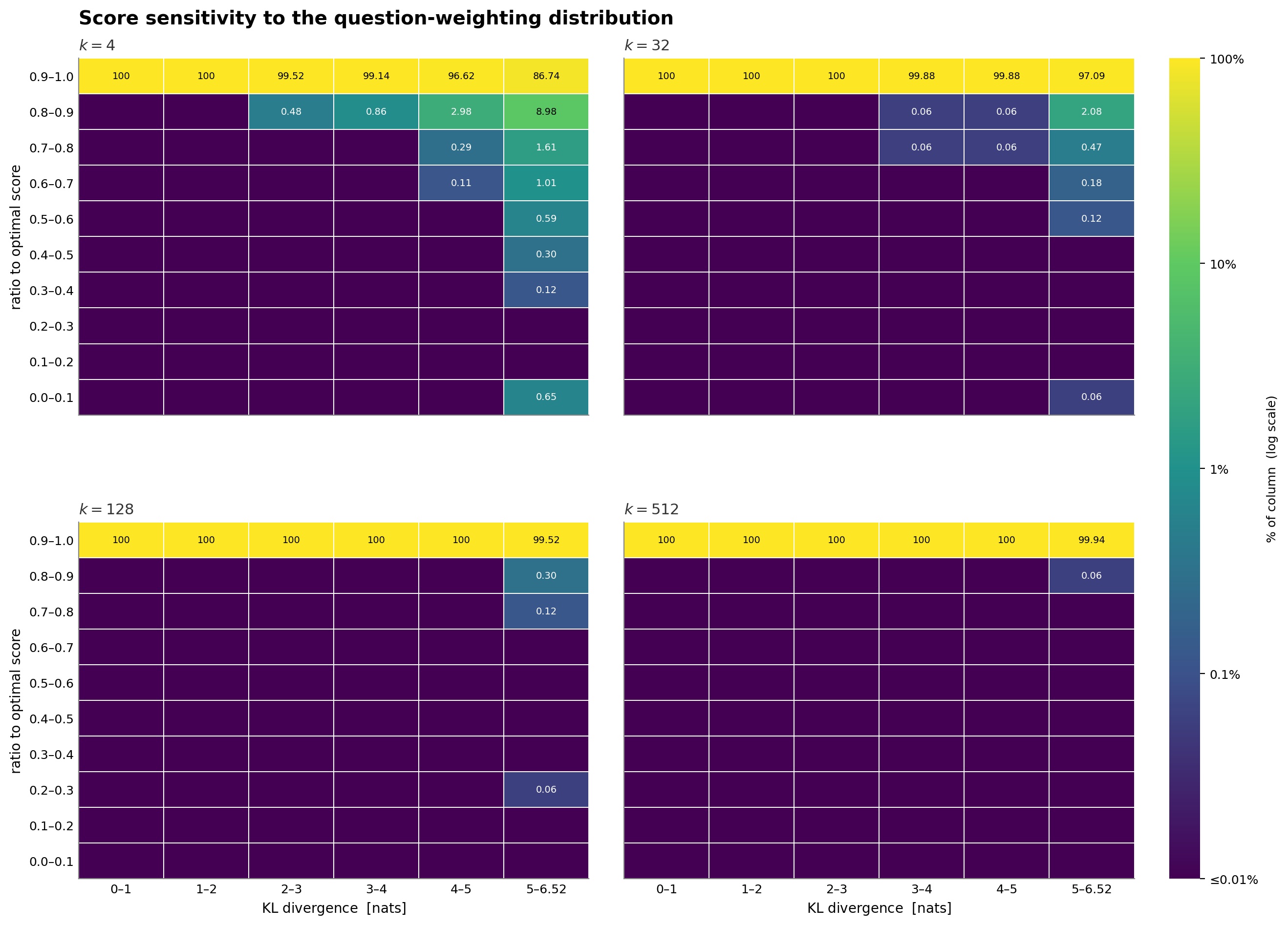}
  \caption{GPT OSS 20B on Hi-ToM ($678$ included questions), constructed as in Figure \ref{fig:RandomH:KGrid}.}
  \label{fig:RandomH:KGrid:HiToM:GPToss}
\end{figure}

\begin{figure}[p]
  \centering
  \includegraphics[width=\linewidth]{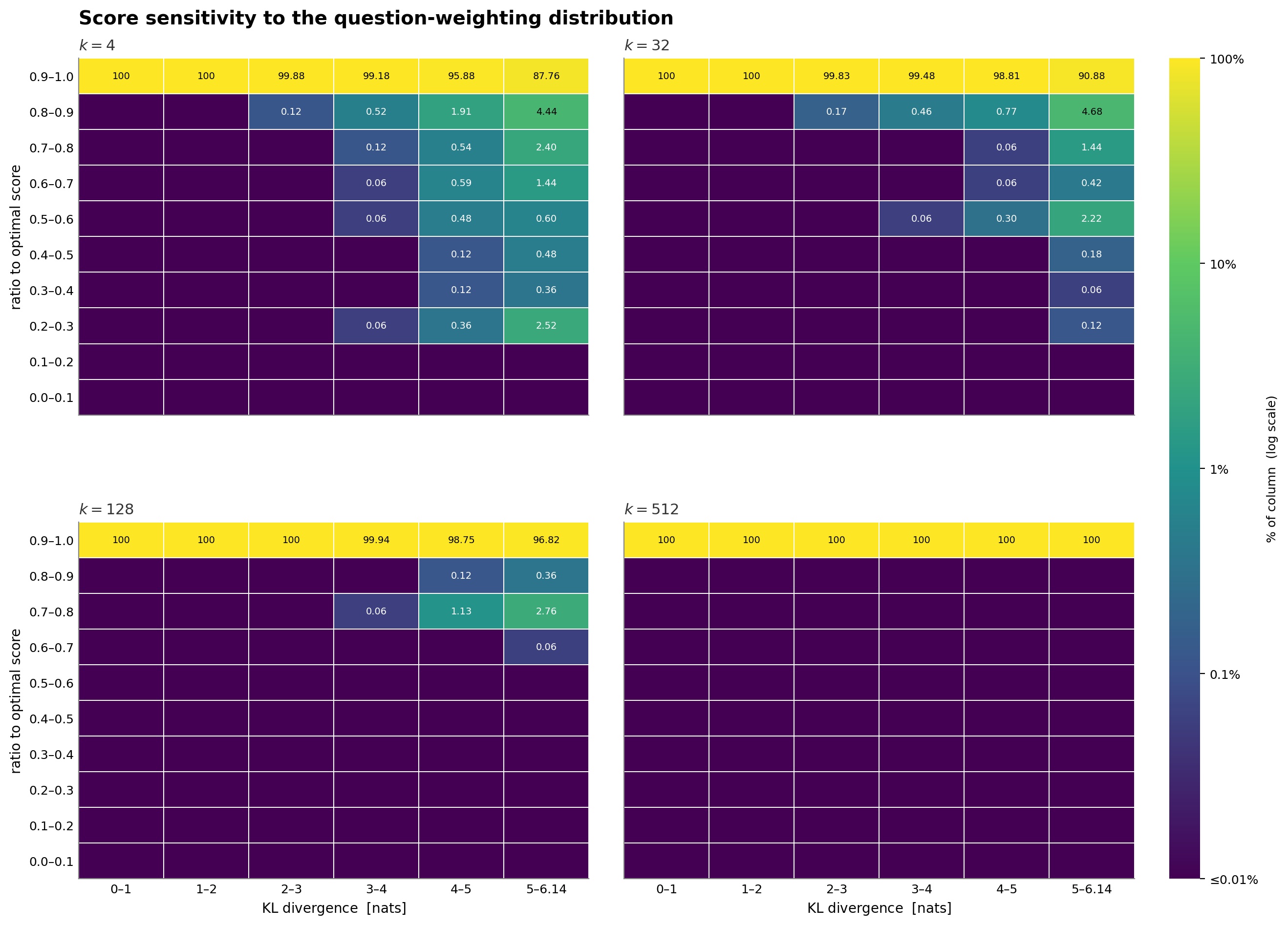}
  \caption{Qwen3-8B on Hi-ToM ($466$ included questions), constructed as in Figure \ref{fig:RandomH:KGrid}.}
  \label{fig:RandomH:KGrid:HiToM:Qwen8B}
  \par\vspace{10pt}

  \centering
  \includegraphics[width=\linewidth]{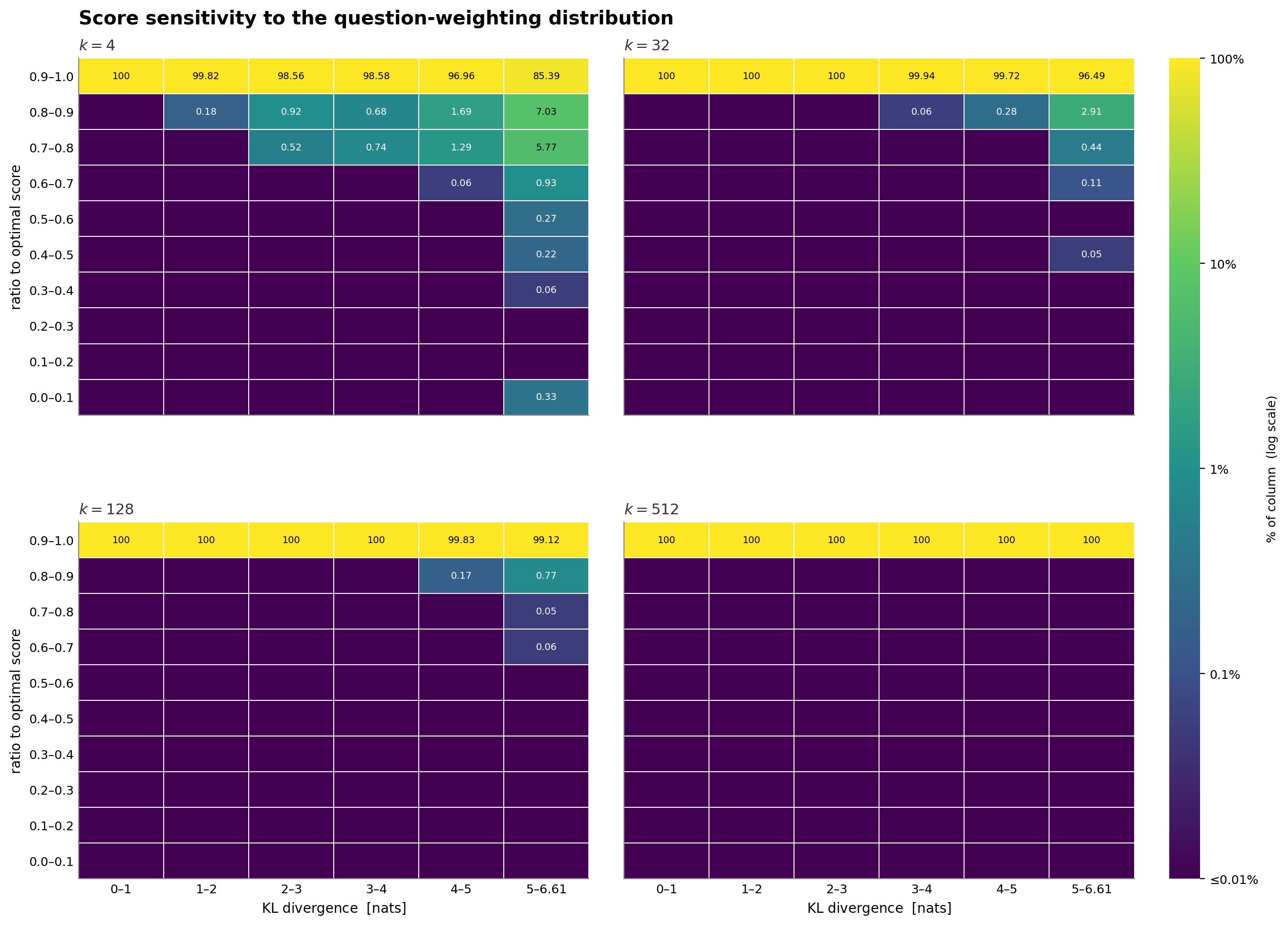}
  \caption{Qwen3-0.6B on Hi-ToM ($743$ included questions), constructed as in Figure \ref{fig:RandomH:KGrid}.}
  \label{fig:RandomH:KGrid:HiToM:Qwen06B}
\end{figure}

The figures plot the $W$'s with $0 \le \mathrm{KL} < \ln n$, and leave out the near one-hot $W$'s with KL at the ceiling.
On LiveCodeBench, these are $97$ $W$'s for GPT OSS 20B and $167$ for Mistral 7B Instruct, all at the ceiling.
The Qwen3-8B figure shows all $10{,}000$.
Of the $10{,}000$ $W$'s of Qwen3-0.6B on LiveCodeBench, the figure leaves out $2{,}005$, all near one-hot $W$'s at the ceiling, and plots the other $7{,}995$.
On Hi-ToM, the figures leave out $392$ $W$'s for GPT OSS 20B and $91$ for Qwen3-0.6B, all at the ceiling, and the six other figures of MATH-500 and Hi-ToM leave out none.
These two figures thus plot $9{,}608$ and $9{,}909$ $W$'s.

\paragraph{Reading the plot.}
Each panel of Figures \ref{fig:RandomH:KGrid} to \ref{fig:RandomH:KGrid:HiToM:Qwen06B} is a column-normalized heatmap.
The $x$-axis is $\mathrm{KL}(W \,\|\, \mathrm{uniform})$ in nats, bucketed into $0$-$1$, $1$-$2$, $2$-$3$, $3$-$4$, $4$-$5$, and $5$ up to the entropy ceiling, which is the right edge: $5$-$6.64$ for GPT OSS 20B, $5$-$5.88$ for Mistral 7B Instruct, and $5$-$6.47$ for Qwen3-8B on LiveCodeBench, and $5$ up to the ceiling for every model on MATH-500 and Hi-ToM.
For Qwen3-0.6B, the ceiling $\ln(85) \approx 4.44$ is below $5$, so its KL bands are $0$-$1$, $1$-$2$, $2$-$3$, $3$-$4$, and $4$-$4.44$.
The $y$-axis is $\mathrm{score}_B / \mathrm{score}_A$, bucketed into ratio bands.
On LiveCodeBench, the ratio bands start at each model's lowest ratio over its four panels, rounded down to a multiple of $0.1$.
They are $0.4$-$0.5$, $\ldots$, $0.9$-$1.0$ for Mistral 7B Instruct, $0.7$-$0.8$, $0.8$-$0.9$, and $0.9$-$1.0$ for GPT OSS 20B and Qwen3-8B, and $0.2$-$0.5$, $0.5$-$0.6$, $\ldots$, $0.9$-$1.0$ for Qwen3-0.6B.
On MATH-500 and Hi-ToM, every heatmap uses the same ten ratio bands of width $0.1$, from $0.0$-$0.1$ to $0.9$-$1.0$, like the equal bands of Figures \ref{fig:minp:RandomH:q15} to \ref{fig:minp:RandomH:mis}.
Each cell
reports the percentage of $W$'s that, conditional on landing in
the cell's KL band, ended up in the cell's ratio band; by
construction every column sums to exactly $100$. Cell labels are
two-decimal percentages, with one exception: the cell shows
$100$ only when \emph{every} $W$ in that column landed in that ratio band.
A column where $9{,}999$ out of $10{,}000$ $W$'s landed in it
reads $99.99$, never $100$. The fill is a log-scaled
viridis from $\le 0.01\%$ (cell floor) to $100\%$.

\subsection{\texorpdfstring{Extended Results}{Extended Results}}
\label{zap:DataDis:RandomH:res}

Figures \ref{fig:RandomH:KGrid}, \ref{fig:RandomH:KGrid:GPToss}, \ref{fig:RandomH:KGrid:Qwen8B}, and \ref{fig:RandomH:KGrid:Qwen06B} show the heatmaps of Mistral 7B Instruct, GPT OSS 20B, Qwen3-8B, and Qwen3-0.6B on LiveCodeBench.
Figures \ref{fig:RandomH:KGrid:MATH:Mistral} to \ref{fig:RandomH:KGrid:MATH:Qwen06B} show them on MATH-500, and Figures \ref{fig:RandomH:KGrid:HiToM:Mistral} to \ref{fig:RandomH:KGrid:HiToM:Qwen06B} on Hi-ToM, in the same order.

\paragraph{\texorpdfstring{Findings on LiveCodeBench.}{Findings on LiveCodeBench.}}
Robustness of the optimal temperature is itself a function of the
pass@$K$ regime. At every $K$, low KL keeps $\mathrm{score}_B / \mathrm{score}_A$ near $1$: in the lowest KL band, at least $99.8\%$ of the $W$'s of each model have a ratio of $0.9$ or more.
What changes with $K$ is the rate at
which high-KL columns slide downward.
In every panel, the top row holds the $W$'s with ratio $\ge 0.9$, i.e., whose $\mathrm{score}_B$ is within $10\%$ of $\mathrm{score}_A$.
\begin{itemize}
  \item At $K = 4$, the top-row occupancies in the $1$-$2$ KL band and the highest band are $100\%$ and $87.47\%$ for GPT OSS 20B, $98.08\%$ and $69.71\%$ for Mistral 7B Instruct, $100\%$ in both for Qwen3-8B, and $96.05\%$ and $55.72\%$ for Qwen3-0.6B.
        For Qwen3-8B at $K = 4$, all $10{,}000$ $W$'s have ratio $\ge 0.9$ (the lowest is $0.905$), so its whole top row reads $100$.
        For Mistral 7B Instruct, the choice at $K = 4$ is brittle.
        The top-row occupancy
        already drops to $99.86\%$ in the lowest KL band, then
        slides monotonically to $69.71\%$ at the entropy ceiling.
        The flat per-question pass curves at small $K$ make the
        argmax easy to move with small reweightings.
  \item At $K = 32$ and $K = 128$, the table below gives the top-row occupancy in the highest KL band for each model.
        For Mistral 7B Instruct, the picture at these two values of $K$ is similar: the top row stays above $98\%$ up through $\mathrm{KL} \le 3$ nats and dips to roughly $84\%$ and $83\%$ respectively in the highest KL band.
\end{itemize}

{\par\ifdefined\nolinenumbers\nolinenumbers\fi %
The following table gives, for each model and $K$, the reference temperature $\tau_A$, its score $\mathrm{score}_A$, and the top-row occupancy (ratio $0.9$-$1.0$, in $\%$) in the highest KL band, under ``top''. From left to right, this band is $5$-$6.64$, $5$-$5.88$, $5$-$6.47$, and $4$-$4.44$ nats.
\begin{center}
\small
\setlength{\tabcolsep}{3pt}
\begin{tabular}{@{}c ccc ccc ccc ccc@{}}
  \toprule
   & \multicolumn{3}{c}{GPT OSS 20B} & \multicolumn{3}{c}{Mistral 7B Instruct} & \multicolumn{3}{c}{Qwen3-8B} & \multicolumn{3}{c}{Qwen3-0.6B} \\
  \cmidrule(lr){2-4} \cmidrule(lr){5-7} \cmidrule(lr){8-10} \cmidrule(lr){11-13}
  $K$ & $\tau_A$ & $\mathrm{score}_A$ & top & $\tau_A$ & $\mathrm{score}_A$ & top & $\tau_A$ & $\mathrm{score}_A$ & top & $\tau_A$ & $\mathrm{score}_A$ & top \\
  \midrule
  $4$   & $0.90$ & $505.86$ & $87.47$ & $0.70$ & $110.44$ & $69.71$ & $1.20$ & $454.64$ & $100$ & $0.70$ & $13.77$ & $55.72$ \\
  $32$  & $0.80$ & $645.46$ & $97.96$ & $0.90$ & $203.60$ & $84.18$ & $1.20$ & $542.34$ & $94.47$ & $0.70$ & $40.06$ & $70.48$ \\
  $128$ & $0.80$ & $701.40$ & $98.12$ & $0.90$ & $267.11$ & $83.31$ & $1.10$ & $585.45$ & $96.64$ & $0.70$ & $57.35$ & $80.16$ \\
  $512$ & $0.90$ & $742.11$ & $98.45$ & $1.00$ & $319.77$ & $87.87$ & $1.10$ & $618.73$ & $98.07$ & $0.70$ & $71.92$ & $86.02$ \\
  \bottomrule
\end{tabular}
\end{center}
\par}

\begin{itemize}
  \item At $K = 512$, the same occupancies are $100\%$ and $98.45\%$ for GPT OSS 20B, $100\%$ and $87.87\%$ for Mistral 7B Instruct, $100\%$ and $98.07\%$ for Qwen3-8B, and $99.94\%$ and $86.02\%$ for Qwen3-0.6B.
        For Mistral 7B Instruct, the choice at $K = 512$ is marginally the most robust.
        The top-row
        occupancies essentially match $K = 32$ and $K = 128$
        across the lower KL bands, and $87.87\%$ of $W$'s in the
        highest KL band still lose under $10\%$.
\end{itemize}

\paragraph{\texorpdfstring{Findings on MATH-500 and Hi-ToM.}{Findings on MATH-500 and Hi-ToM.}}
On MATH-500, the top-row occupancy in the highest KL band is at least $92.31\%$ for every model and $K$.
It is lowest for Mistral 7B Instruct at every $K$, with $93.43\%$, $97.48\%$, $92.31\%$, and $95.36\%$ at $K = 4$, $32$, $128$, and $512$.
On Hi-ToM, it is between $78.31\%$ for Mistral 7B Instruct and $87.76\%$ for Qwen3-8B at $K = 4$, and it rises at $K = 32$ and again at $K = 128$ for every model.
At $K = 512$, it is between $99.94\%$ for Mistral 7B Instruct and GPT OSS 20B and $100\%$ for Qwen3-8B and Qwen3-0.6B.

\FloatBarrier

\section{\texorpdfstring{Sensitivity to Reweighting Outside a Protected Hardness Window}{Sensitivity to Reweighting Outside a Protected Hardness Window}}
\label{zap:DataDis:Prot}

In Appendix \ref{zap:DataDis:RandomH}, we reweighted all questions at random. In this appendix, we test the Goldilocks sampling effect directly. For each $K$, we protect the questions whose hardness is within $D$ decades of $K$ by keeping their weights at $W_q = 1$, and let the weight of every other question range over $[1/\beta, \beta]$. We then find $\beta^*$, the smallest $\beta$ at which a temperature that loses at least $10\%$ under uniform weights becomes the best. We show $\beta^*$ against $D$ for each dataset, model, and $K$ in Figures \ref{fig:Prot:D:LCB}, \ref{fig:Prot:D:MATH}, and \ref{fig:Prot:D:HiToM}. Figure \ref{fig:Prot:D:Main}(a) in Section \ref{sec:gold} shows the curves of Mistral 7B Instruct on LiveCodeBench at a $10\%$ loss.

\subsection{\texorpdfstring{Statistical Analysis Approach}{Statistical Analysis Approach}}
\label{zap:DataDis:Prot:The}

\paragraph{\texorpdfstring{Test.}{Test.}}
For a question $q$, let $s_q(\tau) = 1 - \bar{P}_q(\tau)^K$ be its pass@$K$ at temperature $\tau$, and for weights $W$ let $S_W(\tau) = \sum_q W_q s_q(\tau)$. As in Appendix \ref{zap:DataDis:RandomH}, $\tau_A$ maximizes the score $S_U$ under uniform weights, and $\mathrm{score}_A = S_U(\tau_A)$. The temperatures that lose at least $10\%$ form the set $C = \{\tau : S_U(\tau) \leq 0.9\,\mathrm{score}_A\}$, and for a $1\%$ loss we replace $0.9$ by $0.99$. The protected window holds the questions with $|\log_{10} H_q - \log_{10} K| \leq D$, which keep $W_q = 1$. The other questions have $W_q \in [1/\beta, \beta]$, and we do not normalize $\sum_q W_q$.

A temperature $\tau_c \in C$ wins under $W$ if $S_W(\tau_c) - S_W(\tau') \geq 10^{-9}$ for every $\tau' \notin C$, so ties do not count. We define $\beta^*$ as the smallest $\beta$ on the grid $1.005^i$, capped at $10^6$, at which some $\tau_c \in C$ wins under some such $W$, and write $\beta^* > 10^6$ when no $\beta$ up to $10^6$ suffices. Since a larger $D$ protects more questions, $\beta^*$ is nondecreasing in $D$. We use $K \in \{4, 32, 128, 512\}$ and $23$ values of $D$ from $0$ to $6$, spaced more finely near $0$, and stop each curve at the first $D$ whose window holds every question.

\paragraph{\texorpdfstring{Data.}{Data.}}
We run the test on all $12$ combinations of our three datasets, LiveCodeBench, MATH-500, and Hi-ToM, with the four models. For every question, $P_q(\tau)$ is the profile fit of Appendix \ref{zap:Stats:Methodology} to the question's own sample counts, and we keep the questions that pass the global $10^{-3}$ peak success-probability filter. On MATH-500 and Hi-ToM, these are the included questions of Appendix \ref{zap:CompResources}. On LiveCodeBench, we use a corrected parsing of the sample counts, which gives $359$, $766$, $677$, and $86$ included questions for Mistral 7B Instruct, GPT OSS 20B, Qwen3-8B, and Qwen3-0.6B, respectively. Appendix \ref{zap:DataDis:RandomH} uses the earlier parsing, with $359$, $766$, $647$, and $85$. The correction changes the profiles of $4$, $18$, $197$, and $3$ questions of these models, respectively, and adds $30$ included questions for Qwen3-8B and one for Qwen3-0.6B.

\paragraph{\texorpdfstring{Exact certification.}{Exact certification.}}
We verify every point with a question outside the window in exact rational arithmetic on the stored values of $s_q(\tau)$. For each finite $\beta^*$, an explicit $W$, with $W_q = 1$ inside the window and every other weight in $[1/\beta^*, \beta^*]$, shows that some $\tau_c \in C$ wins, and a dual certificate shows that no $\tau_c \in C$ wins one grid step below. Such a certificate is a set of nonnegative weights $\lambda_{\tau'}$ on the temperatures $\tau' \notin C$ that sum to $1$. With $g_q = \sum_{\tau' \notin C} \lambda_{\tau'} \left(s_q(\tau_c) - s_q(\tau')\right)$, no $W$ allowed at $\beta$ lets $\tau_c$ win if
\begin{equation}
\colorbox{white}{$\displaystyle \sum_{q \text{ protected}} g_q + \sum_{q \text{ outside}} \max\left(\beta g_q, \frac{g_q}{\beta}\right) < 10^{-9}$} \ .
\label{zap:eq:Prot:Dual}
\end{equation}
For each point with $\beta^* > 10^6$ and questions outside the window, such certificates show that no $\tau_c \in C$ wins at $\beta = 10^6$. Over the $12$ combinations of dataset and model, this covers all $1{,}129$ finite points and all $373$ points with $\beta^* > 10^6$ and questions outside the window. An independent implementation re-checks every certificate in exact arithmetic, and every finite $\beta^*$ and every verdict $\beta^* > 10^6$ is unchanged when the pass@$K$ values are computed exactly from the fitted pass rates. These counts are those of the fitted pass rates. When we resample the sample counts of every question and temperature and refit them, the $90\%$ interval of $\beta^*$ spans, in the median, less than a factor of $2$ on each side only for $\beta^*$ below about $10^3$, and the number of points above $10^6$ has a $90\%$ interval of $365$ to $518$.

\paragraph{\texorpdfstring{The loss of the chosen temperature.}{The loss of the chosen temperature.}}
The value of $\beta^*$ shows when a temperature that loses at least $10\%$ can become the best, but not how much a reweighting can cost $\tau_A$. For the same windows and weights, we therefore also compute the worst-case loss
\begin{equation}
\colorbox{white}{$\displaystyle R(\beta, D) = \max_{W} \left( 1 - \frac{S_W(\tau_A)}{\max_{\tau} S_W(\tau)} \right)$} \ ,
\label{zap:eq:Prot:Loss}
\end{equation}
over the $W$ with $W_q = 1$ inside the window and $W_q \in [1/\beta, \beta]$ outside it. We compute $R$ exactly on the double-precision values of $s_q(\tau)$, for $K \in \{1, 4, 16, 32, 64, 128, 256, 512, 1024\}$, $\beta \in \{1.5, 2, 3, 10, 30, 100, 1000\}$, and $D \in \{0, 0.25, 0.5, 1, 1.5, 2, 3\}$.

\paragraph{\texorpdfstring{One side at a time.}{One side at a time.}}
The Goldilocks weight $\alpha e^{1-\alpha}$ of Equation \ref{eq:Gold:Dev}, with $\alpha = K/H_q$, falls exponentially in $\alpha$ only for questions much easier than $K$, where $\alpha \to \infty$. For questions harder than $K$, where $\alpha \leq 1$, it lies between $\alpha$ and $e\alpha$, and each further decade of hardness divides it by $10\, e^{-0.9\alpha}$, a factor between $4$ and $10$. We therefore also reweight one side of $K$ at a time. For $d$ from $0$ to $3$, only the questions whose hardness at their best temperature is more than $d$ decades above $K$, or only those more than $d$ decades below it, take weights in $[1/\beta, \beta]$, and every other question keeps $W_q = 1$. We find $\beta^*$ as before, with the same exact certification, for $K \in \{4, 32, 128, 512\}$ at losses of $10\%$ and $1\%$, and for every power of $2$ from $K = 1$ to $1024$ at losses of $1\%$, $5\%$, $10\%$, and $20\%$.

\paragraph{\texorpdfstring{Tilting the mix.}{Tilting the mix.}}
We also tilt every weight by hardness, with $W_q = 10^{\gamma(\log_{10} H_q - \log_{10} K)}$, which multiplies the weight by $10^{\gamma}$ per decade of hardness and favors harder questions for $\gamma > 0$. We use the $132$ combinations of dataset, model, and $K$, for every power of $2$ from $K = 1$ to $1024$.

\paragraph{\texorpdfstring{Natural sub-datasets.}{Natural sub-datasets.}}
We then test shifts that occur naturally. We take the sub-datasets that the labels of each dataset define, namely the subjects and levels of MATH-500, the order, deception, story length, and prompting type of Hi-ToM, and the difficulty, platform, and version of LiveCodeBench, together with their pairwise intersections. We keep those with at least $20$ questions, which gives $448$ combinations of sub-dataset and model, and use $K \in \{1, 4, 16, 64, 256, 1024\}$. On each sub-dataset, we compare the temperature chosen on the full dataset with the best temperature for the sub-dataset alone.

\paragraph{\texorpdfstring{The role of $K$.}{The role of K.}}
Since $s_q(\tau) = 1 - e^{-K/H_q(\tau)}$, multiplying $K$ by a factor has exactly the same effect on this test as dividing every hardness by that factor. How $\beta^*$ depends on $K$ is therefore a matter of where $K$ sits among the hardnesses of the questions.

\begin{figure}[tbp]
{\par
  \centering
  \includegraphics[width=\linewidth]{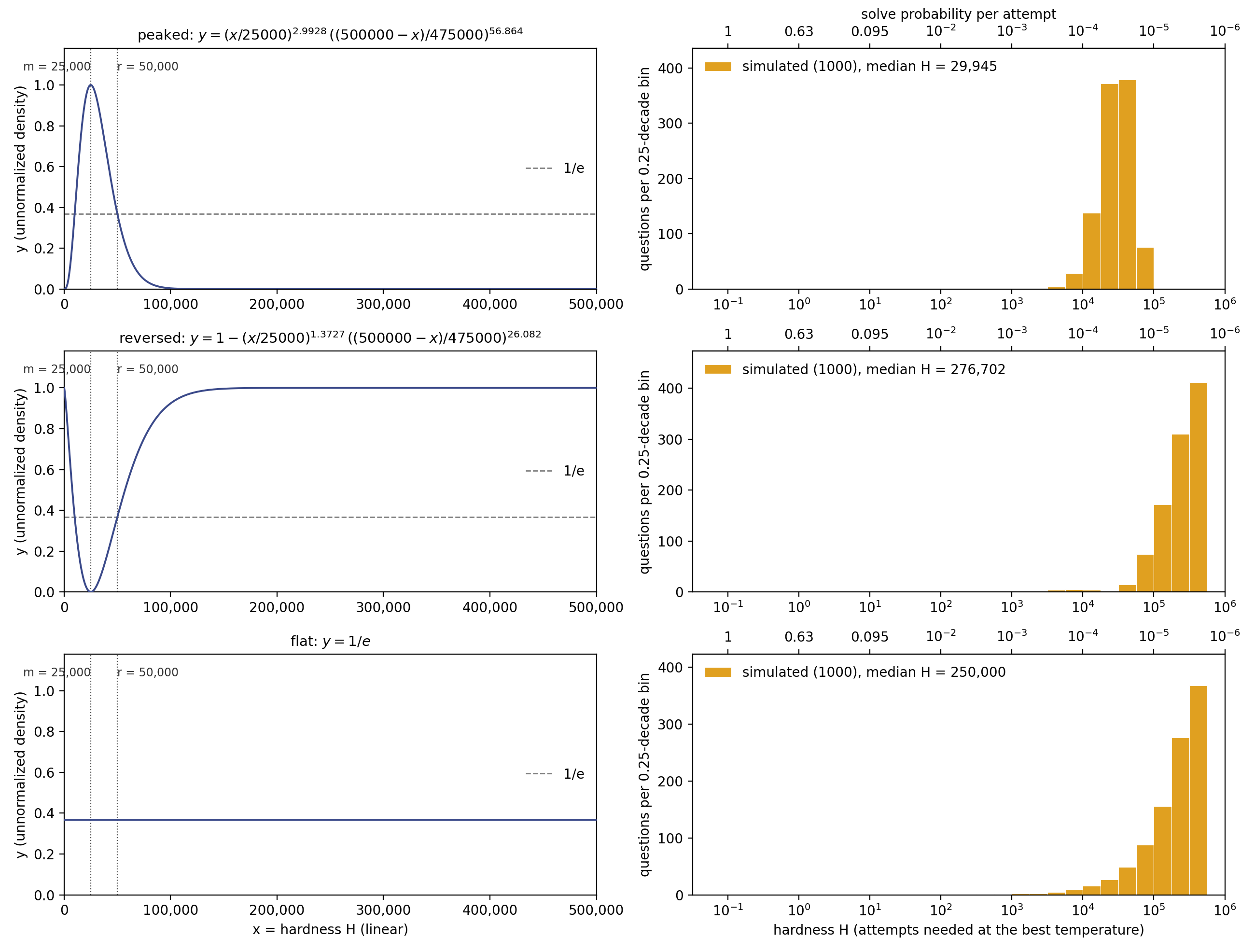}
  \caption{The three simulated hardness shapes, peaked (top), reversed (middle), and flat (bottom).}
  \label{fig:Prot:Sim:Shapes}
\par}
\end{figure}

\paragraph{\texorpdfstring{Simulated hardness distributions.}{Simulated hardness distributions.}}
The hardness of our measured questions stays below $3 \times 10^3$, so to test larger budgets, we repeat the test on three simulated datasets of $1000$ questions each. Their hardness follows one of three shapes on a linear hardness axis, up to a cutoff of $500{,}000$ (Figure \ref{fig:Prot:Sim:Shapes}). The peaked shape has its maximum at $H = 25{,}000$, the reversed shape has its minimum there, both equal $1/e$ at $H = 50{,}000$, and the flat shape is constant. We place the $i$-th question at the $(i - 0.5)/1000$ quantile of each shape, normalized as a density, which puts $966$, $985$, and $980$ of the $1000$ questions above $H = 10^4$.

In order of hardness, the simulated questions take their temperature profiles in turn from the $109$ Mistral 7B Instruct LiveCodeBench questions with hardness between $10^2$ and $10^3$. We scale each template $T$ in hardness by $\bar{P}_i(\tau) = \bar{P}_T(\tau)^{H_T/H_i}$, which gives $H_i(\tau) = (H_i/H_T)\,H_T(\tau)$ at every temperature. We use $K \in \{4, 32, 512, 10^4, 10^5, 10^6\}$ and $D$ up to $7.5$, and leave the test otherwise unchanged.

\paragraph{\texorpdfstring{Reading the plots.}{Reading the plots.}}
In Figures \ref{fig:Prot:D:LCB}, \ref{fig:Prot:D:MATH}, and \ref{fig:Prot:D:HiToM}, open markers on the top row of each panel have $\beta^* > 10^6$, and the last one of each curve is the first $D$ whose window holds every question. In Figure \ref{fig:Prot:Sim:Shapes}, the left panels show each shape on a linear hardness axis up to the cutoff of $500{,}000$, with $H = 25{,}000$ and $H = 50{,}000$ marked, and the right panels show the hardness of the $1000$ simulated questions of each shape, in bins of a quarter decade.

\subsection{\texorpdfstring{Extended Results}{Extended Results}}
\label{zap:DataDis:Prot:res}

\paragraph{\texorpdfstring{$\beta^*$ against $D$.}{beta* against D.}}
Over the $88$ curves with at least five finite points, the median least-squares slope of $\log_{10}\beta^*$ against $D$ for each dataset and loss lies between $1.36$ and $2.39$ decades per decade of $D$, and at $K = 4$ and a $10\%$ loss the slopes lie between $0.9$ and $2.5$. These slopes are averages for windows centred at $K$, with hardness measured at each question's best temperature. In $42$ of the $88$ curves, a quadratic term in $D$ has a nominal $p$-value below $0.01$ and bends the fit by more than $0.25$ decades from its chord, and in $35$ of these the curve steepens as $D$ grows.

\begin{figure}[p]
{\par
  \centering
  \begin{subfigure}[t]{0.48\textwidth}
    \centering
    \includegraphics[width=\linewidth]{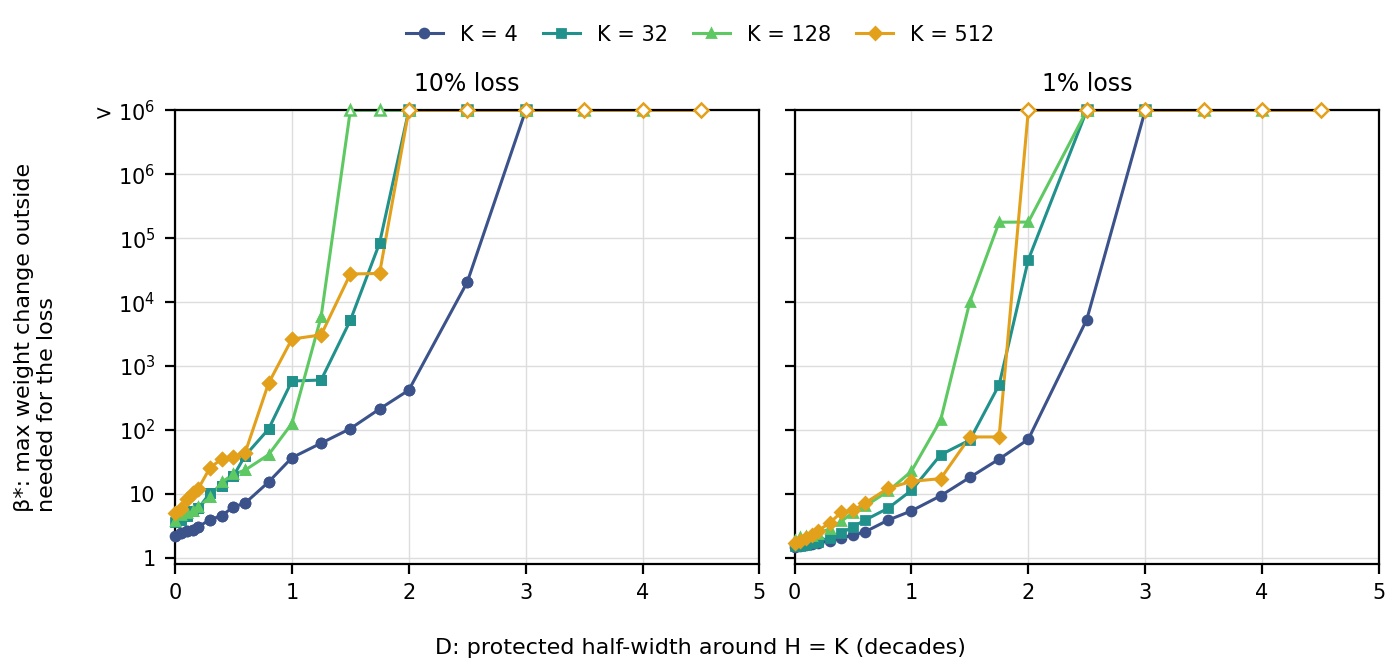}
    \caption{Mistral 7B Instruct, $n = 359$}
  \end{subfigure}\hfill  \begin{subfigure}[t]{0.48\textwidth}
    \centering
    \includegraphics[width=\linewidth]{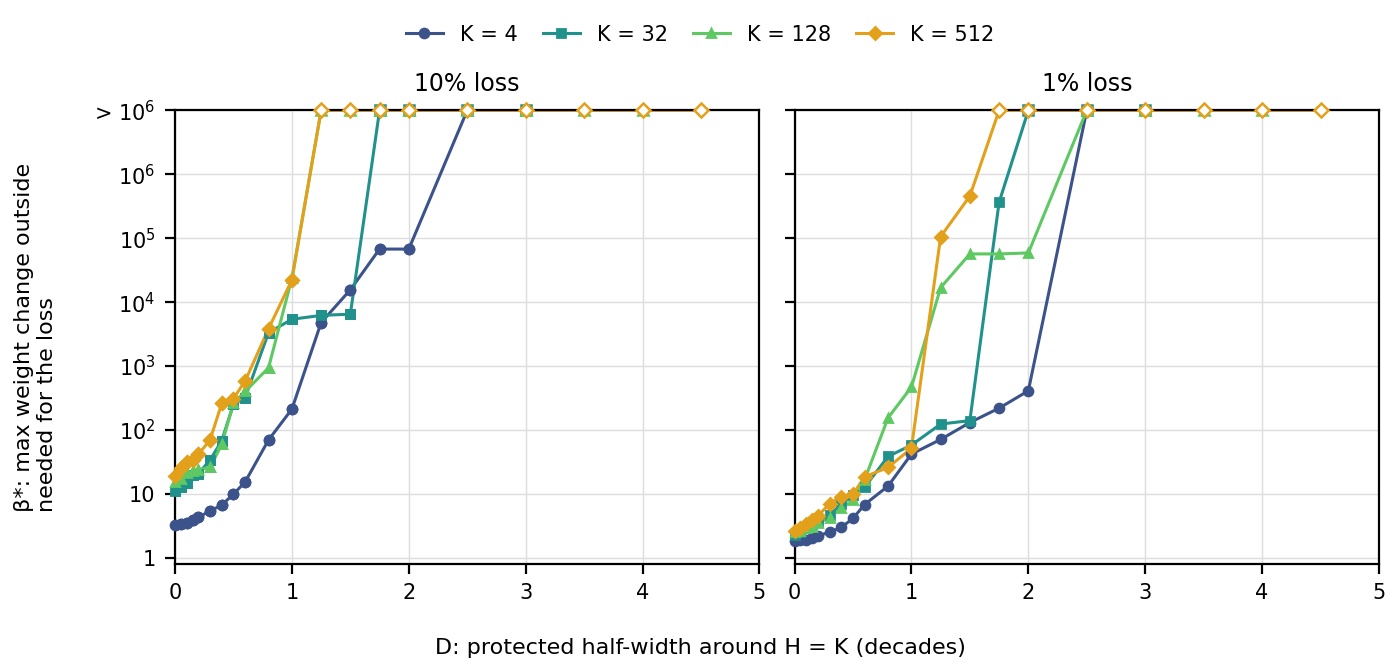}
    \caption{GPT OSS 20B, $n = 766$}
  \end{subfigure}
  \vspace{8pt}

  \begin{subfigure}[t]{0.48\textwidth}
    \centering
    \includegraphics[width=\linewidth]{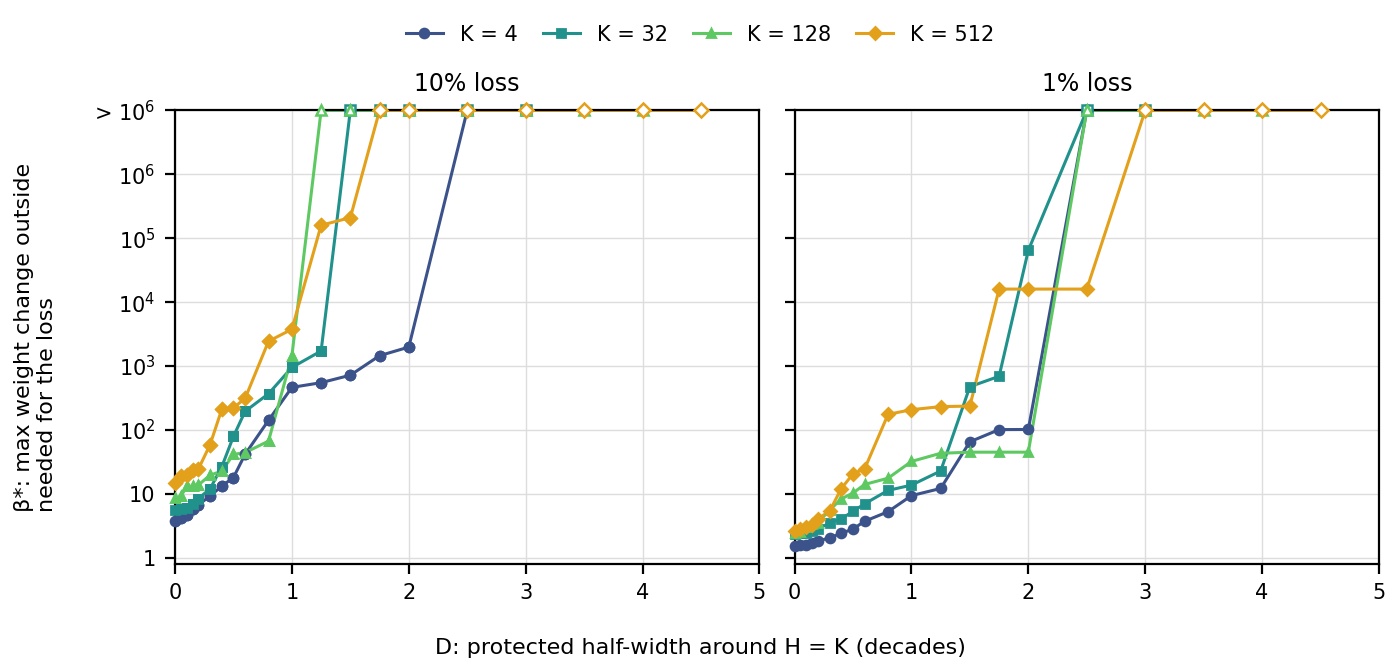}
    \caption{Qwen3-8B, $n = 677$}
  \end{subfigure}\hfill  \begin{subfigure}[t]{0.48\textwidth}
    \centering
    \includegraphics[width=\linewidth]{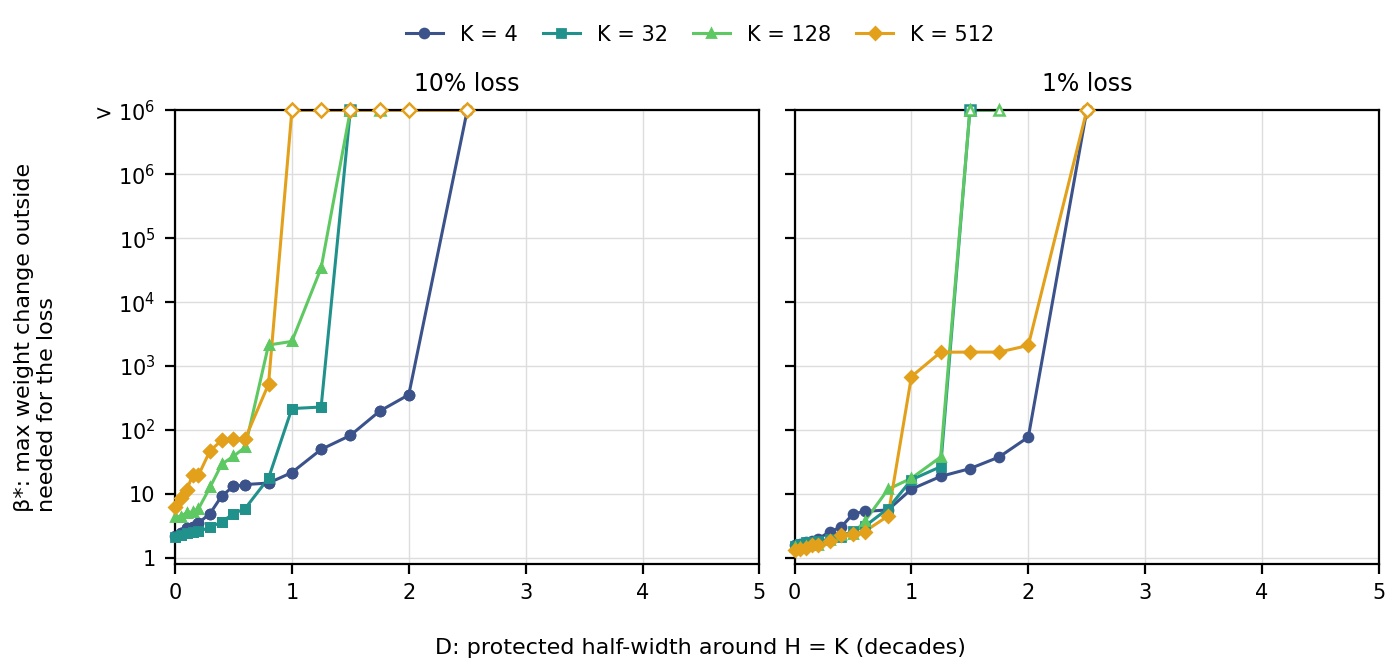}
    \caption{Qwen3-0.6B, $n = 86$}
  \end{subfigure}
  \caption{$\beta^*$ against $D$ on LiveCodeBench, at a $10\%$ loss (left in each panel) and a $1\%$ loss (right), with one curve per $K$. $n$ is the number of included questions.}
  \label{fig:Prot:D:LCB}
\par}

{\par
  \centering
  \begin{subfigure}[t]{0.48\textwidth}
    \centering
    \includegraphics[width=\linewidth]{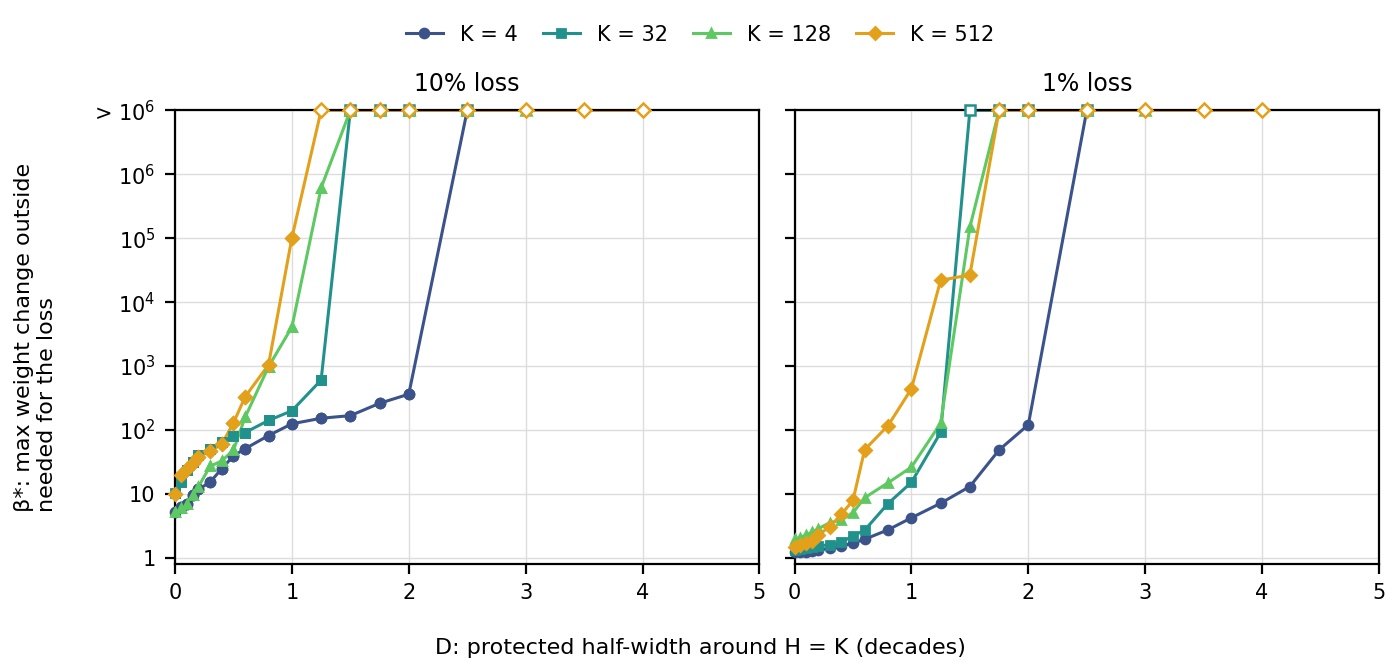}
    \caption{Mistral 7B Instruct, $n = 455$}
  \end{subfigure}\hfill  \begin{subfigure}[t]{0.48\textwidth}
    \centering
    \includegraphics[width=\linewidth]{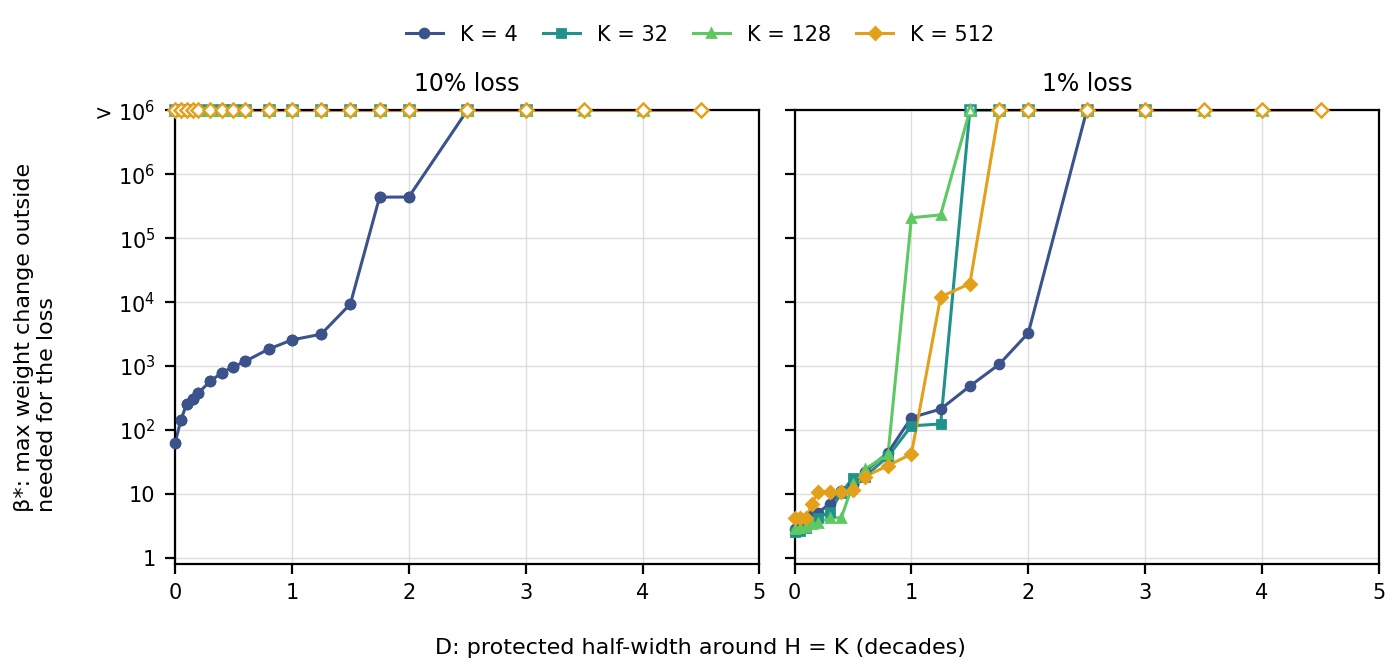}
    \caption{GPT OSS 20B, $n = 500$}
  \end{subfigure}
  \vspace{8pt}

  \begin{subfigure}[t]{0.48\textwidth}
    \centering
    \includegraphics[width=\linewidth]{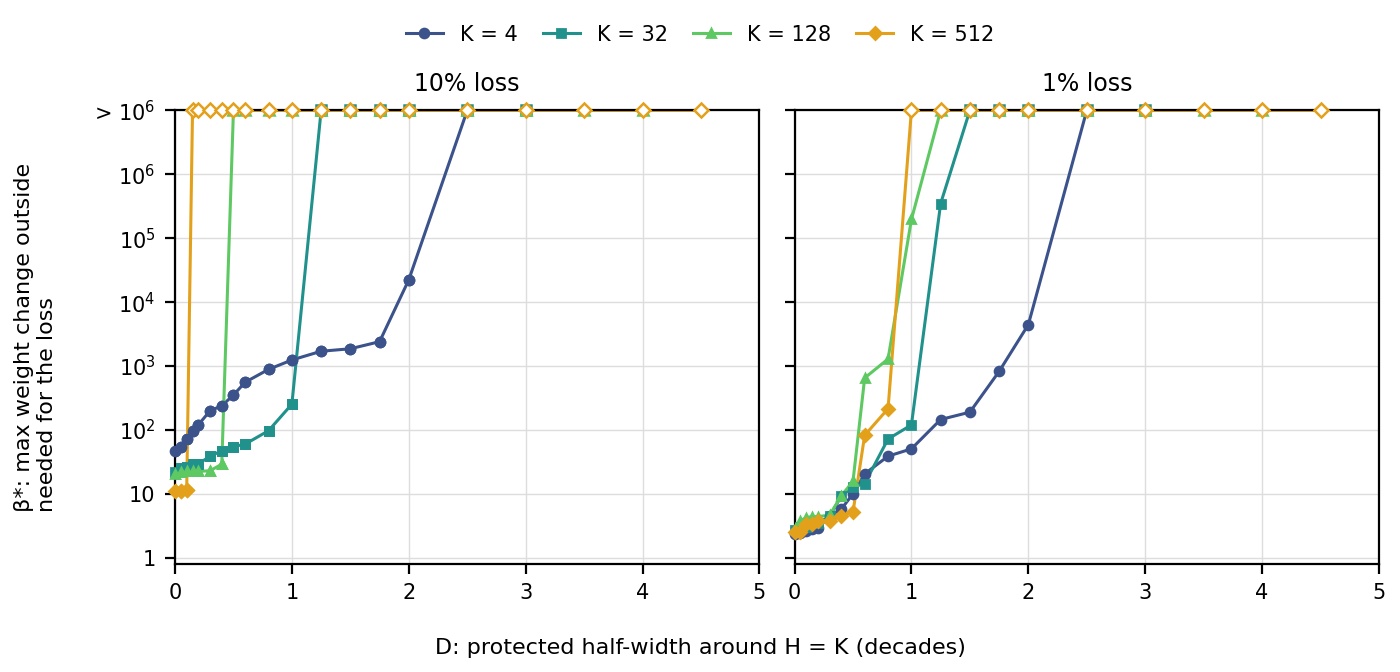}
    \caption{Qwen3-8B, $n = 496$}
  \end{subfigure}\hfill  \begin{subfigure}[t]{0.48\textwidth}
    \centering
    \includegraphics[width=\linewidth]{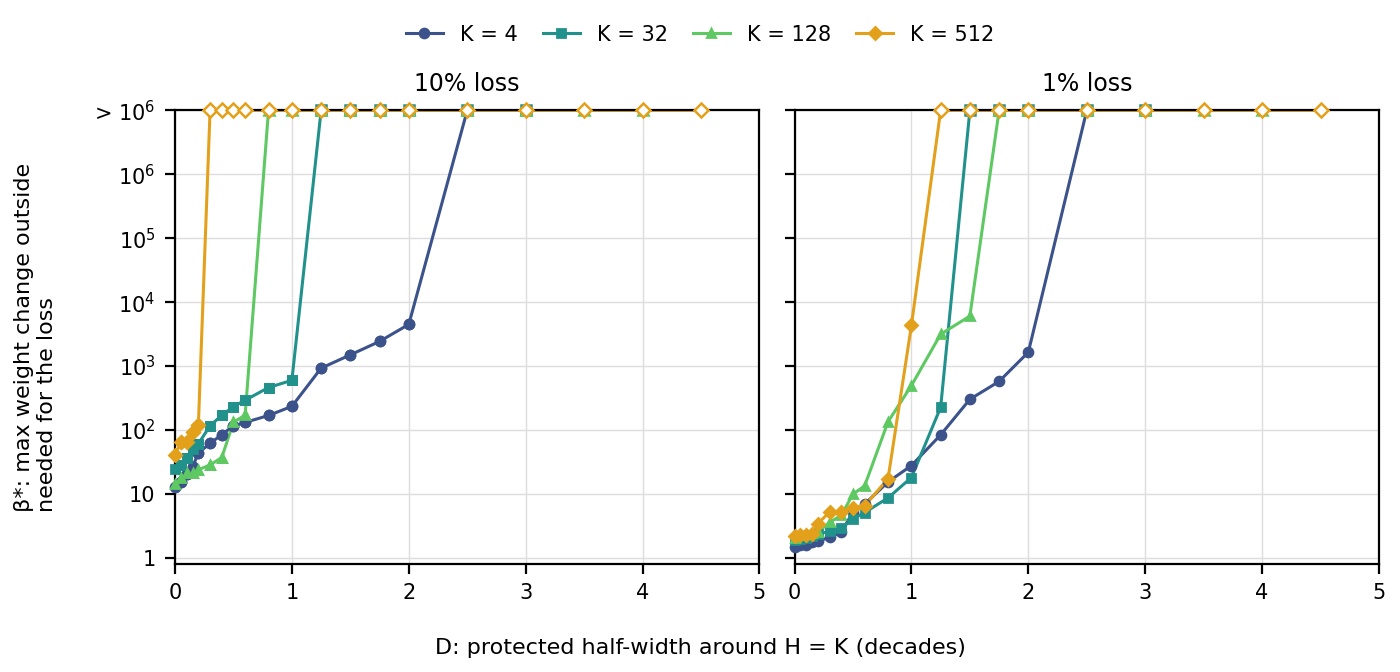}
    \caption{Qwen3-0.6B, $n = 492$}
  \end{subfigure}
  \caption{$\beta^*$ against $D$ on MATH-500, drawn as in Figure \ref{fig:Prot:D:LCB}. $n$ is the number of included questions. For GPT OSS 20B at a $10\%$ loss, the curves of $K = 32$, $128$, and $512$ have no reachable loss with the fitted pass rates and lie on the top row.}
  \label{fig:Prot:D:MATH}
\par}
\end{figure}

\begin{figure}[tbp]
{\par
  \centering
  \begin{subfigure}[t]{0.48\textwidth}
    \centering
    \includegraphics[width=\linewidth]{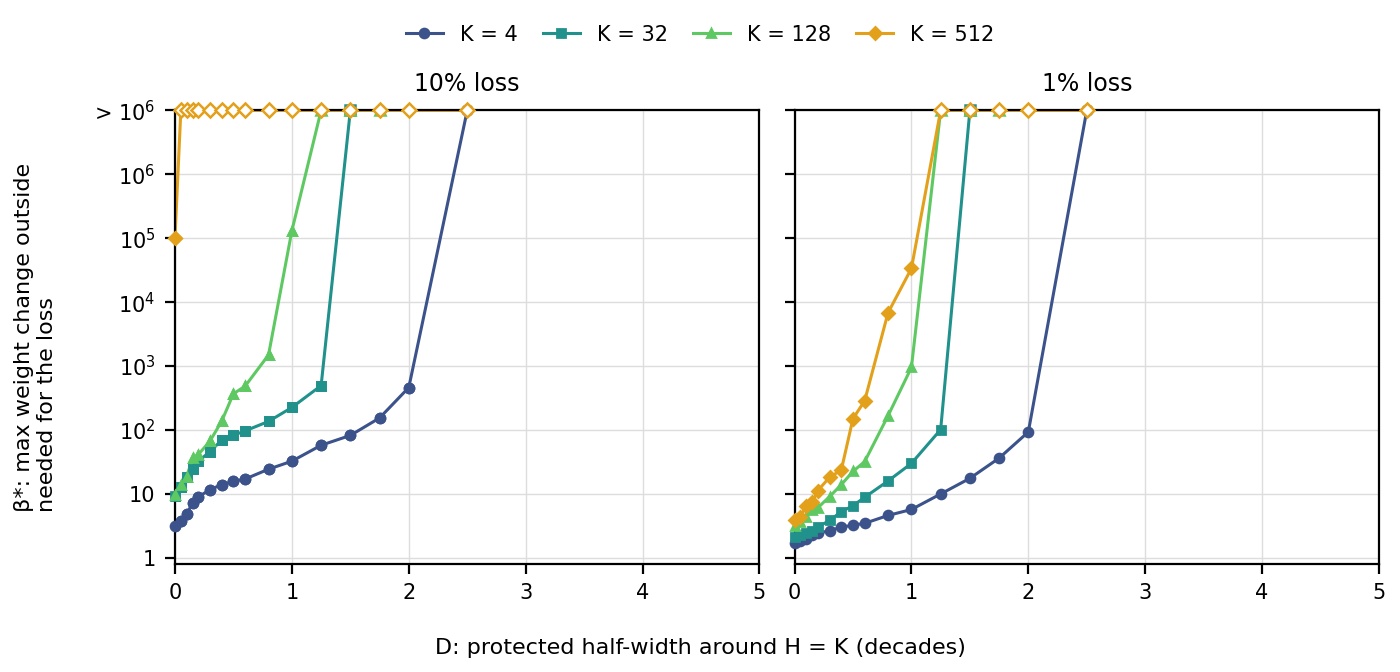}
    \caption{Mistral 7B Instruct, $n = 615$}
  \end{subfigure}\hfill  \begin{subfigure}[t]{0.48\textwidth}
    \centering
    \includegraphics[width=\linewidth]{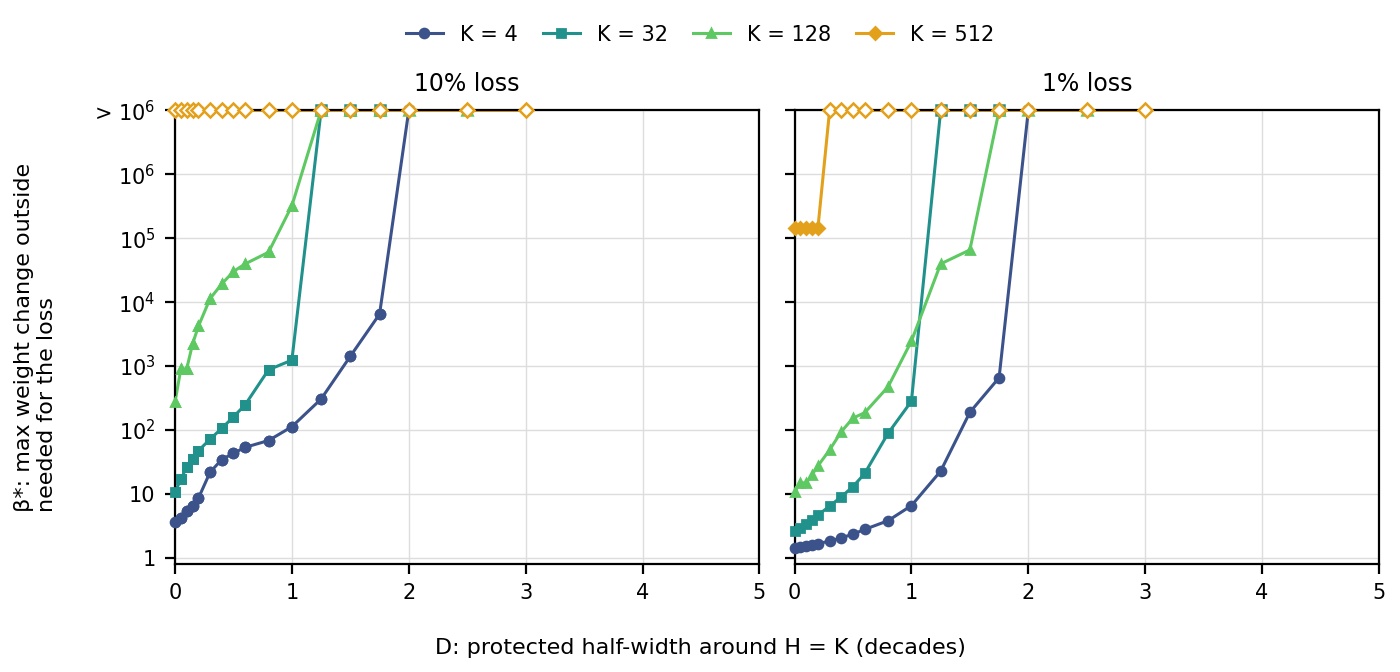}
    \caption{GPT OSS 20B, $n = 678$}
  \end{subfigure}
  \vspace{8pt}

  \begin{subfigure}[t]{0.48\textwidth}
    \centering
    \includegraphics[width=\linewidth]{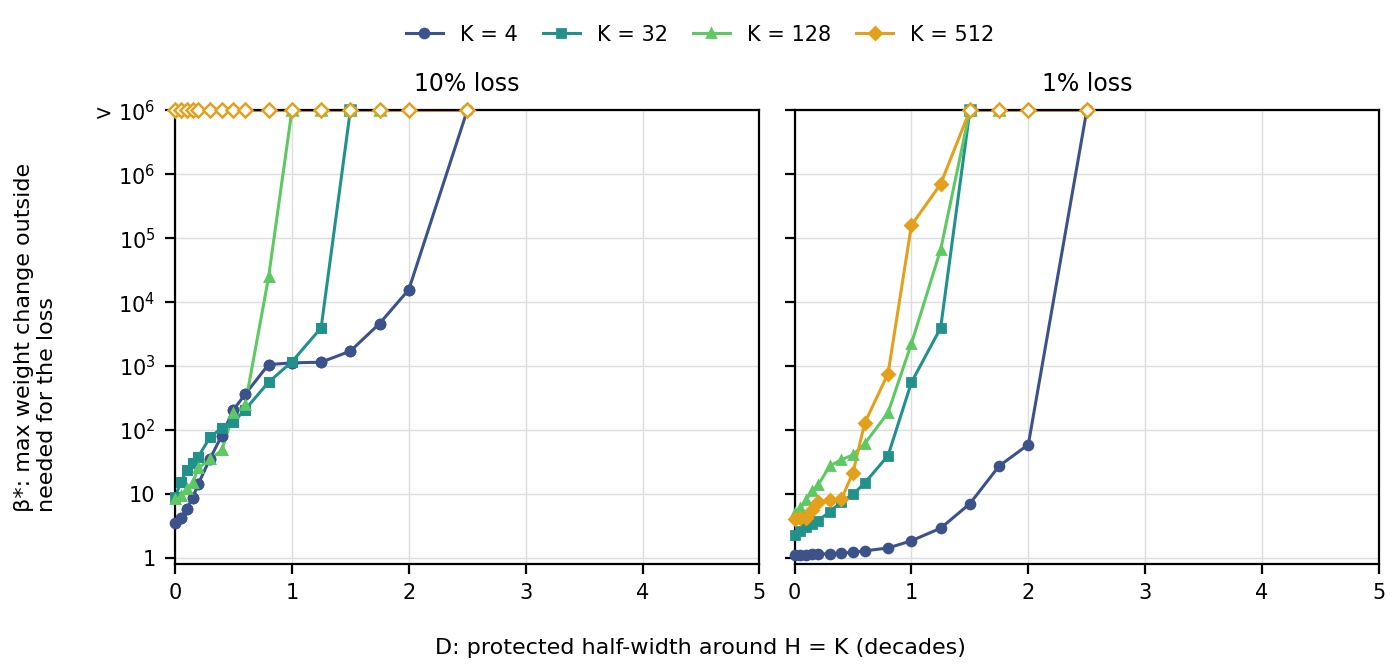}
    \caption{Qwen3-8B, $n = 466$}
  \end{subfigure}\hfill  \begin{subfigure}[t]{0.48\textwidth}
    \centering
    \includegraphics[width=\linewidth]{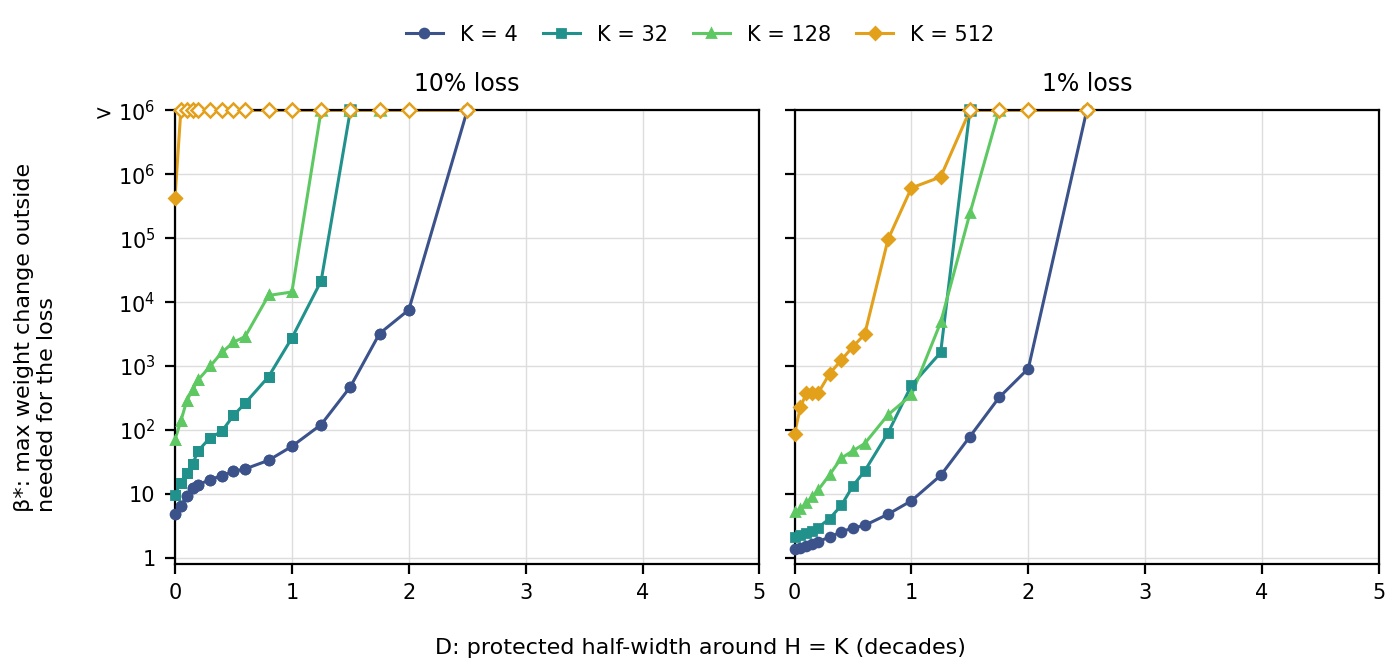}
    \caption{Qwen3-0.6B, $n = 743$}
  \end{subfigure}
  \caption{$\beta^*$ against $D$ on Hi-ToM, drawn as in Figure \ref{fig:Prot:D:LCB}. $n$ is the number of included questions. For GPT OSS 20B at a $10\%$ loss, with pass@$K$ computed exactly from the fitted pass rates, the curve of $K = 512$ needs $\beta > 10^{12}$ at every $D$ and lies on the top row.}
  \label{fig:Prot:D:HiToM}
\par}
\end{figure}

\paragraph{\texorpdfstring{Points with $\beta^* > 10^6$.}{Points with beta* > 1e6.}}
Of the $373$ points with $\beta^* > 10^6$ and questions outside the window, $248$ have no finite $\beta^*$ for the pass@$K$ values as stored in double precision, which we prove with certificates for which Equation \ref{zap:eq:Prot:Dual} holds at every $\beta$. When the pass@$K$ values are instead computed exactly from the fitted pass rates, $117$ of them still have no finite $\beta^*$, and the other $131$ need $\beta > 10^{12}$. The remaining $125$ points have a finite $\beta^*$ above $10^6$. For GPT OSS 20B at a $10\%$ loss, with the fitted pass rates, no temperature in $C$ can win under any weights on MATH-500 at $K = 32$, $128$, and $512$, and on Hi-ToM at $K = 512$ a win needs $\beta > 10^{12}$ at every $D$. When the sample counts are resampled, with pass@$K$ stored in double precision, the curve of MATH-500 at $K = 32$ has no reachable loss in $9$ of $20$ resamples, and those at $K = 128$ and $512$ in $18$ of $20$. In Figures \ref{fig:Prot:D:MATH} and \ref{fig:Prot:D:HiToM}, these four curves lie on the top row. In Figure \ref{fig:Prot:D:HiToM}, the curve of Qwen3-8B at $K = 512$ and a $10\%$ loss has a reachable loss and $\beta^* > 10^6$ at every $D$, so it lies on the top row as well.

\paragraph{\texorpdfstring{The loss of the chosen temperature.}{The loss of the chosen temperature.}}
At $\beta = 10$ and $D = 1.5$, $R$ is at most $0.94\%$ on all $12$ combinations of dataset and model at every $K$ on the grid from $4$ to $1024$. With no protected window, $R$ is $18\%$ to $36\%$ in the median combination at each of these $K$, and more than $56\%$ on LiveCodeBench with Qwen3-0.6B. The value of $0.94\%$ is not stable. When we resample the questions of each dataset, the $90\%$ interval of this maximum is $0.96\%$ to $4.10\%$, and when we resample the sample counts it is $0.84\%$ to $1.14\%$. Between grid points, it reaches $1.34\%$ at $K = 5$. A pool that we construct by adversarial search, with all ten of its summary statistics within the ranges of our datasets, loses $60.1\%$ at $K = 512$ with $D = 1.5$. In each of $20$ resamples of the questions and $20$ resamples of the sample counts, the smallest $D$ that keeps $R$ at most $1\%$ on all $12$ combinations at every $K$ on the grid from $4$ to $1024$ is at most $2.36$, so $D = 2.5$ suffices in all of them. On the original data, with $D = 2.5$, $R$ is exactly $0$ at every $K$ on the grid from $4$ to $1024$ and at most $0.18\%$ at every integer $K$ from $4$ to $1024$. This is a property of our datasets rather than a guarantee. At $K = 1$, $D = 1.5$ allows $R = 7.9\%$. Finally, $R$ can be large even below $\beta^*$, through temperatures that lose less than $10\%$ under uniform weights. At a $10\%$ loss, $R$ reaches $10\%$ at $339$ of the $3{,}219$ grid points with $\beta < \beta^*$ that we compared.

\paragraph{\texorpdfstring{One side at a time.}{One side at a time.}}
For $K \in \{4, 32, 128, 512\}$, reweighting only the harder questions makes a temperature in $C$ the best at some $\beta^* \leq 10^6$ at $97.5\%$ of the $315$ points with $d \leq 1$ and at least five questions outside. As $d$ grows, $\log_{10}\beta^*$ rises by a median of $1.20$ decades per decade of $d$, with an interquartile range of $0.87$ to $1.44$ over the $67$ curves with at least three finite points, and by $1.12$ over the $368$ such curves of the larger grid. Reweighting only the easier questions still flips the choice at $22$ of the $43$ points with $d = 1$ at a $10\%$ loss, counting the $4$ curves on which no weighting reaches the loss when pass@$K$ is stored in double precision. From $d = 2.5$, no $\beta \leq 10^6$ yields a $10\%$ loss at any of the $34$ points, and a $1\%$ loss is reached at only $1$ of $34$. At these four values of $K$, each of the $28$ flips with $d \geq 1$ at a $10\%$ loss promotes a temperature colder than $\tau_A$. In the median flip, all of the gain comes from questions that are easy at their own best temperature but within a decade of $K$ at $\tau_A$.

For deciding which of the easier questions can be reweighted without effect, hardness at a question's best temperature is therefore a poor yardstick. When we measure it at $\tau_A$ instead, reweighting only the questions more than $1.5$ decades easier than $K$ there, whose pass@$K$ at $\tau_A$ exceeds $1 - 2 \cdot 10^{-14}$, by any factor up to $10^6$, never makes a temperature that loses $1\%$, $5\%$, $10\%$, or $20\%$ the best on our $12$ combinations of dataset and model, at any $K$ we test from $1$ to $1024$. This holds exactly, both on the double-precision values of $s_q(\tau)$ and with pass@$K$ computed exactly from the fitted pass rates. At $K \in \{4, 32, 128, 512\}$ and at losses of $1\%$ and $10\%$, it also holds in each of $20$ resamples of the sample counts. It holds at all $4{,}177$ such points of simulated pools built from our measured profiles, and up to $K = 1024$ it fails only in simulated pools whose questions each have their own steep best temperature. At $K = 4096$, one pool built from measured profiles flips at $1.5$ decades. With $1$ decade instead of $1.5$, it still holds at a $10\%$ loss at $101$ of $102$ points, but at a $1\%$ loss $60$ of the $102$ points flip.

Part of this is a theorem. Write $M_A(\tau_c) = \sum_q (s_q(\tau_c) - s_q(\tau_A))$ over the questions that keep their weights, and $E_A = \sum_q e^{-K/H_q(\tau_A)}$ over the reweighted questions. A reweighted question can favor $\tau_c$ over $\tau_A$ by at most its failure probability $e^{-K/H_q(\tau_A)}$ at $\tau_A$. So if $M_A(\tau_c) < 0$ for every $\tau_c \in C$, no $\tau_c \in C$ becomes the best for $\beta \leq \min_{\tau_c \in C} |M_A(\tau_c)| / E_A$. If each of the $n_{\mathrm{out}}$ reweighted questions is more than $d$ decades easier than $K$ at $\tau_A$, then $E_A < n_{\mathrm{out}}\, e^{-10^d}$. For $K \in \{4, 32, 128, 512\}$ and $d \geq 1$, the condition $M_A < 0$ holds at all $133$ points at a $10\%$ loss at which some question is reweighted, and this bound alone exceeds $10^6$ at $95$ of them. Without the condition the statement can fail. In a constructed pool of two groups of questions, a temperature that loses more than $10\%$ becomes the best by down-weighting reweighted questions on which it is worse.

When the reweighted questions are easy at every temperature that competes, a bound follows from the loss alone. Let $\ell$ be the loss, so that $S_U(\tau_c) \leq (1 - \ell)\, S_U(\tau_A)$ for every $\tau_c \in C$, and suppose that we reweight only $n_{\mathrm{out}} \geq 1$ questions, each more than $d$ decades easier than $K$ both at $\tau_A$ and at every $\tau \in C$, so that $s_q(\tau) > 1 - e^{-10^d}$ at all of them. Every reweighted question then has $|s_q(\tau_c) - s_q(\tau_A)| < e^{-10^d}$, and $|W_q - 1| \leq \beta - 1$, so every allowed $W$ gives
\begin{equation}
\colorbox{white}{$\displaystyle\begin{aligned} S_W(\tau_c) - S_W(\tau_A) &\leq -\ell\, S_U(\tau_A) + (\beta - 1)\, n_{\mathrm{out}}\, e^{-10^d} \\ &< n_{\mathrm{out}} \left( (\beta - 1)\, e^{-10^d} - \frac{\ell}{1 - \ell} \left(1 - e^{-10^d}\right) \right) \ , \end{aligned}$}
\label{zap:eq:Prot:Easy}
\end{equation}
since $S_U(\tau_A) \geq S_U(\tau_c)/(1 - \ell) > n_{\mathrm{out}} (1 - e^{-10^d})/(1 - \ell)$. The right-hand side is at most $0$ whenever $\beta \leq 1 + \frac{\ell}{1 - \ell}\,(e^{10^d} - 1)$, so for these $\beta$ no $\tau_c \in C$ beats $\tau_A$, and therefore $\beta^* > 1 + \frac{\ell}{1 - \ell}\,(e^{10^d} - 1)$. This bound is doubly exponential in $d$, and it exceeds $10^6$ for every $d \geq 1.21$ at a $10\%$ loss and every $d \geq 1.27$ at a $1\%$ loss. On our data, however, such questions exist at only $24$ of the $133$ points above.

\begin{figure}[tbp]
{\par
  \centering
  \begin{subfigure}[t]{0.8\textwidth}
    \centering
    \includegraphics[width=\linewidth]{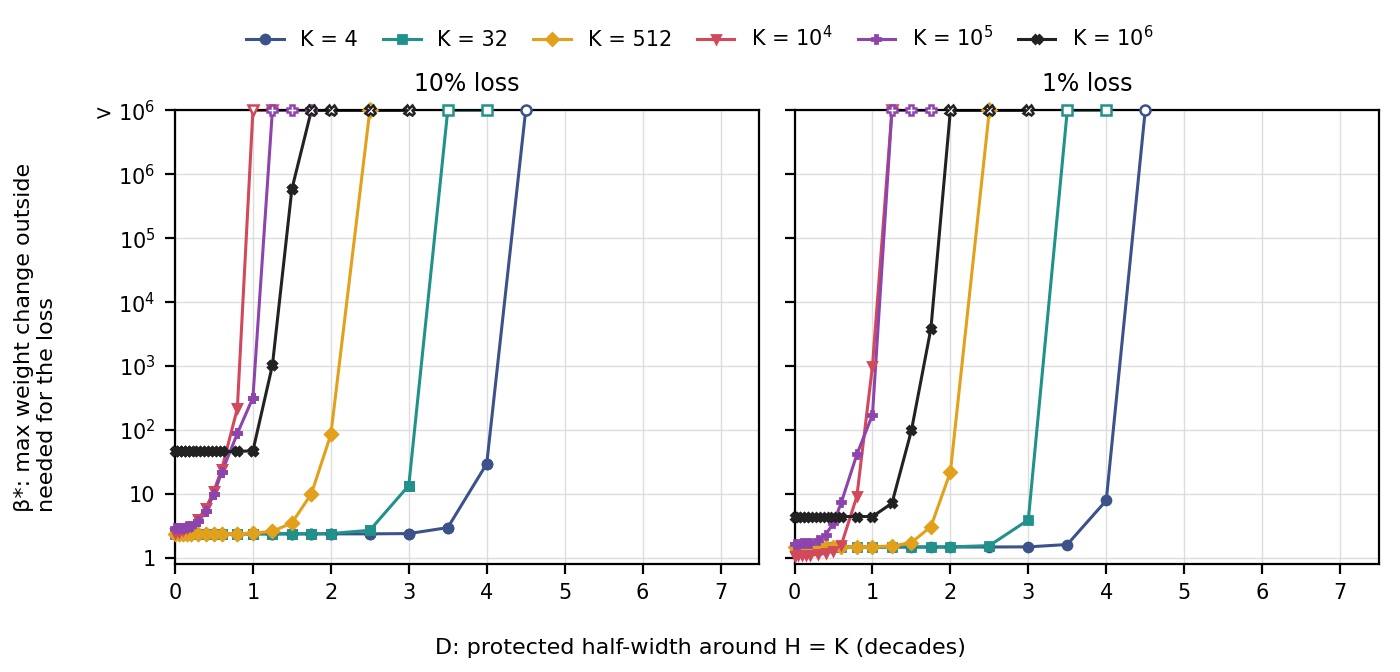}
    \caption{Peaked}
  \end{subfigure}
  \vspace{4pt}

  \begin{subfigure}[t]{0.8\textwidth}
    \centering
    \includegraphics[width=\linewidth]{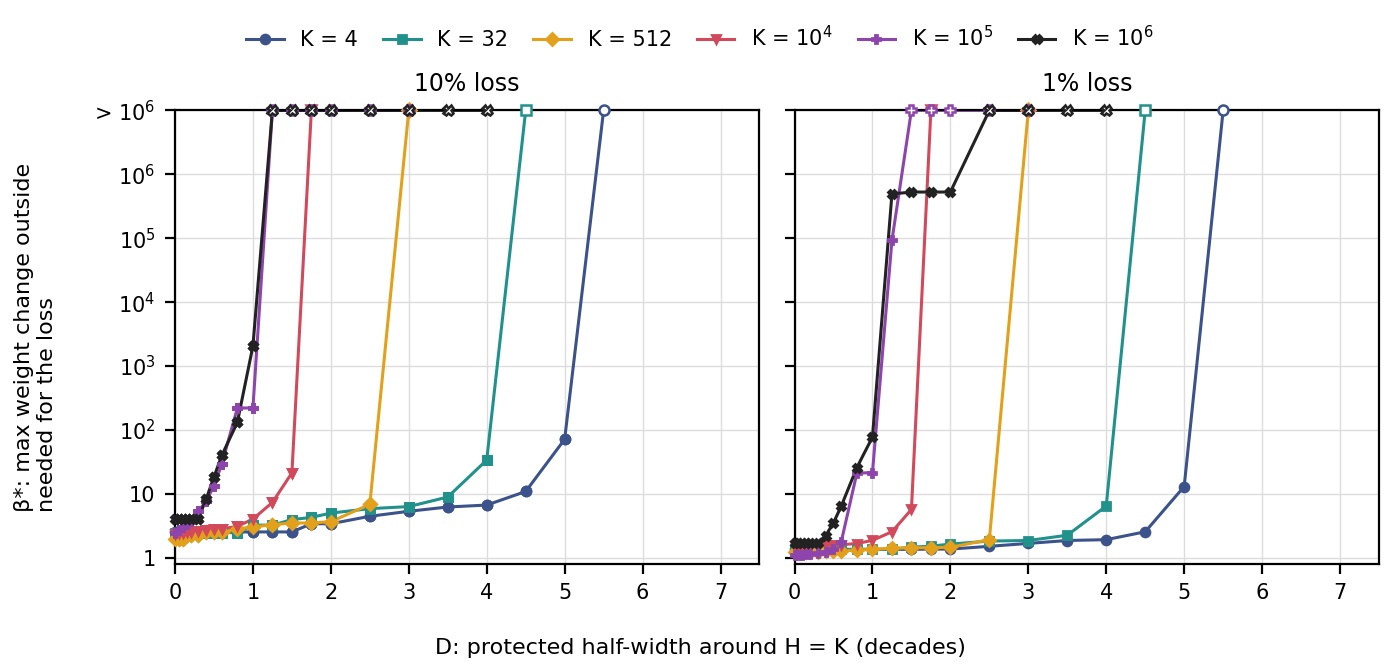}
    \caption{Reversed}
  \end{subfigure}
  \vspace{4pt}

  \begin{subfigure}[t]{0.8\textwidth}
    \centering
    \includegraphics[width=\linewidth]{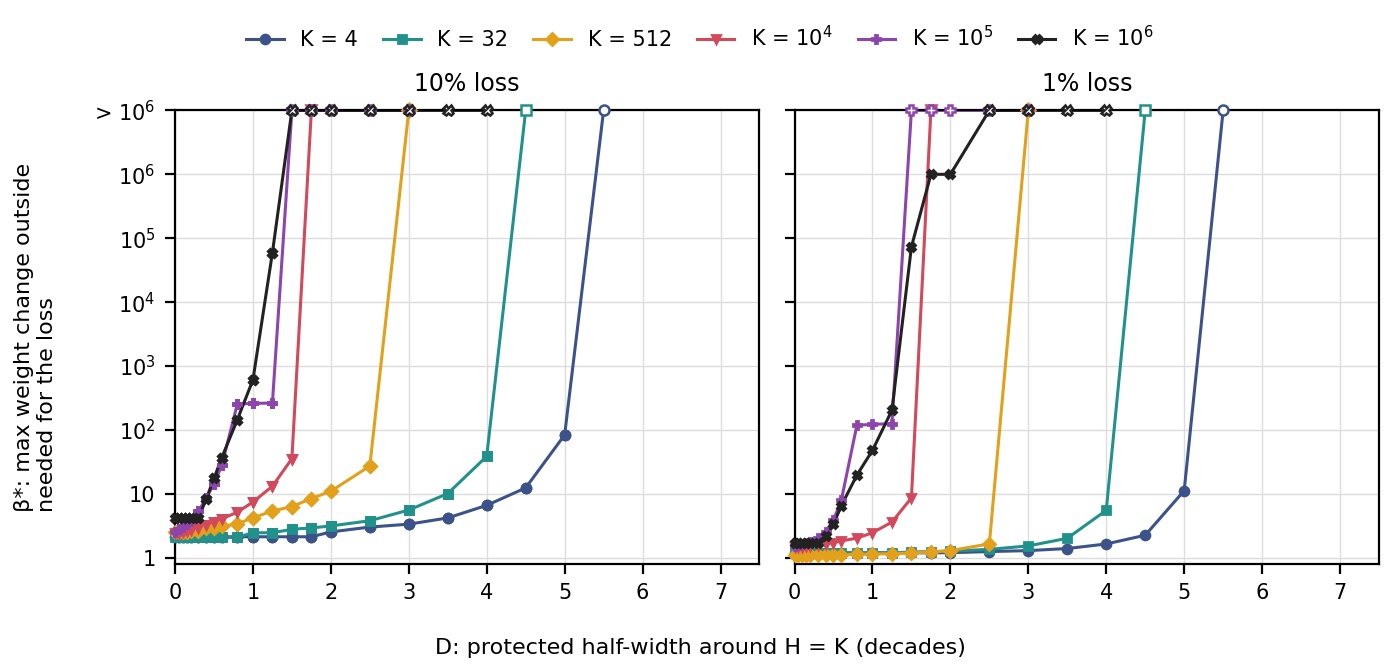}
    \caption{Flat}
  \end{subfigure}
  \caption{$\beta^*$ against $D$ for the three simulated shapes, at a $10\%$ loss (left in each panel) and a $1\%$ loss (right), with one curve per $K$, drawn as in Figure \ref{fig:Prot:D:LCB}. Each shape has $1000$ simulated questions, with hardness up to a cutoff of $500{,}000$ and temperature profiles from Mistral 7B Instruct.}
  \label{fig:Prot:Sim}
\par}
\end{figure}

\paragraph{\texorpdfstring{Tilting the mix.}{Tilting the mix.}}
As measured, at $\gamma = 0.25$ the best temperature moves colder in none of the $132$ combinations, and at $\gamma = -0.25$ it moves hotter in none. When we resample the questions or the sample counts, at least $129$ of the $132$ keep each direction in every resample, but all $132$ do so in only $4$ to $9$ of $20$. For $\gamma$ from $-1$ to $1$, the best temperature is nondecreasing in $\gamma$ in $124$ of the $132$ combinations. In the other $8$, all on MATH-500, it falls as $\gamma$ grows under tilts of more than a factor of $1.8$ per decade toward easier questions or at least $3.2$ toward harder ones. In resamples, such falls also occur on LiveCodeBench and Hi-ToM. Toward harder questions, $\tau_A$ first loses $1\%$ at a median tilt of a factor of $17$ per decade, with $90\%$ intervals of about $11$ to $23$ in resamples. This median is about $3$ at $K = 1$ and about $2{,}100$ at $K = 1024$, and higher at $K = 1024$ than at $K = 1$ in every resample, but the tilt needed is positively rank-correlated with $K$ in only $9$ of the $12$ combinations of dataset and model. Toward easier questions, $\tau_A$ loses $1\%$ at some tilt up to a factor of $10^4$ per decade in only $23$ of the $132$ combinations, and in at most $30$ in every resample. This direction holds approximately for moderate tilts by hardness within a dataset, but not across natural sub-datasets. There, in the $1{,}526$ cases in which both the best temperature and the median hardness of the sub-dataset differ from those of the full dataset, the best temperature moves the way the hardness shift predicts in only $807$, and on LiveCodeBench and MATH-500 it more often moves the other way.

\paragraph{\texorpdfstring{Natural sub-datasets.}{Natural sub-datasets.}}
The temperature chosen on the full dataset loses at least $10\%$ in only $15$ of the $2{,}688$ cases, or $0.56\%$, all at $K \leq 4$, and at least $1\%$ in $12.5\%$ of them. Its median loss is below $0.1\%$ on every dataset. The share of cases that lose at least $1\%$ falls from $24.8\%$ at $K = 1$ to $4.5\%$ at $K = 64$, and is $5.8\%$ and $5.1\%$ at $K = 256$ and $1024$. In each of $20$ resamples of the questions of each dataset, fewer than $2\%$ of the cases lose at least $10\%$, more than $80\%$ of these are at $K \leq 4$, and the median loss stays below $0.25\%$ on every dataset.

Reweighting whole groups is much weaker than reweighting single questions. On MATH-500 and Hi-ToM, no weighting of the levels of one factor, at any ratio, makes a temperature that loses $10\%$ the best on the stored values, in all $144$ cases. Weights on single questions do so in $122$ of these cases, with no protected window and a median $\beta^*$ of $7.5$. On LiveCodeBench, some weighting of the levels of one factor makes a temperature that loses $10\%$ the best on the stored values in $6$ of the $72$ cases, and each of these involves a group of fewer than $20$ questions, or one or two questions without a difficulty label.

\paragraph{\texorpdfstring{The role of $K$.}{The role of K.}}
Over $K$ from $1$ to $1024$, $\beta^*$ at $D = 0$ tends to grow with $K$, with a median Spearman correlation of $0.90$ over the $48$ combinations of dataset, model, and loss, but it never falls as $K$ grows in only $7$ of them. Pooling the $12$ combinations of dataset and model and the $11$ values of $K$ at each loss, $\beta^*$ at $D = 0$ follows the share of questions easier than $K$, with Spearman correlations of $0.76$ to $0.82$ over the four losses, more closely than $\log_{10} K$, with $0.51$ to $0.63$.

\paragraph{\texorpdfstring{Simulated hardness distributions.}{Simulated hardness distributions.}}
For $K \leq 512$, which lies below the bulk of the simulated questions, the window holds few or no questions until $D$ reaches that bulk. Until then, $\beta^*$ stays within a factor of about $2$ of its value at $D = 0$, which is between $2.0$ and $2.5$ at a $10\%$ loss and between $1.1$ and $1.5$ at a $1\%$ loss, and it then rises steeply (Figure \ref{fig:Prot:Sim}). For $K = 10^5$ and $10^6$, the least-squares slope of $\log_{10}\beta^*$ against $D$ lies between $1.1$ and $3.5$ decades per decade, and $8$ of these $12$ curves rise slowly at first and steeply later. For these two values of $K$, $\beta^*$ exceeds $10^6$ by $D = 1.25$ to $2.5$, with $5$ to $172$ questions still outside the window.

\section{min\_p Sampling on Hi-ToM}
\label{zap:MinP}

The temperature sweeps of the main text and of Appendix \ref{zap:CompResources} vary $\tau$ alone. In this appendix, we fix $\tau$ and vary \texttt{min\_p}, a sampler parameter that removes unlikely tokens. A token is dropped if its probability is below \texttt{min\_p} times that of the most likely token, and $\texttt{min\_p} = 0$ is plain sampling. We ask whether one \texttt{min\_p} value is enough at a fixed temperature, whether $\tau = 2.0$ does better than $\tau = 1.5$ with \texttt{min\_p}, and whether the best \texttt{min\_p} value does better than the best plain temperature. We study this on Hi-ToM, with Qwen3-0.6B at both temperatures and Mistral 7B Instruct at $\tau = 2.0$ alone.

{\par
\paragraph{Setup.}
We sampled with \texttt{vLLM} 0.19.0, \texttt{top\_p} $= 1$, and \texttt{top\_k} $= 0$, on the NVIDIA GPUs listed below, with all values of a question at $\tau = 2.0$ on one GPU. For Qwen3-0.6B, the chat template, token limit, and precision are those of Appendix \ref{zap:CompResources}, and the sampler, sample counts, and GPUs are those given here. For Mistral 7B Instruct, the prompts, token limit, precision, and grader are those of that appendix.
\begin{center}
\small
\begin{tabular}{@{}lcccl@{}}
  \toprule
  Model & $\tau$ & kept questions & $\nu$ & GPUs (questions) \\
  \midrule
  Qwen3-0.6B & $1.5$ & $648$ & $500$ & RTX 5090, with one L40S in the screen \\
  Qwen3-0.6B & $2.0$ & $648$ & $400$ & RTX 5090 ($351$), L40S ($270$), RTX 4090 ($27$) \\
  Mistral 7B Instruct & $2.0$ & $617$ & $1500$ & RTX 5090 ($330$), L40S ($287$) \\
  \bottomrule
\end{tabular}
\end{center}
The grid at $\tau = 1.5$ has the $10$ values $\texttt{min\_p} \in \{0, 0.001, 0.003, 0.01, 0.03, 0.1, 0.2, 0.3, 0.5, 0.7\}$, and that at $\tau = 2.0$ has the $5$ values $\texttt{min\_p} \in \{0.0056, 0.0128, 0.0316, 0.0721, 0.178\}$. Since \texttt{vLLM} applies \texttt{min\_p} after temperature, a token survives if and only if its untempered probability is at least $\texttt{min\_p}^{\tau}$ times that of the most likely token. Within $1\%$, the $\tau = 2.0$ values have the untempered cutoffs of the $\tau = 1.5$ values $0.001$, $0.003$, $0.01$, $0.03$, and $0.1$, which run from $3.2 \cdot 10^{-5}$ to $0.032$. As a result, each such pair keeps nearly the same tokens, and $\tau = 2.0$ only samples more evenly among them. Because \texttt{min\_p} can only remove diversity, we chose $\tau = 1.5$ at or above the large-$K$ optimum of plain sampling for Qwen3-0.6B, $1.35$ to $1.5$ at $K = 256$ in the sweep of Appendix \ref{zap:Multiple:Comp:res}, and $\tau = 2.0$ to test whether a higher temperature does better.

For Qwen3-0.6B, we screened all 1,200 questions with $100$ samples at each $\tau = 1.5$ value, and kept a question if and only if the pass rate of its best value was below $0.30$. In a second stage, every kept question received $400$ samples at each value of both grids. For Mistral 7B Instruct, we used the $617$ questions kept by its first stage in Appendix \ref{zap:Datasets}, with no further screen. Pass rates are instance-weighted, and outside the ratio-to-uniform results we use the unbiased pass@$K$ estimator.
\par}

\begin{figure}[!htbp]
{\par
  \centering
  \includegraphics[width=0.85\linewidth]{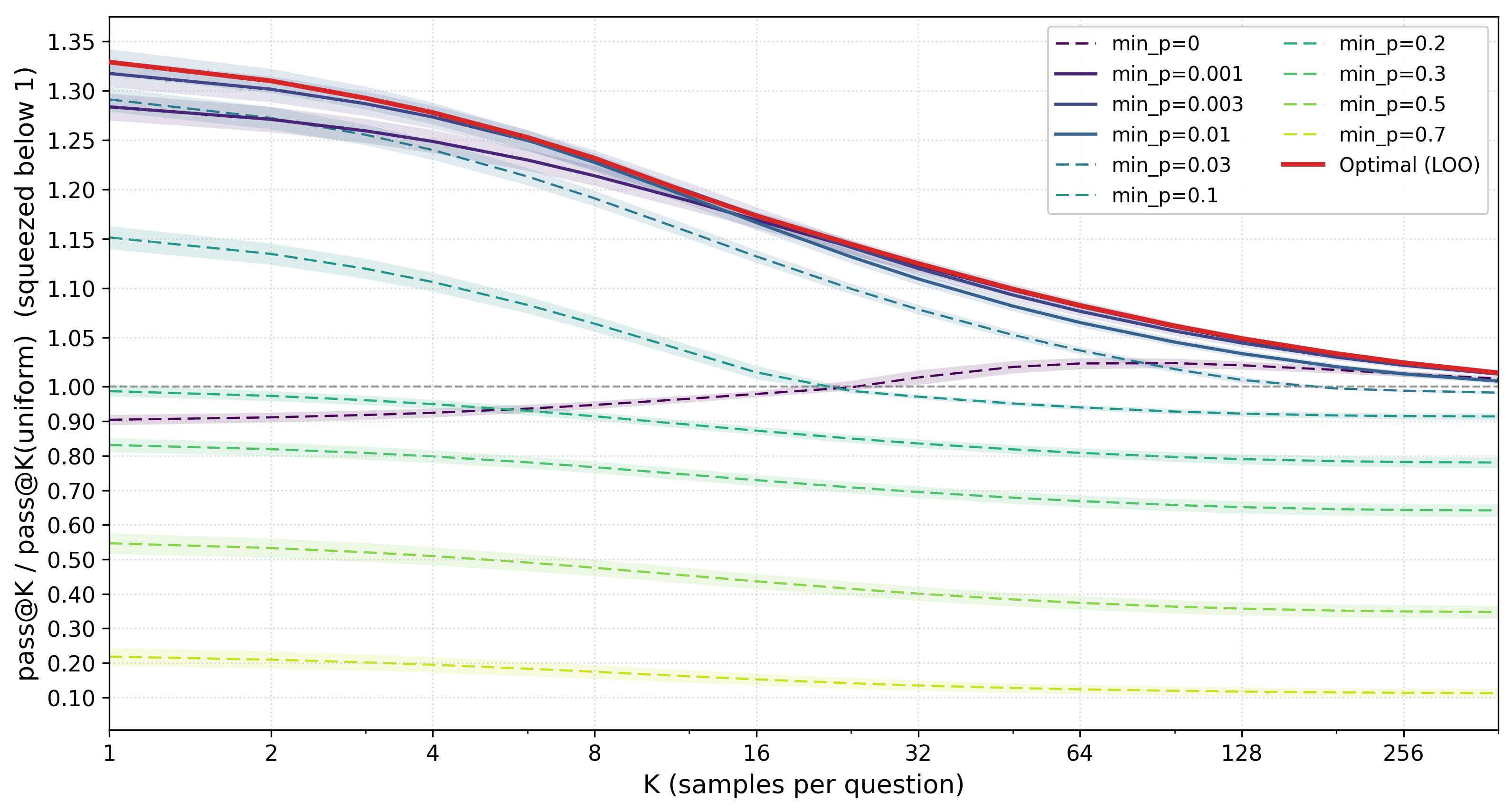}
  \caption{Ratio to uniform pass@$K$, $\mathrm{rtu}(K)$, of Qwen3-0.6B on Hi-ToM at $\tau = 1.5$, defined as in Figure \ref{fig:rtu:grid} with \texttt{min\_p} in place of the temperature. Colored curves are single-\texttt{min\_p} allocations, each with its paired-bootstrap $\pm 1\sigma$ band. Solid curves show the values that are the dominant choice in the optimal allocation at some $K$, namely $0.001$, $0.003$, and $0.01$, and dashed curves show the other values. The red trace is the LOO optimum. The population is the $648$ kept questions, with $\nu = 500$, and the vertical axis is compressed below $1$.}
  \label{fig:rtu:minp}
\par}
\end{figure}

\paragraph{Results at $\tau = 1.5$.}
One \texttt{min\_p} value is enough for Qwen3-0.6B at $\tau = 1.5$. Figure \ref{fig:rtu:minp} and the table below show pass@$K$ as a ratio to the uniform scheme of Section \ref{sec:Multiple:Comp}, which spreads $K$ equally over the $10$ values, so a ratio above $1$ is better. We compute the ratios from measured pass rates, because the degree-$10$ profile fit of Appendix \ref{zap:Stats:Methodology} is underdetermined on $10$ grid values. At every $K$ from $1$ to $384$, the leave-one-out (LOO) optimum over all values is within $0.003$ of the best single value in this ratio. The fixed value $0.003$ is within $0.012$ of it at these $K$, with the largest gap at $K = 1$, and exceeds the uniform scheme.

Plain sampling is below the uniform scheme until $K \approx 25$--$30$. The best single value decreases as $K$ grows, from $0.01$ for $K \le 6$ to $0.003$ for $K = 8$--$16$ and $0.001$ for $K \ge 24$. Values of $0.2$ and above lose to the uniform scheme at every $K$, with ratios of at most $0.99$, so the table stops at $0.2$. The best single value at each $K$ is in bold, and the bootstrap standard deviation of every single-value entry is at most $0.030$.

{\par
\begin{center}
\small
\begin{tabular}{@{}lcccc@{}}
  \toprule
  \texttt{min\_p} & $K = 1$ & $K = 16$ & $K = 64$ & $K = 384$ \\
  \midrule
  $0$ & $0.904$ & $0.979$ & $1.024$ & $1.008$ \\
  $0.001$ & $1.284$ & $1.169$ & $\mathbf{1.081}$ & $\mathbf{1.014}$ \\
  $0.003$ & $1.318$ & $\mathbf{1.174}$ & $1.076$ & $1.013$ \\
  $0.01$ & $\mathbf{1.329}$ & $1.166$ & $1.065$ & $1.006$ \\
  $0.03$ & $1.291$ & $1.132$ & $1.037$ & $0.983$ \\
  $0.1$ & $1.151$ & $1.015$ & $0.940$ & $0.914$ \\
  $0.2$ & $0.987$ & $0.873$ & $0.808$ & $0.780$ \\
  \midrule
  Optimal (LOO) & $1.329$ & $1.173$ & $1.082$ & $1.014$ \\
  uniform pass@$K$ & $7.32\%$ & $58.38\%$ & $85.32\%$ & $97.94\%$ \\
  \bottomrule
\end{tabular}
\end{center}
In absolute terms, with $\nu = 500$, the best single value raises pass@$1$ from $6.62\%$ at $\texttt{min\_p} = 0$ to $9.73\%$, a gain of $3.11 \pm 0.14$ percentage points (pp), or $47\%$, where $\pm$ is the paired standard error over questions. It gains $11.34 \pm 0.51$ pp in pass@$16$ and $4.63 \pm 0.45$ pp in pass@$64$, and the fixed value $0.003$ gains $3.03$, $11.34$, and $4.21$ pp at these $K$. In the second stage, every value from $0.001$ up shortens the mean output by $35\%$ to $46\%$, and the share of samples at the $1000$-token limit falls from $6.4\%$ to at most $0.5\%$.
\par}

\paragraph{$\tau = 2.0$ against $\tau = 1.5$.}
With \texttt{min\_p}, $\tau = 2.0$ does better than $\tau = 1.5$ for Qwen3-0.6B at each $K \in \{1, 4, 16, 64\}$. We compare the best value at each temperature in the table below, picked per $K$ on the $648$ questions and their $400$ second-stage samples per value. By a decision rule fixed in advance, $\tau = 2.0$ is better at $K$ if and only if the gap exceeds $2$ paired standard errors and is positive in at least $95\%$ of $400$ split-half checks. A check picks both best values on a random half of the questions and measures the gap on the other half. Worse is the mirror image, and anything else is no detectable difference.

{\par
\begin{center}
\small
\begin{tabular}{@{}rccccrl@{}}
  \toprule
  & \multicolumn{2}{c}{best at $\tau = 2.0$} & \multicolumn{2}{c}{best at $\tau = 1.5$} & & \\
  \cmidrule(lr){2-3} \cmidrule(lr){4-5}
  $K$ & \texttt{min\_p} & pass@$K$ & \texttt{min\_p} & pass@$K$ & gap (pp) & result for $\tau = 2.0$ \\
  \midrule
  $1$ & $0.0316$ & $10.42\%$ & $0.01$ & $9.71\%$ & $+0.71 \pm 0.11$ & better \\
  $4$ & $0.0316$ & $33.72\%$ & $0.01$ & $31.51\%$ & $+2.22 \pm 0.28$ & better \\
  $16$ & $0.0316$ & $72.54\%$ & $0.003$ & $68.97\%$ & $+3.57 \pm 0.38$ & better \\
  $64$ & $0.0128$ & $95.11\%$ & $0.001$ & $92.92\%$ & $+2.20 \pm 0.37$ & better \\
  $256$ & $0.0128$ & $99.73\%$ & $0.001$ & $99.25\%$ & $+0.48 \pm 0.23$ & borderline \\
  $400$ & $0.0128$ & $100.00\%$ & $0.001$ & $99.69\%$ & $+0.31 \pm 0.22$ & no detectable difference \\
  \bottomrule
\end{tabular}
\end{center}
All $400$ split-half gaps are positive at each $K \in \{1, 4, 16, 64\}$. At $K = 256$, the gap passes the $2$-standard-error test, but the share of positive splits lies around the $95\%$ bar and depends on the split seed, so the result is borderline. At $K = \nu$, pass@$K$ is the share of questions with any correct sample, so it saturates.
\par}

\begin{figure}[!htbp]
{\par
  \centering
  \includegraphics[width=0.9\linewidth]{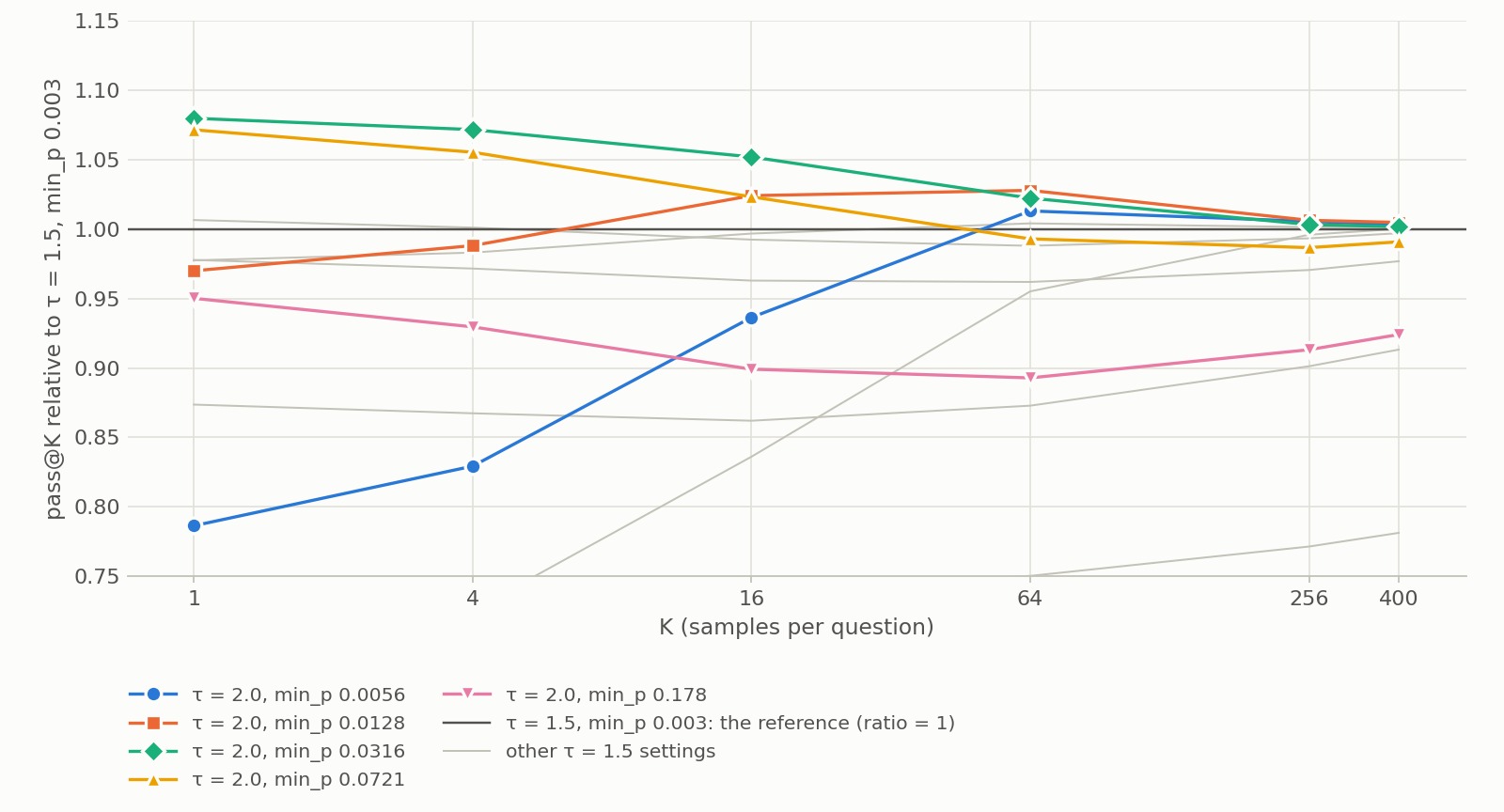}
  \caption{Ratio of pass@$K$ to that of $\tau = 1.5$ with $\texttt{min\_p} = 0.003$, for Qwen3-0.6B on Hi-ToM. Colored curves are the $5$ values at $\tau = 2.0$, and the dark line at $1$ is the reference. Faint curves are the other $\tau = 1.5$ values. Those of $\texttt{min\_p} = 0$ and $0.2$ are cut at the lower edge, and those of $0.3$, $0.5$, and $0.7$ lie entirely below the plotted range, so $6$ of the $9$ are visible. Points are ratios of two means over the $648$ questions, with $400$ second-stage samples and no error bars.}
  \label{fig:minp:T2}
\par}
\end{figure}

A single setting carries most of the gain. Figure \ref{fig:minp:T2} shows each $\tau = 2.0$ value relative to $\tau = 1.5$ with $\texttt{min\_p} = 0.003$, the fixed value above, so a curve above the dark line at $1$ is ahead of this setting. Against it, the value $0.0316$ gains between $0.77 \pm 0.10$ and $3.57 \pm 0.38$ pp at each $K \in \{1, 4, 16, 64\}$, and $0.31 \pm 0.34$ pp at $K = 256$, which is not significant.

The best \texttt{min\_p} value is higher at $\tau = 2.0$ than at $\tau = 1.5$ at every evaluated $K$. In untempered terms, the best cutoff is the same at both temperatures at $K = 1$ and $4$, namely $10^{-3}$, and higher at $\tau = 2.0$ at each evaluated $K$ from $16$ to $400$. Between values with the same untempered cutoff, $\tau = 2.0$ gains $0.7$ to $4.2$ pp at each $K \in \{1, 4, 16, 64\}$ at the three cutoffs of $10^{-3}$ and above. At $1.6 \cdot 10^{-4}$, the result is mixed, from $-0.38 \pm 0.29$ pp at $K = 4$ to $+2.56 \pm 0.37$ pp at $K = 64$. At the smallest cutoff, $3.2 \cdot 10^{-5}$, $\tau = 2.0$ loses up to $4.83 \pm 0.36$ pp at $K = 1$, $4$, and $16$. With $\texttt{min\_p} = 0.0316$, the outputs at $\tau = 2.0$ stay close in length to those of the best values at $\tau = 1.5$, at $251.1$ against at most $222.8$ tokens per sample.

{\par
\paragraph{Comparison with the best plain temperature.}
For Qwen3-0.6B, the best \texttt{min\_p} value also does better than the best plain temperature of the sweep of Appendix \ref{zap:Multiple:Comp:res}. This holds at every evaluated $K$ at $\tau = 2.0$, by the decision rule above, with at least $99\%$ of the $400$ split-half gaps positive. At $\tau = 1.5$, it is ahead at each $K \in \{1, 4, 16, 64, 256\}$, and the gap at $K = 400$ is not significant. We give these gaps in the table below, on the $642$ questions that both the screen and the sweep's first stage kept. To leave out the samples that selected the questions, we score this study on its $400$ second-stage samples and the sweep on its $1000$.
\begin{center}
\small
\begin{tabular}{@{}rccc@{}}
  \toprule
  $K$ & best plain $\tau$ & gap at $\tau = 2.0$ (pp) & gap at $\tau = 1.5$ (pp) \\
  \midrule
  $1$ & $1.275$ & $+1.52 \pm 0.10$ & $+0.79 \pm 0.07$ \\
  $4$ & $1.275$ & $+4.49 \pm 0.28$ & $+2.22 \pm 0.19$ \\
  $16$ & $1.275$ & $+7.28 \pm 0.43$ & $+3.68 \pm 0.29$ \\
  $64$ & $1.35$ & $+4.49 \pm 0.41$ & $+2.27 \pm 0.34$ \\
  $256$ & $1.35$ & $+1.02 \pm 0.22$ & $+0.54 \pm 0.22$ \\
  $400$ & $1.35$ & $+0.56 \pm 0.19$ & $+0.25 \pm 0.16$ \\
  \bottomrule
\end{tabular}
\end{center}
At $K = 1$, the best values reach $10.37\%$ at $\tau = 2.0$ and $9.65\%$ at $\tau = 1.5$, against $8.86\%$ for the best plain temperature and $1.83\%$ for plain $\tau = 2.0$ in the sweep. The split-half gap at $\tau = 1.5$ is positive in all $400$ splits at each $K \in \{1, 4, 16, 64\}$, where the fixed value $0.001$, not re-picked per $K$, exceeds the best plain temperature by at least $6.5$ standard errors. On the $642$ questions, the two setups agree where they overlap: plain sampling at $\tau = 1.5$ in this study differs from $\tau = 1.5$ in the sweep by $-0.04 \pm 0.06$ pp in pass@$1$, and by less than one standard error at each evaluated $K$.
\par}

\begin{figure}[!htbp]
{\par
  \centering
  \includegraphics[width=0.9\linewidth]{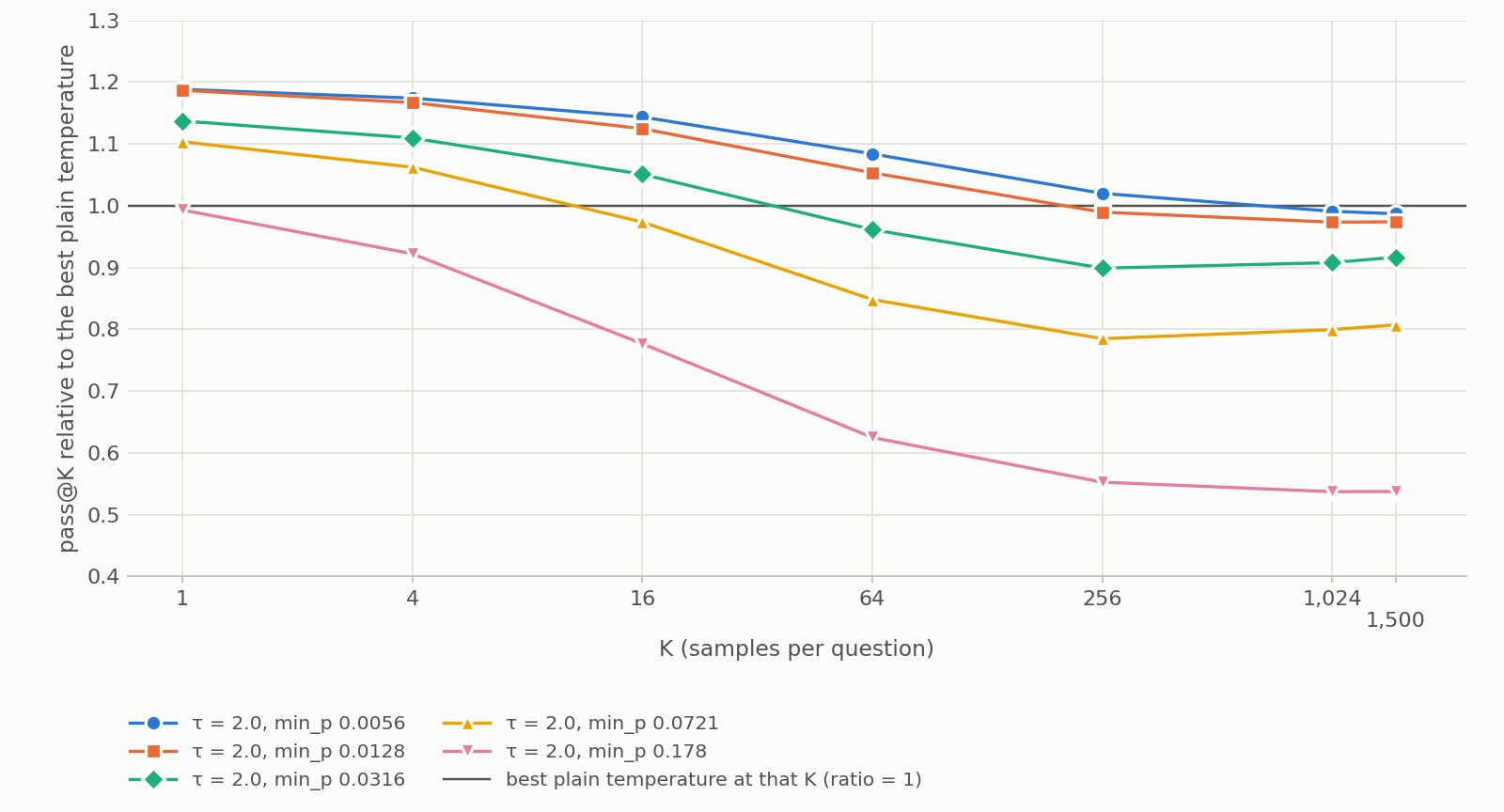}
  \caption{Ratio of pass@$K$ with \texttt{min\_p} at $\tau = 2.0$ to that of the best plain temperature of Appendix \ref{zap:Multiple:Comp:res}, for Mistral 7B Instruct on Hi-ToM. Colored curves are the $5$ values, and the dark line at $1$ is the best plain temperature at each $K$, picked on the same questions. Points are ratios of two means over the $617$ questions, with $\nu = 1500$ and no error bars.}
  \label{fig:minp:mistral}
\par}
\end{figure}

{\par
\paragraph{Mistral 7B Instruct.}
At $\tau = 2.0$, the best \texttt{min\_p} value does better for Mistral 7B Instruct than the best plain temperature at each $K \in \{1, 4, 16, 64, 256\}$, with relative gains of $19\%$ at $K = 1$ and $14\%$ at $K = 16$. At $K = 1024$ and $1500$, there is no detectable difference, by the same decision rule, fixed in advance. We take as reference the best of the $31$ plain temperatures of Appendix \ref{zap:CompResources}, picked per $K$, and score both sides on all $\nu = 1500$ samples, which for the reference include the $1000$ of the first stage. In Figure \ref{fig:minp:mistral}, a curve above the dark line at $1$ is ahead of this reference.
\begin{center}
\small
\begin{tabular}{@{}rcccrl@{}}
  \toprule
  & best value, $0.0056$ & \multicolumn{2}{c}{best plain $\tau$} & & \\
  \cmidrule(lr){3-4}
  $K$ & pass@$K$ & $\tau$ & pass@$K$ & gap (pp) & result for \texttt{min\_p} \\
  \midrule
  $1$ & $7.33\%$ & $1.575$ & $6.17\%$ & $+1.16 \pm 0.06$ & better \\
  $4$ & $23.72\%$ & $1.575$ & $20.20\%$ & $+3.52 \pm 0.17$ & better \\
  $16$ & $53.19\%$ & $1.575$ & $46.52\%$ & $+6.68 \pm 0.31$ & better \\
  $64$ & $78.56\%$ & $1.725$ & $72.48\%$ & $+6.08 \pm 0.43$ & better \\
  $256$ & $91.60\%$ & $1.8$ & $89.81\%$ & $+1.79 \pm 0.49$ & better \\
  $1024$ & $97.12\%$ & $2.0$ & $98.00\%$ & $-0.88 \pm 0.58$ & no detectable difference \\
  $1500$ & $97.89\%$ & $2.0$ & $99.19\%$ & $-1.30 \pm 0.60$ & no detectable difference \\
  \bottomrule
\end{tabular}
\end{center}
\par}

All $400$ split-half gaps are positive at each evaluated $K \le 256$. However, at $K = 1024$ and $1500$, there is no detectable difference from plain $\tau = 2.0$, as the gap at $K = 1024$ is within $2$ standard errors, and at $K = 1500$, $87\%$ of the splits are negative, below the $95\%$ bar.

The best value is the smallest of the grid, $0.0056$, at every $K$, and pass@$K$ falls as \texttt{min\_p} rises. Since no value between $0$ and $0.0056$ was tested, the grid cannot locate the best cutoff. The value $0.0128$ ties $0.0056$ only at $K \le 4$. For Qwen3-0.6B, on its $648$ questions with at most $1000$ new tokens, the best value at $\tau = 2.0$ is $0.0316$ or $0.0128$ at every evaluated $K$. Against plain sampling at $\tau = 2.0$, the value $0.0056$ gains $3.41 \pm 0.14$ pp at $K = 1$, from $3.92\%$, and $15.57$ and $11.42$ pp at $K = 16$ and $64$. Output length falls as \texttt{min\_p} rises, from $449$ tokens per sample at $0.0056$, with $76\%$ at the $500$-token limit, to $238$ at $0.178$, with $3\%$.

\begin{figure}[!htbp]
{\par
  \centering
  \begin{subfigure}[t]{0.85\textwidth}
    \centering
    \includegraphics[width=\linewidth]{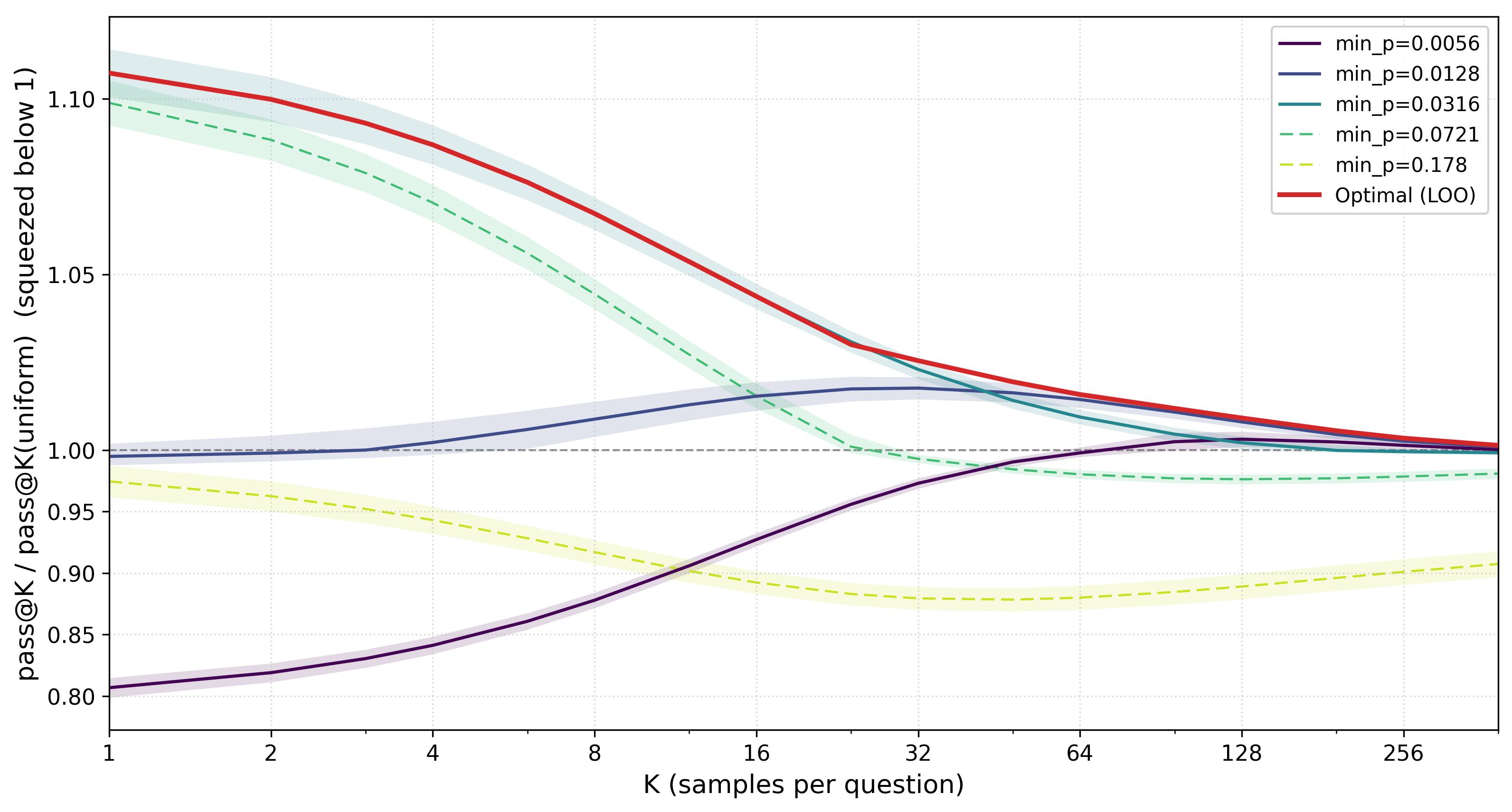}
    \caption{Qwen3-0.6B, $n = 648$, $\nu = 400$}
    \label{fig:minp:rtuT2:qwen}
  \end{subfigure}
  \vspace{8pt}

  \begin{subfigure}[t]{0.85\textwidth}
    \centering
    \includegraphics[width=\linewidth]{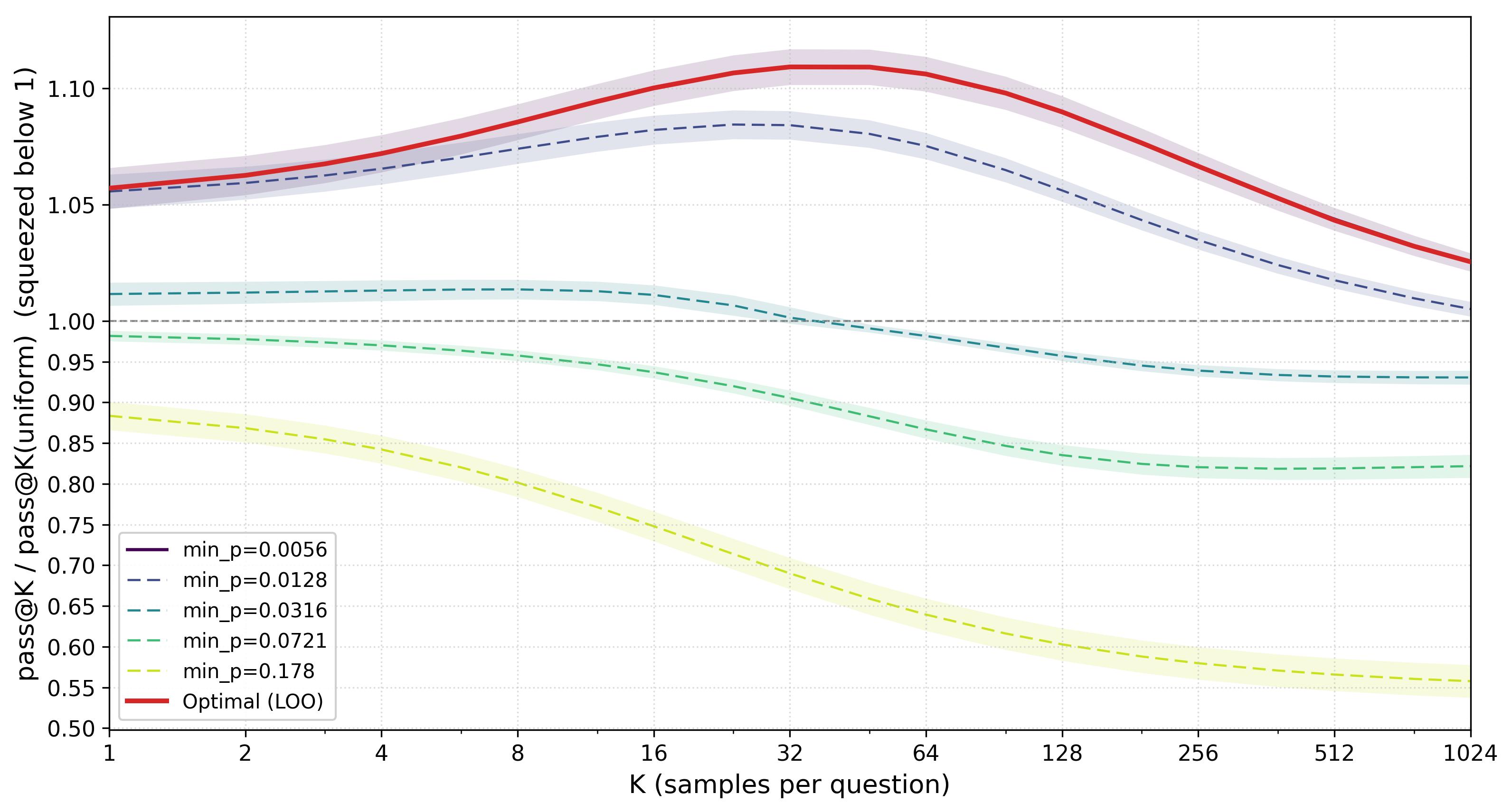}
    \caption{Mistral 7B Instruct, $n = 598$, $\nu = 1500$}
    \label{fig:minp:rtuT2:mistral}
  \end{subfigure}
  \caption{Ratio to uniform pass@$K$, $\mathrm{rtu}(K)$, on Hi-ToM at $\tau = 2.0$, defined as in Figure \ref{fig:rtu:minp} and computed from measured pass rates. $n$ is the number of included questions. Colored curves are single-\texttt{min\_p} allocations, each with its paired-bootstrap $\pm 1\sigma$ band. Solid curves show the values that are the dominant choice in the optimal allocation at some $K$, dashed curves the others, and the red trace is the LOO optimum. $K$ runs to $384$ for Qwen3-0.6B and to $1024$ for Mistral 7B Instruct. For Mistral 7B Instruct, $19$ questions have a best pass rate below the $0.1\%$ inclusion floor of the paper's analysis, $8$ of them with no correct sample, and the curve of $0.0056$ lies under the red trace. The vertical axis is compressed below $1$.}
  \label{fig:minp:rtuT2}
\par}
\end{figure}

\paragraph{Ratio to uniform at $\tau = 2.0$.}
Figure \ref{fig:minp:rtuT2} shows $\mathrm{rtu}(K)$ at $\tau = 2.0$, defined as in Figure \ref{fig:rtu:minp}, where the uniform scheme spreads $K$ equally over the $5$ values. For Qwen3-0.6B, $0.0316$ is the best single value at every $K \le 32$, with a ratio of $1.107$ at $K = 1$, and $0.0128$ is the best from $K = 48$ on. The LOO optimum puts all samples on $0.0316$ at every $K \le 16$, and it is within $0.004$ of the best single value at every $K$. The value $0.178$ is below the uniform scheme at every $K$, with a largest ratio of $0.974$. For Mistral 7B Instruct, $0.0056$ is the best single value at every $K$, and the optimal allocation puts all samples on it at every $K \le 384$, so its curve lies under the red trace. Its ratio rises from $1.057$ at $K = 1$ to $1.109$ at $K = 32$ and falls to $1.025$ at $K = 1024$, while that of $0.178$ falls from $0.884$ to $0.558$. The bootstrap standard deviation of every single-value entry is at most $0.013$ for Qwen3-0.6B and $0.021$ for Mistral 7B Instruct.

\paragraph{Question profiles and hardness.}
We repeat the analyses of Figures \ref{fig:ManyQ}, \ref{fig:AveTemp:grid}, \ref{fig:RangeSize:grid}, and \ref{fig:relC:grid} with \texttt{min\_p} in place of the temperature, on measured pass rates, since a profile fit is underdetermined on these grids. The hardness $H_q$ is Equation \ref{eq:def:Hardness} at the best value of the question on the grid, and the efficient range $E_q$ is the set of contiguous grid values around the peak whose pass rate is at least half the peak, so $|E_q|$ counts grid values. Grid values are spaced equally on every \texttt{min\_p} axis, and a sliding mean over \texttt{min\_p} is a mean of grid position. All questions were kept as hard, so $H_q$ starts near $2$, and the easy side of the temperature figures has no counterpart here. For Mistral 7B Instruct, the hardness axes leave out the $19$ questions below the inclusion floor, which leaves $598$.

\begin{figure}[!htbp]
{\par
  \centering
  \begin{subfigure}[t]{0.48\textwidth}
    \centering
    \includegraphics[width=\linewidth]{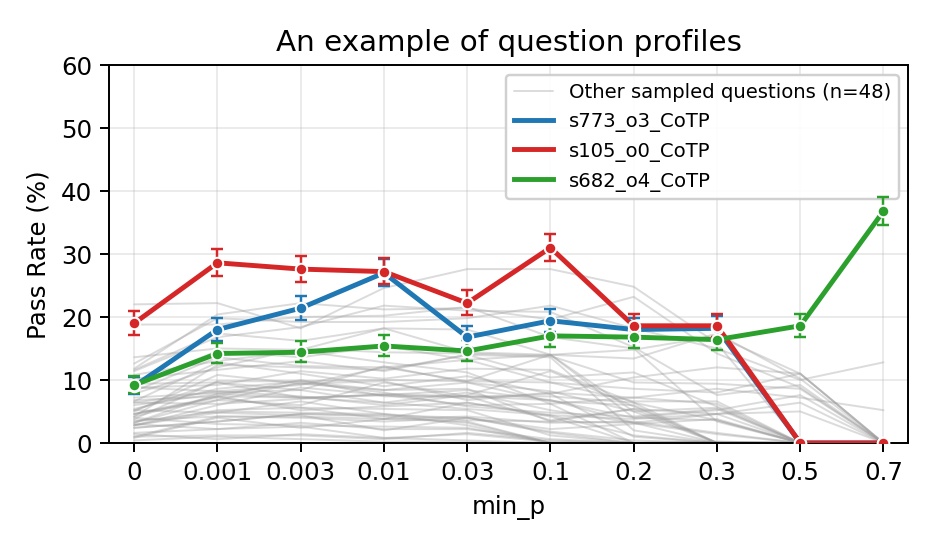}
    \caption{Qwen3-0.6B, $\tau = 1.5$}
    \label{fig:minp:profiles:q15}
  \end{subfigure}\hfill  \begin{subfigure}[t]{0.48\textwidth}
    \centering
    \includegraphics[width=\linewidth]{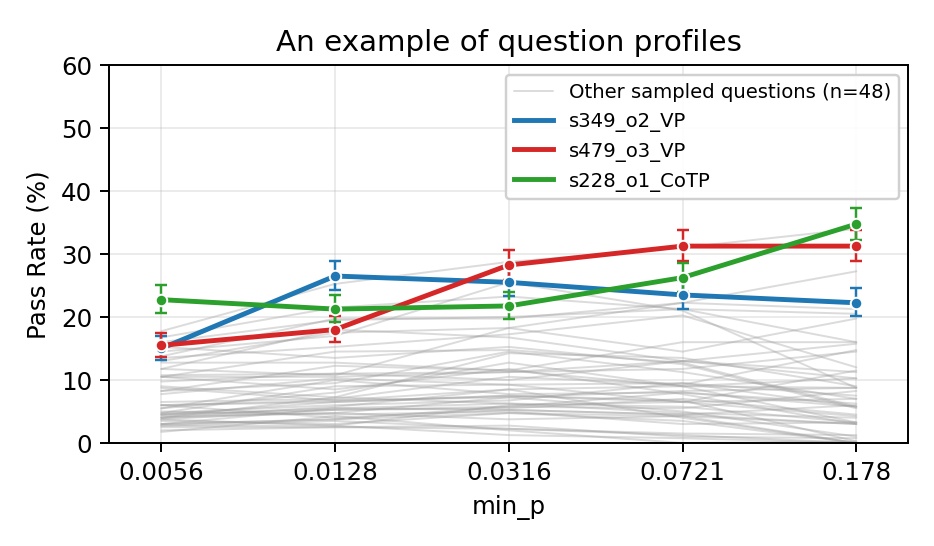}
    \caption{Qwen3-0.6B, $\tau = 2.0$}
    \label{fig:minp:profiles:q20}
  \end{subfigure}
  \vspace{8pt}

  \begin{subfigure}[t]{0.48\textwidth}
    \centering
    \includegraphics[width=\linewidth]{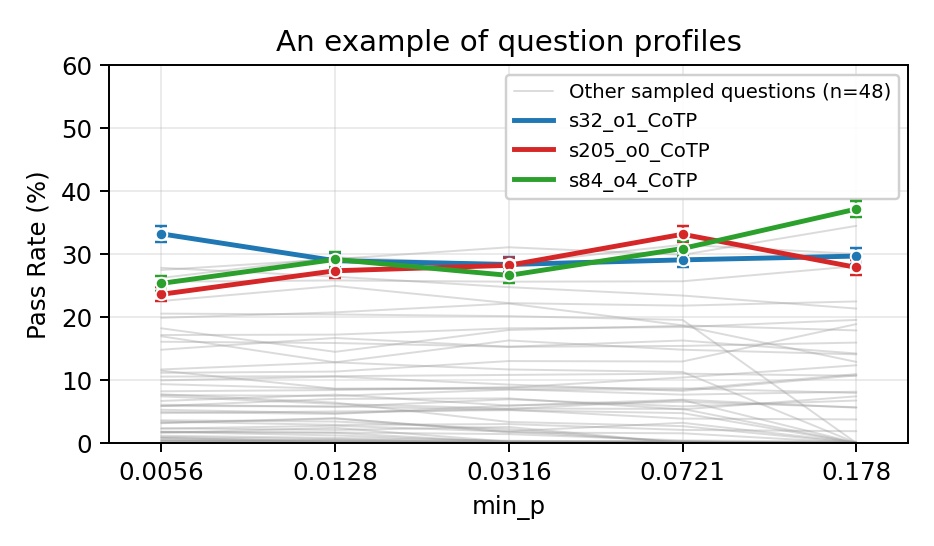}
    \caption{Mistral 7B Instruct, $\tau = 2.0$}
    \label{fig:minp:profiles:mistral}
  \end{subfigure}
  \caption{\texttt{min\_p} profiles of $51$ Hi-ToM questions in each study, the counterpart of Figure \ref{fig:ManyQ}. Lines join measured pass rates, with no fit, and grid values are spaced equally on the horizontal axis. In each third of the grid, with $4$, $3$, and $3$ values at $\tau = 1.5$ and $2$, $2$, and $1$ at $\tau = 2.0$, we highlight, among the questions that peak there, the one with the most correct samples at its peak, with the peak at the smallest \texttt{min\_p} among tied values and ties between questions broken by id. Bars are exact Clopper-Pearson $1\sigma$ intervals, and the vertical axis stops at $60\%$. The $48$ grey questions are drawn at random, with a fixed seed, from the other questions of the study.}
  \label{fig:minp:profiles}
\par}
\end{figure}

In Figure \ref{fig:minp:profiles}, the smallest value, $0$ at $\tau = 1.5$ and $0.0056$ at $\tau = 2.0$, samples at a fixed $\tau$, so the profiles are not pinned at $0$ or $1$ at the left edge. A temperature profile is pinned at $0$ or $1$ at $\tau = 0$, where sampling is greedy. The peaks spread over the whole grid. At $\tau = 1.5$, $33$ of the $648$ questions peak at $\texttt{min\_p} = 0$ and $24$ at $0.7$, although $576$ have no correct sample at $0.7$. At $\tau = 2.0$, $133$ of the $648$ Qwen3-0.6B questions and $119$ of the $598$ Mistral 7B Instruct questions peak at the largest value, $0.178$. Each highlighted question is the easiest one peaking in its third, at $26.5\%$ to $37.1\%$.

A hard question's peak rests on a few correct samples, and its $H_q$ comes from the same maximum. We therefore describe a trend only if it passes two checks. First, in $400$ random splits of the samples of every cell into halves, with the peak or range from one half and $H_q$ from the other, it keeps its sign in at least $95\%$ of the splits. Second, its slope lies outside the central $95\%$ of the slopes of a null, on the side of its sign. In the null, every question has the same profile shape, the pass rate of each grid value pooled over all questions of the study, scaled so that its mean over the grid is the question's own mean pass rate and capped at $1$, with binomial draws at the same $\nu$, in $400$ replicates, from which $H_q$, the peak, and $E_q$ are recomputed. The first check alone is not enough, because noise grows with $H_q$, and on the null data, with $40$ splits in each of $100$ replicates, it passes in up to all of them.

\begin{figure}[!htbp]
{\par
  \centering
  \begin{subfigure}[t]{0.48\textwidth}
    \centering
    \includegraphics[width=0.75\linewidth]{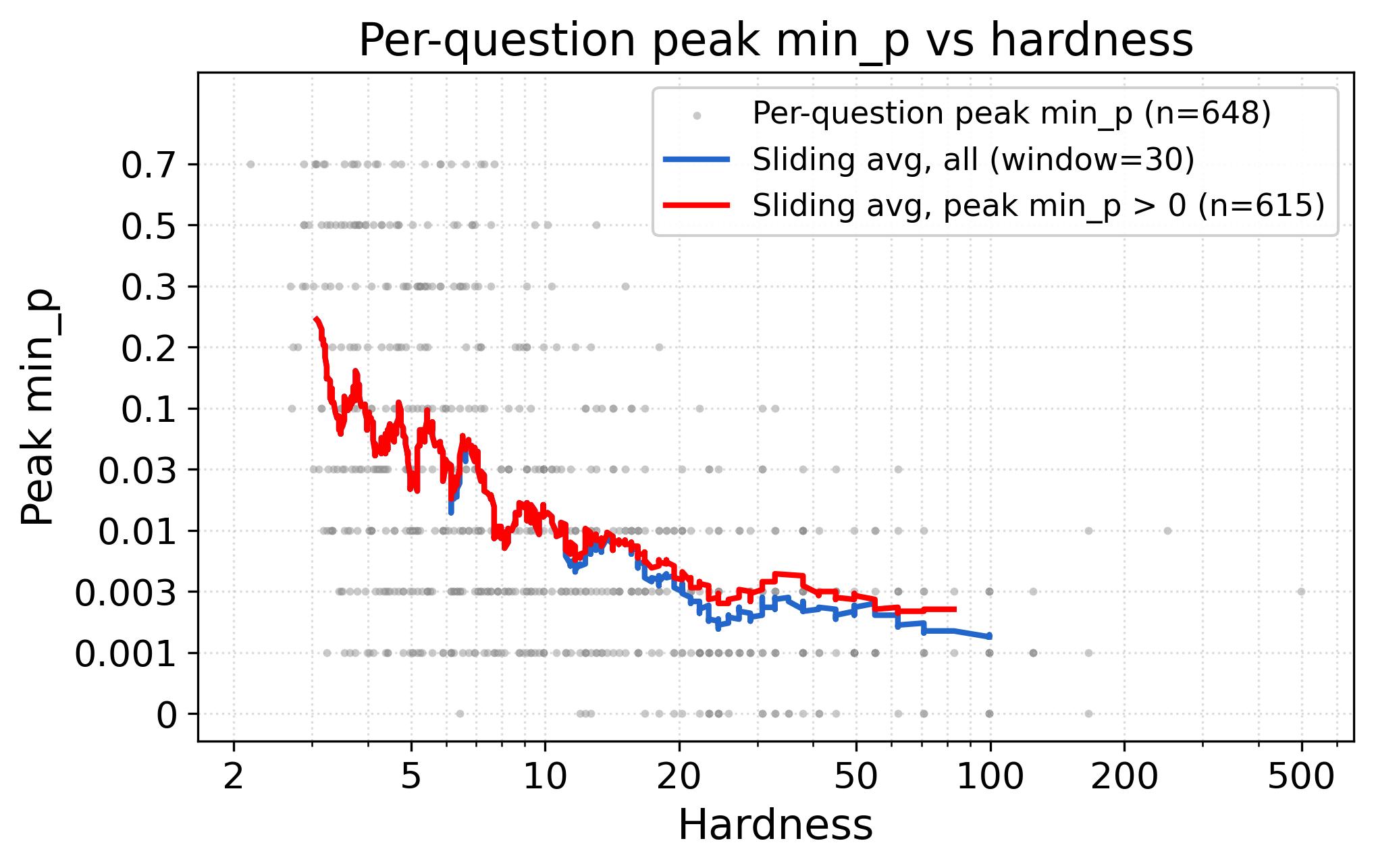}
    \caption{Qwen3-0.6B, $\tau = 1.5$}
    \label{fig:minp:Peak:q15}
  \end{subfigure}\hfill  \begin{subfigure}[t]{0.48\textwidth}
    \centering
    \includegraphics[width=0.75\linewidth]{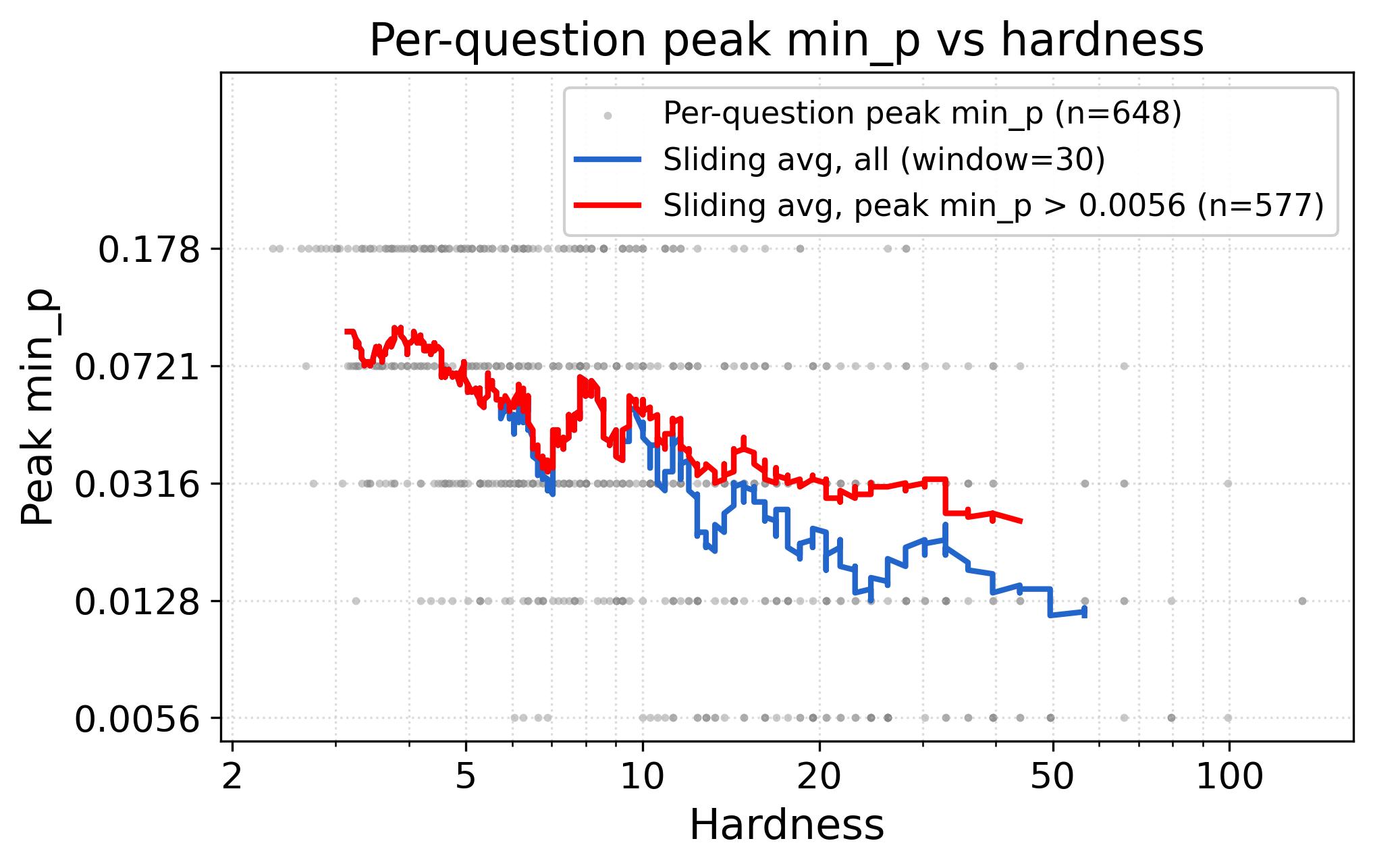}
    \caption{Qwen3-0.6B, $\tau = 2.0$}
    \label{fig:minp:Peak:q20}
  \end{subfigure}
  \vspace{8pt}

  \begin{subfigure}[t]{0.48\textwidth}
    \centering
    \includegraphics[width=0.75\linewidth]{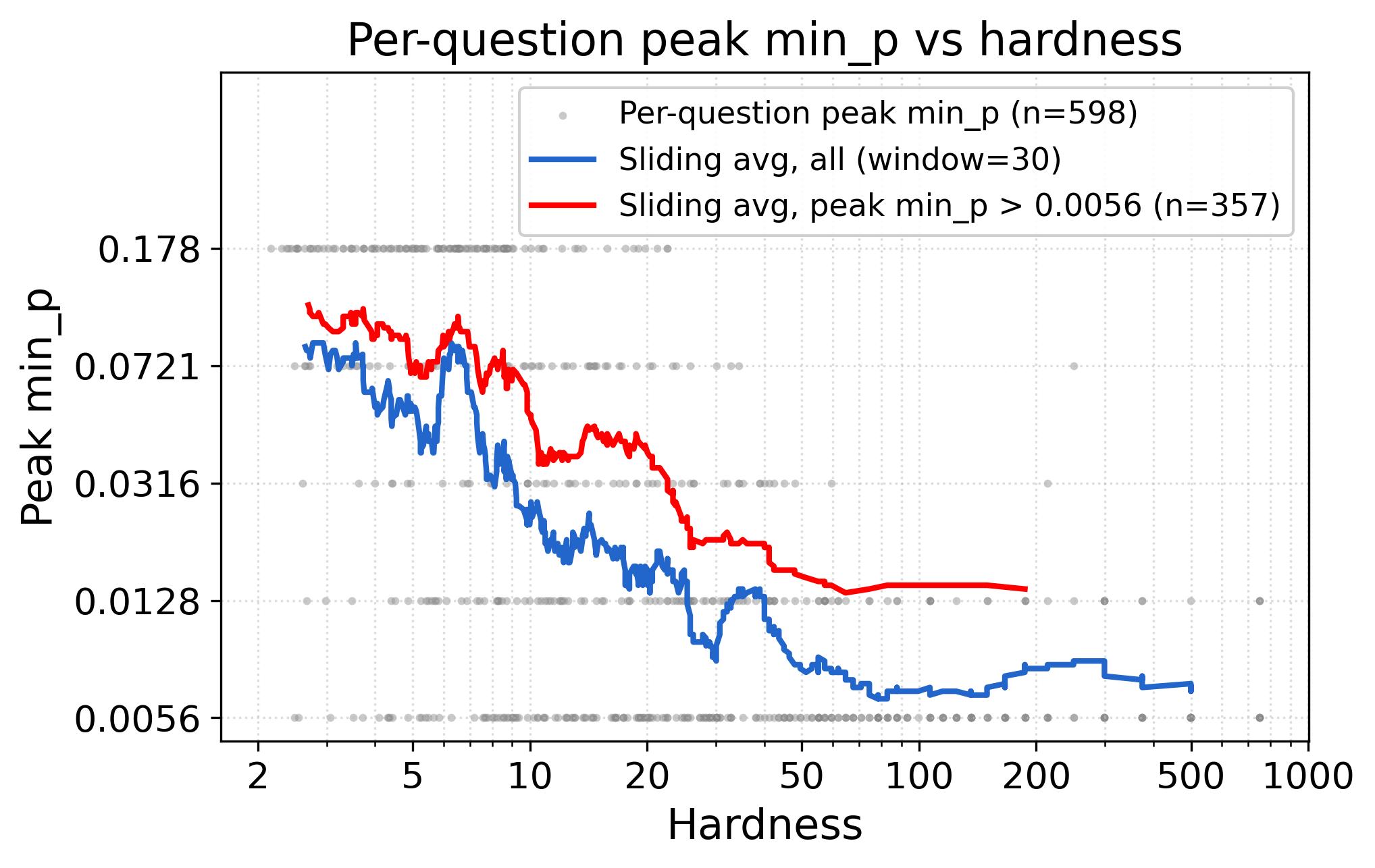}
    \caption{Mistral 7B Instruct, $\tau = 2.0$}
    \label{fig:minp:Peak:mistral}
  \end{subfigure}
  \caption{Per-question peak \texttt{min\_p} against hardness, with sliding averages over all questions (blue) and over the questions whose peak lies above the smallest grid value (red), the counterpart of Figure \ref{fig:AveTemp:grid}. The peak is the \texttt{min\_p} of the largest measured pass rate, the smallest among tied values. Grid values are spaced equally on the vertical axis, so the averages are means of grid position. The window is $30$ questions, and the averages are drawn only where the window is full. No further low-hardness points are dropped, since $H_q$ takes discrete values here.}
  \label{fig:minp:Peak}
\par}
\end{figure}

In Figure \ref{fig:minp:Peak}, harder questions peak at smaller \texttt{min\_p} in all three studies, that is, with fewer tokens removed. The OLS slope of the peak grid position on $\log_{10} H_q$ is $-3.10$ grid steps per decade at $\tau = 1.5$, and $-1.98$ and $-1.51$ at $\tau = 2.0$ for Qwen3-0.6B and Mistral 7B Instruct. It keeps its sign in all $400$ splits, and no null replicate reaches it. The central $95\%$ of the null slopes lies within $[-0.60, 0.54]$ in every study. Without the questions that peak at the smallest value, the slopes are $-2.86$, $-1.35$, and $-1.48$, and they pass both checks. The peak temperature trends of Figure \ref{fig:AveTemp:grid} rise gently with hardness once near-zero peaks are excluded; on these hard questions, \texttt{min\_p} moves the same way, toward more diverse sampling.

\begin{figure}[!htbp]
{\par
  \centering
  \begin{subfigure}[t]{0.48\textwidth}
    \centering
    \includegraphics[width=0.75\linewidth]{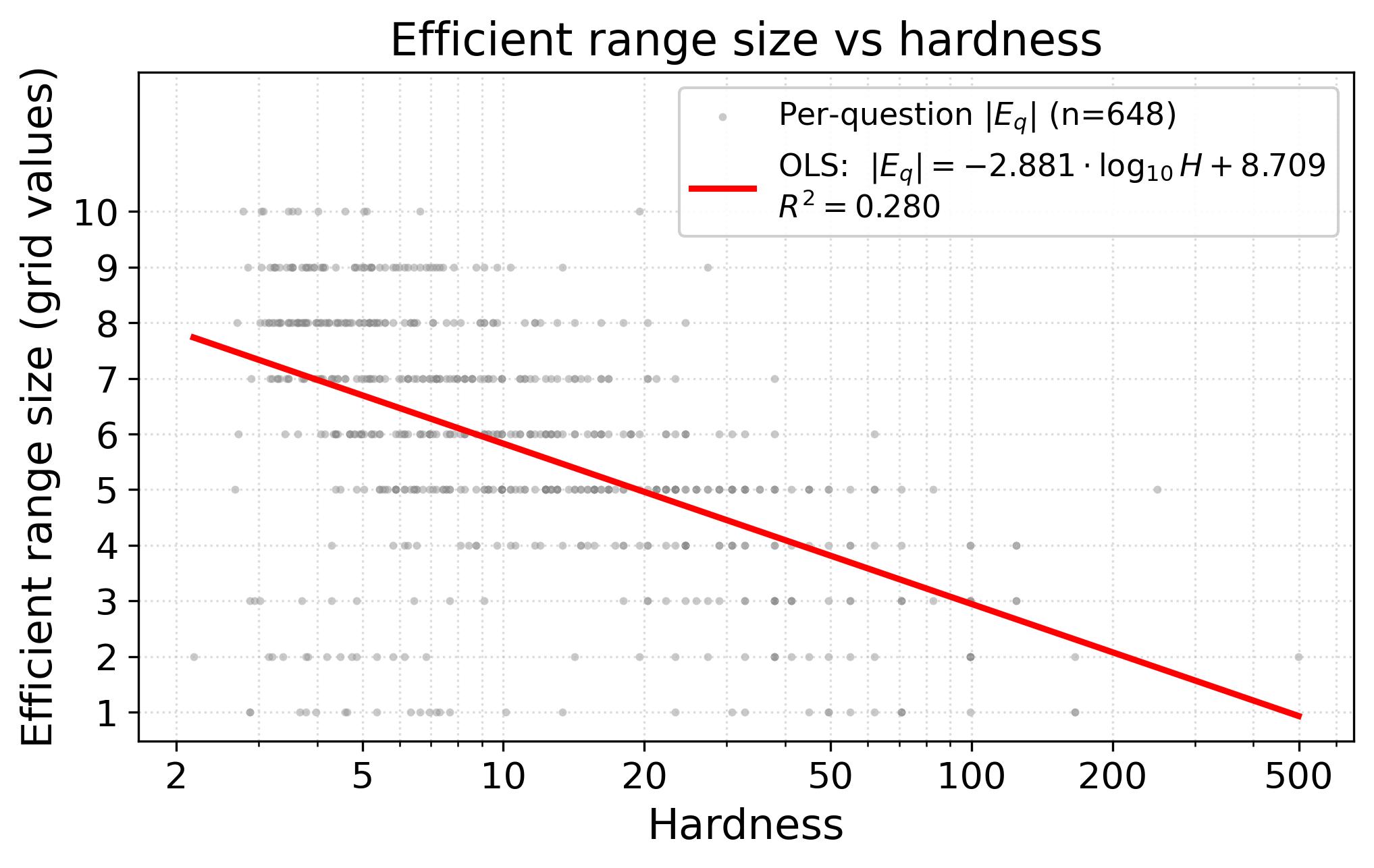}
    \caption{Qwen3-0.6B, $\tau = 1.5$}
    \label{fig:minp:RangeSize:q15}
  \end{subfigure}\hfill  \begin{subfigure}[t]{0.48\textwidth}
    \centering
    \includegraphics[width=0.75\linewidth]{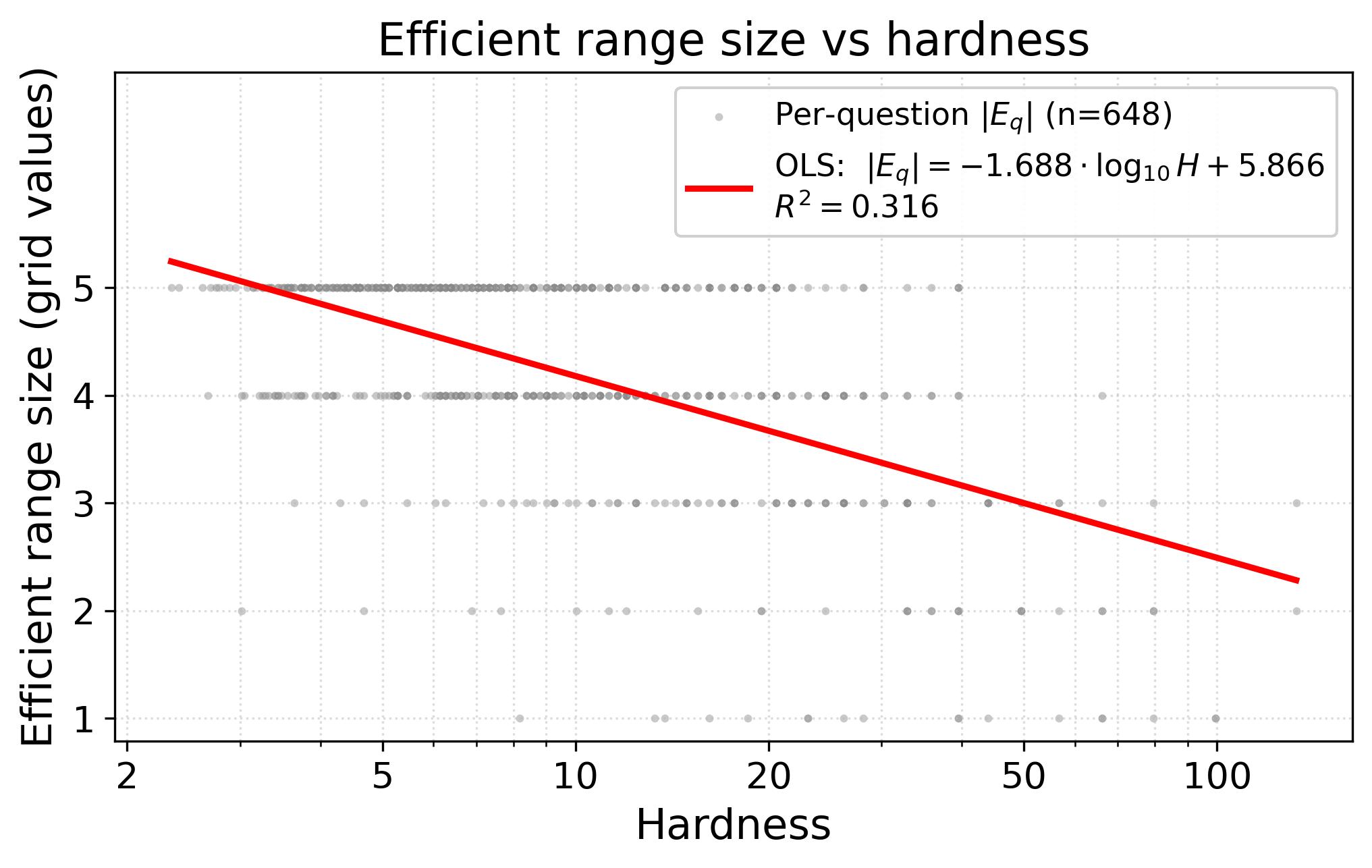}
    \caption{Qwen3-0.6B, $\tau = 2.0$}
    \label{fig:minp:RangeSize:q20}
  \end{subfigure}
  \vspace{8pt}

  \begin{subfigure}[t]{0.48\textwidth}
    \centering
    \includegraphics[width=0.75\linewidth]{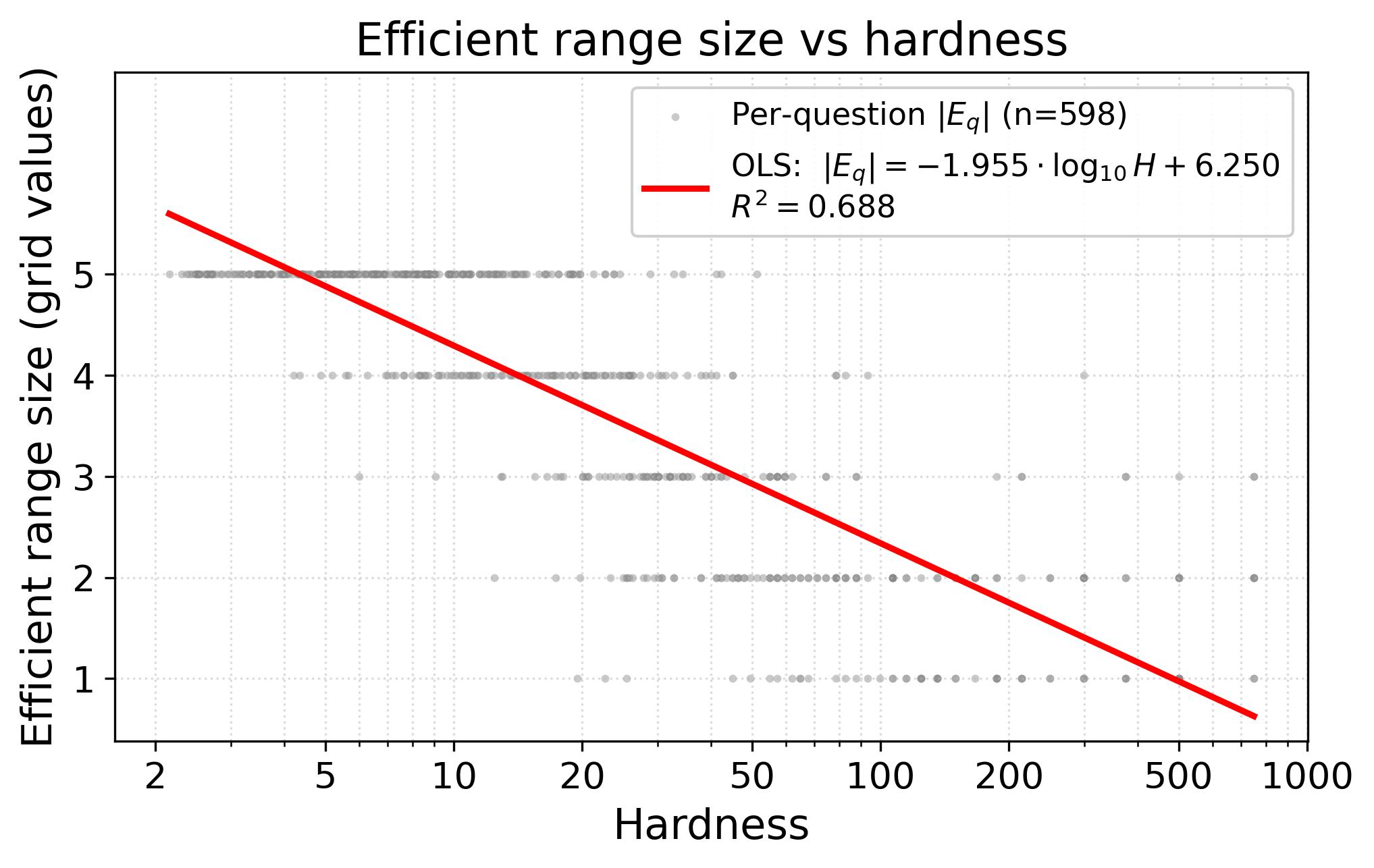}
    \caption{Mistral 7B Instruct, $\tau = 2.0$}
    \label{fig:minp:RangeSize:mistral}
  \end{subfigure}
  \caption{Per-question efficient range size against hardness, with the OLS fit on $\log_{10} H_q$, the counterpart of Figure \ref{fig:RangeSize:grid}. Here $|E_q|$ is the number of grid values in the efficient range, since the $10$ and $5$ values of these grids do not support the interpolated $t_R - t_L$.}
  \label{fig:minp:RangeSize}
\par}
\end{figure}

In Figure \ref{fig:minp:RangeSize}, the OLS slopes of the efficient range size on $\log_{10} H_q$ are $-2.88$, $-1.69$, and $-1.96$ grid values per decade. Noise alone also gives negative slopes, and the null gives median slopes of $-3.25$, $-1.43$, and $-1.20$. At $\tau = 2.0$, the range narrows with hardness beyond noise for both models, clearly for Mistral 7B Instruct, where no null replicate reaches its slope, and narrowly for Qwen3-0.6B, where $2$ of $400$ do. At $\tau = 1.5$, the measured slope is flatter than that of $398$ of the $400$ null replicates, but this null's common shape is wider than the profiles of the easiest questions, which overstates the narrowing from noise, so the test cannot tell a trend from noise there.

\begin{figure}[!htbp]
{\par
  \centering
  \begin{subfigure}[t]{0.48\textwidth}
    \centering
    \includegraphics[width=0.85\linewidth]{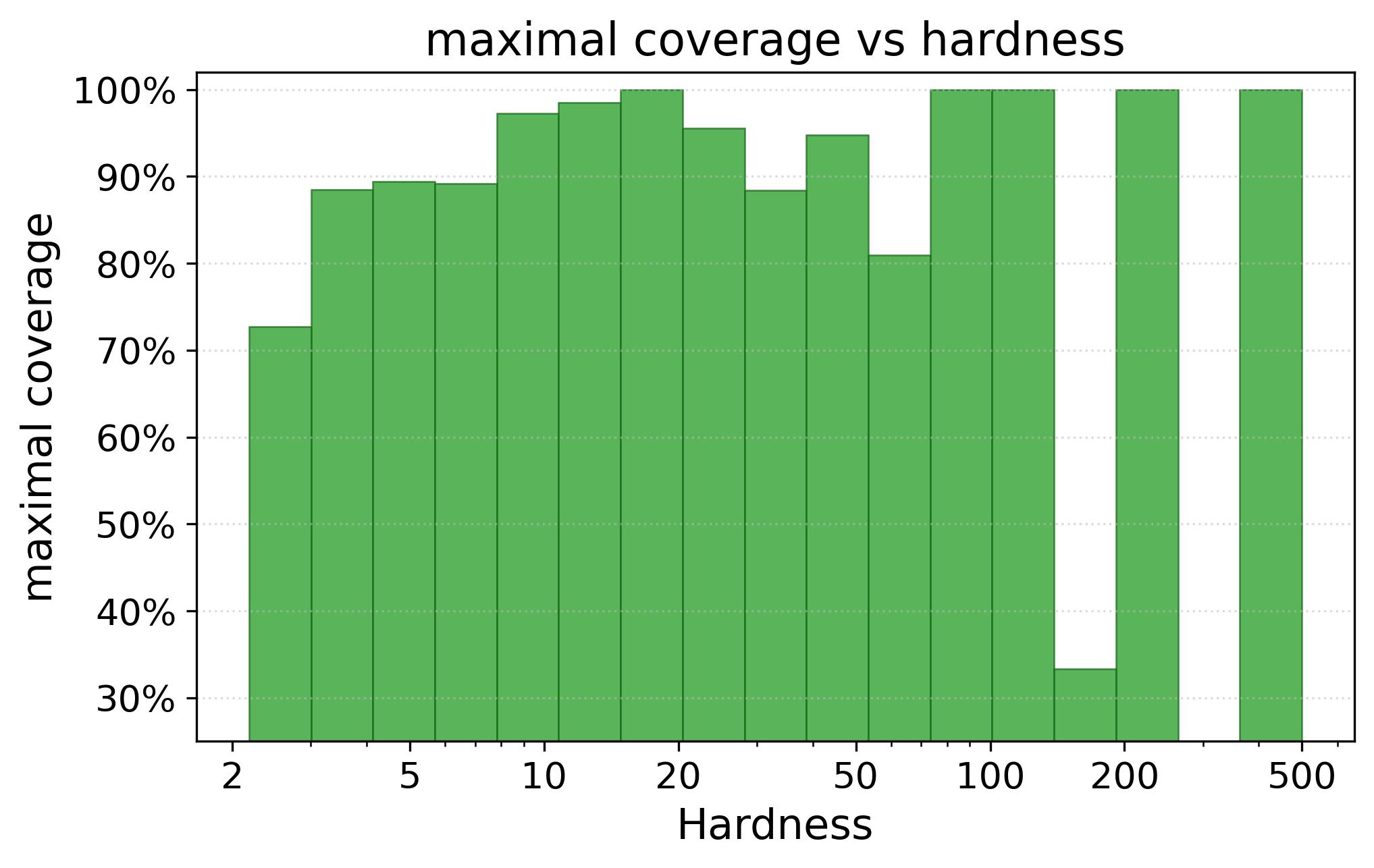}
    \caption{Qwen3-0.6B, $\tau = 1.5$}
    \label{fig:minp:relC:q15}
  \end{subfigure}\hfill  \begin{subfigure}[t]{0.48\textwidth}
    \centering
    \includegraphics[width=0.85\linewidth]{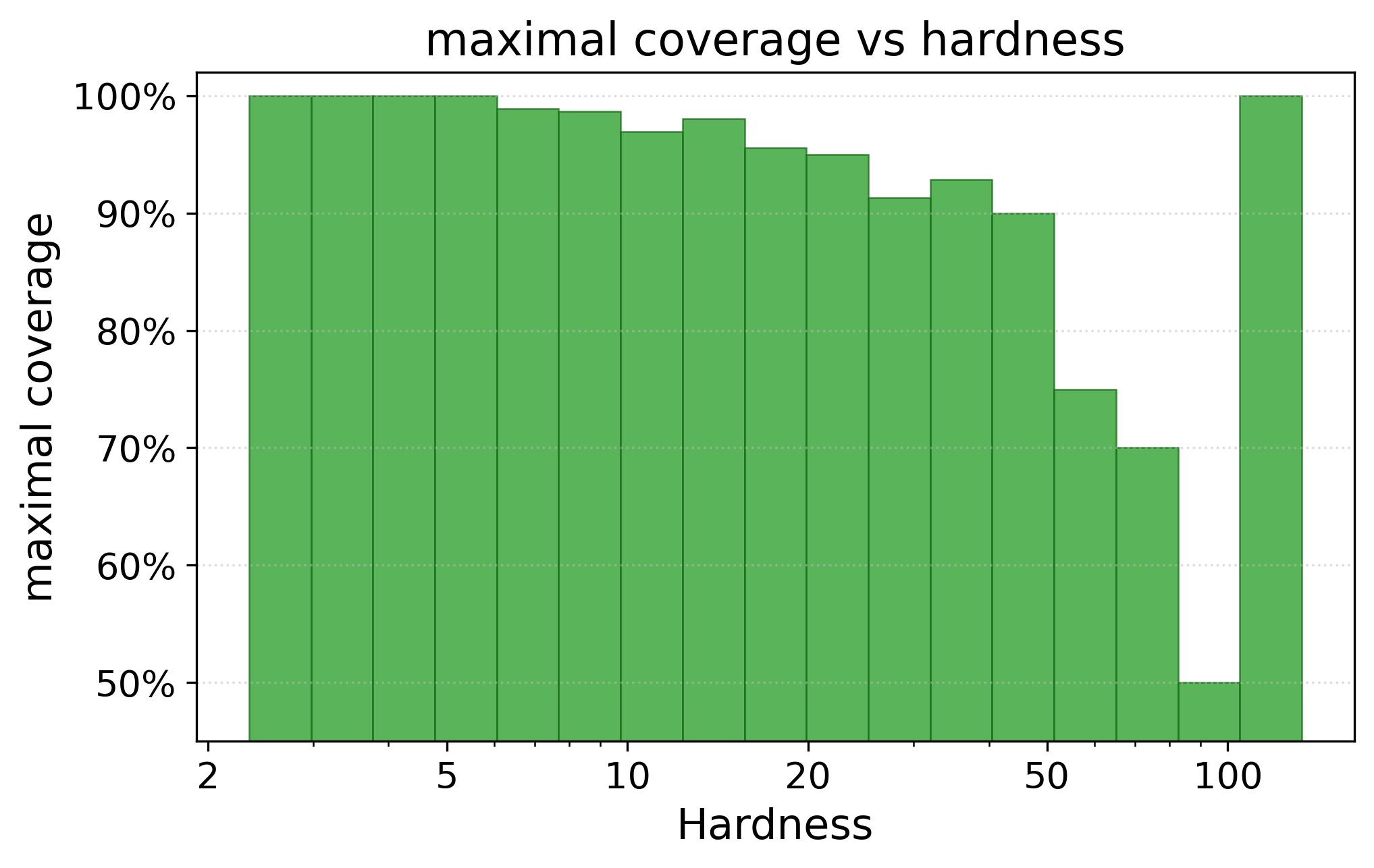}
    \caption{Qwen3-0.6B, $\tau = 2.0$}
    \label{fig:minp:relC:q20}
  \end{subfigure}
  \vspace{8pt}

  \begin{subfigure}[t]{0.48\textwidth}
    \centering
    \includegraphics[width=0.85\linewidth]{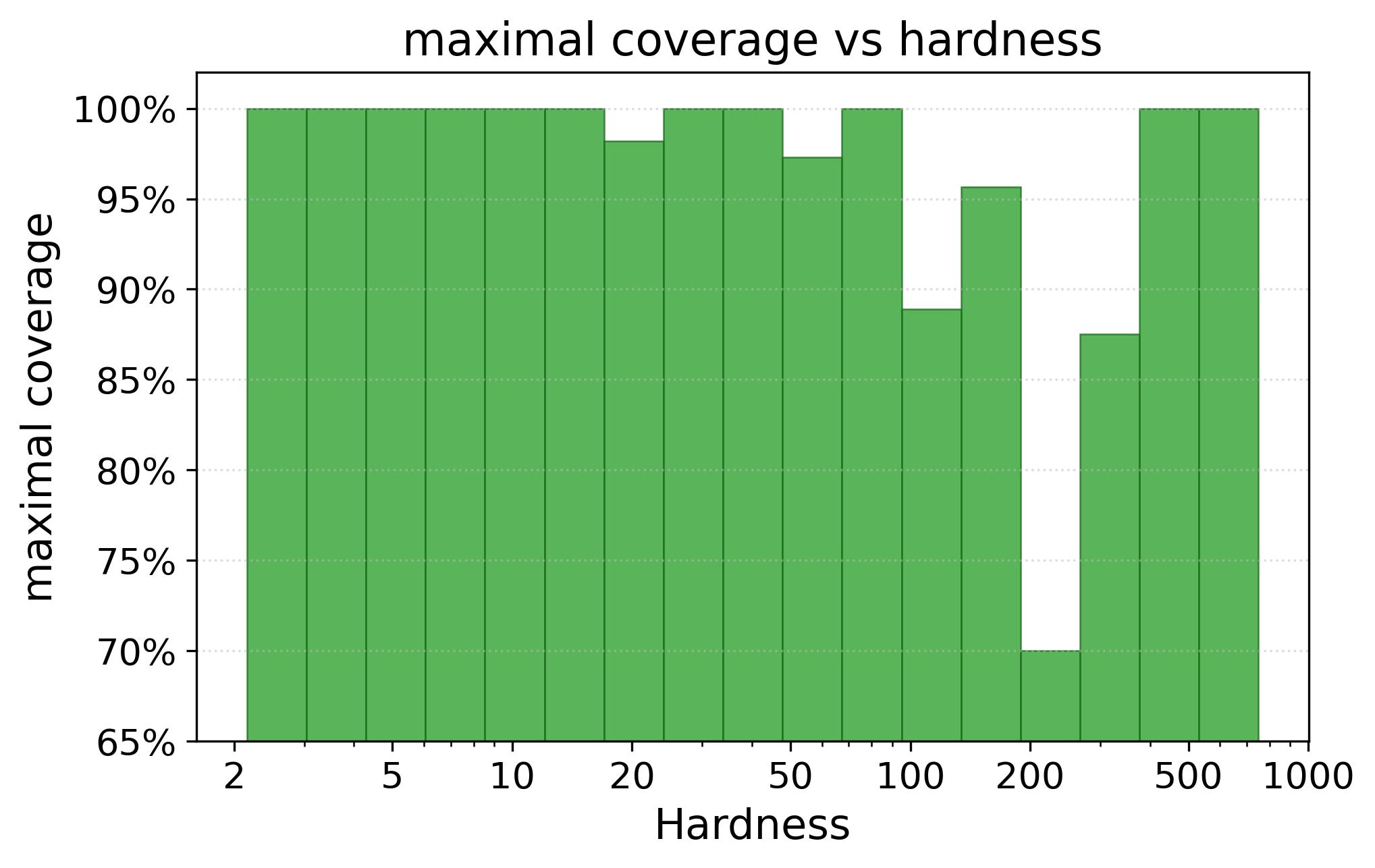}
    \caption{Mistral 7B Instruct, $\tau = 2.0$}
    \label{fig:minp:relC:mistral}
  \end{subfigure}
  \caption{Maximal coverage against hardness, the counterpart of Figure \ref{fig:relC:grid}. Questions are placed in $17$ log-uniform hardness bins, and the bar height is the largest share of the in-bin questions whose efficient ranges $E_q$ hold one grid value. Empty bins have no bar, and the bins wholly above $H = 100$ hold $9$, $2$, and $68$ questions in the three studies.}
  \label{fig:minp:relC}
\par}
\end{figure}

Figure \ref{fig:minp:relC} shows no trend beyond noise in any study. Among bins with at least $10$ questions, the lowest bar is $0.73$ at $\tau = 1.5$ and $0.70$ for both models at $\tau = 2.0$. Any fall of the bars with $H_q$ is within what noise alone produces, as the null slope of the bars on $\log_{10} H_q$, an unweighted OLS fit over all non-empty bins at their geometric centres, is at least as steep as the measured one in all $400$ replicates at $\tau = 1.5$ and for Mistral 7B Instruct, and in $41$ of $400$ for Qwen3-0.6B at $\tau = 2.0$. At $\tau = 1.5$, the split-half check fails as well.

\paragraph{Reweighting the questions.}
We repeat the random reweighting of Appendix \ref{zap:DataDis:RandomH} with \texttt{min\_p} in place of the temperature, using the same Dirichlet smooth-base sampler, seed, and bands, with $10{,}000$ random $W$'s per study reused across its four $K$ panels. The grid has $10$ or $5$ values. A question's pass@$K$ at each value is the unbiased estimator of its sample counts, and its peak pass rate, which sets the $0.1\%$ filter and $H$, is its largest measured rate on the grid. In Appendix \ref{zap:DataDis:RandomH}, both come from the fitted profiles. The largest $K$ is $512$ for Mistral 7B Instruct, as in Appendix \ref{zap:DataDis:RandomH}, and $256$ for Qwen3-0.6B, since $\nu = 500$ at $\tau = 1.5$ and $\nu = 400$ at $\tau = 2.0$. The $0.1\%$ peak filter keeps all $648$ questions of Qwen3-0.6B and $598$ of the $617$ of Mistral 7B Instruct, leaving out the $8$ with no correct sample and the $11$ whose best value has one correct sample in $1500$, so the entropy ceilings are $\ln(648) \approx 6.47$ and $\ln(598) \approx 6.39$ nats. All of these questions were kept as hard, with $H$ between $2.15$ and $749.5$ at their best value, so this test holds no easy questions.

\begin{figure}[!htbp]
{\par
  \centering
  \includegraphics[width=0.85\linewidth]{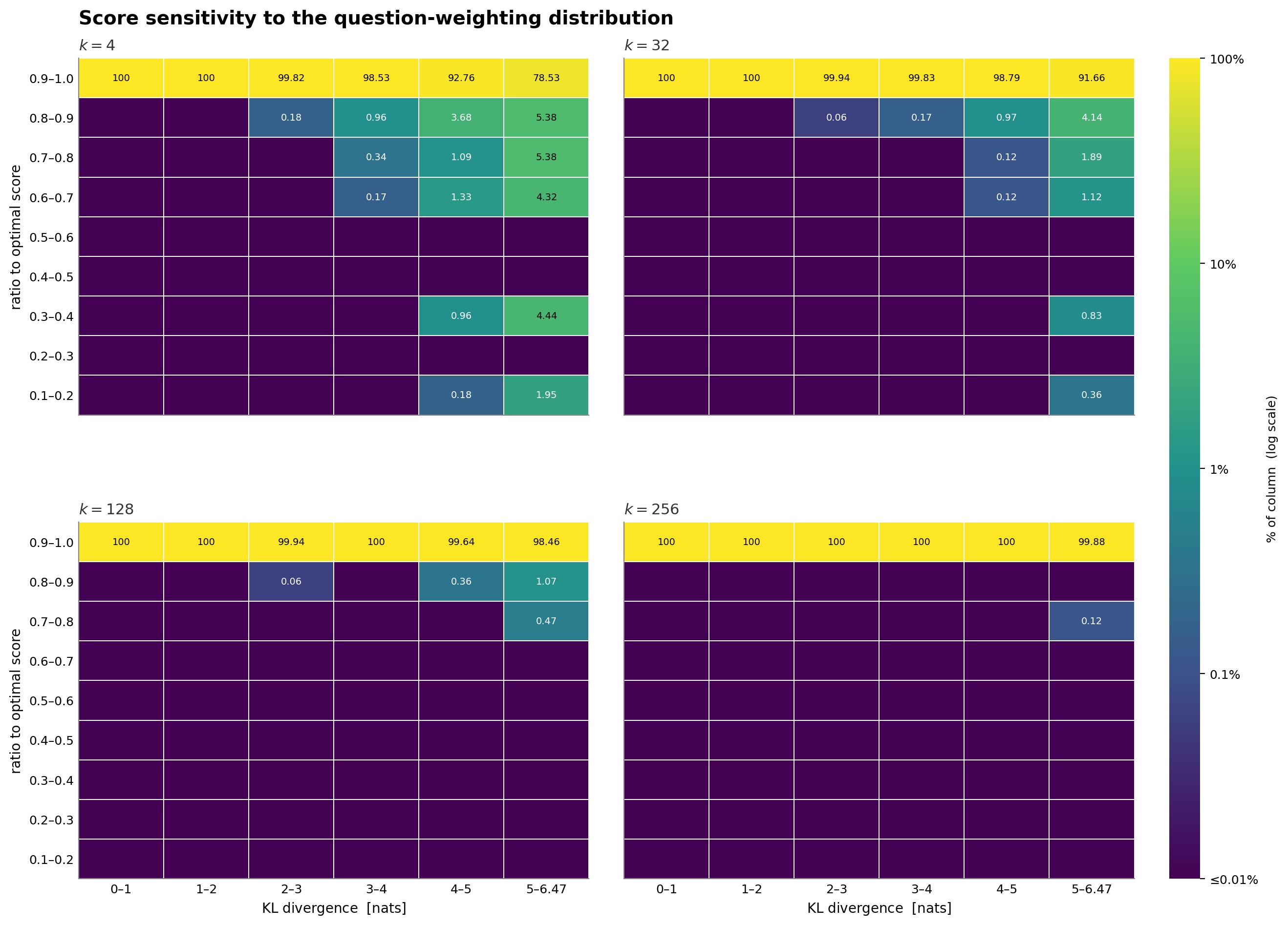}
  \caption{Distribution of $\mathrm{score}_B / \mathrm{score}_A$ as a function of $\mathrm{KL}(W \,\|\, \mathrm{uniform})$ for Qwen3-0.6B on Hi-ToM at $\tau = 1.5$ ($648$ included questions, $\nu = 500$, $K \in \{4, 32, 128, 256\}$, entropy ceiling $\ln(648) \approx 6.47$ nats), constructed as in Figure \ref{fig:RandomH:KGrid} with \texttt{min\_p} in place of the temperature.}
  \label{fig:minp:RandomH:q15}
\par}
\end{figure}

\begin{figure}[!htbp]
{\par
  \centering
  \includegraphics[width=0.85\linewidth]{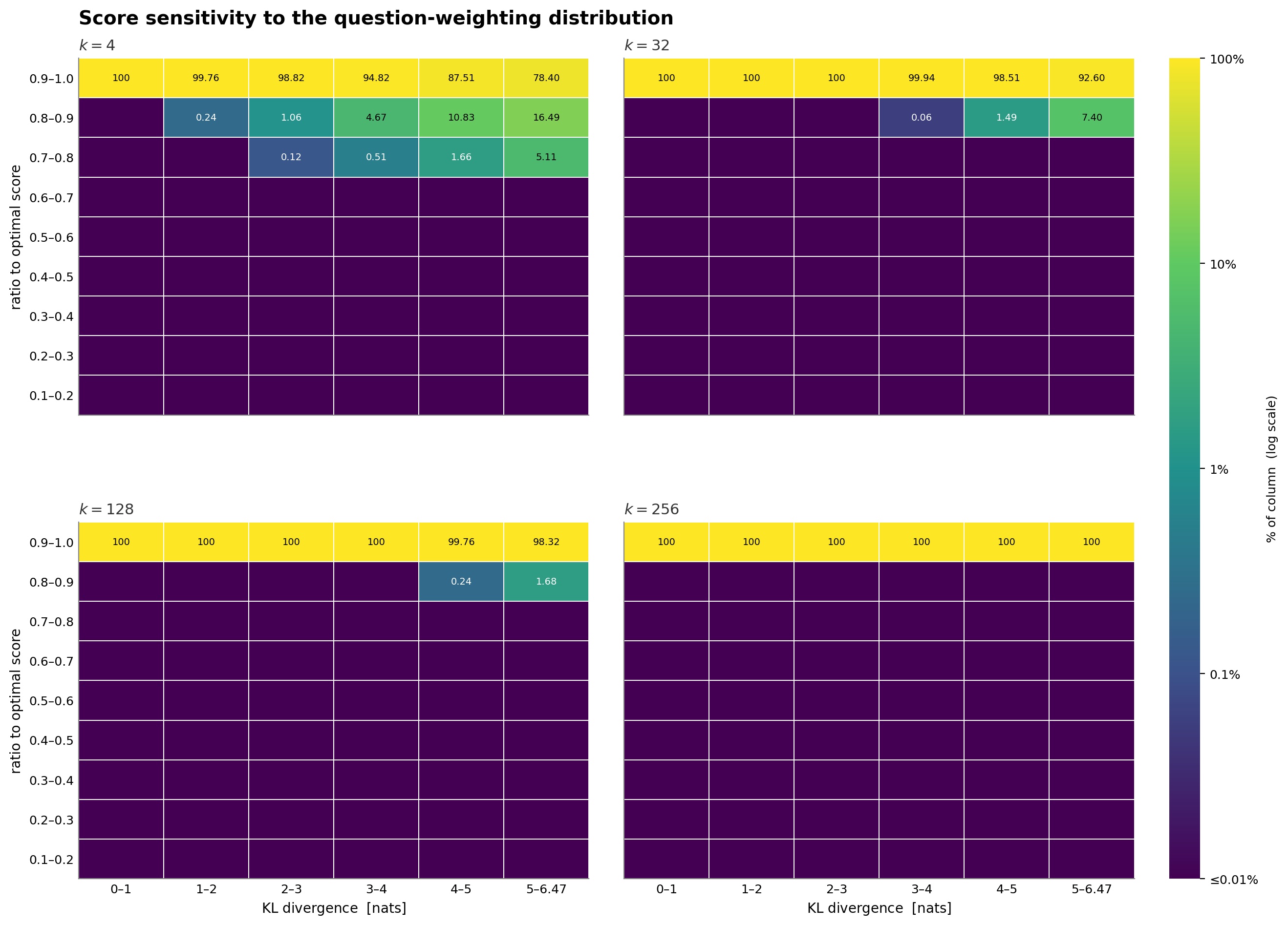}
  \caption{Distribution of $\mathrm{score}_B / \mathrm{score}_A$ as a function of $\mathrm{KL}(W \,\|\, \mathrm{uniform})$ for Qwen3-0.6B on Hi-ToM at $\tau = 2.0$ ($648$ included questions, $\nu = 400$, $K \in \{4, 32, 128, 256\}$, entropy ceiling $\ln(648) \approx 6.47$ nats), constructed as in Figure \ref{fig:RandomH:KGrid} with \texttt{min\_p} in place of the temperature.}
  \label{fig:minp:RandomH:q20}
\par}
\end{figure}

\begin{figure}[!htbp]
{\par
  \centering
  \includegraphics[width=0.85\linewidth]{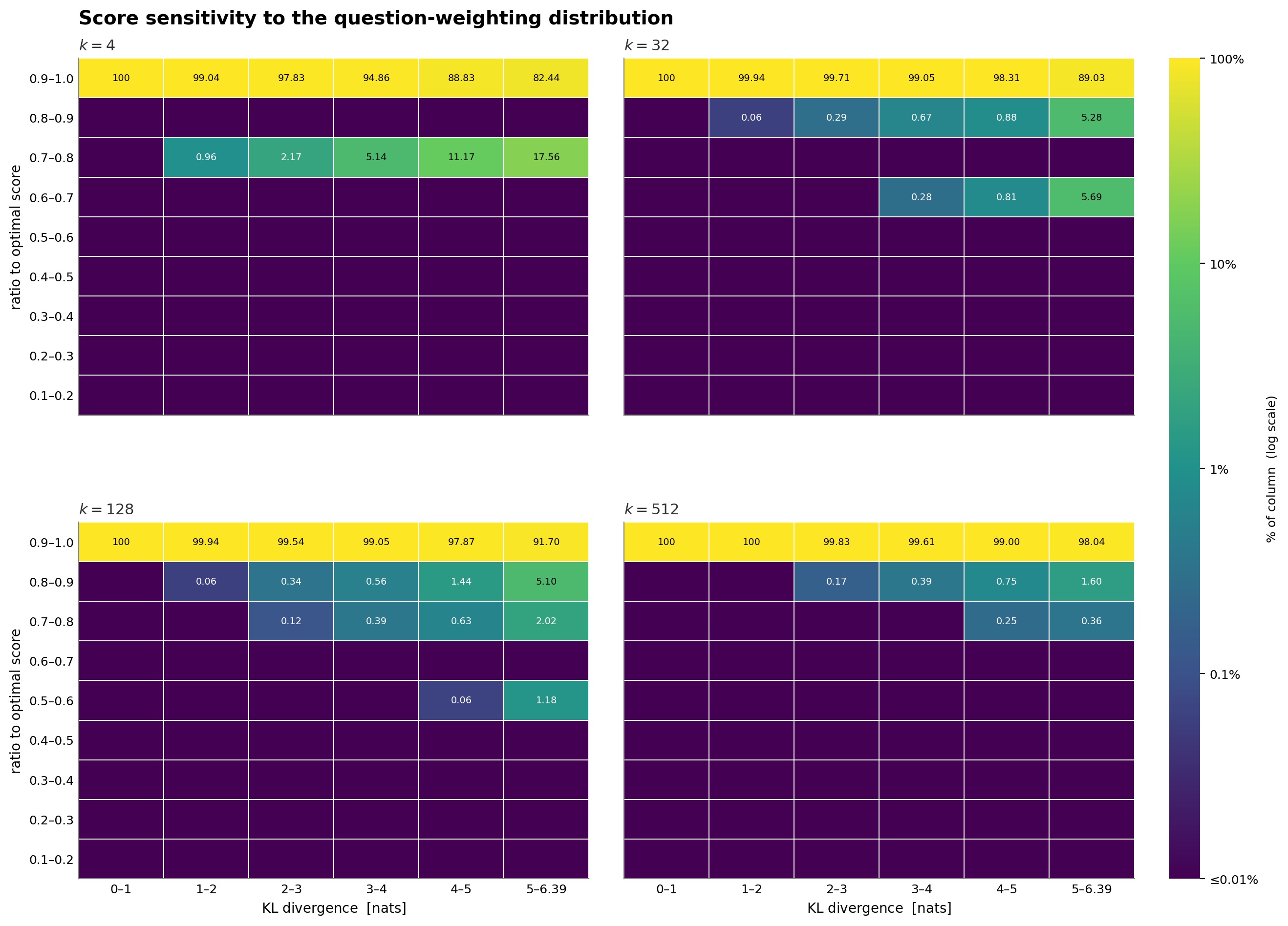}
  \caption{Distribution of $\mathrm{score}_B / \mathrm{score}_A$ as a function of $\mathrm{KL}(W \,\|\, \mathrm{uniform})$ for Mistral 7B Instruct on Hi-ToM at $\tau = 2.0$ ($598$ included questions, $\nu = 1500$, $K \in \{4, 32, 128, 512\}$, entropy ceiling $\ln(598) \approx 6.39$ nats), constructed as in Figure \ref{fig:RandomH:KGrid} with \texttt{min\_p} in place of the temperature.}
  \label{fig:minp:RandomH:mis}
\par}
\end{figure}

Figures \ref{fig:minp:RandomH:q15}, \ref{fig:minp:RandomH:q20}, and \ref{fig:minp:RandomH:mis} show that low KL costs little. In the $0$-$1$ band, every $W$ has a ratio of at least $0.9$ at every $K$ in all three studies. In the highest band, the top-row occupancy rises with $K$ in each study, from $78.53\%$, $78.40\%$, and $82.44\%$ at $K = 4$ to $99.88\%$, $100\%$, and $98.04\%$ at the largest $K$, for Qwen3-0.6B at $\tau = 1.5$ and $\tau = 2.0$ and Mistral 7B Instruct, respectively. For Qwen3-0.6B at $\tau = 2.0$ and $K = 256$, every value is within $10\%$ of the best, with a lowest ratio of $0.9075$, so the whole top row reads $100$. Here a label of $100$ means that every $W$ of the column has a ratio of at least $0.9$.

Here $\tau_A$ and $\tau_B$ are \texttt{min\_p} values of the grid, $10$ at $\tau = 1.5$ and $5$ at $\tau = 2.0$, so each panel's ratios take at most $10$ or $5$ distinct values. Every panel shows all $10{,}000$ $W$'s, since none has a computed KL below $0$ or at the ceiling. Every panel has the same nine ratio bands, from $0.1$-$0.2$ to $0.9$-$1.0$, and its six KL bands are drawn at equal width, so all cells have the same size. The lowest band that holds a $W$ is $0.1$-$0.2$, $0.7$-$0.8$, and $0.5$-$0.6$ for Qwen3-0.6B at $\tau = 1.5$ and $\tau = 2.0$ and Mistral 7B Instruct, respectively, and $163$, $178$, and $162$ of the $216$ cells of each figure hold no $W$.

\begin{figure}[!htbp]
{\par
  \centering
  \begin{subfigure}[t]{0.48\textwidth}
    \centering
    \includegraphics[width=0.85\linewidth]{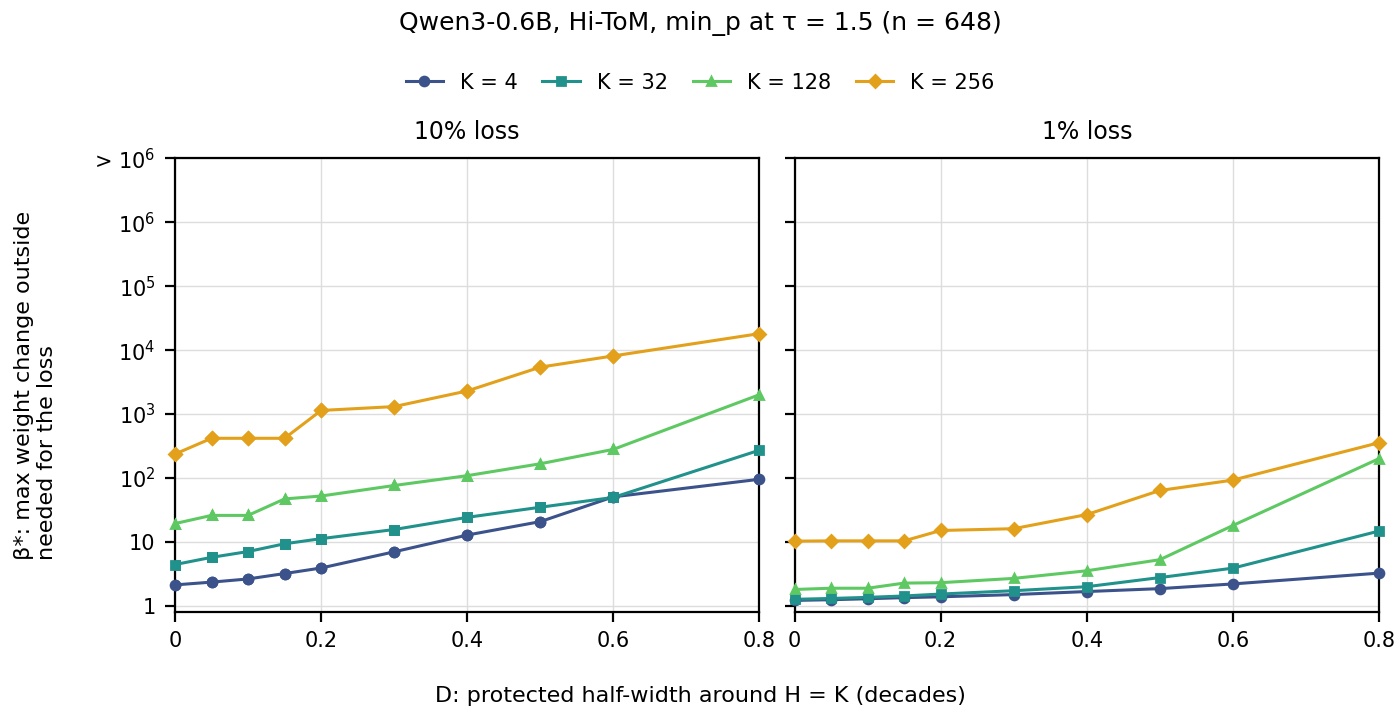}
    \caption{Qwen3-0.6B, $\tau = 1.5$}
  \end{subfigure}\hfill  \begin{subfigure}[t]{0.48\textwidth}
    \centering
    \includegraphics[width=0.85\linewidth]{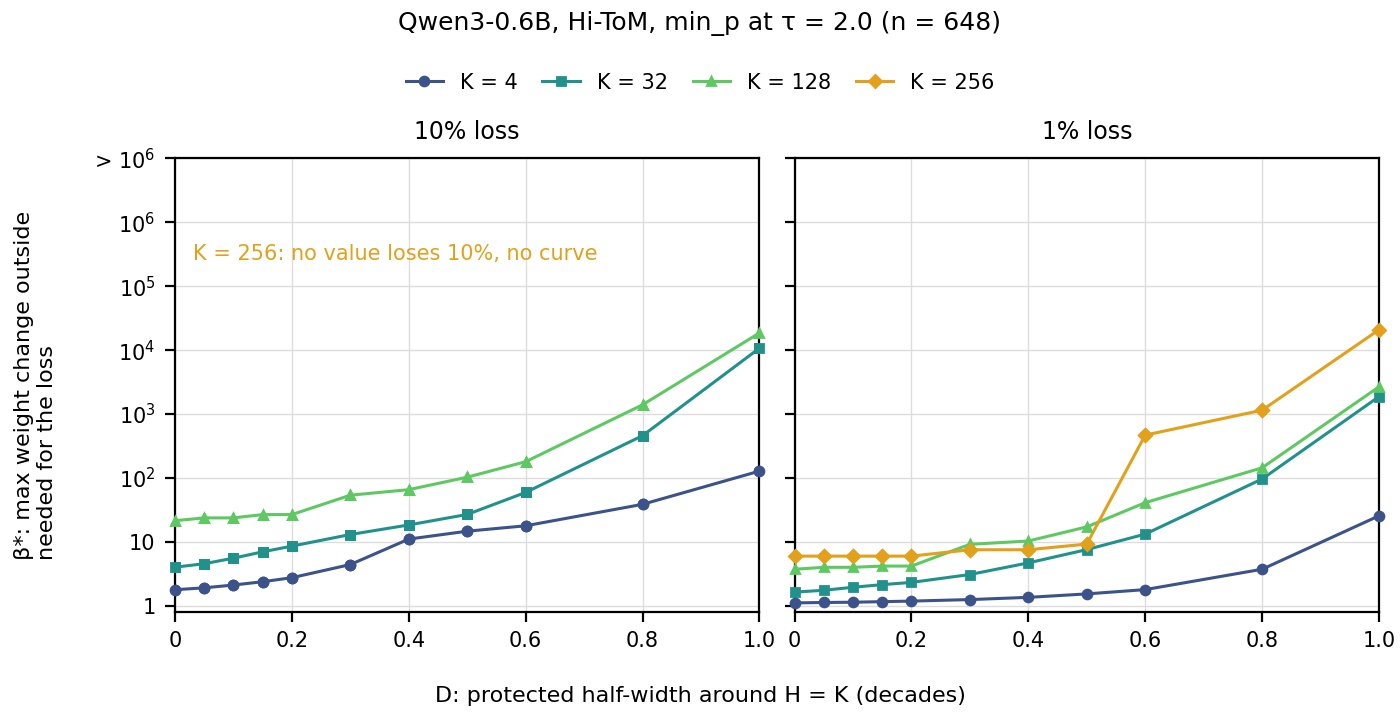}
    \caption{Qwen3-0.6B, $\tau = 2.0$}
  \end{subfigure}
  \vspace{8pt}

  \begin{subfigure}[t]{0.48\textwidth}
    \centering
    \includegraphics[width=0.85\linewidth]{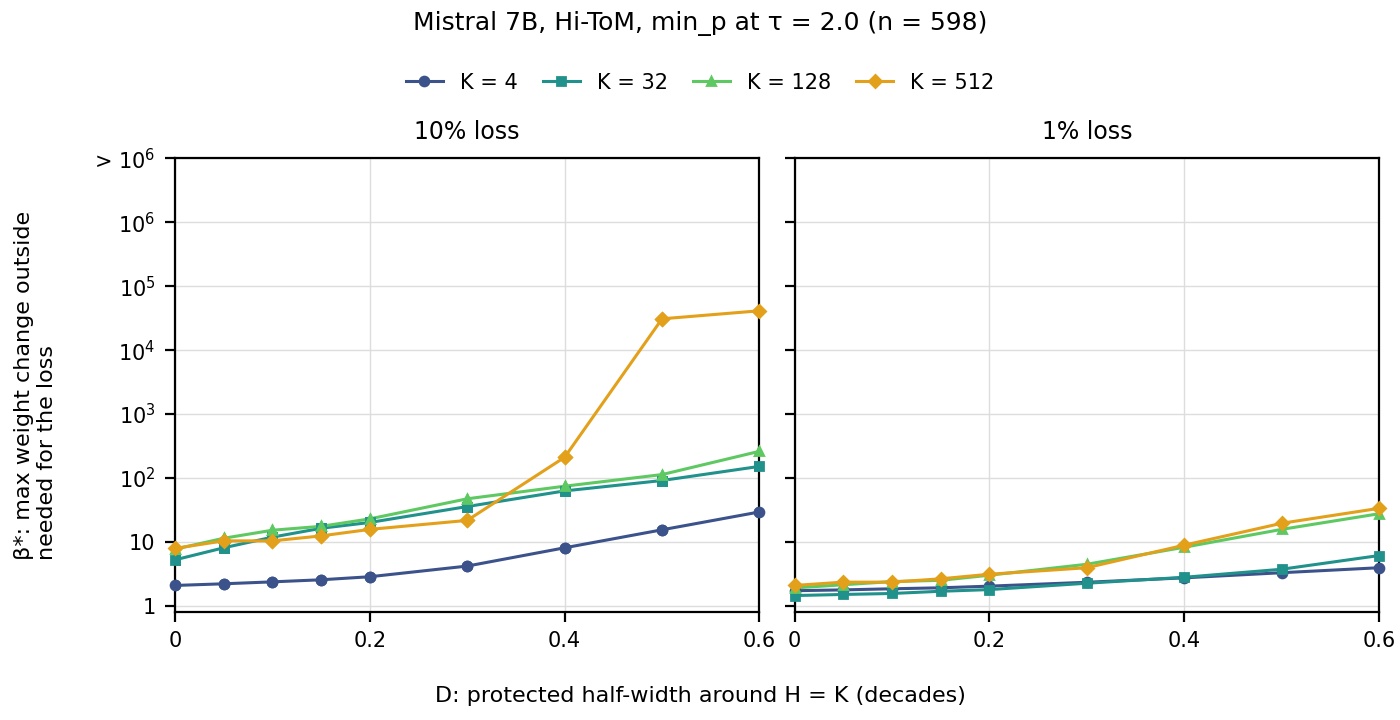}
    \caption{Mistral 7B Instruct, $\tau = 2.0$}
  \end{subfigure}
  \caption{$\beta^*$ against $D$ on Hi-ToM with \texttt{min\_p} in place of the temperature, drawn as in Figure \ref{fig:Prot:D:LCB}, for $K \in \{4, 32, 128, 256\}$ for Qwen3-0.6B and $K \in \{4, 32, 128, 512\}$ for Mistral 7B Instruct. Each panel shows $D$ up to the last grid value before any of its curves exceeds $10^6$: $0.8$, $1.0$, and $0.6$ in (a), (b), and (c). Hardness is measured at each question's best value, and pass@$K$ is the unbiased estimator of the sample counts. For Qwen3-0.6B at $\tau = 2.0$ and a $10\%$ loss, no value loses $10\%$ at $K = 256$, so $C$ is empty and the panel has no curve for it. Every other curve has a reachable loss at $D = 0$.}
  \label{fig:minp:Prot}
\par}
\end{figure}

\ifdefined\nolinenumbers\nolinenumbers\fi %
\paragraph{Reweighting outside a protected window.}
We also run the test of Appendix \ref{zap:DataDis:Prot} with \texttt{min\_p} in place of the temperature, with the same $D$ grid, losses, $\beta$ grid, and exact certification, on the questions and pass@$K$ values of the reweighting above. The hardness $H_q$ of a question is measured at its best value. Figure \ref{fig:minp:Prot} shows $\beta^*$ against $D$.

With no protected window, a weighting that changes no weight by more than a factor of $1.78$ to $237$ makes a value that loses $10\%$ the best, and a factor of $1.11$ to $10.27$ suffices for a $1\%$ loss. $\beta^*$ exceeds $10^6$ from $D = 0.8$ to $1.75$ at a $10\%$ loss and from $D = 1.25$ to $2.0$ at a $1\%$ loss. When the hardness and pass@$K$ of each question come from separate random halves of its samples, with the inclusion floor applied to the hardness half and the sweep run without exact certification, this $D$ is unchanged or one grid step away in $192$ of $200$ cases, $10$ splits of each of the $20$ curves with $K \le \nu / 2$. In the other $8$ cases, all for Mistral 7B Instruct at a $10\%$ loss, it moves by more than one grid step. The largest are at $K = 4$, where this $D$ is $0.8$ on all samples and splits give $2.0$ and $2.5$, and at $K = 512$, where a split gives $0.5$ against $1.25$. Since the hardness of these questions spans less than $2.6$ decades, the first $D$ whose window holds every question is $1.25$ to $2.5$, and $6$ of the $23$ curves reach the top row only at that $D$.

\paragraph{Limitations.}
The study covers two models, one dataset, and two temperatures, and the best combination of $\tau$ and \texttt{min\_p} may lie at an untested temperature, such as one above $2.0$ or between $1.5$ and $2.0$. The $648$ and $617$ questions are those kept as hard, so their absolute pass rates are not benchmark scores. Since every best value above is an in-sample maximum, its margin is optimistic, and the standard errors ignore the pick. The split-half checks guard against this in the comparisons between temperatures and with the best plain temperature, and the results of the decision rule depend on the split seed in two cases, both at $K = 256$: between the two temperatures, and between $\tau = 1.5$ and the best plain temperature. We compared one-dimensional slices, \texttt{min\_p} at fixed $\tau$ and $\tau$ at $\texttt{min\_p} = 0$, and did not run \texttt{min\_p} at the best plain temperatures, $1.275$ to $1.35$ for Qwen3-0.6B and, at $K \le 256$, $1.575$ to $1.8$ for Mistral 7B Instruct. For Qwen3-0.6B, these come from measured pass rates on the $642$ questions. The table of Appendix \ref{zap:Multiple:Comp:res} uses fitted profiles on all $743$ questions kept by its first stage.

\ifdefined\nolinenumbers\nolinenumbers\fi %
For Qwen3-0.6B, this study and the sweep also differ in hardware, seeds, and $\nu$. On all 1,200 questions, plain sampling in the $\tau = 1.5$ screen passes $18.78\%$ of its samples against $18.48\%$ for $\tau = 1.5$ in the sweep. We have not resolved this gap of $0.30$ pp (paired $z = 2.70$). It is $+0.35$ pp on the 1,090 questions that the screen ran on RTX 5090 GPUs and $-0.19$ pp on the 110 on the L40S GPU, too few to tell the two GPUs apart.

For Mistral 7B Instruct, the reference samples combine the $1000$ that selected the questions, on RTX 5090 GPUs, with $500$ more, drawn on L40S GPUs or from a continuation of the first stage's run. Selecting on low observed rates biases the reference's pass rates slightly down, while the \texttt{min\_p} samples are fresh, so the selection mildly favors \texttt{min\_p}. We have not measured the size of this bias. It can act only on questions near the $0.30$ threshold, mostly at small $K$. Over all $1500$ samples, $30$ of the $617$ questions have a best single-temperature pass rate of at least $0.25$, and none of $0.30$ or more.

For $\tau = 2.0$ against $\tau = 1.5$ and for Mistral 7B Instruct, the results of the decision rule hold on the questions of each GPU type. For Qwen3-0.6B, the $27$ RTX 4090 questions are enough to decide at $K = 16$ and too few at the other $K$, and the borderline result at $K = 256$ becomes no detectable difference. For Mistral 7B Instruct, the L40S gap at $K = 256$ just clears $2$ standard errors.

\FloatBarrier

\section{Expanded Proofs}

In this section, we present the expanded proofs of several results from earlier sections and appendices. We begin with the proofs corresponding to Appendix \ref{zap:sec:Ideal}, which are included here to avoid unnecessary clutter.

\subsection{Proofs of the Ideal Sampling Method Approximation Theorems}

 \label{zap:sec:Ideal:exp}

We start with the proof of Theorem \ref{zap:the:TempDis}.

\begin{proof}[\textbf{Proof - Theorem \ref{zap:the:TempDis}}]
    
Let $\mathcal Q_{\mathrm{good}}$ denote the set of questions for which Equations \ref{zap:eq:the:TempDis:lip} and \ref{zap:eq:the:TempDis:truncation} hold. By the assumption,
\begin{equation}
\Pr_{q\sim\mathcal Q}\left(q\in\mathcal Q_{\mathrm{good}}\right)\geq 1-\rho \ .
\end{equation}
Fix arbitrary temperatures $\tau_1,\ldots,\tau_K\geq0$. We define a corresponding set of discrete temperatures $\gamma_1,\ldots,\gamma_K\in\mathcal G$ as follows. If $\tau_k\leq T$, choose $\gamma_k\in\mathcal G$ such that $|\tau_k-\gamma_k|\leq\Delta$. This is possible by the definition of the grid in Equation \ref{zap:eq:the:TempDis:Dis}. If $\tau_k>T$, set $\gamma_k=M\Delta$.

Now fix some $q\in\mathcal Q_{\mathrm{good}}$. If $\tau_k\leq T$, then by Equation \ref{zap:eq:the:TempDis:lip},
\begin{equation}
\bar P_q(\gamma_k)
\leq
\exp\left(L|\tau_k-\gamma_k|\right)\bar P_q(\tau_k)
\leq
\exp(L\Delta)\bar P_q(\tau_k) \ .
\end{equation}
If $\tau_k>T$, then by Equation \ref{zap:eq:the:TempDis:truncation},
\begin{equation}
\bar P_q(\gamma_k)
=
\bar P_q(M\Delta)
\leq
\exp(\alpha)\bar P_q(\tau_k) \ .
\end{equation}
Therefore, for every $k$, we have the common bound
\begin{equation}
\bar P_q(\gamma_k)
\leq
\exp(L\Delta+\alpha)\bar P_q(\tau_k) \ .
\end{equation}
Multiplying over $k=1,\ldots,K$, we get
\begin{equation}
\prod_{k=1}^{K}\bar P_q(\gamma_k)
\leq
\exp\left(K(L\Delta+\alpha)\right)
\prod_{k=1}^{K}\bar P_q(\tau_k) \ .
\end{equation}

We can now average over $q\sim\mathcal Q$. Using the definition of $\bar {\mathcal{P}}$ from Equation \ref{eq:MultiTempProb:Basic}, and splitting the expectation over $\mathcal Q_{\mathrm{good}}$ and its complement, we obtain
\begin{equation}
\bar{\mathcal{P}}(\gamma_1,\ldots,\gamma_K)
=
\mathbb E_{q\sim\mathcal Q}
\prod_{k=1}^{K}\bar P_q(\gamma_k) \ .
\end{equation}
On the good set, we use the multiplicative bound above. On the complement of the good set, we use only the fact that $0\leq \bar P_q(\tau)\leq1$, and hence
$\prod_{k=1}^{K}\bar P_q(\gamma_k)\leq1$. Therefore,
\begin{equation}
\bar{\mathcal{P}}(\gamma_1,\ldots,\gamma_K)
\leq
\exp\left(K(L\Delta+\alpha)\right)
\mathbb E_{q\sim\mathcal Q}
\prod_{k=1}^{K}\bar P_q(\tau_k)
+
\Pr_{q\sim\mathcal Q}\left(q\notin\mathcal Q_{\mathrm{good}}\right) \ .
\end{equation}
Since $\Pr(q\notin\mathcal Q_{\mathrm{good}})\leq\rho$, this gives
\begin{equation}
\bar{\mathcal{P}}(\gamma_1,\ldots,\gamma_K)
\leq
\exp\left(K(L\Delta+\alpha)\right)
\bar{\mathcal{P}}(\tau_1,\ldots,\tau_K)
+
\rho \ .
\end{equation}

Since $\tau_1,\ldots,\tau_K$ were arbitrary, we may take the infimum over all continuous temperatures. Using the definitions of $\bar{\mathcal{P}}^{\star}$ and $\bar{\mathcal{P}}_{\mathcal G}^{\star}$ in Equation \ref{eq:ideal}, we obtain
\begin{equation}
\bar{\mathcal{P}}_{\mathcal G}^{\star}
\leq
\exp\left(K(L\Delta+\alpha)\right)\bar{\mathcal{P}}^{\star}
+
\rho \ .
\end{equation}
Rearranging gives
\begin{equation}
\bar{\mathcal{P}}_{\mathcal G}^{\star}-\bar{\mathcal{P}}^{\star}
\leq
\left(
\exp\left(K(L\Delta+\alpha)\right)-1
\right)\bar{\mathcal{P}}^{\star}
+
\rho \ .
\end{equation}
Finally, since $\mathcal G\subseteq[0,\infty)$, the discrete optimization is over a restricted set of temperatures. Hence,
\begin{equation}
\bar{\mathcal{P}}^{\star}
\leq
\bar{\mathcal{P}}_{\mathcal G}^{\star} \ .
\end{equation}
Combining the last two displays proves that
\begin{equation}
0
\leq
\bar{\mathcal{P}}_{\mathcal G}^{\star}-\bar{\mathcal{P}}^{\star}
\leq
\left(e^{K(L\Delta+\alpha)}-1\right)\bar{\mathcal{P}}^{\star}
+
\rho \ ,
\end{equation}
as required.
\end{proof}

We can now present the proof of the other main result of this section, Theorem \ref{zap:lem:FiniteQ}, thereby completing all appendix proofs relevant to our analysis:

\begin{proof}[\textbf{Proof - Theorem \ref{zap:lem:FiniteQ}}]
The proof follows the standard uniform convergence argument, but with a variance-sensitive bound. The main idea is to compare every grid assignment to the best grid assignment under the population objective, and then show that a bad population assignment is unlikely to look better on the sampled questions.

Let $\gamma=(\gamma_1,\ldots,\gamma_K)$ denote an arbitrary assignment of grid temperatures. For each sampled question $q_i$, define
\begin{equation}
X_i(\gamma)
=
\prod_{k=1}^{K}
\bar P_{q_i}(\gamma_k) \ .
\end{equation}
Since $0\leq \bar P_q(\tau)\leq 1$, we have $0\leq X_i(\gamma)\leq 1$. Moreover, by Equations \ref{eq:ProbFailTot} and \ref{eq:lemFinite:emp},
\begin{equation}
\mathbb E_{q_i\sim\mathcal Q} X_i(\gamma)
=
\bar {\mathcal{P}}_{\mathcal Q}(\gamma_1,\ldots,\gamma_K),
\qquad
\frac{1}{n}\sum_{i=1}^{n}X_i(\gamma)
=
\bar {\mathcal{P}}_{\mathcal Q'}(\gamma_1,\ldots,\gamma_K) \ .
\end{equation}
Let also $\gamma^{\star}=(\gamma_1^{\star},\ldots,\gamma_K^{\star})$ be an assignment minimizing the population objective over the grid. Thus, by Equation \ref{eq:ideal},
\begin{equation}
\bar{\mathcal{P}}_{\mathcal Q}(\gamma^{\star})
=
\bar{\mathcal{P}}_{\mathcal G}^{\star} \ .
\end{equation}
For any grid assignment $\gamma$, we can define its population excess over the best grid assignment by
\begin{equation}
R(\gamma)
=
\bar{\mathcal{P}}_{\mathcal Q}(\gamma)
-
\bar{\mathcal{P}}_{\mathcal G}^{\star} \ .
\end{equation}
The goal is to show that the empirical minimizer cannot have a large value of $R(\gamma)$.

To do so, we compare $\gamma$ directly to $\gamma^{\star}$. Define:
\begin{equation}
Y_i(\gamma)
=
X_i(\gamma)-X_i(\gamma^{\star}) \ .
\end{equation}
Then:
\begin{equation}
\mathbb E_{q_i\sim\mathcal Q}Y_i(\gamma)
=
R(\gamma) \ .
\end{equation}
Also, since $0\leq X_i(\gamma),X_i(\gamma^{\star})\leq1$, we have:
\begin{equation}
Y_i(\gamma)^2
=
\left(X_i(\gamma)-X_i(\gamma^{\star})\right)^2
\leq
X_i(\gamma)+X_i(\gamma^{\star}) \ .
\end{equation}
Therefore:
\begin{equation}
\mathrm{Var}\left(Y_i(\gamma)\right)
\leq
\mathbb E Y_i(\gamma)^2
\leq
\bar{\mathcal{P}}_{\mathcal Q}(\gamma)
+
\bar{\mathcal{P}}_{\mathcal Q}(\gamma^{\star})
=
R(\gamma)+2\bar{\mathcal{P}}_{\mathcal G}^{\star} \ .
\end{equation}

We can now apply Bernstein's inequality to the variables $Y_i(\gamma)$. For a fixed assignment $\gamma$, the event that $\gamma$ looks at least as good as $\gamma^{\star}$ on the sampled questions is:
\begin{equation}
\bar{\mathcal{P}}_{\mathcal Q'}(\gamma)
\leq
\bar {\mathcal{P}}_{\mathcal Q'}(\gamma^{\star}) \ ,
\end{equation}
or equivalently:
\begin{equation}
\frac{1}{n}
\sum_{i=1}^{n}Y_i(\gamma)
\leq
0 \ .
\end{equation}
Since the expectation of this empirical average is $R(\gamma)$, Bernstein's inequality gives
\begin{equation}
\Pr\left(
\frac{1}{n}
\sum_{i=1}^{n}Y_i(\gamma)
\leq
0
\right)
\leq
\exp\left(
-
\frac{
nR(\gamma)^2
}{
4\bar{\mathcal{P}}_{\mathcal G}^{\star}+\frac{10}{3}R(\gamma)
}
\right) \ .
\end{equation}
This means that all that remains is to make this bound uniform over all possible grid assignments. Since the objective is invariant to the order of the $K$ temperatures, it is enough to count unordered assignments. The number of multisets of size $K$ from the $M+1$ grid points is
\begin{equation}
\binom{K+M}{M} \ .
\end{equation}
Using a union bound over these assignments, and recalling the definition of $\Lambda$ in Equation \ref{eq:lem:Finite:Wiz}, we get that with probability at least $1-\rho'$, every assignment $\gamma$ satisfying:
\begin{equation}
\frac{
nR(\gamma)^2
}{
4\bar{\mathcal{P}}_{\mathcal G}^{\star}+\frac{10}{3}R(\gamma)
}
>
\Lambda
\end{equation}
has empirical value strictly larger than that of $\gamma^{\star}$.

Now apply this to the empirical minimizer $\hat\gamma=(\hat\gamma_1,\ldots,\hat\gamma_K)$ from Equation \ref{eq:lemFinite:empMin}. Since $\hat\gamma$ minimizes the empirical objective, it satisfies:
\begin{equation}
\bar{\mathcal{P}}_{\mathcal Q'}(\hat\gamma)
\leq
\bar{\mathcal{P}}_{\mathcal Q'}(\gamma^{\star}) \ .
\end{equation}
Therefore, on the high-probability event above, $\hat\gamma$ cannot satisfy the previous strict inequality. Hence:
\begin{equation}
\frac{
nR(\hat\gamma)^2
}{
4\bar{\mathcal{P}}_{\mathcal G}^{\star}+\frac{10}{3}R(\hat\gamma)
}
\leq
\Lambda \ .
\end{equation}
Solving this quadratic inequality gives:
\begin{equation}
R(\hat\gamma)
\leq
\frac{5\Lambda}{3n}
+
\sqrt{
\frac{
4\bar{\mathcal{P}}_{\mathcal G}^{\star}\Lambda
}{n}
+
\frac{
25\Lambda^2
}{9n^2}
} \ .
\end{equation}
Using $\sqrt{a+b}\leq \sqrt a+\sqrt b$, we obtain:
\begin{equation}
R(\hat\gamma)
\leq
2
\sqrt{
\frac{
\bar{\mathcal{P}}_{\mathcal G}^{\star}\Lambda
}{n}
}
+
\frac{10\Lambda}{3n} \ .
\end{equation}

Finally, by Equation \ref{eq:lemFinite:popPlug}, we have:
\begin{equation}
R(\hat\gamma)
=
\bar{\mathcal{P}}'_{\mathcal Q}
-
\bar{\mathcal{P}}_{\mathcal G}^{\star} \ .
\end{equation}
Since $\bar{\mathcal{P}}_{\mathcal G}^{\star}$ is the minimum of the population objective over the grid, we also have:
\begin{equation}
0
\leq
\bar{\mathcal{P}}'_{\mathcal Q}
-
\bar{\mathcal{P}}_{\mathcal G}^{\star} \ .
\end{equation}
Combining the last two displays proves that, with probability at least $1-\rho'$,
\begin{equation}
0
\leq
\bar{\mathcal{P}}'_{\mathcal Q}
-
\bar{\mathcal{P}}_{\mathcal G}^{\star}
\leq
2
\sqrt{
\frac{
\bar{\mathcal{P}}_{\mathcal G}^{\star}\Lambda
}{n}
}
+
\frac{10\Lambda}{3n} \ .
\end{equation}
Which complete our proof.
\end{proof}

\subsection{Lemma \ref{zap:lem:TempSat} Proof}

\label{zap:lem:TempSat:proof}

We will now prove Lemma \ref{zap:lem:TempSat}. The argument is a direct application of a standard idea from convex geometry. The key step is to move beyond viewing a temperature as a scalar and instead represent it by the vector it induces over the sampled questions.

For each candidate temperature $\tau_j$, define its log-failure vector across the $n$ questions by
\begin{equation}
a_j
=
\left(
\log \bar{P}_{1j},
\ldots,
\log \bar{P}_{nj}
\right)
\in \mathbb{R}^n .
\end{equation}
Thus, for the finite optimization problem, each temperature corresponds to a point in $\mathbb{R}^n$, where the $i$-th coordinate records its log-failure probability on the $i$-th question.

Now consider a relaxed $K$-attempt strategy. Such a strategy assigns a total mass of $K$ across the candidate temperatures. Equivalently, it produces a total log-failure profile $z$, obtained by averaging the temperature profiles and scaling by $K$. The set of all achievable vectors $z$ is therefore exactly the scaled convex hull
\begin{equation}
K \cdot \operatorname{conv}\{a_1,\ldots,a_l\}.
\end{equation}
This is the geometric core of the argument: the optimization is no longer over temperatures, but over points $z$ obtained by mixing their log-failure profiles.

The objective depends on the strategy only through $z$. For $z=(z_1,\ldots,z_n)$, the average failure probability is
\begin{equation}
F(z)
=
\frac{1}{n}\sum_{i=1}^{n}\exp(z_i).
\end{equation}
Hence, the relaxed problem becomes
\begin{equation}
\min_{z\in K\cdot \operatorname{conv}\{a_1,\ldots,a_l\}} F(z).
\end{equation}

Let $z^*$ be an optimal solution. Since $F$ is convex and differentiable, first-order optimality implies that the gradient of $F$ at $z^*$ defines a supporting hyperplane to the feasible polytope. Equivalently, $z^*$ lies on an exposed face of the polytope; see, for example, \cite{rockafellar1970convex}.

Moreover, the gradient of $F$ is strictly positive in every coordinate and thus nonzero. Consequently, the supporting hyperplane is nondegenerate, and the exposed face containing $z^*$ has dimension at most $n-1$.

We now apply Carathéodory's theorem within this face \cite{caratheodory1907,barvinok2002course}. While the standard theorem guarantees a representation using at most $d+1$ points in $\mathbb{R}^d$, here we operate within a set of dimension at most $n-1$. Therefore, $z^*$ can be expressed as a convex combination of at most $n$ of the vectors $\{a_j\}$.

That is, there exist indices $j_1,\ldots,j_m$ with $m\leq n$, and nonnegative weights $\lambda_1,\ldots,\lambda_m$ satisfying
\begin{equation}
\sum_{r=1}^{m}\lambda_r = 1,
\end{equation}
such that
\begin{equation}
z^*
=
\sum_{r=1}^{m}\lambda_r K a_{j_r}.
\end{equation}

Translating this representation back into the language of temperature strategies, we obtain an alternative optimal strategy that achieves the same log-failure profile $z^*$ using only the temperatures $\tau_{j_1},\ldots,\tau_{j_m}$. Since the objective depends only on $z^*$, this strategy attains the same optimal value.

Therefore, there always exists an optimal strategy that uses at most $n$ distinct temperatures, where $n$ is the number of sampled questions. 

This completes the proof.

\subsection{\texorpdfstring{Proofs of the Short-Sampling Effect Propositions}{Proofs of the Short-Sampling Effect Propositions}}

\label{zap:sec:short:proof}

In this section we present the proofs corresponding to Appendix \ref{zap:sec:short}, which are included here to avoid unnecessary clutter.

We start with the proof of the first lemma of the Appendix.

\begin{proof}[\textbf{Proof - Proposition \ref{claim:short:W}}]
The proof is again only an algebraic extension of the symmetric two-question
case. The only difference is that the two groups of questions no longer have the same weight. A fraction $r$ of the questions behaves like the first question, and a fraction $1-r$ behaves like the second question.

As before, using the compromise temperature assigns all $K$ samples to the same temperature. Therefore, its failure probability remains exactly the same as in the unweighted case:
\begin{equation}
    \bar{\mathcal{P}}_{\mathrm{comp}}
    =
    \left(1-P_C\right)^K \ .
\end{equation}
Now consider the two-expert strategy, where we allocate $K'$ samples to
the first expert temperature and $K-K'$ samples to the second expert temperature.

Since a fraction $r$ of the questions belongs to the first group and a fraction $1-r$ belongs to the second group, the weighted average failure probability is
\begin{equation}
    \bar{\mathcal{P}}_{\mathrm{expert}}(K')
    =
    r\bar{P}_E^{K'}
    +
    (1-r)\bar{P}_E^{K-K'} \ .
    \label{eq:weighted:expert:failure}
\end{equation}
Then we can optimize this expression over $K'$ in the continuous relaxation. The minimum is obtained when the two weighted contributions in Equation \ref{eq:weighted:expert:failure} are balanced. Indeed, differentiating
with respect to $K'$ gives
\begin{equation}
    r\bar{{P}}_E^{K'}
    =
    (1-r)\bar{{P}}_E^{K-K'} \ .
    \label{eq:weighted:balance}
\end{equation}
This equation has a simple interpretation. At the optimum, neither side should dominate the total failure. If one group contributed much more to the weighted failure than the other, we could move a small amount of sampling mass toward that group and improve the strategy.

Substituting the balancing condition from Equation \ref{eq:weighted:balance} back into Equation \ref{eq:weighted:expert:failure}, we obtain
\begin{equation}
    \bar{\mathcal{P}}_{\mathrm{expert}}^{\star}
    =
    2\sqrt{r(1-r)}\bar{P}_E^{\frac{K}{2}} \ .
    \label{eq:weighted:expert:optimal}
\end{equation}
This is the key point. In the original symmetric proposition, the expert failure term is $\bar{P}_E^{\frac{K}{2}}$. In the weighted setting, after optimizing the allocation between the two expert temperatures, the same term is multiplied by $2\sqrt{r(1-r)}$. Hence the weighted problem is equivalent to the original one after replacing $\bar{P}_E$ by an effective failure probability $\widetilde{\bar{P}}_E$ satisfying:
\begin{equation}
    \widetilde{\bar{P}}_E^{\frac{K}{2}}
    =
    2\sqrt{r(1-r)}\bar{P}_E^{\frac{K}{2}} \ .
\end{equation}
Taking both sides to the power $\frac{2}{K}$ gives:
\begin{equation}
    \widetilde{\bar{P}}_E
    =
    \sqrt[K]{4r(1-r)}\bar{P}_E \ .
\end{equation}
This proves the exact failure-probability substitution in Equation \ref{zap:eq:Severus}.

It remains to translate the same statement into the first-order success-probability language. Using the standard low-probability approximation,\:
\begin{equation}
    \bar{P}_E^{\frac{K}{2}}
    =
    (1-P_E)^{\frac{K}{2}}
    =
    e^{-P_E\frac{K}{2}}
    \left(
    1+O(KP_E^2)
    \right) \ ,
\end{equation}
Equation \ref{eq:weighted:expert:optimal} becomes
\begin{equation}
    \bar{\mathcal{P}}_{\mathrm{expert}}^{\star}
    =
    2\sqrt{r(1-r)}
    e^{-P_E\frac{K}{2}}
    \left(
    1+O(KP_E^2)
    \right) \ .
\end{equation}
Since:
\begin{equation}
    2\sqrt{r(1-r)}
    =
    e^{\frac{1}{2}\log(4r(1-r))} \ ,
\end{equation}
we may rewrite this as:
\begin{equation}
    \bar{\mathcal{P}}_{\mathrm{expert}}^{\star}
    =
    \exp\left(
    -\frac{K}{2}
    \left(
    P_E-\frac{1}{K}\log\left(4r(1-r)\right)
    \right)
    \right)
    \left(
    1+O(KP_E^2)
    \right) \ .
\end{equation}
Thus, to first order, the weighted case has exactly the same form as Proposition \ref{claim:short}, but with the effective substitution
\begin{equation}
    P_E
    \mapsto
    P_E-\frac{1}{K}\log\left(4r(1-r)\right) \ .
\end{equation}
This is precisely Equation \ref{zap:eq:snape}.

Therefore, optimizing the allocation between the two expert temperatures reduces the weighted setting to the same short-sampling comparison as before, with the modified effective expert strength. 

This proves the proposition.

\end{proof}

Secondly, we will now present the two gaussians case proof, as the way this proof works will also be generalized to other more complicated cases.

\begin{proof}[\textbf{Proof - Proposition \ref{claim:short:Gaussian}}]
The proof is again a direct comparison between the compromise temperature and the
best possible two-temperature strategy. The only point that requires care is that
the two expert temperatures should not be assumed to lie exactly at the Gaussian
peaks. Instead, we allow them to move, and ask when moving away from the compromise
temperature actually improves the objective.

By symmetry, we place the compromise temperature at $0$, and write the two
Gaussian peaks as $-a$ and $a$. Thus,
\begin{equation}
    P_1(\tau)
    =
    A\exp\left(
    -\frac{(\tau+a)^2}{2\sigma^2}
    \right),
    \qquad
    P_2(\tau)
    =
    A\exp\left(
    -\frac{(\tau-a)^2}{2\sigma^2}
    \right) \ .
\end{equation}
The peak probability is therefore:
\begin{equation}
    P_E=A \ ,
\end{equation}
while the compromise probability is the value of either Gaussian at the
intersection point:
\begin{equation}
    P_C
    =
    A\exp\left(
    -\frac{a^2}{2\sigma^2}
    \right) \ .
\end{equation}
Hence:
\begin{equation}
    \frac{P_E}{P_C}
    =
    \exp\left(
    \frac{a^2}{2\sigma^2}
    \right) \ .
    \label{eq:claim:gaussian:ratio:proof}
\end{equation}

We now allow the two expert temperatures to move symmetrically around the
compromise temperature. Write them as $-x$ and $x$. The case $x=0$ corresponds to
using the compromise temperature twice. Therefore, the two-expert strategy
outperforms the compromise temperature exactly when some $x>0$ gives a larger
first-order success probability than $x=0$.

For a fixed value of $x$, each question receives half of the samples at one
temperature and half at the other. By symmetry, the effective first-order success probability is
\begin{equation}
    S(x)
    =
    \frac{A}{2}
    \left(
    \exp\left(
    -\frac{(a-x)^2}{2\sigma^2}
    \right)
    +
    \exp\left(
    -\frac{(a+x)^2}{2\sigma^2}
    \right)
    \right) \ .
\end{equation}
At the compromise temperature, this becomes:
\begin{equation}
    S(0)
    =
    A\exp\left(
    -\frac{a^2}{2\sigma^2}
    \right)
    =
    P_C \ .
\end{equation}

We now compare $S(x)$ with $S(0)$. Factoring out the compromise value gives
\begin{equation}
    \frac{S(x)}{S(0)}
    =
    \exp\left(
    -\frac{x^2}{2\sigma^2}
    \right)
    \cosh\left(
    \frac{ax}{\sigma^2}
    \right) \ .
    \label{eq:claim:gaussian:Sratio}
\end{equation}

First suppose that $a\leq \sigma$. Using the standard inequality
\begin{equation}
    \cosh(u)\leq \exp\left(\frac{u^2}{2}\right) \ ,
\end{equation}
Equation \ref{eq:claim:gaussian:Sratio} gives
\begin{equation}
    \frac{S(x)}{S(0)}
    \leq
    \exp\left(
    -\frac{x^2}{2\sigma^2}
    +
    \frac{a^2x^2}{2\sigma^4}
    \right)
    =
    \exp\left(
    -\frac{x^2}{2\sigma^2}
    \left(
    1-\frac{a^2}{\sigma^2}
    \right)
    \right)
    \leq 1 \ .
\end{equation}
Thus, when $a\leq\sigma$, moving away from the compromise temperature cannot
improve the objective. In this case, the best two-temperature strategy collapses to the compromise temperature.

Now suppose that $a>\sigma$. Expanding Equation
\ref{eq:claim:gaussian:Sratio} around $x=0$, we obtain:
\begin{equation}
    \frac{S(x)}{S(0)}
    =
    1
    +
    \frac{x^2}{2\sigma^2}
    \left(
    \frac{a^2}{\sigma^2}-1
    \right)
    +
    O(x^4) \ .
\end{equation}
Since $a>\sigma$, the coefficient of $x^2$ is positive. Therefore, for sufficiently small $x>0$, we have:
\begin{equation}
    S(x)>S(0) \ .
\end{equation}
Hence, the optimized two-expert strategy strictly outperforms the compromise temperature exactly when:
\begin{equation}
    a>\sigma \ .
\end{equation}

It remains only to translate this condition into the ratio between the peak probability $P_E$ and the compromise probability $P_C$. By Equation \ref{eq:claim:gaussian:ratio:proof}, the condition $a>\sigma$ is equivalent to:
\begin{equation}
    \frac{P_E}{P_C}
    =
    \exp\left(
    \frac{a^2}{2\sigma^2}
    \right)
    >
    e^{1/2}
    =
    \sqrt{e} \ .
\end{equation}
Equivalently,
\begin{equation}
    P_E>\sqrt{e}P_C \simeq 1.65P_C \ .
\end{equation}

This proves the proposition.
\end{proof}

\newpage

\section*{NeurIPS Paper Checklist}

\begin{enumerate}

\item {\bf Claims}
    \item[] Question: Do the main claims made in the abstract and introduction accurately reflect the paper's contributions and scope?
    \item[] Answer: \answerYes{} %
    \item[] Justification: The main claims in the abstract and introduction are aligned with the actual contributions: profile structure, hardness, efficient-temperature overlap, the ideal multi-temperature allocation, and the dulling effects are all developed in the main text and supported in the appendix. 

\item {\bf Limitations}
    \item[] Question: Does the paper discuss the limitations of the work performed by the authors?
    \item[] Answer: \answerYes{} %
    \item[] Justification: The paper includes a dedicated limitations discussion in Appendix~\ref{zapp:limitations}. It discusses empirical coverage, the gap between the theoretical assumptions and full real-world generality, the partial nature of the short-sampling theory, and the need for broader validation.

\item {\bf Theory assumptions and proofs}
    \item[] Question: For each theoretical result, does the paper provide the full set of assumptions and a complete (and correct) proof?
    \item[] Answer: \answerYes{} %
    \item[] Justification: The formal results state their assumptions and are accompanied by proofs or derivations in the Appendix.

    \item {\bf Experimental result reproducibility}
    \item[] Question: Does the paper fully disclose all the information needed to reproduce the main experimental results of the paper to the extent that it affects the main claims and/or conclusions of the paper (regardless of whether the code and data are provided or not)?
    \item[] Answer: \answerYes{} %
    \item[] Justification: The paper provides the information needed to reproduce the main empirical conclusions: the pass@$K$ formulation is defined in Section~\ref{sec:passK}, the models and LiveCodeBench V1-V6 setup are stated in Section~\ref{sec:Notations}, and the appendix gives the temperature grids, sample counts, filtering rule, fitting procedure, optimization formulation, and pass@$K$ tables.
For the temperature-allocation comparison on MATH-500, Appendix~\ref{zap:CompResources} gives the dataset source, prompts, temperature grids, sample counts, token limits, and grading rules.
For the comparison on Hi-ToM, Appendix~\ref{zap:CompResources} gives the same information, together with the first stage that selects each model's questions. Our code, at \url{https://github.com/avivorly/one-temperature}, implements the sampling, grading, pass@$K$ estimation and temperature-allocation steps; see Appendix~\ref{zap:CompResources}.

\item {\bf Open access to data and code}
    \item[] Question: Does the paper provide open access to the data and code, with sufficient instructions to faithfully reproduce the main experimental results, as described in supplemental material?
    \item[] Answer: \answerYes{} %
    \item[] {\par\raggedright Justification: The code is available at \url{https://github.com/avivorly/one-temperature}. It covers sampling with vLLM over a temperature grid, grading, the unbiased pass@$K$ estimator, and the temperature-allocation methods with held-out evaluation, and its README gives the exact commands. All benchmarks and models are public; their sources, versions and licenses are listed under the licenses item.\par}

\item {\bf Experimental setting/details}
    \item[] Question: Does the paper specify all the training and test details (e.g., data splits, hyperparameters, how they were chosen, type of optimizer) necessary to understand the results?
    \item[] Answer: \answerYes{} %
    \item[] Justification: The experimental setting is described in the main text and in Appendix~\ref{zap:CompResources} and Appendix~\ref{zap:QuestionsProfiles}. 
The MATH-500 and Hi-ToM settings are in Appendix~\ref{zap:CompResources}.
The \texttt{min\_p} setting and results are in Appendix~\ref{zap:MinP}.

\item {\bf Experiment statistical significance}
    \item[] Question: Does the paper report error bars suitably and correctly defined or other appropriate information about the statistical significance of the experiments?
    \item[] Answer: \answerYes{} %
    \item[] Justification: The paper reports all of the uncertainty information and explains how it is computed.

\item {\bf Experiments compute resources}
    \item[] Question: For each experiment, does the paper provide sufficient information on the computer resources (type of compute workers, memory, time of execution) needed to reproduce the experiments?
    \item[] Answer: {\answerYes{}} %
    \item[] Justification: Appendix~\ref{zap:CompResources} names the GPU types (NVIDIA L40S, A40, GeForce RTX 5090, and GeForce RTX 4090) and the \texttt{vLLM} inference engine, and gives the number of samples per question and temperature for each model and dataset, the GPU memory, and about 14,000 GPU-hours in total.
The \texttt{min\_p} sampling of Appendix~\ref{zap:MinP} ran on NVIDIA L40S, GeForce RTX 5090, and GeForce RTX 4090 GPUs with \texttt{vLLM} 0.19.0, as reported there.
    
\item {\bf Code of ethics}
    \item[] Question: Does the research conducted in the paper conform, in every respect, with the NeurIPS Code of Ethics \url{https://neurips.cc/public/EthicsGuidelines}?
    \item[] Answer: \answerYes{} %
    \item[] Justification: The work studies existing models on three public benchmarks (LiveCodeBench, MATH-500, and Hi-ToM) and does not introduce human-subject experiments, private data, or deployment decisions about individuals.

\item {\bf Broader impacts}
    \item[] Question: Does the paper discuss both potential positive societal impacts and negative societal impacts of the work performed?
    \item[] Answer: \answerNA{} %
    \item[] Justification: This is foundational work: it analyses how the sampling temperature affects pass@$K$ for existing public models and benchmarks, and it releases no new model, dataset or capability. Its practical effect is limited to choosing sampling temperatures more efficiently, which can lower the inference compute needed for a given pass@$K$. We see no direct negative societal impact beyond those of the models it studies.
    
\item {\bf Safeguards}
    \item[] Question: Does the paper describe safeguards that have been put in place for responsible release of data or models that have a high risk for misuse (e.g., pre-trained language models, image generators, or scraped datasets)?
    \item[] Answer: \answerNA{} %
    \item[] Justification: Our work does not involve releasing any models or datasets that pose a risk of misuse.

\needspace{5\baselineskip}%
\item {\bf Licenses for existing assets}
    \item[] Question: Are the creators or original owners of assets (e.g., code, data, models), used in the paper, properly credited and are the license and terms of use explicitly mentioned and properly respected?
    \item[] Answer: \answerYes{}{} %
    \item[] {\par\sloppy Justification: Each asset is cited and named with its creator, version, URL and license, and we use each only for research evaluation under that license. Models, all Apache-2.0 per their Hugging Face model cards: Mistral AI's Mistral-7B-Instruct-v0.3 \cite{jiang2023mistral}, \url{https://huggingface.co/mistralai/Mistral-7B-Instruct-v0.3}; the Qwen team's Qwen3-8B and Qwen3-0.6B \cite{yang2025qwen3}, \url{https://huggingface.co/Qwen/Qwen3-8B} and \url{https://huggingface.co/Qwen/Qwen3-0.6B}; and OpenAI's gpt-oss-20b \cite{openai2025gptoss}, \url{https://huggingface.co/openai/gpt-oss-20b}. Benchmarks: LiveCodeBench \cite{jain2024livecodebench}, release\_v6 (V1--V6) of \url{https://huggingface.co/datasets/livecodebench/code_generation_lite} (dataset card: \texttt{cc}), with code at \url{https://github.com/LiveCodeBench/LiveCodeBench} (MIT); MATH-500 \cite{hendrycks2021measuring,lightman2023lets}, \url{https://huggingface.co/datasets/HuggingFaceH4/MATH-500} at revision \texttt{6e4ed1a} (no license stated), which is the PRM800K test split of MATH, from \url{https://github.com/openai/prm800k} and \url{https://github.com/hendrycks/math} (both MIT); and Hi-ToM \cite{he2023hitom}, the 1,200-question \texttt{Hi-ToM\_data.json} of \url{https://github.com/ying-hui-he/Hi-ToM_dataset} (Apache-2.0). Inference engine: vLLM \cite{kwon2023vllm}, \url{https://github.com/vllm-project/vllm} (Apache-2.0), version 0.19.0 for the runs of Appendix~\ref{zap:MinP}. We scraped no data: LiveCodeBench states that it collected its problems from LeetCode, AtCoder and Codeforces, and we used its released files. We re-package and release none of these assets.\par}

\item {\bf New assets}
    \item[] Question: Are new assets introduced in the paper well documented and is the documentation provided alongside the assets?
    \item[] Answer: \answerNA{} %
    \item[] Justification: The paper does not introduce a new dataset.

\item {\bf Crowdsourcing and research with human subjects}
    \item[] Question: For crowdsourcing experiments and research with human subjects, does the paper include the full text of instructions given to participants and screenshots, if applicable, as well as details about compensation (if any)? 
    \item[] Answer: \answerNA{} %
    \item[] Justification: The research does not involve crowdsourcing experiments or research with human subjects

\item {\bf Institutional review board (IRB) approvals or equivalent for research with human subjects}
    \item[] Question: Does the paper describe potential risks incurred by study participants, whether such risks were disclosed to the subjects, and whether Institutional Review Board (IRB) approvals (or an equivalent approval/review based on the requirements of your country or institution) were obtained?
    \item[] Answer: \answerNA{} %
    \item[] Justification: The research does not involve crowdsourcing, study participants, or other human-subject research.

\item {\bf Declaration of LLM usage}
    \item[] Question: Does the paper describe the usage of LLMs if it is an important, original, or non-standard component of the core methods in this research? Note that if the LLM is used only for writing, editing, or formatting purposes and does \emph{not} impact the core methodology, scientific rigor, or originality of the research, declaration is not required.
    \item[] Answer: \answerNA{} %
    \item[] Justification: While LLMs were used in preparing this work, their role was limited to editing, assisting with technical proofs, and supporting non-fundamental coding tasks. They were not used to formulate the research ideas, develop the main contributions, or discover the theorems. The final technical content and claims were produced, reviewed, and verified by the authors.

\end{enumerate}

\end{document}